\documentclass[runningheads]{llncs}
\usepackage{graphicx}
\usepackage{amsmath,amssymb} 
\usepackage{color}
\usepackage[width=122mm,left=12mm,paperwidth=146mm,height=193mm,top=12mm,paperheight=217mm]{geometry}
\usepackage[table,usenames,dvipsnames]{xcolor}%
\usepackage{epsfig}
\usepackage{epstopdf} 
\usepackage{bm}
\usepackage{caption,subcaption}
\usepackage{multirow} 
\usepackage[normalem]{ulem}
\usepackage{mdframed} 
\usepackage{wrapfig} 
\usepackage{hyperref}
\hypersetup{
    colorlinks=true,
    linkcolor=blue,
    citecolor=blue,
    urlcolor=blue
}
\usepackage{placeins}
\usepackage[symbol]{footmisc}

\newcommand{\pz}{\phantom{0}}%

\definecolor{dgreen}{rgb}{0.0,0.6,0.0} 
\definecolor{dred}{rgb}{0.6,0.0,0.0} 
\definecolor{BrickRed}{rgb}{0.72,0.0,0.0}%
\definecolor{grey}{rgb}{0.6,0.6,0.6}%

\newcommand{\vx}{{\textbf{x}}}

\newcommand{\vw}{{\textbf{w}}}
\newcommand{\vy}{{\textbf{y}}}
\newcommand{\vu}{{\textbf{u}}}
\newcommand{\vv}{{\textbf{v}}}

\newcommand{\etall}{{et al.}} 

\begin{document}
\pagestyle{headings}
\mainmatter
\def\ECCV18SubNumber{1377}  

\title{Uncertainty Estimates and Multi-Hypotheses Networks for Optical Flow} 

\titlerunning{Uncertainty Estimates and Multi-Hypotheses Networks for Optical Flow}



\author{Eddy Ilg\textsuperscript{*} \and 
\"Ozg\"un \c{C}i\c{c}ek\textsuperscript{*} \and
Silvio Galesso\textsuperscript{*} \and 
Aaron Klein \and 
Osama Makansi \and 
Frank Hutter \and 
Thomas Brox}
\institute{
University of Freiburg, Germany \\
\email{\{ilg,cicek,galessos,kleinaa,makansio,fh,brox\}@cs.uni-freiburg.de}
}

\authorrunning{E. Ilg, \"O. \c{C}i\c{c}ek, S. Galesso, A. Klein, O. Makansi, F. Hutter and T. Brox}




\maketitle

\begin{abstract}
Optical flow estimation can be formulated as an end-to-end supervised learning problem, which yields estimates with a superior accuracy-runtime tradeoff compared to alternative methodology. 
In this paper, we make such networks estimate their local uncertainty about the correctness of their prediction, which is vital information when building decisions on top of the estimations. For the first time we compare several strategies and techniques to estimate uncertainty in a large-scale computer vision task like optical flow estimation. 
Moreover, we introduce a new network architecture utilizing the Winner-Takes-All loss and show that this can provide complementary hypotheses and uncertainty estimates efficiently with a single forward pass and without the need for sampling or ensembles. 
Finally, we demonstrate the quality of the different uncertainty estimates, which is clearly above previous confidence measures on optical flow and allows for interactive frame rates. 
\end{abstract}

\section{Introduction}
\renewcommand{\thefootnote}{\fnsymbol{footnote}}
\footnotetext[1]{equal contribution}
Recent research has shown that deep networks typically outperform handcrafted approaches in computer vision in terms of accuracy and speed.
Optical flow estimation is one example: FlowNet \cite{flownet,flownet2} yields high accuracy optical flow at interactive frame rates, which is relevant for many applications in the automotive domain or for activity understanding. 

\begin{figure}[t]
  \begin{center}
      \newcommand{\galleryRowKitti}[3]{
\includegraphics[width=0.32\textwidth]{images/kitti/#1} & \includegraphics[width=0.32\textwidth]{images/kitti/#2} & \includegraphics[width=0.32\textwidth]{images/kitti/#3} \tabularnewline
}
\resizebox{\linewidth}{!}{%
{%
\setlength{\tabcolsep}{2.4pt}%
\begin{tabular}{ccc}%
\galleryRowKitti{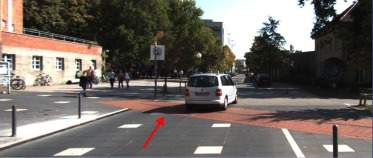}{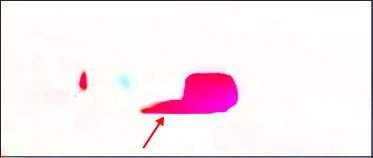}{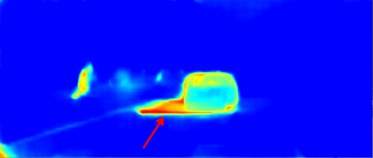}%
\end{tabular}%
}%
}

  \end{center}
\caption{Joint estimation of optical flow and its uncertainty. \textbf{Left:} Image from a KITTI 2015 sequence. \textbf{Middle:} Estimated optical flow. \textbf{Right:} The estimated uncertainty (visualized as heatmap) marks the optical flow in the shadow of the car as unreliable (pointed by the red arrow), contrary to the car itself, which is estimated with higher certainty. Marked as most reliable is the optical flow for the static background.
\label{fig:teaser}
}
\end{figure}

A valid critique of learning-based approaches is their black-box nature: since all parts of the problem are learned from data, there is no strict understanding on how the problem is solved by the network. Although FlowNet 2.0 \cite{flownet2} was shown to generalize well across various datasets, there is no guarantee that it will also work in different scenarios that contain unknown challenges.
In real-world scenarios, such as control of an autonomously driving car, an erroneous decision can be fatal; thus it is not possible to deploy such a system without information about how reliable the underlying estimates are. 
We should expect an additional estimate of the network's own uncertainty, such that the network can highlight hard cases where it cannot reliably estimate the optical flow or where it must decide among multiple probable hypotheses; see Figure~\ref{fig:teaser}.  
However, deep networks in computer vision typically yield only their single preferred prediction rather than the parameters of a distribution. 

The first contribution of this paper is an answer to the open question which of the many approaches for  uncertainty estimation, most of which have been applied only to small problems so far, are most efficient for high-resolution encoder-decoder regression networks. We provide a comprehensive study of empirical ensembles, predictive models, and predictive ensembles. 
The first category comprises frequentist methods, the second one relies on the estimation of a parametric output distribution, and the third one combines the properties of the previous two.
We implemented these approaches for FlowNet using the common MC dropout technique~\cite{baysianDropout}, the less common Bootstrapped Ensembles~\cite{deepEnsembles} and snapshot ensembles~\cite{snapshotEnsembles}. 
We find that in general all these approaches yield surprisingly good uncertainty estimates, where the best performance is achieved with uncertainty estimates derived from Bootstrapped Ensembles of predictive networks.

While such ensembles are a good way to obtain uncertainty estimates, they must run multiple networks to create sufficiently many samples. This drawback increases the computational load and memory footprint at training and test time linearly with the number of samples, such that these approaches are not applicable in real-time.

As a second contribution, we utilize a multi-headed network inspired by~\cite{rupprecht} that yields multiple hypotheses in a single network without the need of sampling.
To obtain the hypotheses, we use the Winner-Takes-All (WTA) loss~\cite{MultipleChoice,MultipleChoiceEnsembles,PhotographicImageSynthesis,rupprecht} to penalize only the best prediction and push the network to make multiple different predictions in case of doubt.
We propose to stack a second network to optimally combine the hypotheses and to estimate the final uncertainty.
This setup yields slightly better uncertainty estimates as Bootstrapped Ensembles, but allows for interactive frame rates. Thus, in this paper, we address all three important aspects for deployment of optical flow estimation in automotive systems: high accuracy inherited from the base network, a measure of reliability, and a fast runtime.



\section{Related Work} 

\textbf{Confidence measures for optical flow.}
While there is a large number of optical flow estimation methods, only few of them provide uncertainty estimates.

\textit{Post-hoc} methods apply post-processing to  already estimated flow fields.
Kondermann \etall~\cite{Kondermann2007} used a learned linear subspace of typical displacement neighborhoods to test the reliability of a model. In their follow-up work~\cite{Kondermann2008}, they proposed a hypothesis testing method based on probabilistic motion models learned from ground-truth data. Aodha \etall~\cite{opticalFlowPAMI2012} trained a binary classifier to predict whether the endpoint error of each pixel is bigger or smaller than a certain threshold and used the predicted classifier's probability as an uncertainty measure. All post-hoc methods ignore information given by the model structure.

\textit{Model-inherent} methods, in contrast, produce their uncertainty estimates using the internal estimation model, i.e., energy minimization models. Bruhn and Weickert~\cite{Bruhn2006} used the inverse of the energy functional as a measure of the deviation from the model assumptions. 
Kybic and Nieuwenhuis ~\cite{bootstrapOpticalFlow} performed bootstrap sampling on the data term of an energy-based method in order to obtain meaningful statistics of the flow prediction.
The most recent work by Wannenwetsch \etall~\cite{probFlow} derived a probabilistic approximation of the posterior of the flow field from the energy functional and computed flow mean and covariance via Bayesian optimization. 
Ummenhofer \etall~\cite{demon} presented a depth estimation CNN that internally uses a  predictor for the deviation of the estimated optical flow from the ground-truth. This yields a confidence map for the intermediate optical flow that is used internally within the network. However, this approach treats flow and confidence separately and there was no evaluation for the reliability of the confidence measure. 

\textbf{Uncertainty estimation with CNNs.} 
Bayesian neural networks (BNNs) have been shown to obtain well-calibrated uncertainty estimates while maintaining the properties of standard neural networks~\cite{neal-1995,mackay}.
Early work~\cite{neal-1995} mostly used Markov Chain Monte Carlo (MCMC) methods to sample networks from the distribution of the weights, where some, for instance Hamiltonian Monte Carlo, can make use of the gradient information provided by the backpropagation algorithm.
More recent methods generalize traditional gradient based MCMC methods to the stochastic mini-batch setting, where only noisy estimates of the true gradient are available~\cite{chen-icml14,welling-icml11}. 
However, even these recent MCMC methods do not scale well to high-dimensional spaces, and since contemporary encoder-decoder networks like FlowNet have millions of weights, they do not apply in this setting.

Instead of sampling, variational inference methods try to approximate the distribution of the weights by a more tractable distribution \cite{graves-nips11,blundell-icml15a}.
Even though they usually scale much better with the number of datapoints and the number of weights than their MCMC counterparts, they have been applied only to much smaller networks ~\cite{lobato-icml15,blundell-icml15a} than in the present paper.

Gal and Ghahramani~\cite{baysianDropout} sampled the weights by using dropout after each layer and estimated the \textit{epistemic} uncertainty of neural networks. In a follow-up work by Kendall and Gal~\cite{visionUncertainties}, this idea was applied to vision tasks, and the \textit{aleatoric} uncertainty (which explains the noise in the observations) and the epistemic uncertainty (which explains model uncertainty) were studied in a joint framework.
We show in this paper, that the dropout strategy used in all previous computer vision applications~\cite{visionUncertainties,Vedaldi} is not the best one per-se, and other strategies yield better results. 

In contrast to Bayesian approaches, such as MCMC sampling, bootstrapping is a frequentist method that is easy to implement and scales nicely to high-dimensional spaces, since it only requires point estimates of the weights. The idea is to train $M$ neural networks independently on $M$ different bootstrapped subsets of the training data and to treat them as independent samples from the weight distribution.
While bootstrapping does not ensure diversity of the models and in the worst case could lead to $M$ identical models, Lakshminarayanan \etall~\cite{deepEnsembles} argued that ensemble model averaging can be seen as dropout averaging. They trained individual networks with random initialization and random data shuffling, where each network predicts a mean and a variance. During test time, they combined the individual model predictions to account for the epistemic uncertainty of the network.
We also consider so-called \emph{snapshot ensembles} \cite{snapshotEnsembles} in our experiments. These are obtained rather efficiently via Stochastic Gradient Descent with warm Restarts (SGDR)~\cite{sgdr}.

\textbf{Multi-hypotheses estimation.}
The loss function for the proposed multi-hypotheses network is an extension of the Winner-Takes-All (WTA) loss from 
Guzman-Rivera~\etall~\cite{MultipleChoice}, who proposed a similar loss function for SSVMs. Lee~\etall~\cite{MultipleChoiceEnsembles} applied the loss to network ensembles and Chen \& Koltun~\cite{PhotographicImageSynthesis} to a single CNN. 
Rupprecht et al.~\cite{rupprecht} showed that the WTA loss leads to a Voronoi tessellation and used it in a single CNN for diverse future prediction and human pose estimation. Chen \& Koltun~\cite{PhotographicImageSynthesis} used the WTA loss for image synthesis.

\section{Uncertainty Estimation with Deep Networks}
\label{sec:generalnetworks}

Assume we have a dataset $\mathcal{D} = \{(\vx_0, \vy_0^{\mathrm{gt}}), \ldots, (\vx_N, \vy_N^{\mathrm{gt}})\}$, which is generated by sampling from a joint distribution $p(\vx,\vy)$. 
In CNNs, it is assumed that there is a unique mapping from $\vx$ to $\vy$ by a function $f_\vw(\vx)$, which is parametrized by weights $\vw$ that are optimized according to a given loss function on $\mathcal{D}$. 

For optical flow, we denote the trained network as a mapping from the input images $\vx=(\mathbf I_1,\mathbf I_2)$ to the output optical flow $\mathbf{y}=(\vu, \vv)$ as $\mathbf{y}=f_\vw(\mathbf I_1,\mathbf I_2)$, 
where $\vu, \vv$ are the x- and y-components of the optical flow. The FlowNet by Dosovitskiy \etall~\cite{flownet} minimizes the per-pixel endpoint error
\begin{equation}
{\rm EPE} = \sqrt{(u - u^{\mathrm{gt}})^2+(v - v^{\mathrm{gt}})^2} \mathrm{\quad,}
\label{eq:epe}
\end{equation} 
where the pixel coordinates are omitted for brevity.
This network, as depicted in Figure~\ref{fig:schematic_fnc_emp}, is fully deterministic and yields only the network's preferred output $\vy=f_\vw(\vx)$. Depending on the loss function, this typically corresponds to the mean of the distribution $p(\vy|\vx,\mathcal D)$. In this paper, we investigate three major approaches to estimate also the variance $\sigma^2$. These are based on the empirical variance of the distribution of an ensemble, a parametric model of the distribution, and a combination of both. The variance in all these approaches serves as an estimate of the uncertainty.

\newcommand{\rulesep}{\hspace*{1mm}}
\begin{figure*}[t]
    \begin{center}
    \resizebox{\linewidth}{!}{%
    \begin{subfigure}[t]{0.10\textwidth}
        \begin{center}
        \includegraphics[height=3.2cm]{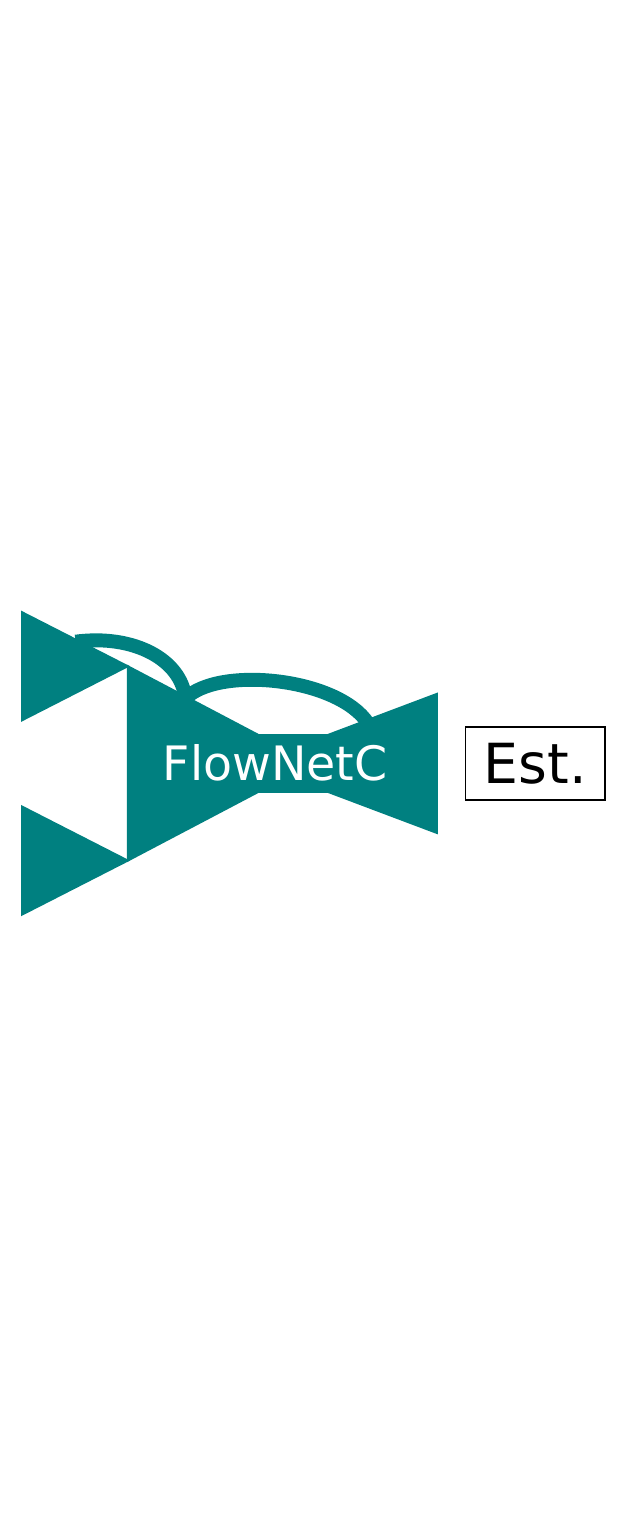}
        \subcaption{
        \label{fig:schematic_fnc_emp}}
        \end{center} 
    \end{subfigure}
    \rulesep
    \begin{subfigure}[t]{0.10\textwidth}
        \begin{center}
        \includegraphics[height=3.2cm]{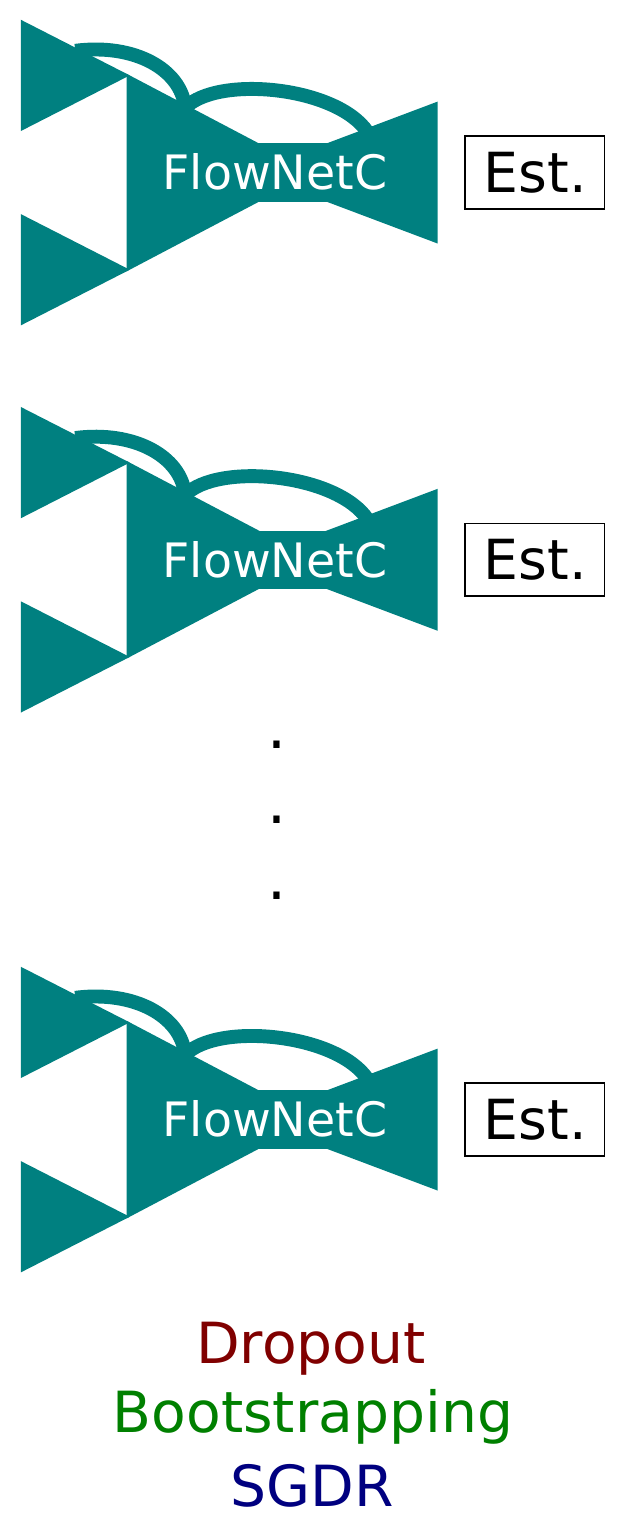}
        \subcaption{
        \label{fig:schematic_emp}}
        \end{center} 
    \end{subfigure}
    \rulesep
    \begin{subfigure}[t]{0.14\textwidth}
        \begin{center} 
        \includegraphics[height=3.2cm]{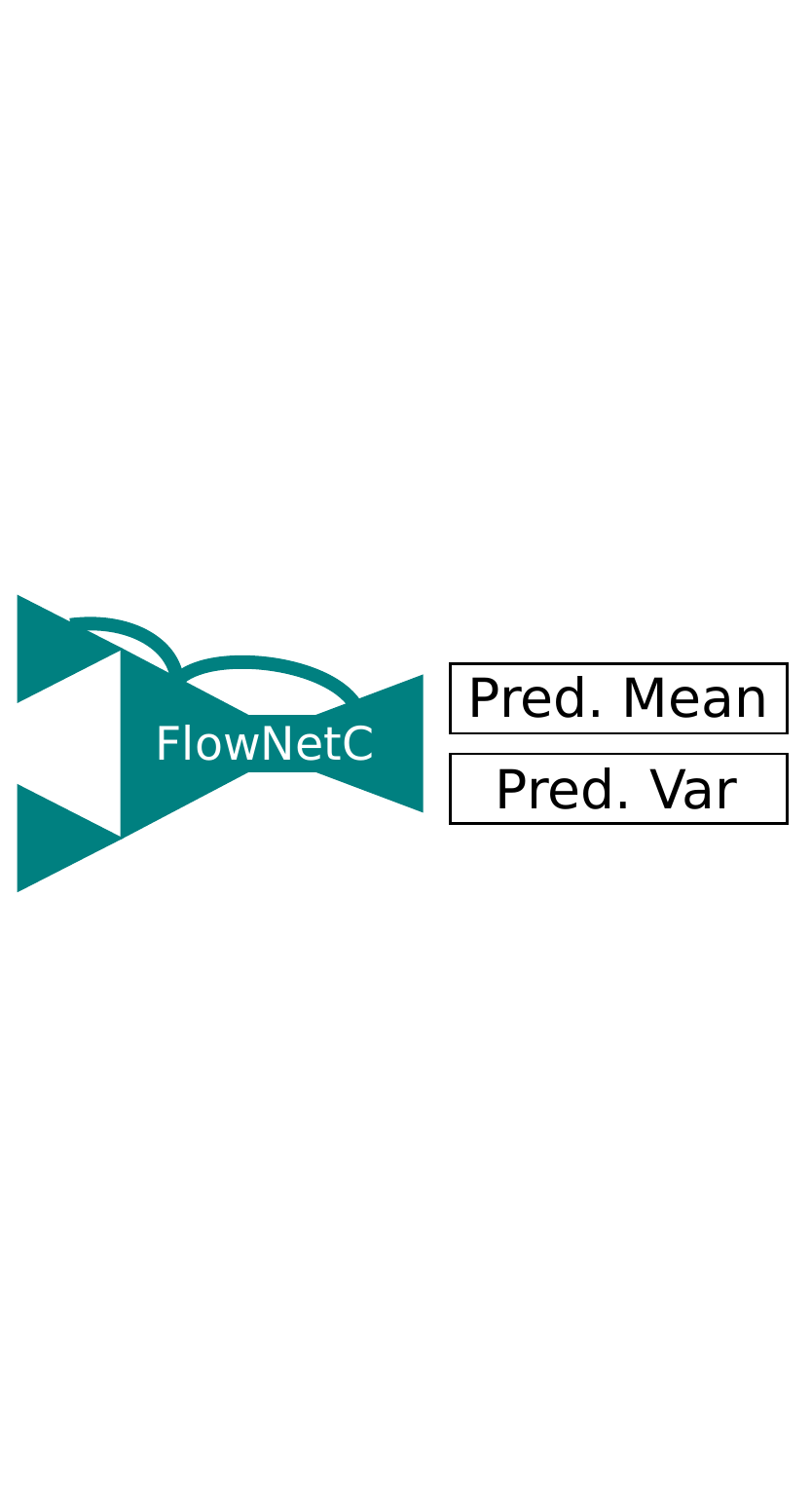}
        \subcaption{
        \label{fig:schematic_fnc_pred}}
        \end{center} 
    \end{subfigure}
    \rulesep
    \begin{subfigure}[t]{0.14\textwidth}
        \begin{center}
        \includegraphics[height=3.2cm]{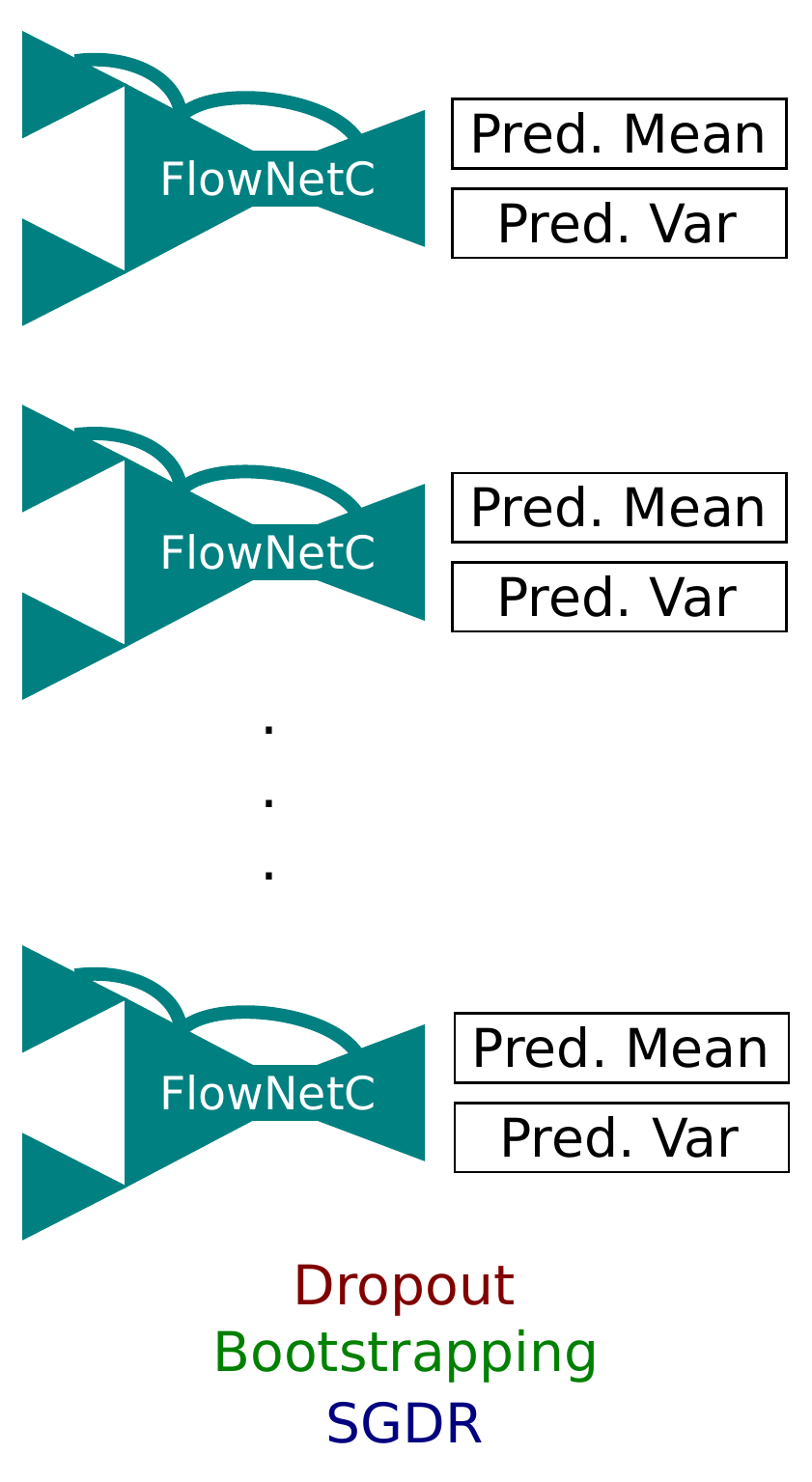}
        \subcaption{
        \label{fig:schematic_pred}}
        \end{center} 
    \end{subfigure}
    \rulesep
    \begin{subfigure}[t]{0.30\textwidth}
        \begin{center}%
        \includegraphics[height=3.2cm]{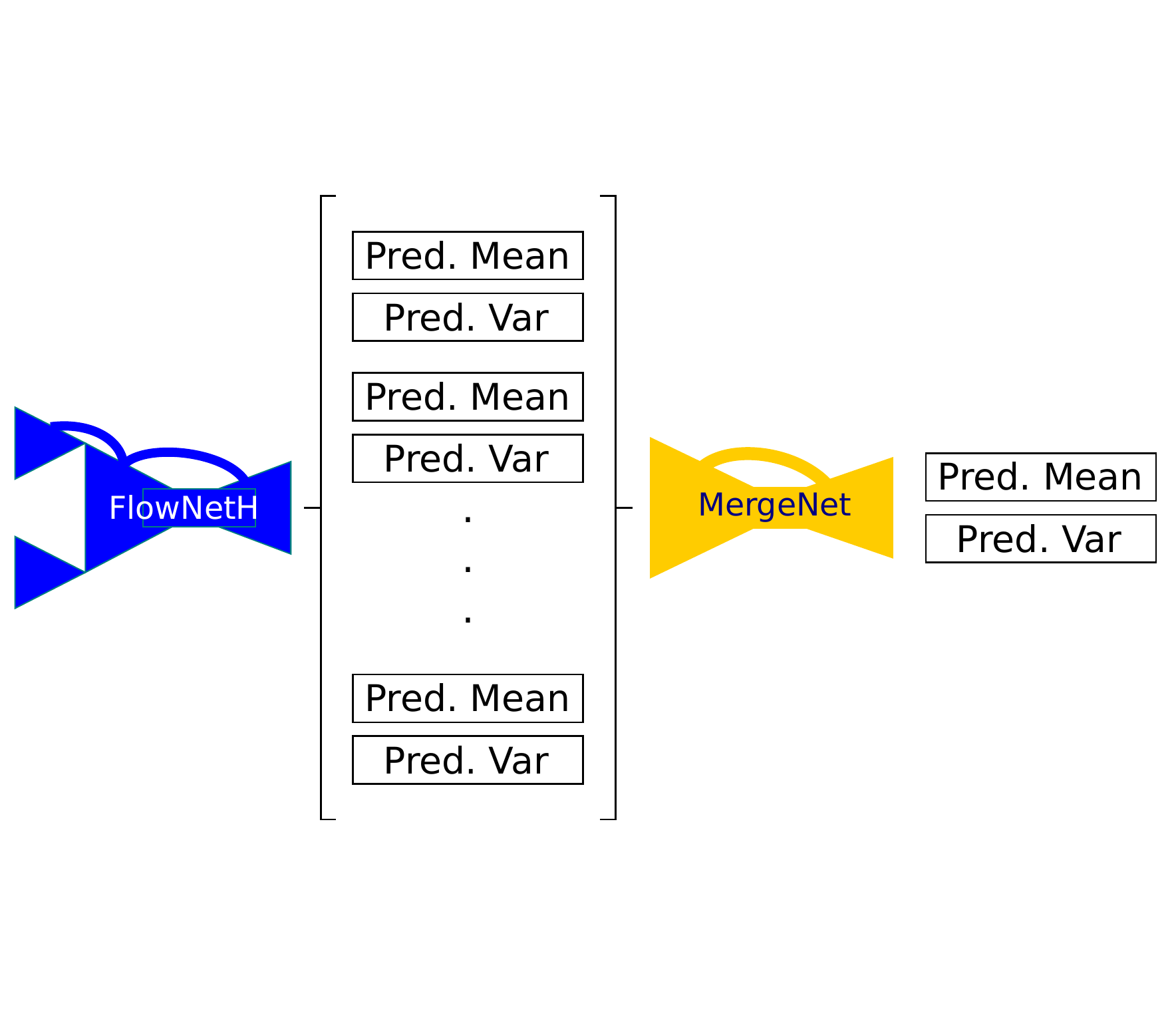}%
        \subcaption{
        \label{fig:schematic_fnh}}%
        \end{center}%
    \end{subfigure}
    }
    \end{center}
   \caption{
        Overview of the networks considered in this paper. 
        \textbf{(a)} FlowNetC trained with EPE. 
        \textbf{(b)} Same network as (a), where an ensemble is built using dropout, bootstrapping or SGDR. 
        \textbf{(c)} FlowNetC trained with -log-likelihood to predict mean and variance. 
        \textbf{(d)} Same network as (c), where an ensemble is built using dropout, bootstrapping or SGDR. 
        \textbf{(e)} FlowNetH trained to predict multiple hypotheses with variances, which are merged to a single distributional output. Only \textbf{(a)} exists in this form for optical flow.
  }
  \label{fig:schematic}
\end{figure*}%

\subsection{Empirical Uncertainty Estimation}
\label{sec:empirical_uncertainty}

A straightforward approach to get variance estimates is to train $M$ different models independently, such that the mean and the variance of the distribution $p(\vy|\vx,\mathcal D)$ can be approximated with the empirical mean and variance of the individual model's predictions. Let $ f_{\vw_i}(\vx)$ denote model $i$ of an ensemble of $M$ models with outputs $\vu_{\vw_{i}}$ and $\vv_{\vw_{i}}$.
We can compute the empirical mean and variance for the $\vu$-component by:
\begin{eqnarray}
    \bm{\mu}_\vu & = & \frac{1}{M} \sum_{i=1}^{M} {\vu_{\vw_{i}}}(\vx)\\
    \bm{\sigma}^2_\vu & = & \frac{1}{M} \sum_{i=1}^{M} ({\vu_{\vw_{i}}}(\vx) - \bm{\mu}_\vu)^2
\end{eqnarray}
and accordingly for the $\mathbf{v}$-component of the optical flow.
Such an ensemble of $M$ networks, as depicted in Figure~\ref{fig:schematic_emp}, can be built in multiple ways. The most common way is via Monte Carlo Dropout~\cite{baysianDropout}. Using dropout also at test time, it is possible to randomly sample from network weights $M$ times to build an ensemble. Alternatively, ensembles of individual networks can be trained with random weight initialization, data shuffling, and bootstrapping as proposed by Lakshminarayanan \etall~\cite{deepEnsembles}. A more efficient way of building an ensemble is to use $M$ pre-converged snapshots of a single network trained with the SGDR~\cite{sgdr} learning scheme, as proposed by Huang \etall~\cite{snapshotEnsembles}. We investigate these three ways of building ensembles for flow estimation and refer to them as Dropout, Bootstrapped Ensembles and SGDR Ensembles, respectively.

\subsection{Predictive Uncertainty Estimation}
\label{sec:predictive_uncertainty}

Alternatively, we can train a network to output the parameters $\bm{\theta}$ of a parametric model of the distribution $p(\vy|\vx,\mathcal D)$ as introduced by Nix and Weigend~\cite{predcnn}.
In the literature, Gaussian distributions (where $\bm{\theta}$ parameterizes the distribution's mean and the variance) are most common, but any type of parametric distribution is possible. 
Such networks can be optimized by maximizing their log-likelihood:
\begin{equation}
 \log p(\mathcal{D} \mid \vw) = \frac{1}{N} \sum_{i=1}^N \log p(\vy_i \mid \bm{\theta}(\vx_i,\vw))
\end{equation}
w.r.t. $\vw$. The predictive distribution for an input $\vx$ is then defined as:
\begin{equation}
p(\vy \mid \vx,\vw) \equiv p(\vy \mid \bm{\theta}(\vx,\vw)).
\end{equation}
While negative log-likelihood of a Gaussian corresponds to $L_2$ loss, FlowNet is trained with an EPE loss, which has more robustness to outliers.
Thus, we model the predictive distribution by a Laplacian, which corresponds to an $L_1$ loss.
The univariate Laplace distribution has two parameters $a$ and $b$ and is defined as:  
\begin{equation} 
\mathcal{L}(u|a,b) = \frac{1}{2b}e^{-\frac{|u-a|}{b}} \mathrm{.}
\end{equation}
As Wannewetsch et al.~\cite{probFlow}, we model the $u$ and $v$ components of the optical flow to be independent.
The approximation yields:
\begin{equation}
\label{eq:laplace}
    \mathcal{L}(u,v|a_u,a_v,b_u,b_v) \approx \mathcal{L}(u|a_u,b_u) \cdot \mathcal{L}(v|a_v,b_v) \mathrm{.}
\end{equation}

We obtain a probabilistic version of FlowNet 
with outputs $a_u$, $a_v$, $b_u$, $b_v$ 
by minimizing the negative log-likelihood of Eq.~\ref{eq:laplace}:
\begin{equation}
\label{eq:laplace_nll}
- \log(\mathcal{L}(u|a_u,b_u) \cdot \mathcal{L}(v|a_v,b_v)) =
 \frac{|u-{a}_u|}{{b}_u} + \log{{b}_u} + \frac{|v-{a}_v|}{{b}_v} + \log{{b}_v}.
\end{equation}

As an uncertainty estimate we use the variance of the predictive distribution, which is $\sigma^2 = 2b^2$ in this case.
This case corresponds to a single FlowNetC predicting flow and uncertainty as illustrated in Figure~\ref{fig:schematic_fnc_pred}.

\subsection{Bayesian Uncertainty Estimation} \label{sec:bayesian_uncertainty}


From a Bayesian perspective, to obtain an estimate of model uncertainty, rather than choosing a point estimate for $\vw$, we would marginalize over all possible values:
 \begin{eqnarray}
 p(\vy \mid \vx,\mathcal{D}) & = & \int p(\vy \mid \vx,\vw) p(\vw \mid \mathcal{D}) d\vw\\
    & = & \int p(\vy \mid \bm{\theta}(\vx,\vw)) p(\vw \mid \mathcal{D}) d\vw.
 \end{eqnarray}    
This integral cannot be computed in closed form, but by sampling $M$ networks $\vw_{i} \sim p(\vw | \mathcal{D})$ from the posterior distribution 
and using a Monte-Carlo approximation~\cite{neal-1995}, we can approximate its mean and variance as:
\begin{equation}
\label{eq:pred_dist_mc}p(\vy \mid \vx,\mathcal{D}) \approx \sum_{i=1}^M p(\vy \mid  \bm{\theta}(\vx,\vw_i)).
\end{equation}
Since every parametric distribution has a mean and a variance, also the distributions predicted by each individual network with weights $\vw_i$ yield a mean $\bm{\mu}_i$ and a variance $\bm{\sigma}_i^2$. The mean and variance of the mixture distribution in Eq. \ref{eq:pred_dist_mc} can then be computed by the law of total variance for the $\vu$-component (as well as for the $\vv$-component) as:

\begin{eqnarray}
    \bm{\mu}_\vu & = & \frac{1}{M} \sum_{i=1}^{M} \bm{\mu}_{\vu,i}\\
    \bm{\sigma}^2_\vu & = & \frac{1}{M} \sum_{i=1}^{M}\Big((\bm{\mu}_{\vu,i} - \bm{\mu}_\vu)^2 + {\bm{\sigma}^2}_{\vu,i}\Big)\quad.
\end{eqnarray}
This again can be implemented as ensembles obtained by predictive variants of dropout~\cite{baysianDropout}, bootstrapping~\cite{deepEnsembles} or SGDR~\cite{snapshotEnsembles}, where the ideas from Section~\ref{sec:empirical_uncertainty} and Section~\ref{sec:predictive_uncertainty} are combined as shown in Figure~\ref{fig:schematic_pred}.

\section{Predicting Multiple Hypotheses within a Single Network}

The methods presented in the Sections~\ref{sec:empirical_uncertainty} and \ref{sec:bayesian_uncertainty} require multiple forward passes to obtain multiple samples with the drawback of a much increased computational cost at runtime. 
In this section, we explain how to apply the Winner-Takes-All (WTA) loss to make multiple predictions within a single network \cite{MultipleChoice,MultipleChoiceEnsembles,PhotographicImageSynthesis,rupprecht} and then subsequenly use a second network to obtain predictions for final flow and uncertainty.
We call these predictions \emph{hypotheses}. 
The WTA loss makes the hypotheses more diverse 
and leads to capturing more different solutions, but does not allow for merging by simply computing the mean as for the ensembles presented in the last section. We propose to use a second network that merges the hypotheses to a single prediction and variance, as depicted in Figure~\ref{fig:schematic_fnh}.

Since a ground-truth is available only for the single true solution, the question arises of how to train a network to predict multiple hypotheses and how to ensure that each hypothesis comprises meaningful information. To this end, we use a loss that punishes only the best among the network output hypotheses $\bm{y}_1,\dots,\bm{y}_M$~\cite{MultipleChoice}. 
Let the loss between a predicted flow vector $\mathbf{y}(i,j)$ and its ground-truth $\mathbf{y}^\mathrm{gt}(i,j)$ at pixel $i,j$ be defined 
by a loss functon $l$.
We minimize: 

\begin{eqnarray}
    L_{hyp} & = & \sum_{i,j} l(
    \bm{y}_{\mathrm{best\_idx}(i,j)}, 
    \bm{y}^{\mathrm{gt}}(i,j)) 
    + \Delta(i,j) \mathrm{\,,}
\end{eqnarray}
where 
$\mathrm{best\_idx}(i,j)$ selects the best hypothesis per pixel according to the ground-truth:
\begin{equation}
    \mathrm{best\_idx}(i,j) = \underset{k}{\mathrm{arg min}} \, \left[{\rm EPE}(\bm{y}_k(i,j), \bm{y^\mathrm{gt}}(i,j))\right] 
    \mathrm{\,.}
\label{eq:best_idx}
\end{equation}
$\Delta =  \Delta_u + \Delta_v$ encourages similar solutions to be from the same hypothesis $k$ via one-sided differences, e.g. for the $\mathbf{u}$ component: 
\begin{equation}
    \begin{split} 
    \Delta_u(i,j) = & 
    \sum_{k;i>1;j}\left| y_{k,u}(i,j) - y_{k,u}(i-1,j) \right| + \\ & 
    \sum_{k;i;j>1}\left| y_{k,u}(i,j) - y_{k,u}(i,j-1) \right|
    \end{split}
    \label{eq:delta}
\end{equation}

For $l$, we either use the endpoint error from Eq.~\ref{eq:epe}
or the negative log-likelihood from Eq.~\ref{eq:laplace_nll}. In the latter case, each hypothesis is combined with an uncertainty estimation and $l$ also operates on a variance $\bm{\sigma}$. Equations $\ref{eq:best_idx}$ and $\ref{eq:delta}$ remain unaffected. For the best index selection we stick to the EPE since it is the main optimization goal.

To minimize $L_{hyp}$, the network must make a prediction close to the ground-truth in at least one of the hypotheses. 
In locations where multiple solutions exist and the network cannot decide for one of them, 
the network will predict several different likely solutions to increase the chance that the true solution is among these predictions.
Consequently, the network will favor making  diverse hypotheses in cases of uncertainty.
In Tables 3 and 4 of the supplemental material we provide visualizations of such hypotheses.

In principle, $L_{hyp}$ could collapse to use only one of the hypotheses' outputs. In this case the other hypotheses would have very high error and would never be selected for back-propagation. However, due to the variability in the data and the stochasticity in training, such collapse is very unlikely. We never observed that one of the hypotheses was not used by the network, and for the oracle merging we observed that all hypotheses contribute more or less equally. We show this diversity in our experiments. 

\section{Experiments} 
\label{sec:experiments}

To evaluate the different strategies for uncertainty estimation while keeping the computational cost tractable, we chose as a base model the FlowNetC architecture from Dosovitsky \etall~\cite{flownet} with  improved training settings by Ilg \etall~\cite{flownet2} and by us. A single FlowNetC shows a larger endpoint error (EPE) than the full, stacked FlowNet 2.0~\cite{flownet2}, but trains much faster. Note that this work aims for uncertainty estimation and not for improving the optical flow over the base model. The use of ensembles may lead to minor improvements of the optical flow estimates due to the averaging effect, but these improvements are not of major concern here. In the end, we will also show results for a large stacked network to demonstrate that the uncertainty estimation as such is not limited to small, simple networks. 

\subsection{Training Details\label{sec:training_settings}}

\begin{wraptable}{r}{42mm}
    \vspace{-1.2cm}
    \begin{center}
    \begin{tabular}{|l|c|c|}
        \hline
        & Iter. & EPE \\
        \hline
        \hline
        FlowNetC~\cite{flownet2} & 600k & 3.77 \\ 
        FlowNetC~\cite{flownet2} & 1.2m & 3.58 \\ 
        \hline 
        FlowNetC ours & 600k & 3.40 \\ 
        \hline
    \end{tabular}
    \caption{
    Optical flow quality on Sintel train clean with the original FlowNetC~\cite{flownet2} and our implementation. 
        \label{tab:FlowNetC}
    }
    \end{center} 
    \vspace{-0.8cm}
\end{wraptable}

In contrast to Ilg \etall~\cite{flownet2}, we use Batch Normalization~\cite{BN} and a continuously dropping cosine learning rate schedule~\cite{sgdr}. This yields shorter training times and improves the results a little; see Table~\ref{tab:FlowNetC}.
We train on FlyingChairs~\cite{flownet} and start with a learning rate of $2e-4$. For all networks, we fix a training budget of $600k$ iterations per network, with an exception for SGDR, where we also evaluate performing some pre-cycles. For SGDR Ensembles, we perform restarts every $75$k iterations. We fix the $T_{mult}$ to $1$, so that each annealing takes the same number of iterations. We experiment with different variants of building ensembles using snapshots at the end of each annealing. We always take the latest $M$ snapshots when building an ensemble. For dropout experiments, we use a dropout ratio of 0.2 as suggested by Kendall \etall~\cite{visionUncertainties}. For Bootstrapped Ensembles, we train $M$ FlowNetC in parallel with bootstrapping, such that each network sees different $67\%$ of the training data. For the final version of our method, we perform an additional training of $250k$ iterations on FlyingThings3D~\cite{dispnet} per network, starting with a learning rate of $2e-5$ that is decaying with cosine annealing. We use the Caffe~\cite{caffe} framework for network training and evaluate all runtimes on an Nvidia GTX 1080Ti.
We will make the source code and the final models publicly available.

For the ensembles, we must choose the size $M$ of the ensemble. The sampling error for the mean and the variance decreases with increasing $M$. However, since networks for optical flow estimation are quite large, we are limited in the tractable sample size and restrict it to $M=8$. We also use $M=8$ for FlowNetH. 

For SGDR there is an additional pre-cycle parameter: snapshots in the beginning have usually not yet converged and the number of pre-cycles is the number of snapshots we discard before building the ensemble. In the supplemental material we show that the later the snapshots are taken, the better the results are in terms of EPE and AUSE. We use $8$ pre-cycles in the following experiments.

\subsection{Evaluation Metrics\label{sec:metrics}}

\begin{figure}[t]
  \begin{center}
      \includegraphics[width=0.75\linewidth]{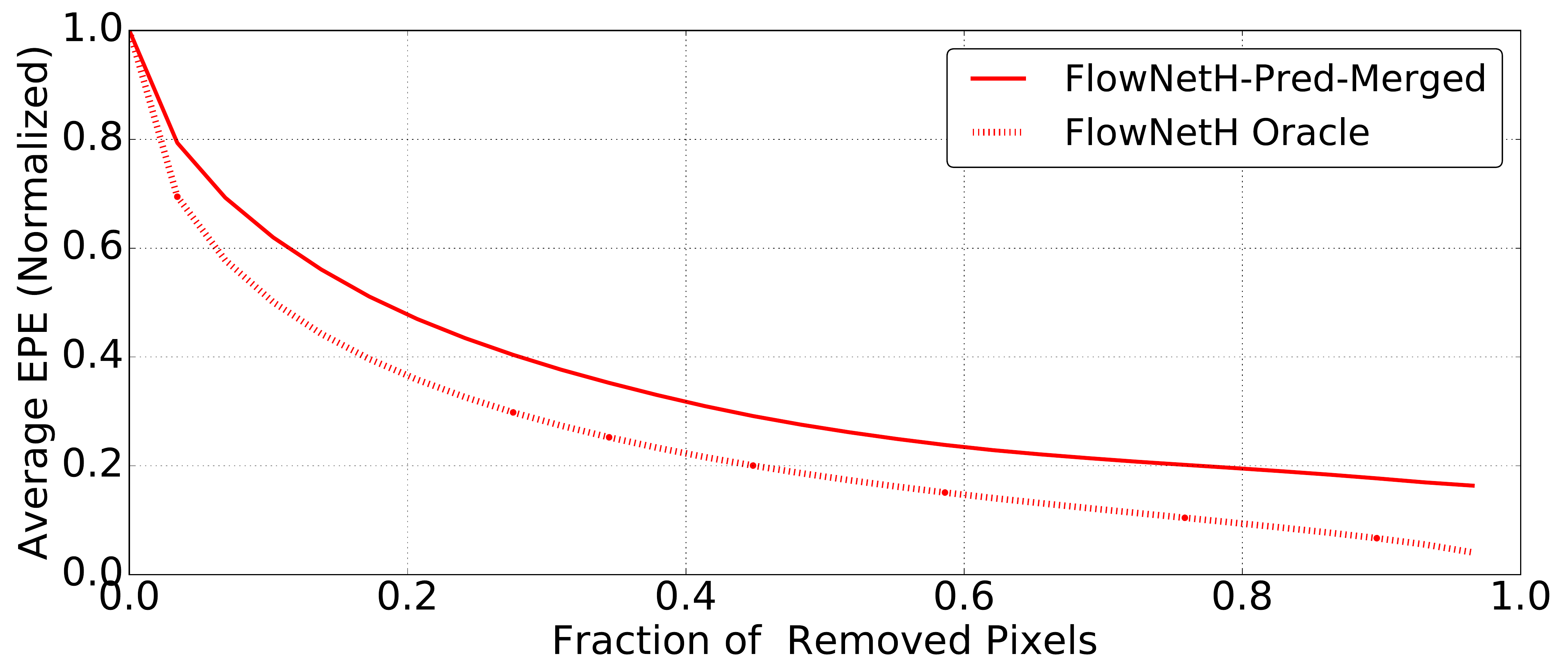}
  \end{center}
\caption{Sparsification plot of FlowNetH-Pred-Merged for the Sintel train clean dataset. The plot shows the average endpoint error (AEPE) for each fraction of pixels having the highest uncertainties removed. The oracle sparsification shows the lower bound by removing each fraction of pixels ranked by the ground-truth endpoint error. Removing 20 percent of the pixels results in halving the average endpoint error.
  \label{fig:sparsification_hyp_sintel}
}
\end{figure}

\textbf{Sparsification Plots.}
To assess the quality of the uncertainty measures, we use so-called sparsification plots, which are commonly used for this purpose~\cite{opticalFlowPAMI2012,probFlow,Kondermann2008,bootstrapOpticalFlow}. Such plots reveal on how much the estimated uncertainty coincides with the true errors. If the estimated variance is a good representation of the model uncertainty, and the pixels with the highest variance are removed gradually, the error should monotonically decrease. Such a plot of our method is shown in Figure~\ref{fig:sparsification_hyp_sintel}. 
The best possible ranking of uncertainties is ranking by the true error to the ground-truth. We refer to this curve as \emph{Oracle Sparsification}. Figure~\ref{fig:sparsification_hyp_sintel} reveals that our uncertainty estimate is very close to this oracle.  

\begin{figure*}[t]
  \begin{center}
  \resizebox{\linewidth}{!}{%
  \begin{subfigure}[b]{0.426\linewidth}
      \centering
      \includegraphics[width=\linewidth]{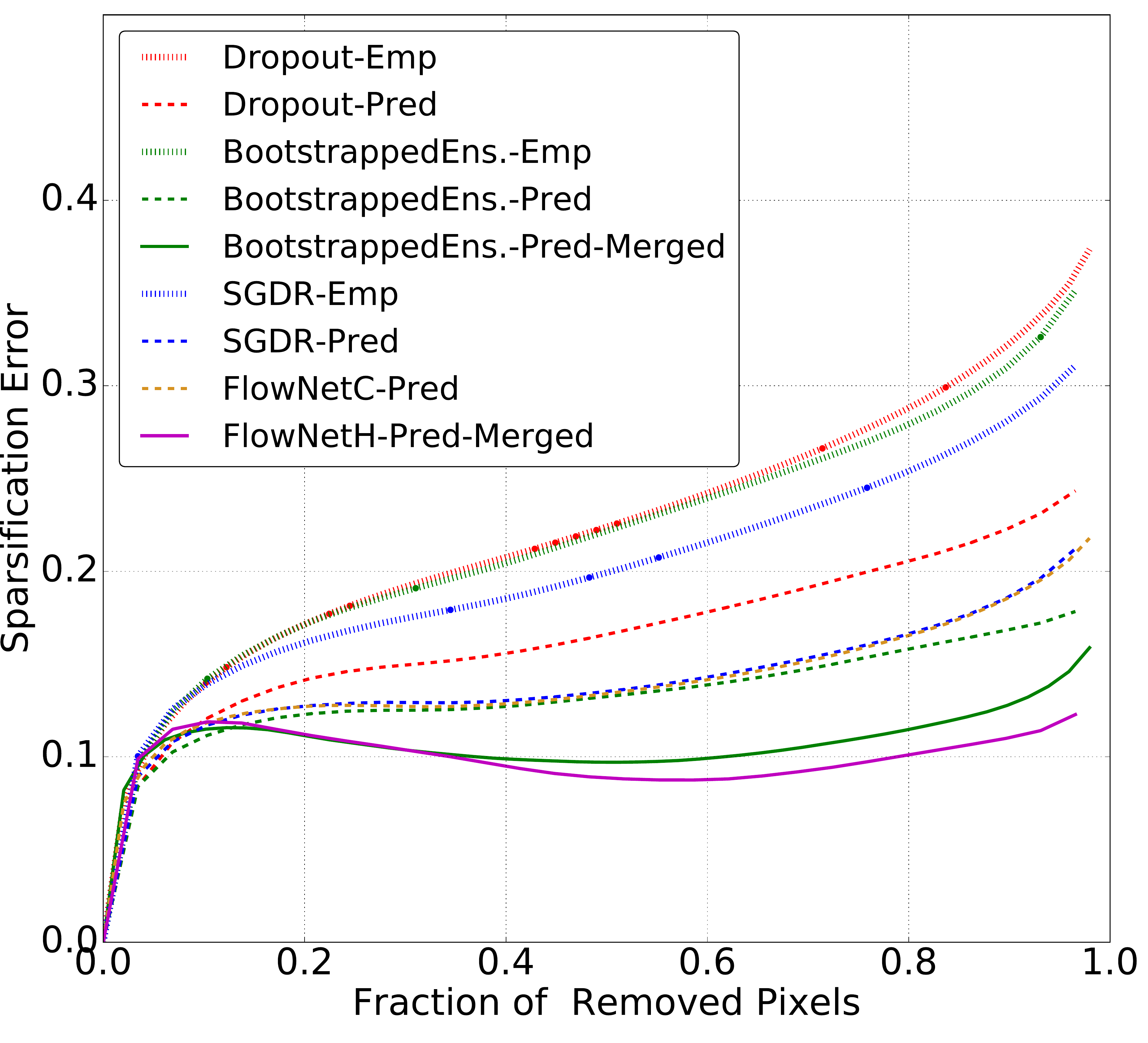}
      \caption{\label{fig:sparsification_all}}
    \end{subfigure}
    \begin{subfigure}[b]{0.574\linewidth}
      \centering
      \includegraphics[width=\linewidth]{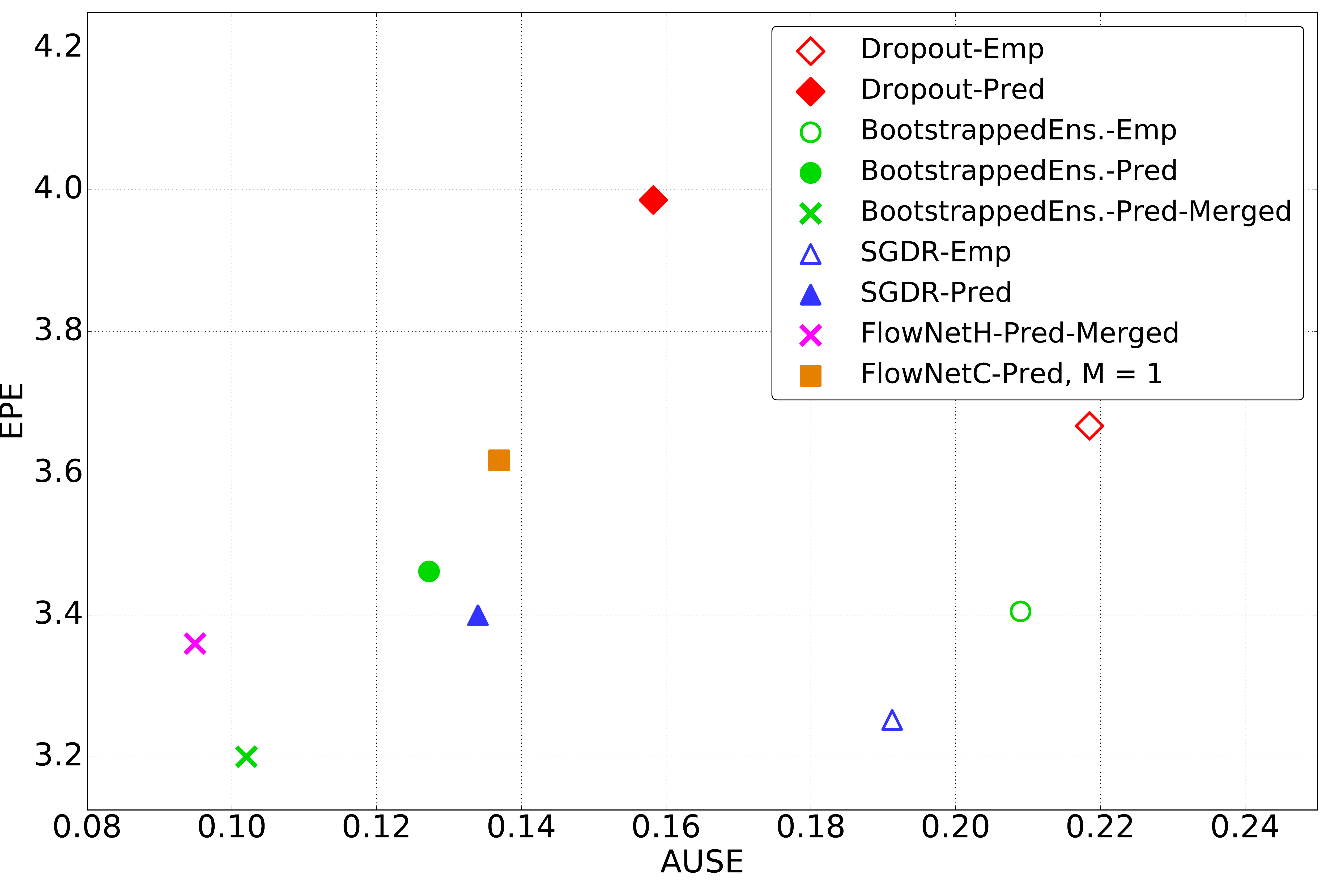}
      \caption{\label{fig:ensemble_size}}
    \end{subfigure}%
    }
  \end{center}
\caption{\textbf{(a)} Sparsification error on the Sintel train clean dataset. The sparsification error (smaller is better) is the proposed measure for comparing the uncertainty estimates among  different methods. FlowNetH-Pred-Merged and BootstrappedEnsemble-Pred-Merged perform best in almost all sections of the plot.
\textbf{(b)} Scatter plot of AEPE vs. AUSE for the tested approaches visualizing some content of Table~\ref{tab:comparison_sintel}.
\label{fig:sizes_shifts_scatter}
}
\end{figure*}

\textbf{Sparsification Error.}
For each approach the oracle is different, hence a comparison among approaches using a single sparsification plot is not possible. To this end, we introduce a measure, which we call \textit{Sparsification Error}. It is defined as the difference between the sparsification and its oracle. Since this measure normalizes the oracle out, a fair comparison of different methods is possible. In Figure~\ref{fig:sparsification_all}, we show sparsification errors for all methods we present in this paper. To quantify the sparsification error with a single number, we use the Area Under the Sparsification Error curve (\emph{AUSE}). 

\begin{table*}[t!]
    \begin{center}
\resizebox{\textwidth}{!}{%
\begin{tabular}{|l||c|c|c|c||c|c|c|c||c|}
    \hline
        & 
        \multicolumn{4}{c||}{empirical (Emp)} &
        \multicolumn{4}{c||}{predictive (Pred)} &
        \\
        \cline{2-10}
        &
        AUSE & EPE & Oracle EPE & Var. &
        AUSE & EPE & Oracle EPE & Var. &
        Runtime 
        \\
    \hline
    \hline
        FlowNetC &
        - & $3.40$ & - & - &
        $0.133$ & $3.62$ & - & - &
        \pz$\mathbf{38}$ms
        \\
    \hline
    \hline
        Dropout &
        $0.212$ & $3.67$ & $2.56$ & 5.05 &
        $0.158$ & $3.99$ & $2.96$ & 3.80 &
        \pz320ms
        \\
    \hline 
        SGDREnsemble & 
        $\mathbf{0.191}$ & $3.25$ & $2.56$ & 3.50 & 
        $0.134$ & $3.40$ & $2.87$ & 1.52 & 
        \pz304ms 
        \\
    \hline 
        BootstrappedEnsemble & 
        $0.209$ & $3.41$ & $2.17$ & 9.52 &
        $0.127$ & $3.46$ & $2.49$ & 6.15 &
        \pz304ms 
        \\
    \hline 
    \hline 
        BootstrappedEnsemble-Merged & 
        \multicolumn{4}{c||}{ } &
        $0.102$ & $3.20$ & $2.49$ & 6.15 & 
        \pz332ms 
        \\        
    \hline
        FlowNetH-Merged & 
        - & $3.50$ & $1.73$ & $\mathbf{83.32}$ &
        $\mathbf{0.095}$ & $3.36$ & $1.89$ & $\mathbf{52.85}$ & 
        \pz60ms
        \\        
    \hline
\end{tabular}
}

    \end{center} 
    \caption{
        Comparison of flow and uncertainty predictions of all proposed methods with $M=8$ on the Sintel train clean dataset. Oracle-EPE is the EPE of the pixel-wise best selection from the samples or hypotheses determined by the ground-truth. Var. is the average empirical variance over the 8 samples or hypotheses. Predictive versions (Pred) generally outperform empirical versions (Emp). Including a merging network increases the performance. 
        FlowNetH-Pred-Merged performs best for predicting uncertainties and has a comparatively low runtime. 
        \label{tab:comparison_sintel}
    }
\end{table*}

\textbf{Oracle EPE.}
For each ensemble, we also compute the hypothetical endpoint error by considering the pixel-wise best selection from each member (decided by the ground-truth). We report this error together with the empirical variances among the members in Table~\ref{tab:comparison_sintel}. 

\subsection{Comparison among Uncertainties from CNNs\label{sec:comparison_cnn}}

\textbf{Nomenclature.} When a single network is trained against the endpoint error, we refer to this single network and the resulting ensemble as empirical (abbreviated as \emph{Emp}; Figures \ref{fig:schematic_fnc_emp} and \ref{fig:schematic_emp}), while when the single network is trained against the negative log-likelihood, we refer to the single network and the ensemble as predictive (\emph{Pred}; Figures \ref{fig:schematic_fnc_pred} and \ref{fig:schematic_pred}). When multiple samples or solutions are merged with a network, we add \emph{Merged} to the name. E.g. FlowNetH-Pred-Merged refers to a FlowNetH that predicts multiple hypotheses and merges them with a network, using the loss for a predictive distribution for both, hypotheses and merging, respectively (Figure \ref{fig:schematic_fnh}).
Table~\ref{tab:comparison_sintel} and Figures~\ref{fig:sparsification_all}, \ref{fig:ensemble_size} show results for all models evaluated in this paper. 

\textbf{Empirical Uncertainty Estimation.}
The results show that uncertainty estimation with empirical ensembles is good, but worse than the other methods presented in this paper. However, in comparison to predictive counterparts, empirical ensembles tend to yield slightly better EPEs, as will be discussed in the following.   

\textbf{Predictive Uncertainty Estimation.}
The estimated uncertainty is better with predictive models than with the empirical ones. 
Even a single FlowNetC with predictive uncertainty yields much better uncertainty estimates than any empirical ensemble in terms of AUSE.
This is because when training against a predictive loss function, the network has the possibility to explain outliers with the uncertainty. This is known as \textit{loss attenuation}~\cite{visionUncertainties}. While the EPE loss tries to enforce correct solutions also for outliers, the log-likelihood loss attenuates them. 
The experiments confirm this effect and show that it is advantageous to let a network estimate its own uncertainty. 

\textbf{Predictive Ensembles.}
Comparing ensembles of predictive networks to a single predictive network shows that a single network is already very close to the predictive ensembles and that the benefit of an ensemble is limited. We attribute this also to loss attenuation: different ensemble members appear to attenuate outliers in a similar manner and induce less diversity, as can be seen by the variance among the members of the ensemble (column 'Var.' in Table~\ref{tab:comparison_sintel}).  

When comparing empirical to predictive ensembles, we can draw the following conclusions: 
\textbf{a.)} 
empirical estimation provides more diversity within the ensemble (variance column in Table ~\ref{tab:comparison_sintel}),  
\textbf{b.)} 
empirical estimation provides lower EPEs and Oracle EPEs, 
\textbf{c.)} 
all empirical setups provide worse uncertainty estimates than predictive setups.

\textbf{Ensemble Types.}
We see that the commonly used dropout~\cite{baysianDropout} technique performs worst in terms of EPE and AUSE, although the differences between the predictive ensemble types are not very large. SGDR Ensembles provide better uncertainties, yet the variance among the samples is the smallest. This is likely because later ensemble members are derived from previous snapshots of the same model. Furthermore, because of the $8$ pre-cycles, SGDR experiments ran the largest number of training iterations, which could be an explanation why they provide a slightly better EPE than other ensembles. 
Bootstrapped Ensembles provide the highest sample variance and the lowest AUSE among the predictive ensembles. 





\textbf{FlowNetH and Uncertainty Estimation with Merging Networks.}
%
Besides FlowNetH we also investigated putting a merging network on top of the predictive Bootstrapped Ensembles. 
Results show that the multi-hypotheses network (FlowNetH-Pred-Merged) is on-par with BootstrappedEnsemble-Pred-Merged in terms of AUSE and EPE. However, including the runtime, FlowNetH-Pred-Merged yields the best trade-off; see Table~\ref{tab:comparison_sintel}. 
Only FlowNetC and FlowNetH-Pred-Merged allow a deployment at interactive frame rates.
Table~\ref{tab:comparison_sintel} also shows that FlowNetH has a much higher sample variance and the lowest oracle EPE. This indicates that it internally has very diverse and potentially useful hypotheses that could be exploited better in the future. For some visual examples, we refer to Tables 3 and 4 in the supplemental material.

\subsection{Comparison to Energy-Based Uncertainty Estimation\label{sec:comparison_sota}}

\begin{figure}[t]
  \begin{center}
   \begin{subfigure}[b]{0.5\linewidth}
     \centering
      \includegraphics[width=\linewidth]{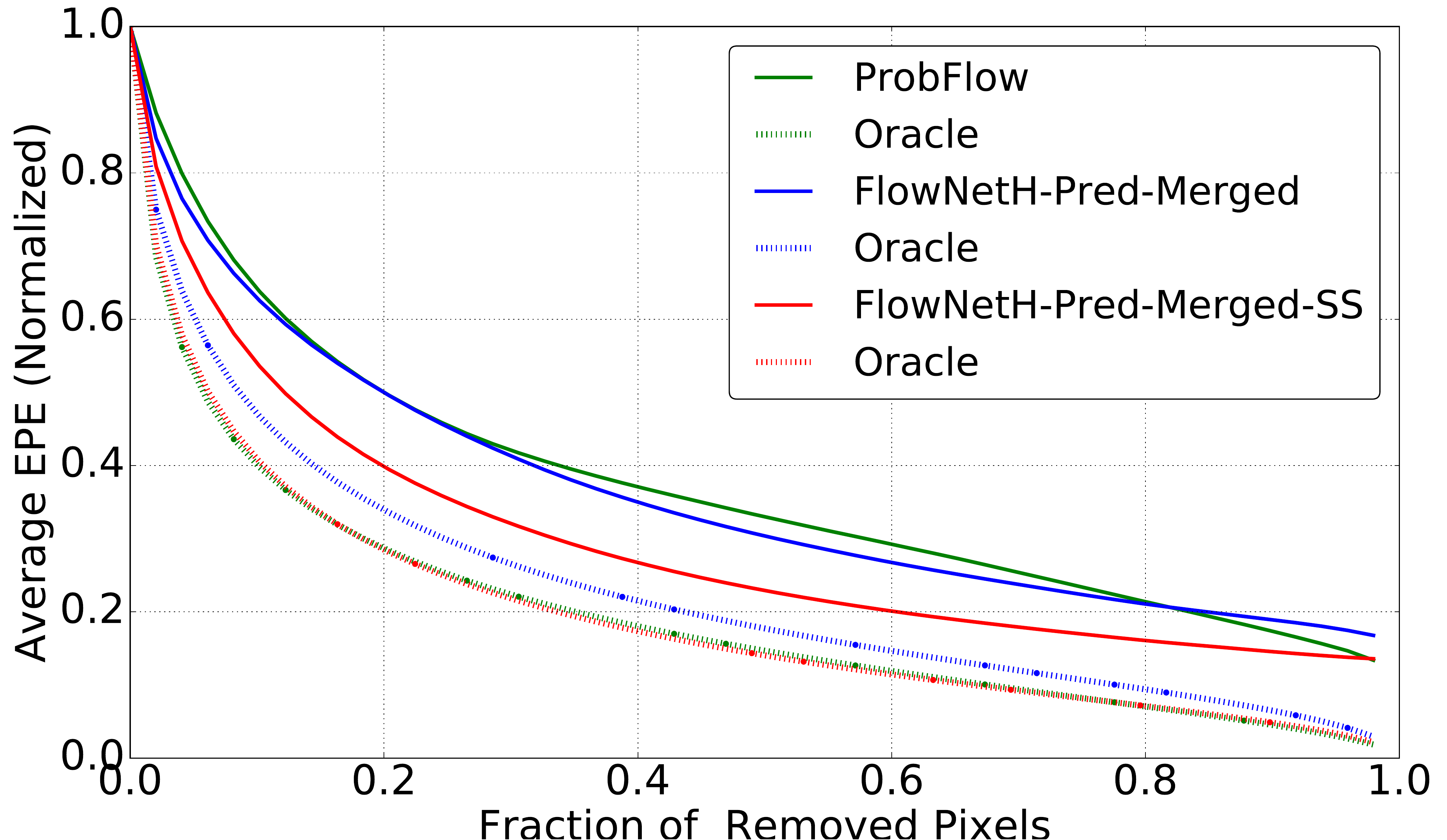}
      \caption{\label{fig:probflow_ours_left}}
   \end{subfigure}%
   \begin{subfigure}[b]{0.5\linewidth}
      \centering
      \includegraphics[width=\linewidth]{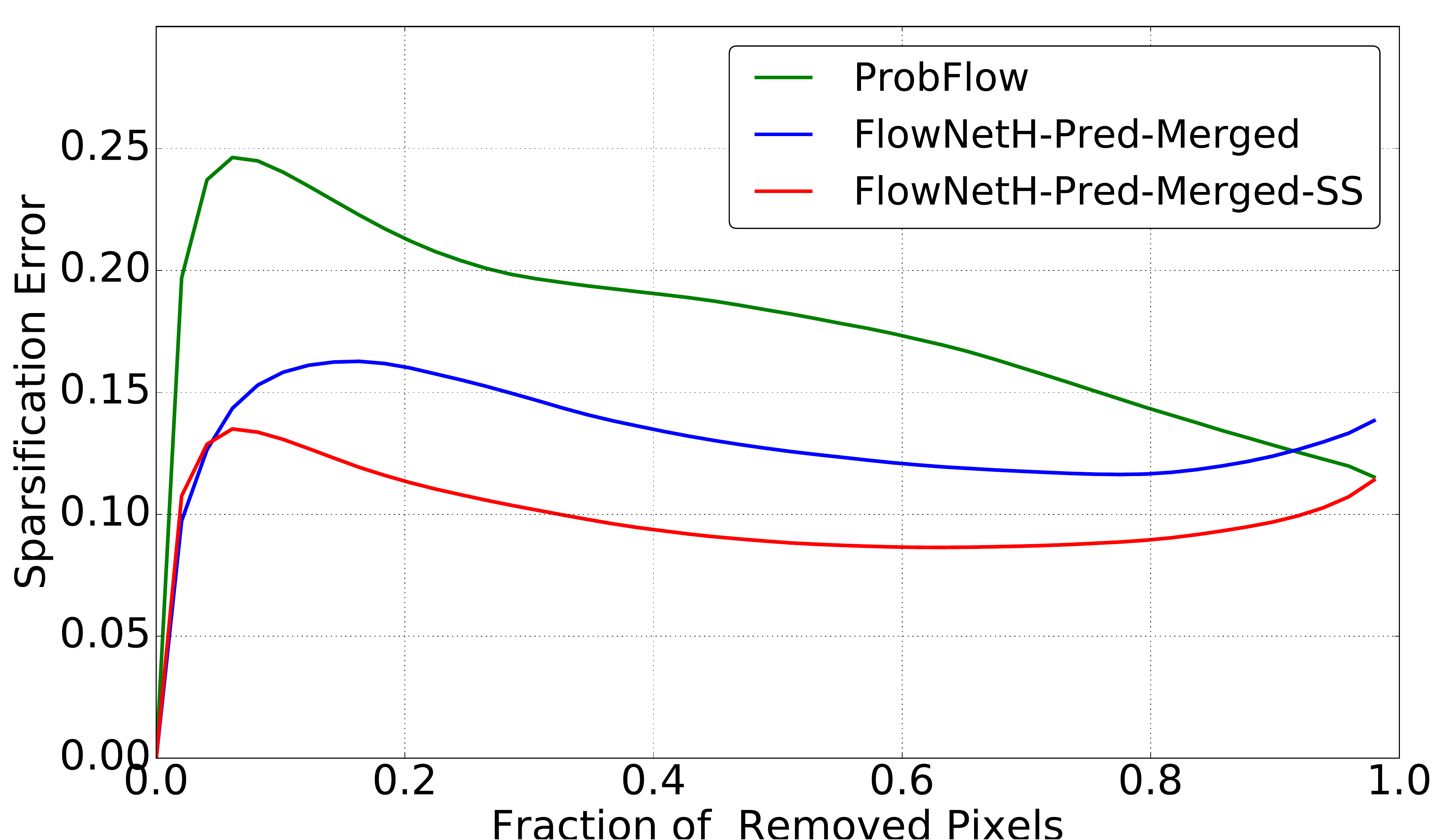}
      \caption{\label{fig:probflow_ours_right}}
      \end{subfigure}
  \end{center}
\caption{Plots of the sparsification curves with their respective oracles \textbf{(a)} and of the sparsification errors \textbf{(b)} for ProbFlow, FlowNetH-Pred-Merged and FlowNetH-Pred-Merged-SS (version with $2$ refinement networks stacked on top) on the Sintel train final dataset. KITTI versions are similar and provided in the supplemental material.\label{fig:probflow_ours}
}
\end{figure}
\begin{table}[t!]
    \begin{center}
    \begin{tabular}{|l||c|c||c|c||c|c||c|}
        \hline
            & \multicolumn{2}{c||}{Sintel Clean} & \multicolumn{2}{c||}{Sintel Final} & \multicolumn{2}{c||}{KITTI} &
            \multirow{2}{*}{runtime}
            \\
            \cline{2-3} 
            \cline{4-5}
            \cline{6-7}
            &
            AUSE & EPE & AUSE & EPE & AUSE & EPE & 
            \\
        \hline
        \hline
            ProbFlow~\cite{probFlow} &
            $0.162$ & $1.87$ & 
            $0.173$ & $3.34$ & 
            $0.466$ & $8.95$ & 
            \pz38.1s$^\dagger$
            \\
        \hline 
        \hline
            FlowNetH-Pred-Merged-FT-KITTI & 
            - & - &
            - & - &
            $\mathbf{0.086}$ & $3.12$ & 
            \pz$\mathbf{60}$ms
            \\
        \hline
        \hline
            FlowNetH-Pred-Merged & 
            $0.117$ & $2.58$ & 
            $0.128$ & $3.78$ & 
            $0.151$ & $7.84$ & 
            \pz$\mathbf{60}$ms
            \\
        \hline
            FlowNetH-Pred-Merged-S & 
            $0.091$ & $2.29$ & 
            $0.098$ & $3.51$ & 
            $0.102$ & $6.86$ & 
            \pz86ms
            \\
        \hline
            FlowNetH-Pred-Merged-SS & 
            $\mathbf{0.089}$ & $2.19$ & 
            $\mathbf{0.096}$ & $3.40$ & 
            $0.091$ & $6.50$ & 
            \pz99ms
            \\
        \hline
    \end{tabular}
    \end{center}
    \caption{
        Comparison of FlowNetH to the state-of-the-art uncertainty estimation method ProbFlow~\cite{probFlow} on the Sintel train clean, Sintel train final and our KITTI 2012+2015 validation split datasets. 
        The '-FT-KITTI' version is trained on FlyingChairs~\cite{flownet} first and then on FlyingThings3D~\cite{dispnet}, as described in Sec.~\ref{sec:training_settings} and subsequently fine-tuned on our KITTI 2012+2015 training split. FlowNetH-Pred-Merged, -S and -SS are all trained with the FlowNet2~\cite{flownet2} schedule described in supplemental material Fig. 6. Our method outperforms ProbFlow in AUSE by a large margin and also in terms of EPE for the KITTI dataset. 
        $^\dagger$runtime taken from~\cite{probFlow}, please see the supplemental material for details on the computation of the ProbFlow outputs.
        \label{tab:final_results}
    }
\end{table}

We compare the favored approach from the previous section (FlowNetH-Pred-Merged) to ProbFlow~\cite{probFlow}, which is an energy minimization approach and currently the state-of-the-art for estimating the uncertainty of optical flow. Figure~\ref{fig:probflow_ours} shows the sparsification plots for the Sintel train final. ProbFlow has almost the same oracle as FlowNetH-Pred-Merged, i.e. the flow field from ProbFlow can equally benefit from sparsification, but the actual sparsification error due to its estimated uncertainty is higher. This shows that FlowNetH-Pred-Merged has superior uncertainty estimates. 
In Table~\ref{tab:final_results} we show that this also holds for the KITTI dataset. FlowNetH outperforms  ProbFlow also in terms of EPE in this case. This shows that the superior uncertainty estimates are not due to a weaker optical flow model, i.e. from obvious mistakes that are easy to predict. 

Table~\ref{tab:final_results} further shows that the uncertainty estimation is not limited to simple encoder-decoder networks, but can also be applied successfully to state-of-the-art stacked networks~\cite{flownet2}. 
To this end, we follow Ilg et al.~\cite{flownet2} and stack refinement networks on top of FlowNetH-Pred-Merged. Different from~\cite{flownet2}, each refinement network yields the residual of the flow field and the uncertainty, as recently proposed by~\cite{crl}. We refer to the network with the 1st refinement network as FlowNetH-Pred-Merged-S and with the second refinement network as FlowNetH-Pred-Merged-SS, since each refinement network is a FlowNetS~\cite{flownet2}. 

The uncertainty estimation is not negatively influenced by the stacking, despite the improving flow fields. This shows again that the uncertainty estimation works reliably notwithstanding if the predicted optical flow is good or bad. 



\begin{figure*}[t]
  \begin{center}%
    \newcommand{\galleryRowSintel}[6]{ \includegraphics[width=0.2\textwidth]{images/#1} & \includegraphics[width=0.2\textwidth]{images/#2} & \includegraphics[width=0.2\textwidth]{images/#3} & \includegraphics[width=0.2\textwidth]{images/#4} & \includegraphics[width=0.2\textwidth]{images/#5} & \includegraphics[width=0.2\textwidth]{images/#6} \tabularnewline
}
  \resizebox{\linewidth}{!}{%
    \setlength{\tabcolsep}{0.7pt}%
    \begin{tabular}{ccc|ccc}%
\galleryRowSintel{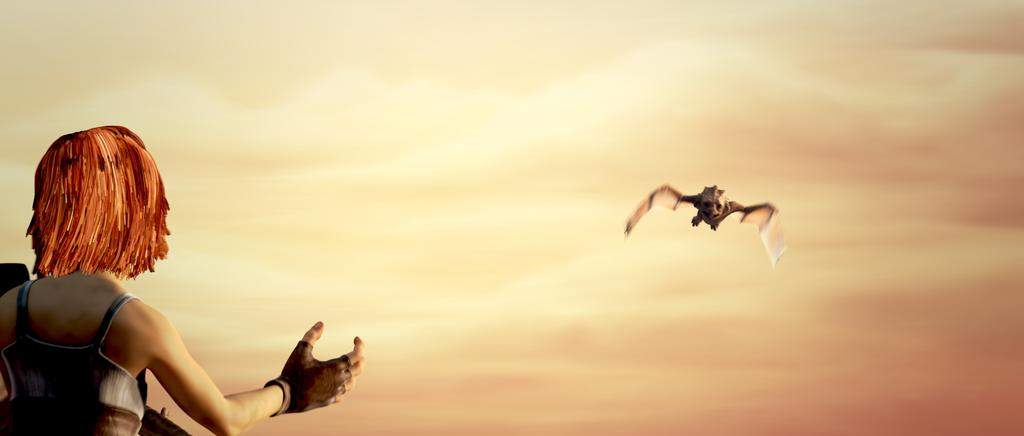}{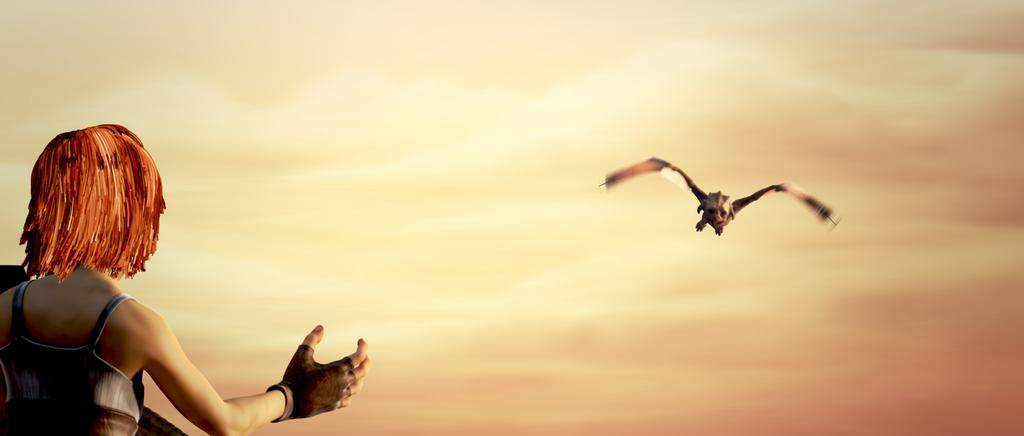}{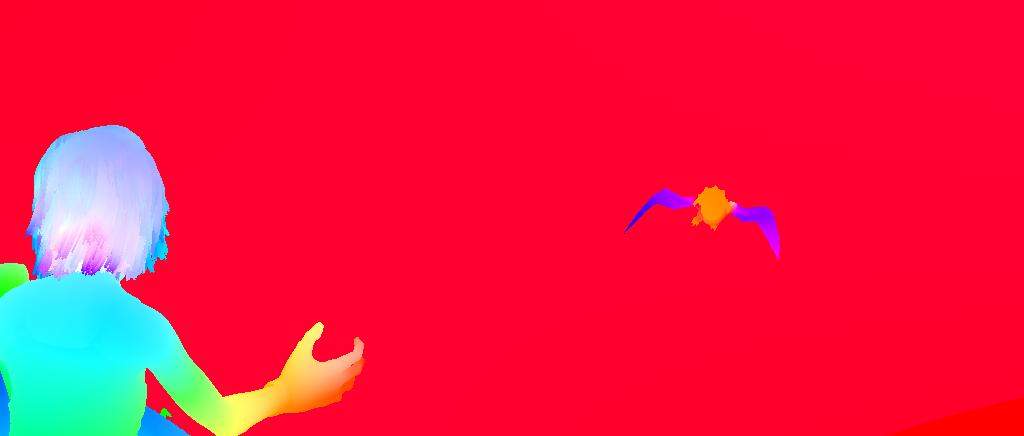}{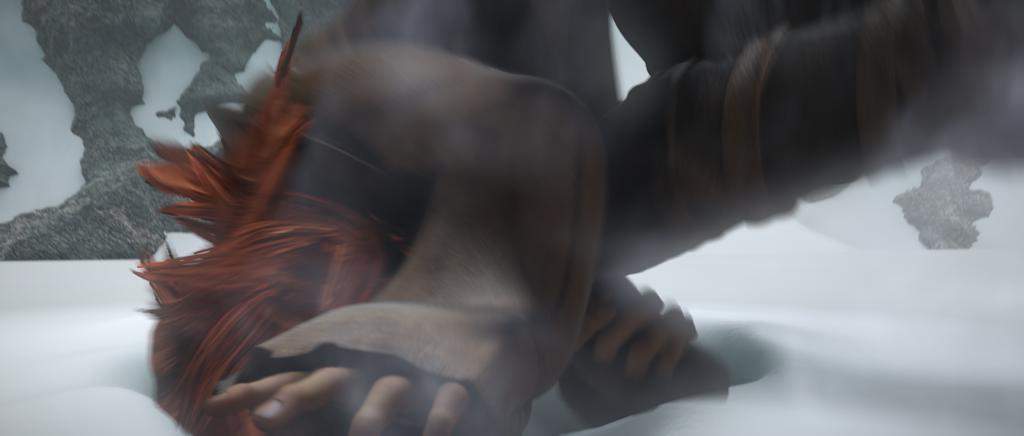}{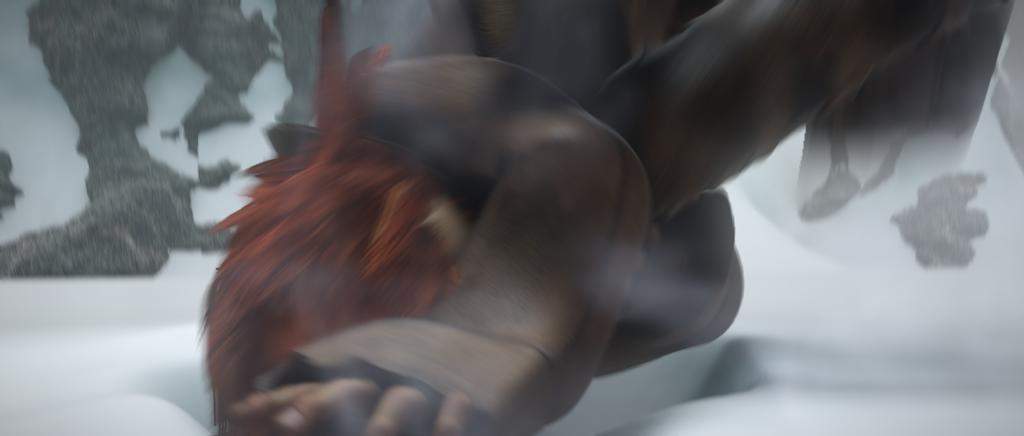}{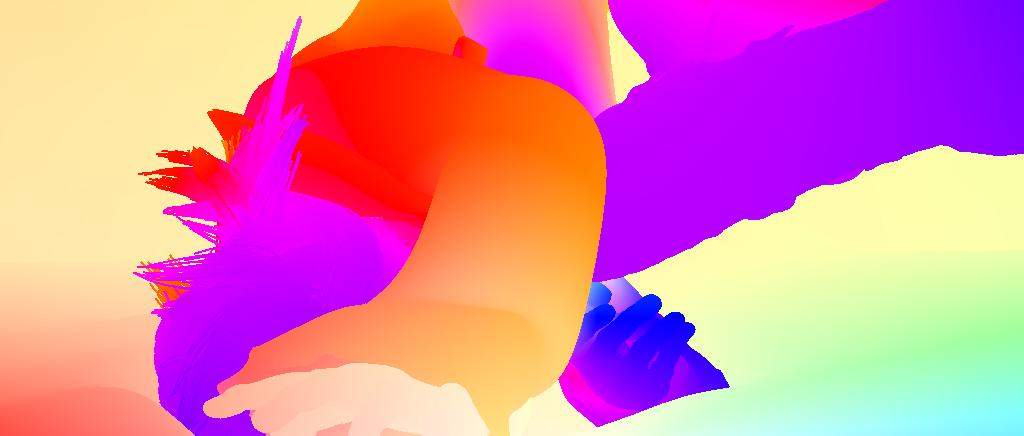}%
\galleryRowSintel{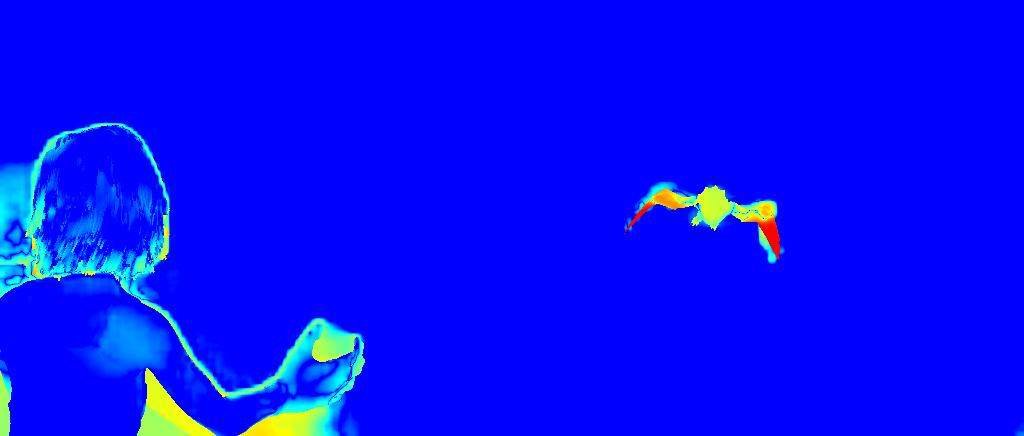}{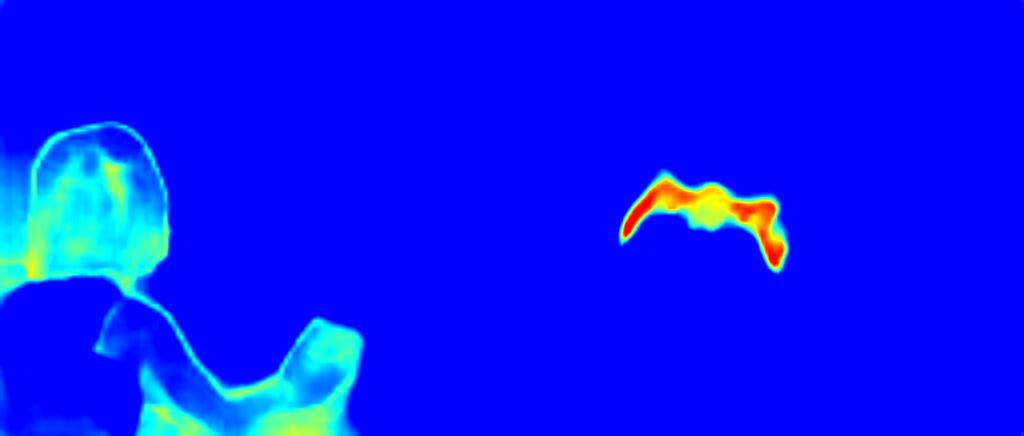}{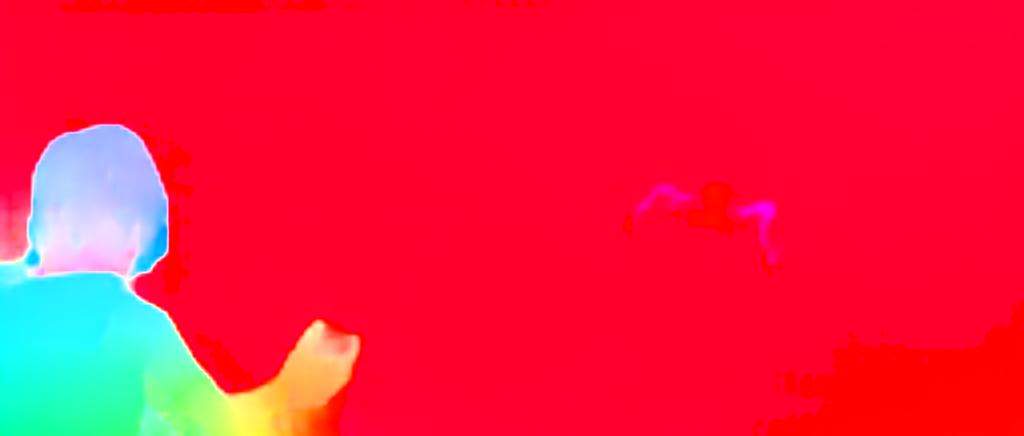}{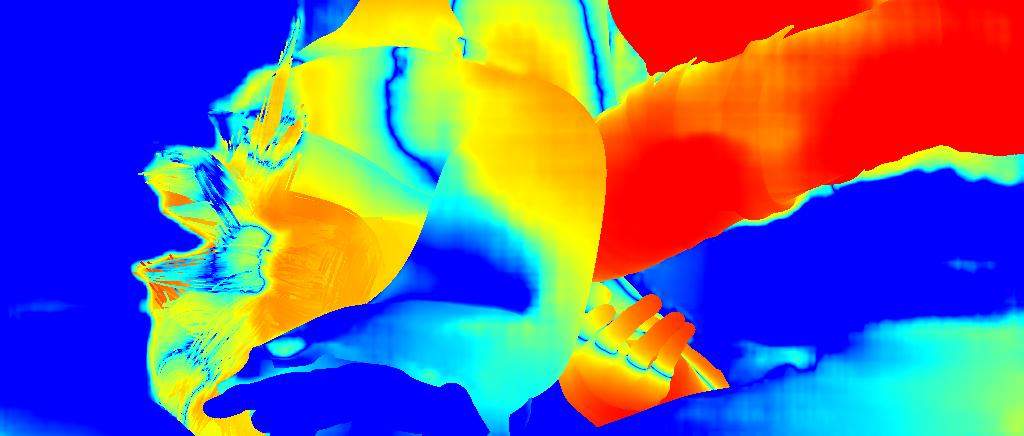}{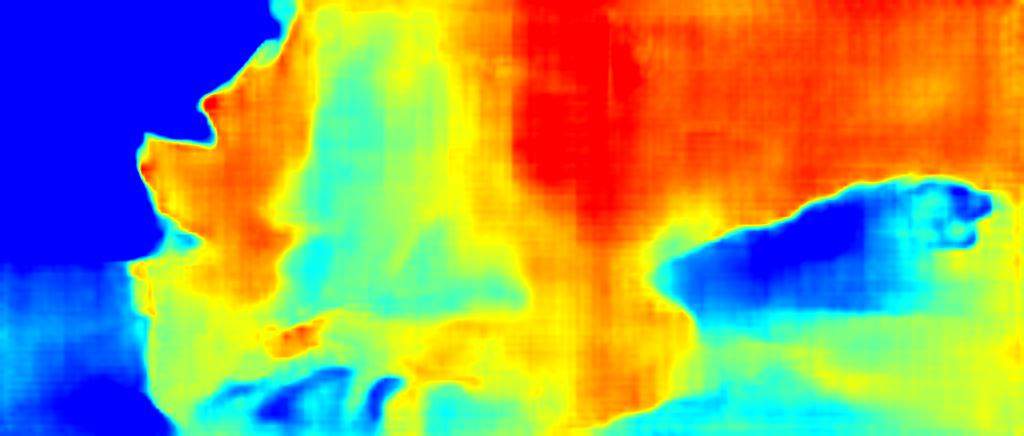}{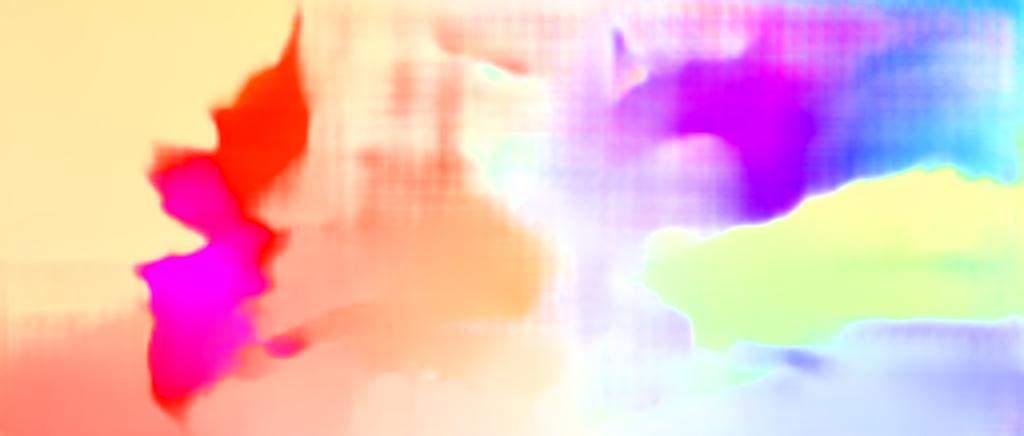}%
\galleryRowSintel{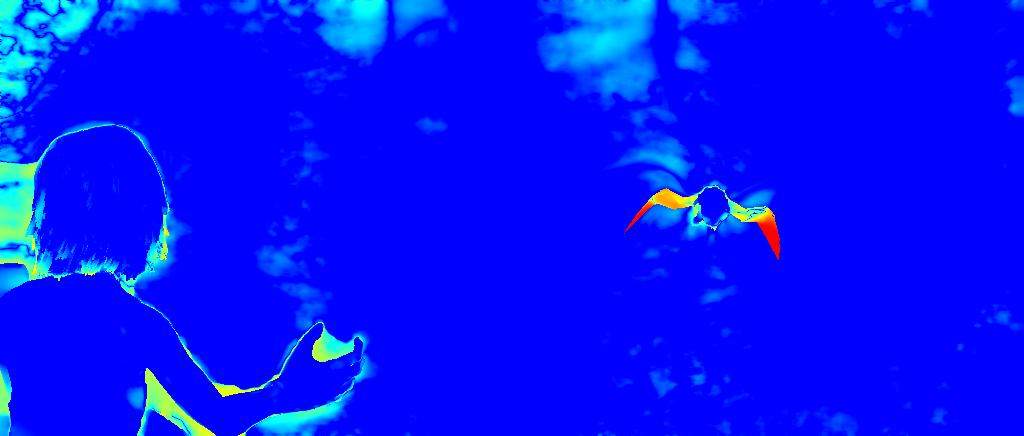}{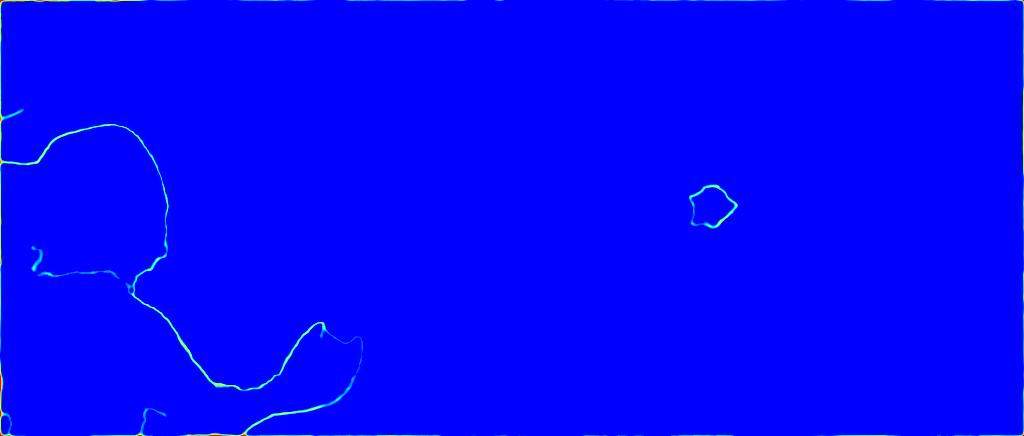}{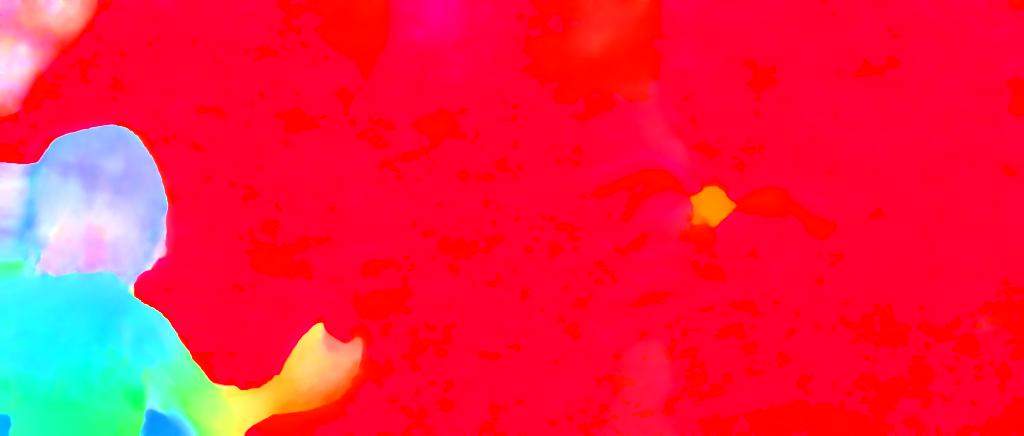}{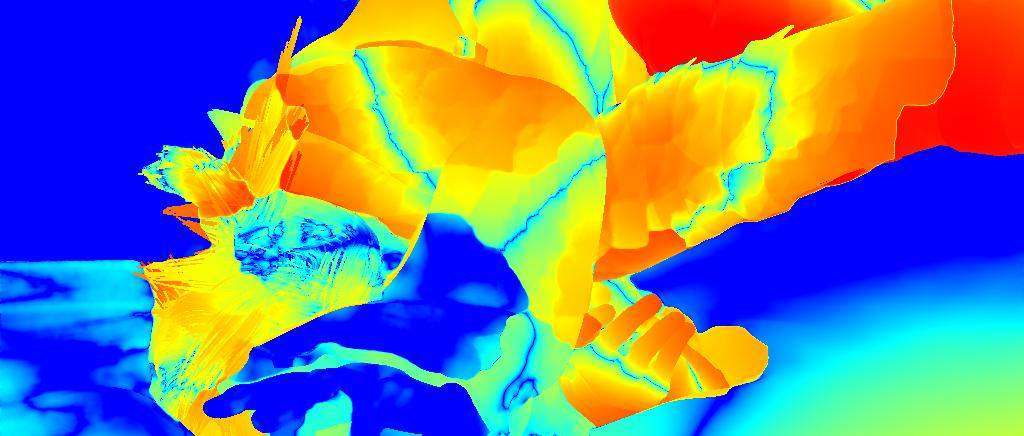}{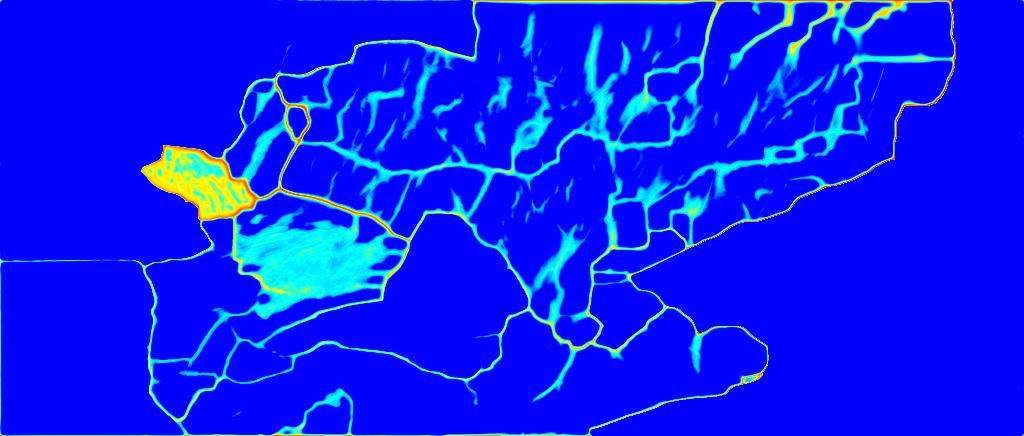}{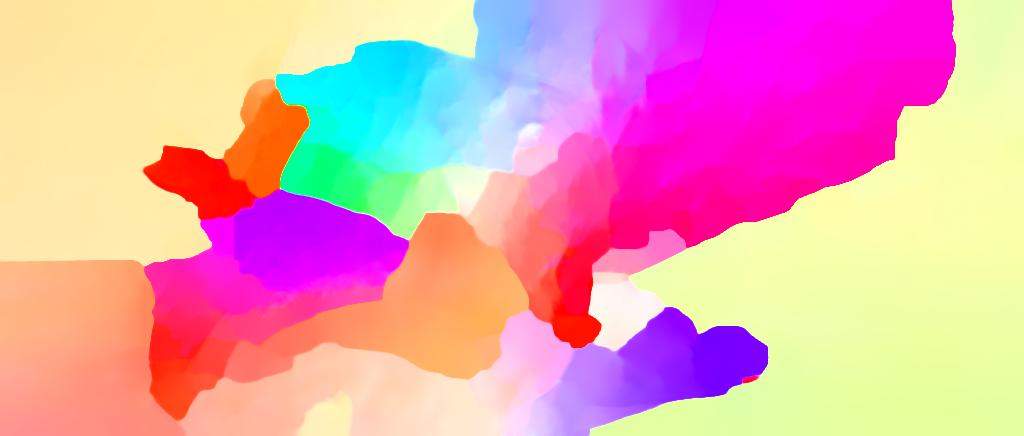}%
    \end{tabular}%
  }%
  \end{center}
  \caption{Comparison between FlowNetH-Pred-Merged and ProbFlow~\cite{probFlow}. The first row shows the image pair followed by its ground-truth flow for two different scenes from the Sintel final dataset. The second row shows FlowNetH-Pred-Merged results: entropy from a Laplace distribution with ground-truth error (we refer to this as \textit{Oracle Entropy} to represent the optimal uncertainty as explained in the supplemental material), predicted entropy and predicted flow. Similar to the second row, the third row shows the results for ProbFlow. Although both methods fail at estimating the motion of the dragon on the left scene and the motion of the arm and the leg in the right scene, our method is better at predicting the uncertainties in these regions.}%
  \label{fig:gallery_sintel}%
\end{figure*}

Figure~\ref{fig:gallery_sintel} shows a qualitative comparison to ProbFlow. Clearly, the uncertainty estimate of FlowNet-Pred-Merged also performs well outside motion boundaries and covers many other causes for brittle optical flow estimates. More results on challenging real-world data are shown in the supplemental video which can also be found on \url{https://youtu.be/HvyovWSo8uE}.  

\section{Conclusion} 

We presented and evaluated several methods to estimate the uncertainty of deep regression networks for optical flow estimation.
We showed that SGDR and Bootstrapped Ensembles perform better than the commonly used dropout technique. Furthermore, we found that a single network can estimate its own uncertainty surprisingly well and that this estimate outperforms every empirical ensemble. 
We believe that these results will apply to many other computer vision tasks, too.
Moreover, we presented a multi-hypotheses network that shows very good performance and is faster than sampling-based approaches and ensembles. 
The fact that networks can estimate their own uncertainty reliably and in real-time is of high practical relevance. Humans tend to trust an engineered method much more than a trained network, of which nobody knows exactly how it solves the task. However, if networks say when they are confident and when they are not, we can trust them a bit more than we do today.

\section*{Acknowledgements} 

We gratefully acknowledge funding by the German Research Foundation (SPP 1527 grants BR 3815/8-1 and HU 1900/3-1, CRC-1140 KIDGEM Z02) and by the Horizon 2020 program of the EU via the ERC Starting Grant 716721 and the project Trimbot2020.


{
\bibliographystyle{splncs04}
\bibliography{egbib}
}

\clearpage

\thispagestyle{empty}
\vspace*{5pt}
\begin{center}
{\Large \bf{
Supplementary Material 
}
}
\end{center}
\vspace*{10pt}

\setcounter{section}{0} 
\setcounter{figure}{0}
\setcounter{table}{0} 
\setcounter{equation}{0} 

\section{Video}
Please see the supplementary video for qualitative results on a number of diverse real-world video sequences and a comparison to ProbFlow~\cite{probFlow}. The video is also available on \url{https://youtu.be/HvyovWSo8uE}.

\section{Color Coding} 
\begin{figure}[t]
\begin{minipage}[b]{\linewidth}
\centering
\includegraphics[width=0.3\linewidth]{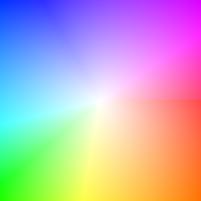}
\caption*{(a)}
\end{minipage}

\begin{minipage}[b]{\linewidth}
\centering
\includegraphics[width=0.3\linewidth]{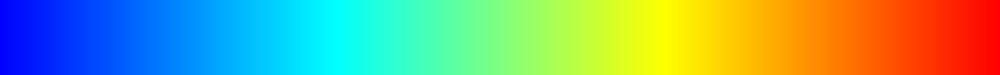}
\caption*{(b)}
\end{minipage}
\caption{\textbf{(a)} Flow field color coding used in this paper. The displacement of every pixel in this illustration is the vector from the center of the square to this pixel. The central pixel does not move. The value is scaled differently for different images to best visualize the most interesting range. \textbf{(b)} The color coding used for displaying the entropy maps, from the lowest value (blue), to the hightst (red).}
\label{fig:flow_key}
\end{figure}

For optical flow visualization we use the color coding of Butler \etall~\cite{sintel}.
The color coding scheme is illustrated in Figure~\ref{fig:flow_key}. 
Hue represents the direction of the displacement vector, while the intensity of the color represents its magnitude.
White color corresponds to no motion.
Because the range of motions is very different in different image sequences, we scale the flow fields before visualization: 
independently for each image pair shown in figures, and independently for each video fragment in the supplementary video.
Scaling is always the same for all methods being compared.

For uncertainty visualizations we show the predicted entropy, which we compute as:
\begin{equation}
    H = log(2b_xe) + log(2b_ye) \mathrm{\,,}
\end{equation}
where $b_x$ and $b_y$ are estimated scale parameters from our Laplace distribution model for $x$ and $y$ dimensions and $e$ is Euler's number. 
To assess the quality of our uncertainty estimations, we compare our estimated entropies against the limiting cases, where $b_x$ and $b_y$ correspond to exactly the estimation errors $|u_{pred}-u_{gt}|$ and $|v_{pred}-v_{gt}|$.
We visualize this as the \textit{Oracle Entropy} in all cases where ground-truth is present. For ProbFlow~\cite{probFlow}, the underlying distribution is Gaussian and therefore we use the entropy of a Gaussian distribution as: 
\begin{equation}
    H = 0.5 * log(2e\sigma_x^2\pi) + 0.5 * log(2e\sigma_y^2\pi) \mathrm{\,,}
    \label{eq:ent_gauss}
\end{equation}
and set $\sigma_x$ and $\sigma_y$ to $|u_{pred}-u_{gt}|$ and $|v_{pred}-v_{gt}|$, respectively. To compare to this oracle entropy we normalize to the same range, but when comparing our method to ProbFlow, we allow to normalize to different ranges to show the most interesting aspects of the entropy. 

\section{Sparsification Plots}
\label{sec:sparsification_types}
Sparsification is a way to assess the quality of uncertainty estimates for optical flow. Already popular in literature~\cite{Bruhn2006,Kondermann2008,bootstrapOpticalFlow,opticalFlowPAMI2012}, it works by progressively discarding percentages of the pixels the model is most uncertain about and verifying whether this corresponds to a proportional decrease in the remaining average endpoint error. To make the results of different experiments comparable, the errors are normalized to $1$.

\begin{figure*}[t]
  \begin{center}
  \begin{subfigure}[b]{0.433\linewidth}
      \centering
      \includegraphics[width=\linewidth]{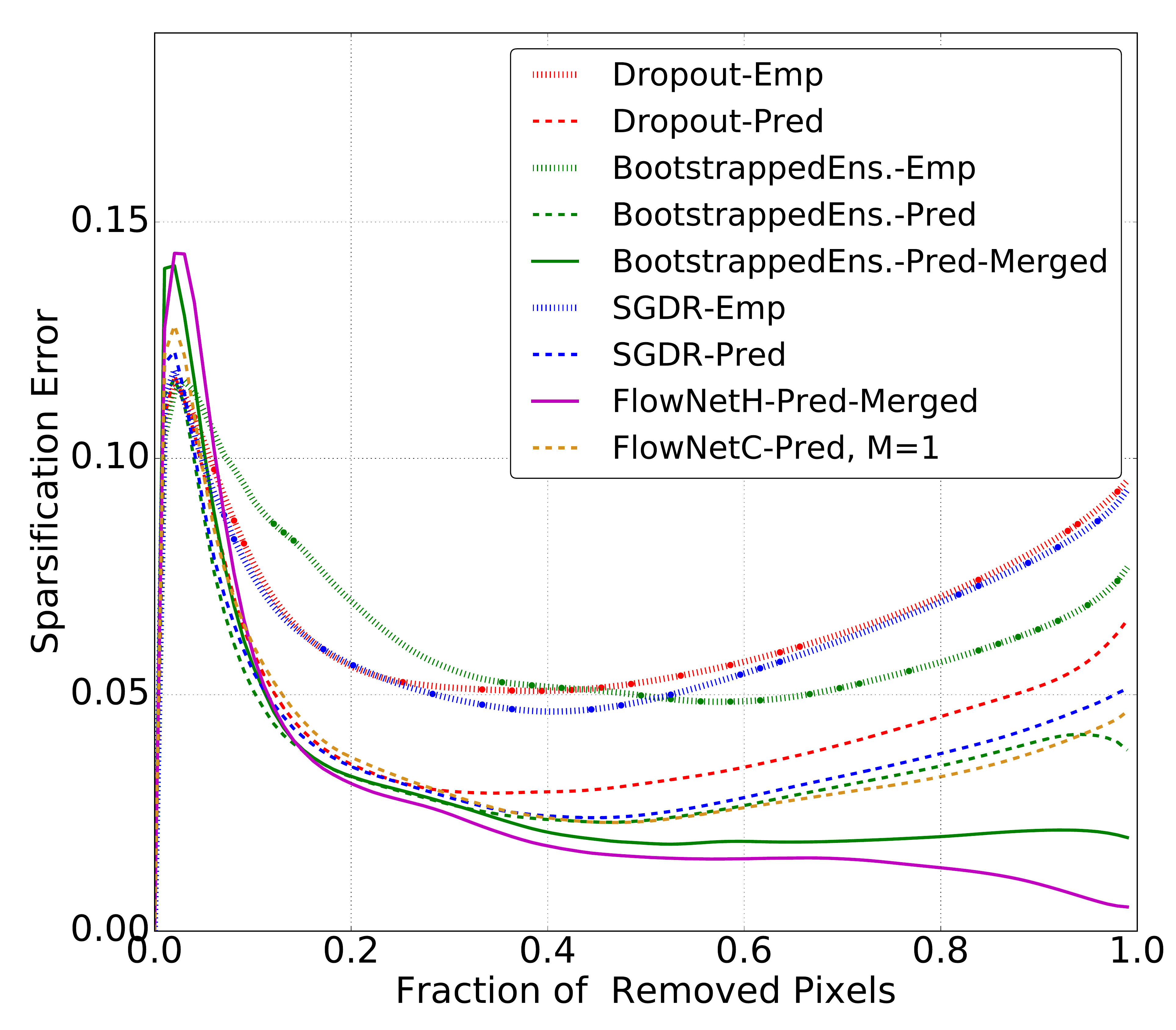}
      \caption{\label{fig:sparsification_all_dataset}}
    \end{subfigure}
    \begin{subfigure}[b]{0.547\linewidth}
      \centering
      \includegraphics[width=\linewidth]{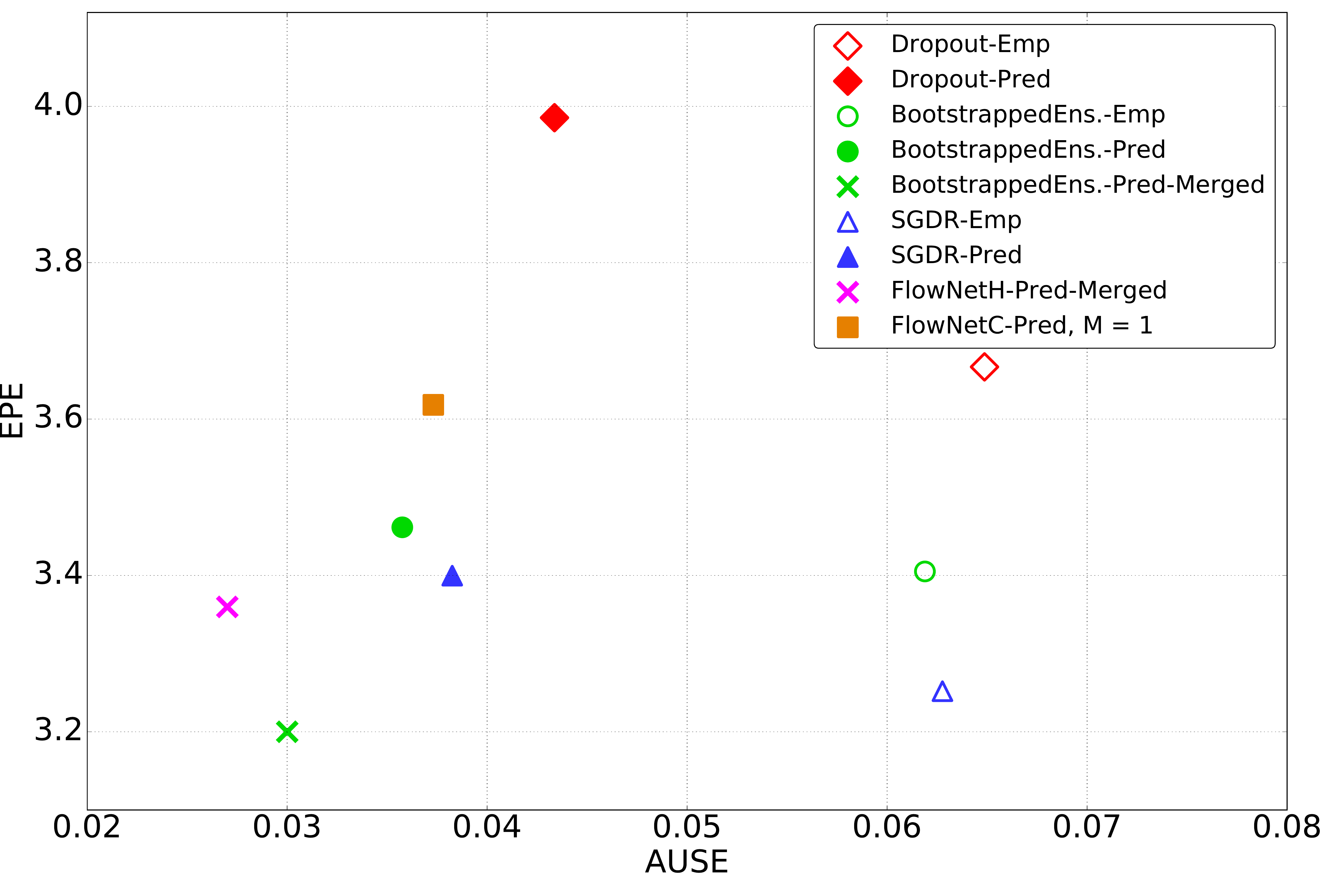}
      \caption{\label{fig:ensemble_size_dataset}}
    \end{subfigure}
  \end{center}
\caption{\textbf{NOTE: In this version we normalize to dataset-wise instead of image-wise.} \textbf{(a)} Sparsification error plots on Sintel train clean dataset. Number of ensemble members are fixed to $M=8$ and number of pre-cycles for SGDR to $8$. We observe that in this case FlowNetH-Pred-Merged and the BootstrappedEnsemble-Pred-Merged perform slightly worse for very high uncertainties, while still showing the best performance for remaining uncertainties. 
\textbf{(b)} Scatter plot of EPE vs. AUSE for proposed ensemble types. For SGDR, we take the last M available snapshots.  The behavior of the different models is not drastically different from the one visible in the per-image sparsification scatter plots in Figure 5 from the main paper. The best performing model in terms of AUSE is FlowNetH-Pred-Merged.
\label{fig:sizes_shifts_scatter}
}
\end{figure*}

\paragraph{Image-wise sparsification.}
The method, including the normalization, is typically applied to images individually and the sparsification plots of all images are then averaged. In the main paper we also follow this procedure. However, this approach weights images where the uncertainty estimation is easy equally to images where the uncertainty estimation is hard. Also, due to the normalization, pixels with very large enpoint error from one image can be treated equally to pixels with very small endpoint error from another image. 

\paragraph{Dataset-wise sparsification.}
Alternatively, one can perform the sparsification on a whole dataset. In this variant, the sparsification is performed first (by ranking across the whole dataset) and normalization is performed last. With this approach, the effect of the outliers is better visible in the sparsification curves, which show larger slopes with respect to the previous version.

In Figures \ref{fig:sparsification_all_dataset}, and \ref{fig:ensemble_size_dataset} we present the figures from the main paper again with the dataset-wise sparsification. In Figure \ref{fig:sparsification_all_dataset} we observe that the FlowNetH-Pred-Merged and BootstrappedEnsemble-Pred-Merged perform slightly worse than other ensembles for very high uncertainties, when sparsified on the whole dataset. As also observed in the main paper the best performing model in terms of AUSE is FlowNetH-Pred-Merged.

\begin{figure*}[t]
  \begin{center}
  \begin{subfigure}[b]{0.49\linewidth}
    \centering
      \includegraphics[width=\linewidth]{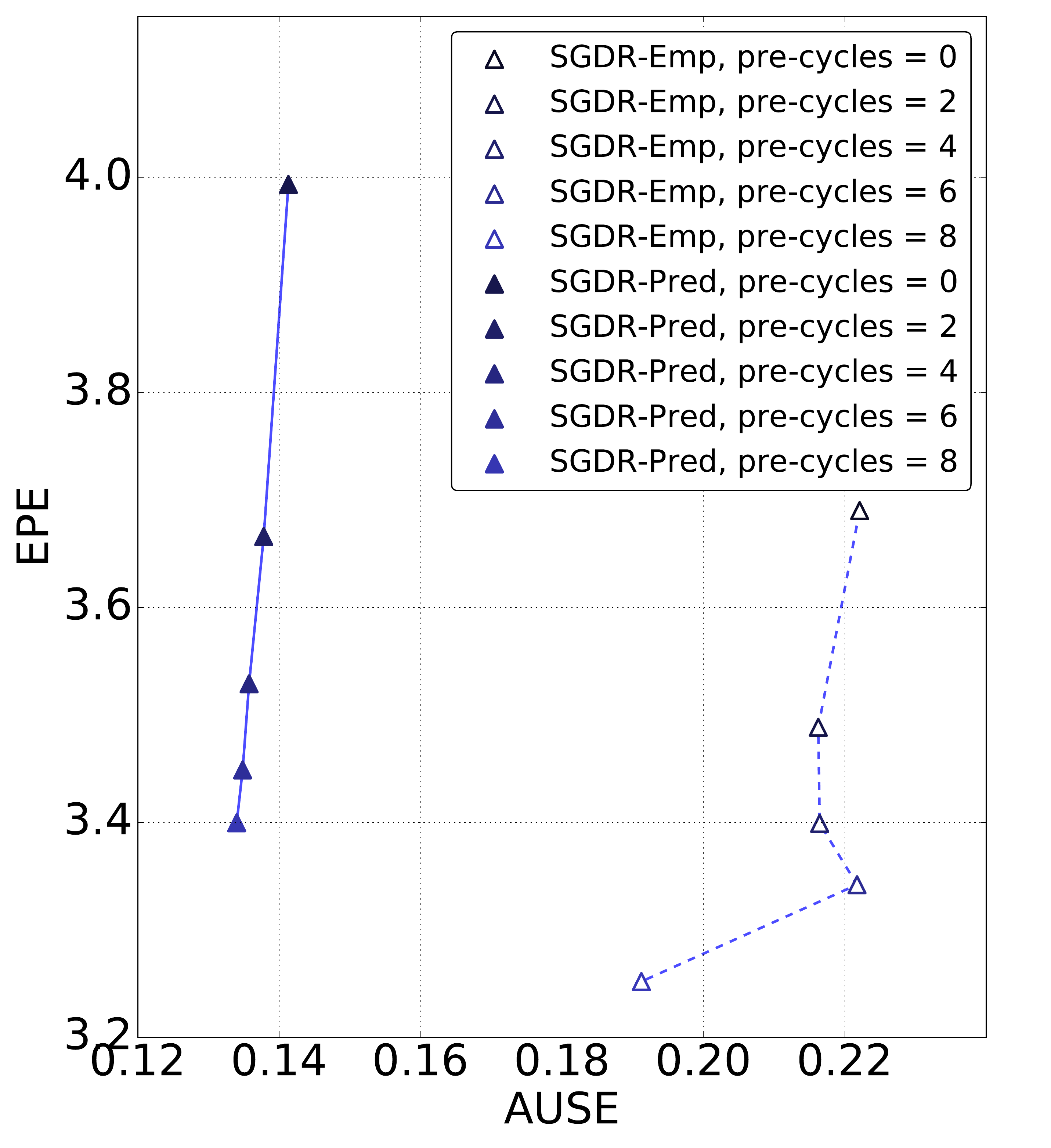}
      \caption{\label{fig:sgdr_offset_per_image_spars}}
  \end{subfigure}
  \begin{subfigure}[b]{0.49\linewidth}
    \centering
      \includegraphics[width=\linewidth]{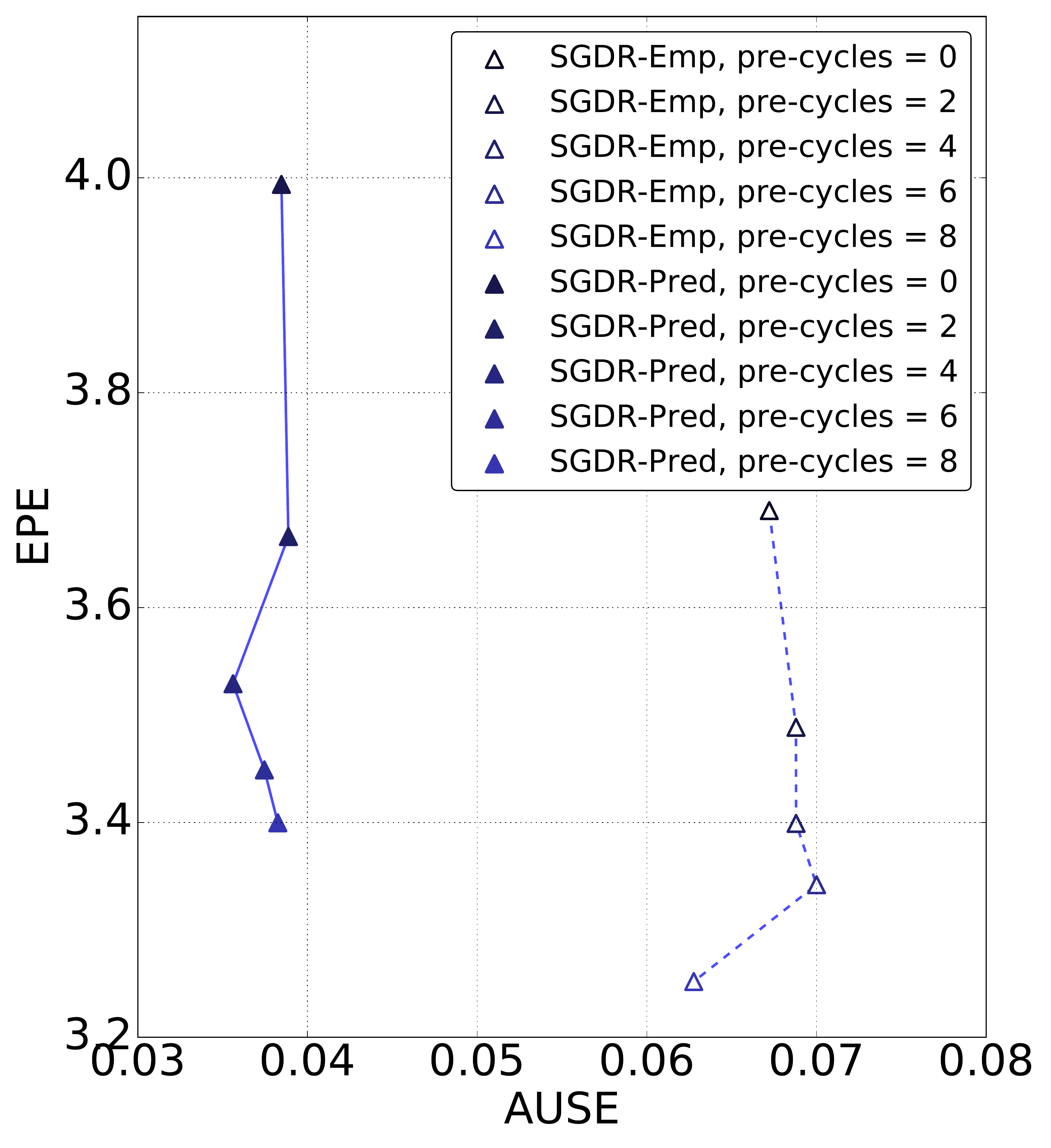}
      \caption{\label{fig:sgdr_offset_whole_dataset_spars}}
  \end{subfigure}
  \end{center}
  \caption{Scatter plot of EPE vs. AUSE showing the effect of different number of pre-cycles for SGDR ensembles with ensemble size $M=8$. \textbf{(a)} shows the plot for the image-wise sparsification and \textbf{(b)} shows the plot for the dataset-wise sparsification, as explained in Section \ref{sec:sparsification_types}. It can be seen that a larger number of pre-cycles always positively affects the EPE without penalizing the AUSE score.\label{fig:sgdr_offset}}
\end{figure*}

\begin{figure}[t]
  \begin{center}
      \includegraphics[width=0.8\linewidth]{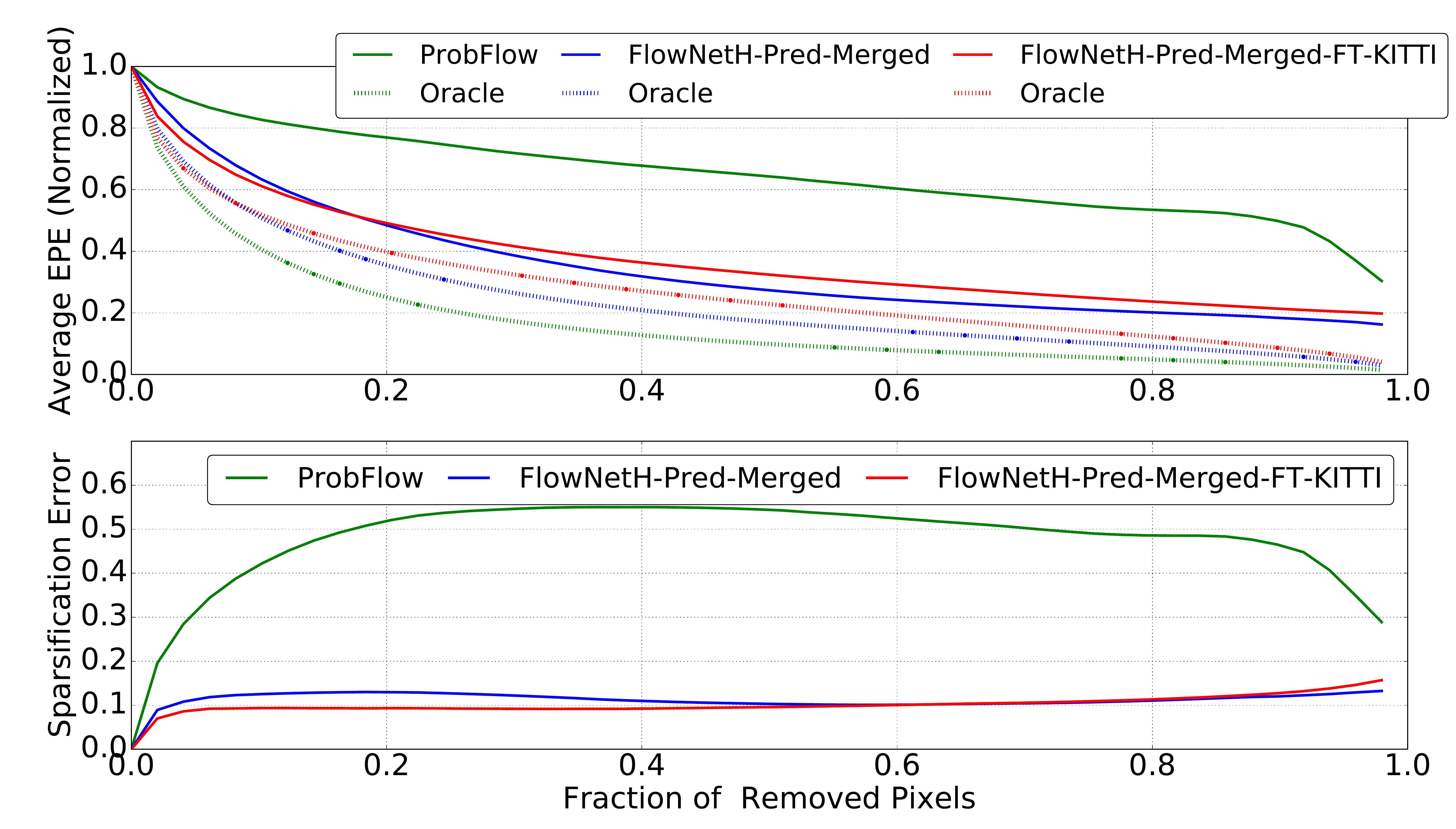}
  \end{center}
\caption{Sparsification and sparsification error plots for ProbFlow~\cite{probFlow}, FlowNetH-Pred-Merged (further trained on the FlyingThings3D dataset for 250k iterations) and FlowNetH-Pred-Merged-FT-KITTI (subsequently fine-tuned on our joint KITTI2012 and 2015 training split). One can observe that fine-tuning does not change the sparsification error drastically, while EPE reduces significantly (see Table~3 in the main paper). We see that FlowNetH outperforms ProbFlow both in terms of EPE and AUSE. 
Note that although the average EPE for ProbFlow is higher, due to the effect of normalization of the oracle sparsification curve, it appears to be the lowest.
  \label{fig:sparsification_hyp}
}
\end{figure}

\section{Effect of Pre-Cycles for SGDR Ensembles}

For SGDR ensembles not only the ensemble size $M$, but also the models discarded from earlier cycles matter (pre-cycles). Therefore, we have further experimented with pre-cycle counts from $0$ to $8$ (with a constant ensemble size of $M = 8$). The scatter plots of EPE vs. AUSE can be seen in Figure~\ref{fig:sgdr_offset}. Figure~\ref{fig:sgdr_offset_per_image_spars} shows the plot where image-wise normalization is used for sparsification, while Figure~\ref{fig:sgdr_offset_whole_dataset_spars} shows the plot for dataset-wise normalization. From both plots we can see that the later the models are taken, the lower the EPE gets without a significant change in the AUSE. When compared to other ensemble types, SGDR ensembles are derived from earlier snapshots and the ones with more pre-cycles are trained for more iterations in total. This might be the reason why they show a lower EPE. However, it also means they can converge more and we actually observe the lowest variance for SGDR ensembles among all the models (see Var. column of Table~2 in the main paper).

\section{Evaluation on KITTI and Comparison to ProbFlow}

We perform the final evaluation of FlowNetH also on the KITTI datasets. We therefore mix KITTI2012 and KITTI2015 and split into $75\%/25\%$ training and test data. In Figure~\ref{fig:sparsification_hyp} and Table~3 from the main paper, we show the performance of our method compared to ProbFlow~\cite{probFlow}. As can be seen from Table~3 in the main paper, fine-tuning significantly reduces the endpoint error, as well as AUSE for FlowNetH-Pred-Merged. This concludes that the quality of the uncertainty estimation of FlowNetH-Pred-Merged is outperforming ProbFlow independent of the flow accuracy. 

\subsection{Details on ProbFlow results}
In order to reproduce ProbFlow results, the ProbFlowFields algorithm contained in the official software package was used. In particular flow initializations were obtained from FlowFields matches~\cite{FlowFields} and subsequently interpolated with EpicFlow \cite{EpicFlow}. For FlowFields on KITTI, we have found the best parameter combination to be $r=5$, $r2=4$, $\epsilon=5, e=4$ and $s=50$ (the search was conducted around the values suggested in the paper, and full parameter optimization was not performed), and on Sintel we employed $r=8$, $r2=6$, $\epsilon=5, e=4$ and $s=50$ as reported by the authors of ProbFlow.
In contrast to~\cite{probFlow} we evaluate on the complete Sintel train set instead of the selected subset from~\cite{probFlow} 
This explains the difference in the resulting EPE comparing to the original paper. 

\section{Training Details for Stacked Networks}

\begin{figure}
  \begin{center}
     \includegraphics[width=\linewidth]{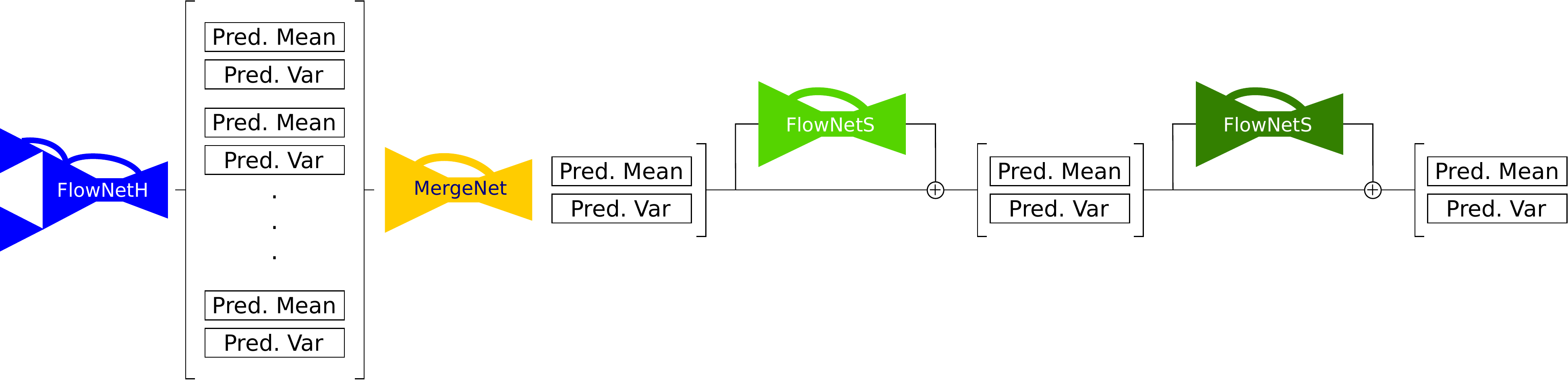}
  \end{center} 
  \caption{
      Illustration of our full uncertainty estimation stack. The first two networks are the FlowNetH and the merging network as described in the paper. The two stacked FlowNetS architecture networks estimate residual refinements. 
      \label{fig:stack_illust}
   }
\end{figure}

\begin{figure}
  \begin{center}
     \includegraphics[width=\linewidth]{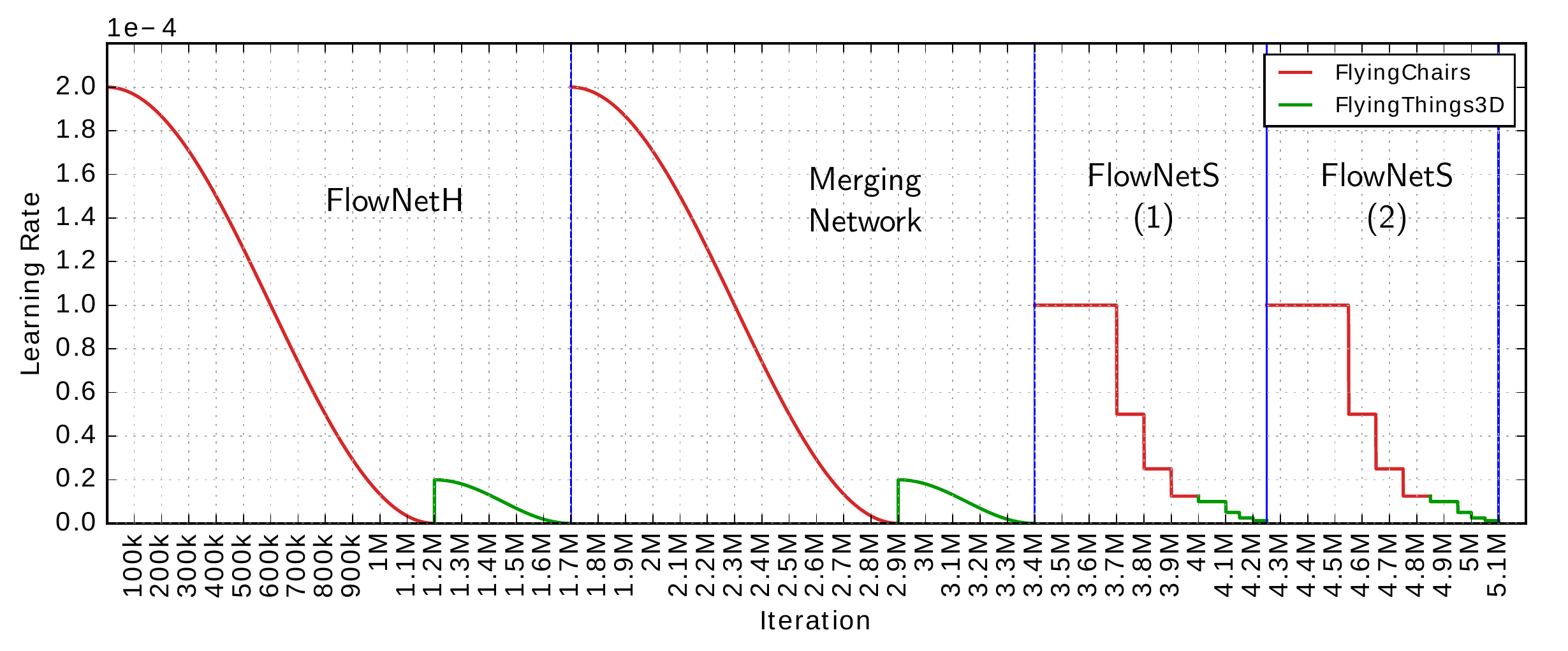}
  \end{center} 
  \caption{
      \label{fig:stack_lr}
      The learning rate curve of our full stack. FlowNetH and Merging network are trained with Batch Normalization and cosine annealing, while refinement networks are trained without Batch Normalization and with the schedules proposed for FlowNet 2.0~\cite{flownet2} scaled to half the number of iterations. 
      As in~\cite{flownet2} the networks are trained step by step, i.e. 
      indicated by each blue line in the figure, the bottom networks are fixed and not trained any more. 
  }
\end{figure}

The work of FlowNet 2.0~\cite{flownet2} presented how several refinement networks could be stacked on top of a FlowNetC~\cite{flownet2} to obtain improved flow fields. 
We also build such a stack for our FlowNetH-Merged. As illustrated in Figure~\ref{fig:stack_illust}, we stack two refinement networks on top. Different from~\cite{flownet2}, each refinement network yields the residual of the flow field and the uncertainty, as recently proposed by~\cite{crl}. Particularly, we find that the residual connections yield much lower convergence times. While we found Batch Normalization and the cosine annealing schedule to be useful for uncertinty estimation networks, we found that refinement networks perform better without Batch Normalization. Therefore, we use the schedule proposed by FlowNet 2.0~\cite{flownet2} scaled down to half the number of iterations and no Batch Normalization for the refinement networks. An overview of the training steps is provided in Figure~\ref{fig:stack_lr}.

\pagebreak 
\section{Qualitative Evaluation} 

\begin{figure*}[t]
  \begin{center}%
    \newcommand{\galleryRowReal}[6]{\includegraphics[width=0.2\textwidth]{images/#1} & \includegraphics[width=0.2\textwidth]{images/#2}  & \includegraphics[width=0.2\textwidth]{images/#3}  & & \includegraphics[width=0.2\textwidth]{images/#4}  & \includegraphics[width=0.2\textwidth]{images/#5}  & \includegraphics[width=0.2\textwidth]{images/#6} \tabularnewline
}
  \resizebox{\linewidth}{!}{%
    \setlength{\tabcolsep}{0.7pt}%
    \begin{tabular}{cccp{2.5mm}ccc}%
\galleryRowReal{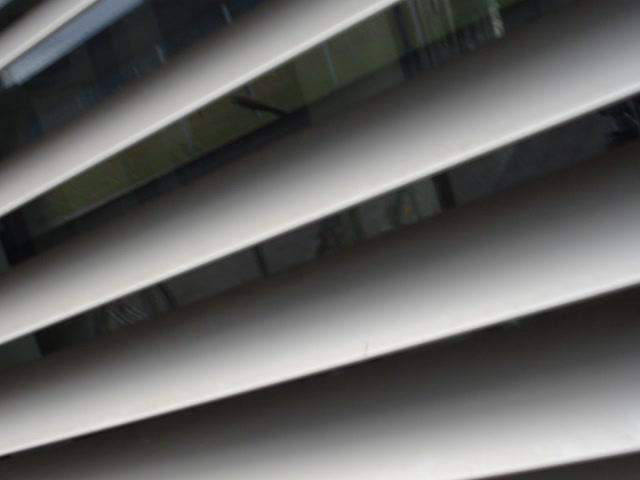}{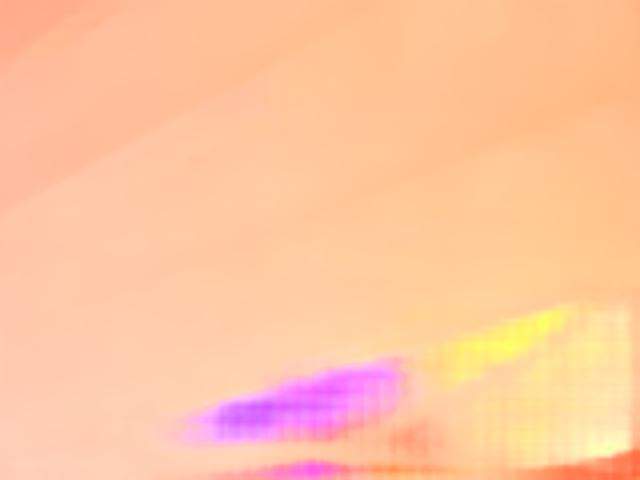}{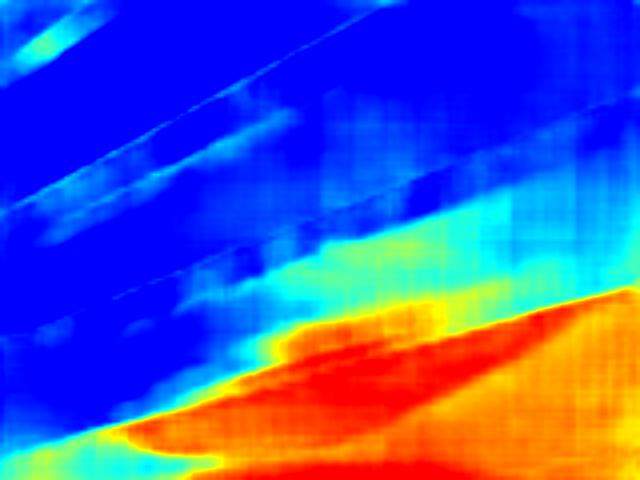}{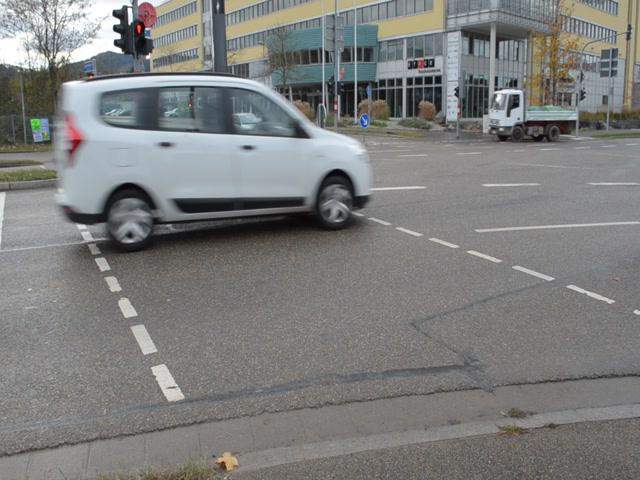}{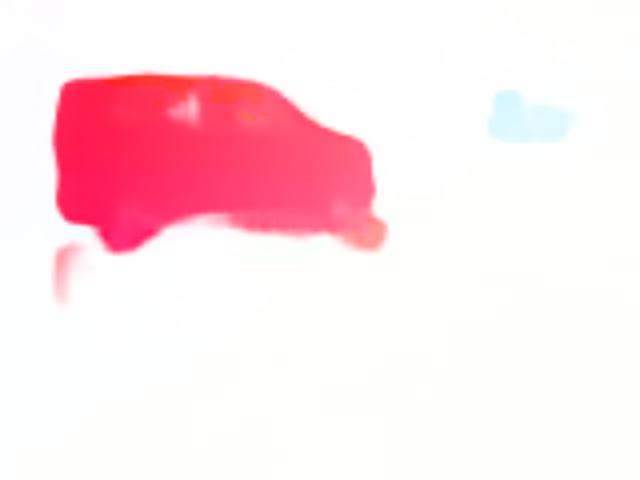}{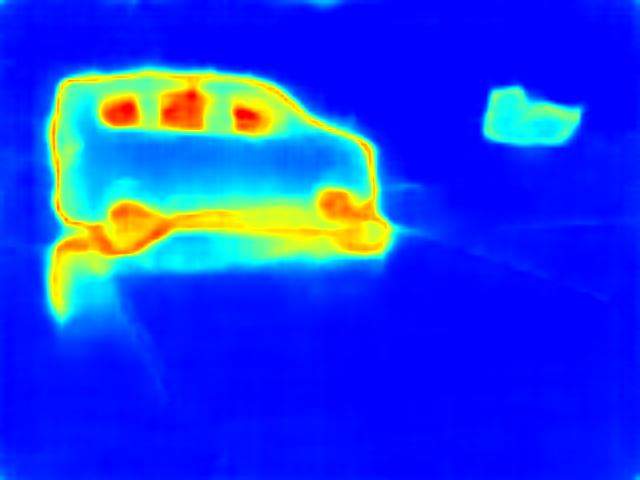}\vspace{3mm}%
\galleryRowReal{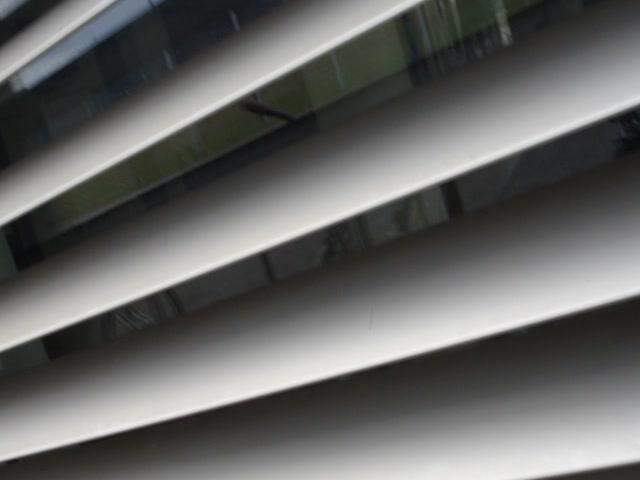}{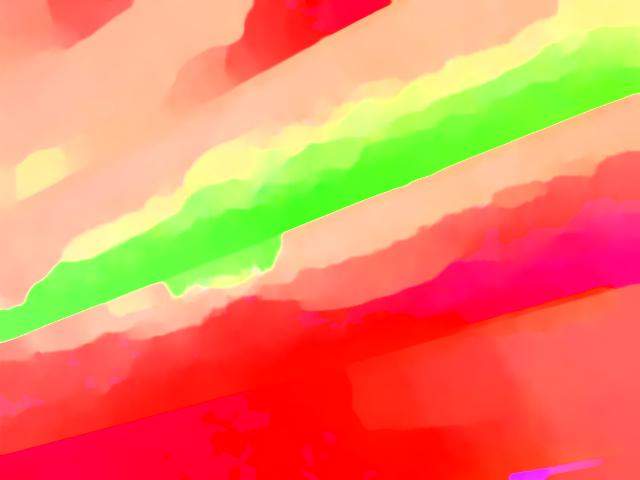}{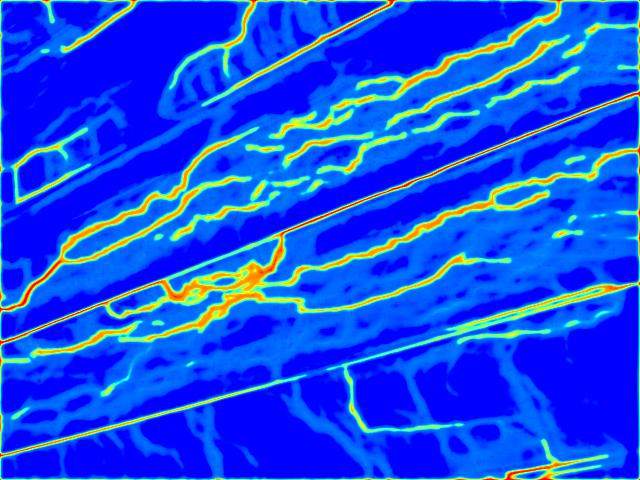}{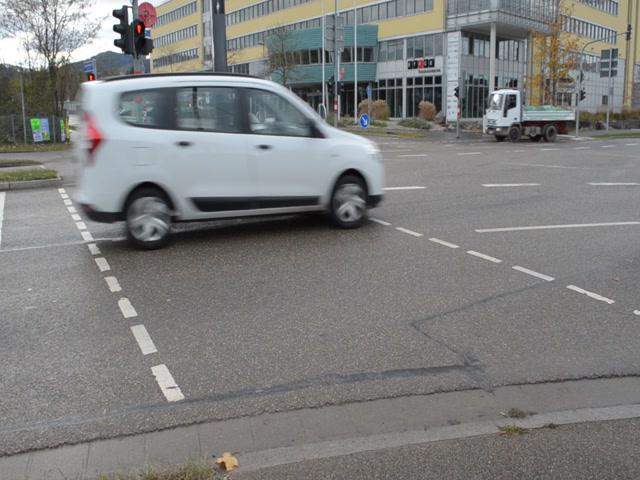}{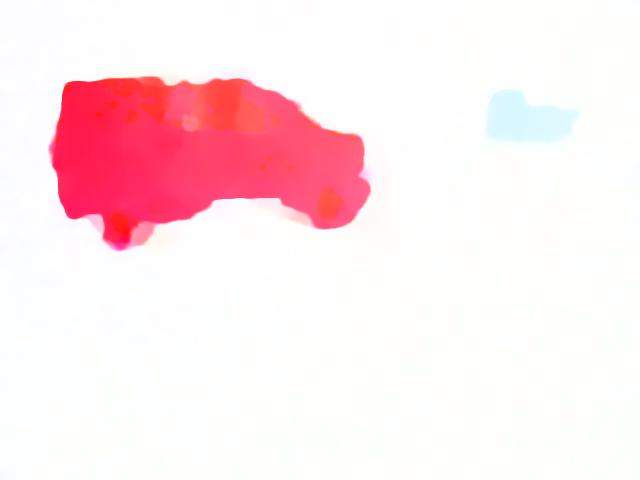}{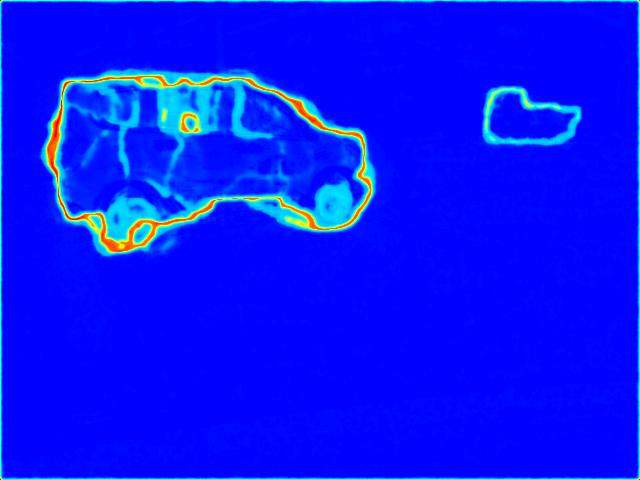}%
\galleryRowReal{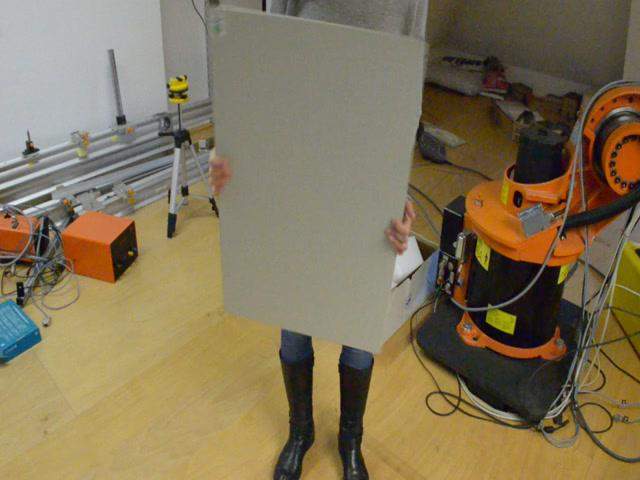}{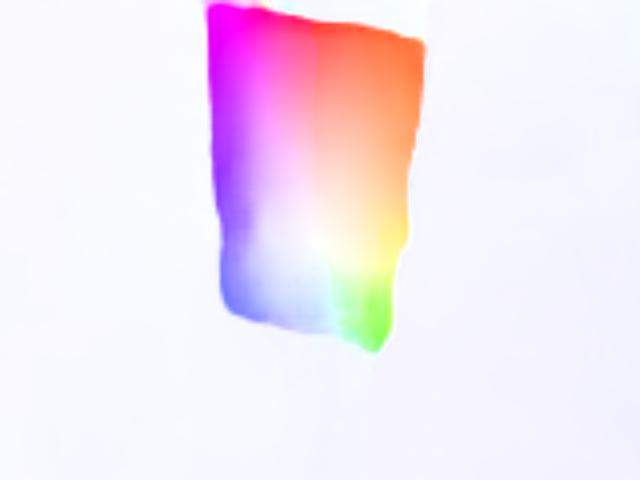}{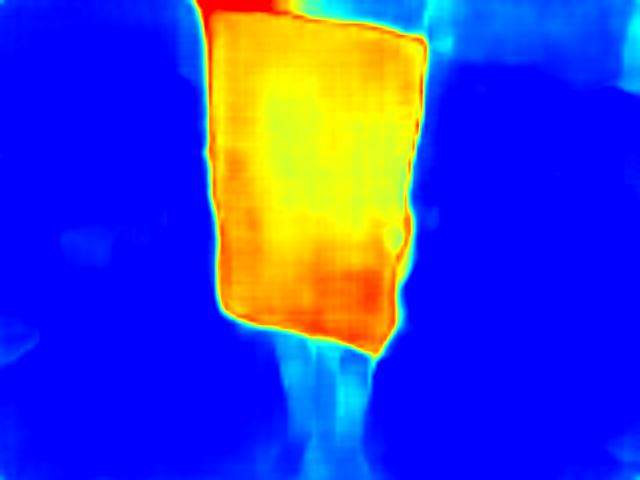}{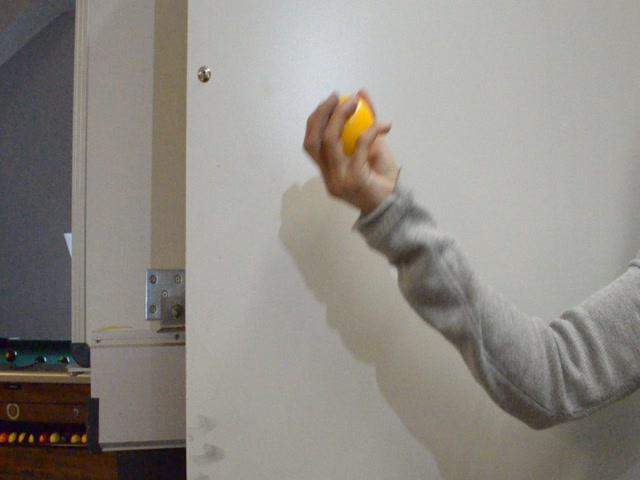}{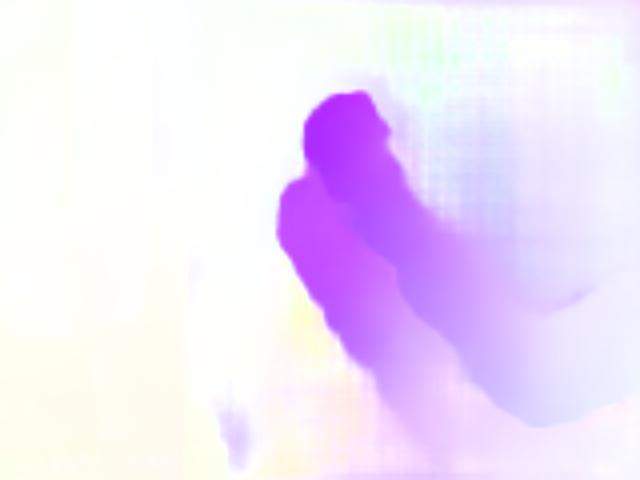}{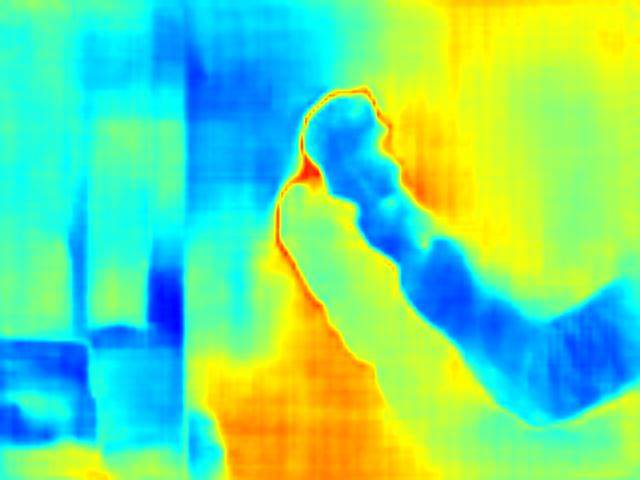}\vspace{3mm}%
\galleryRowReal{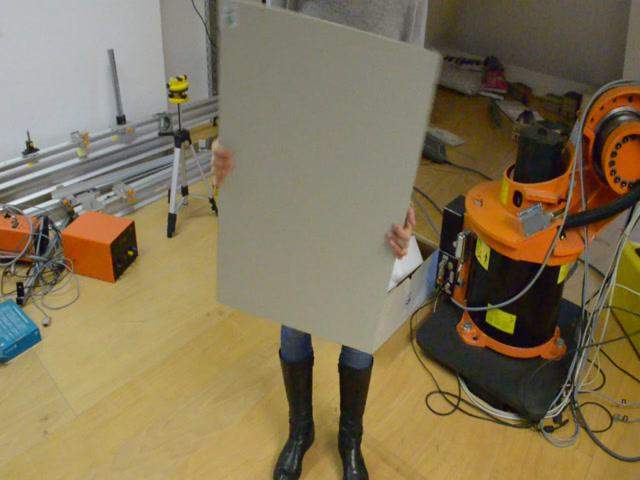}{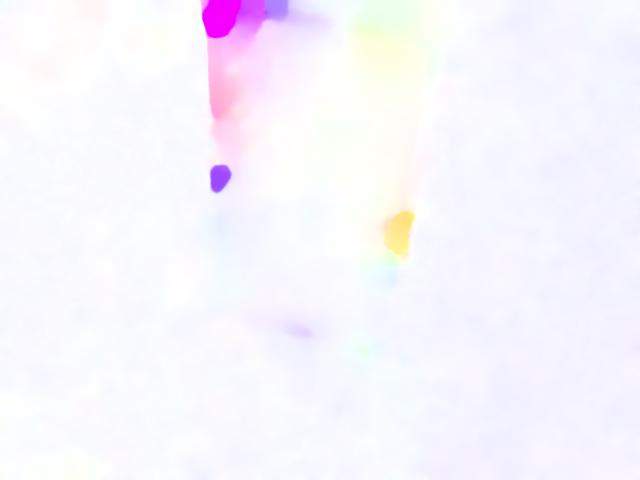}{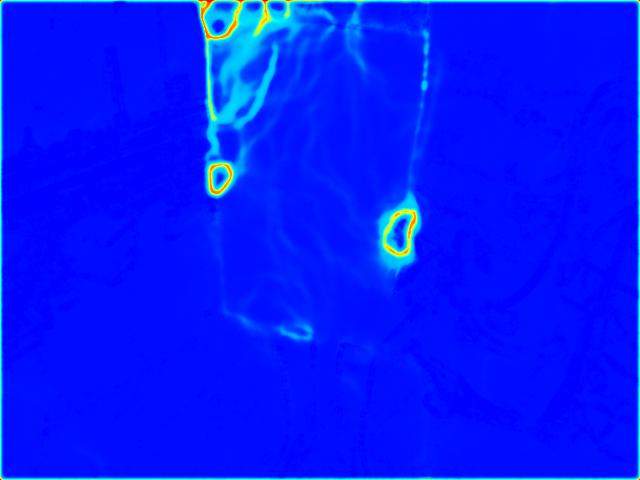}{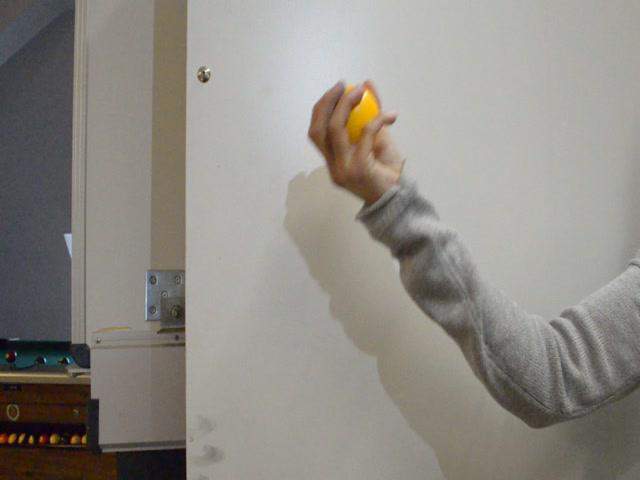}{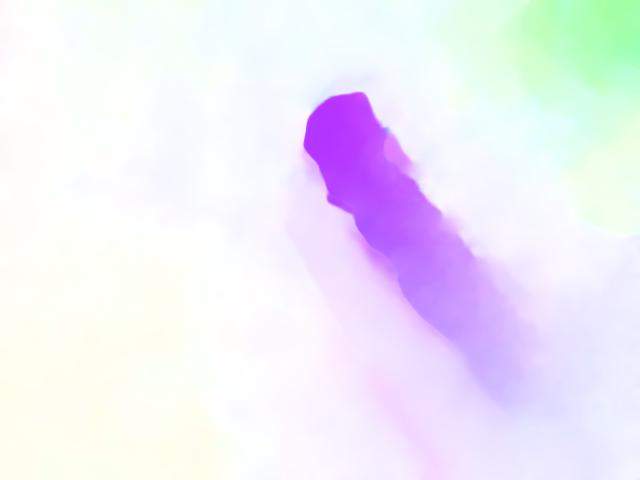}{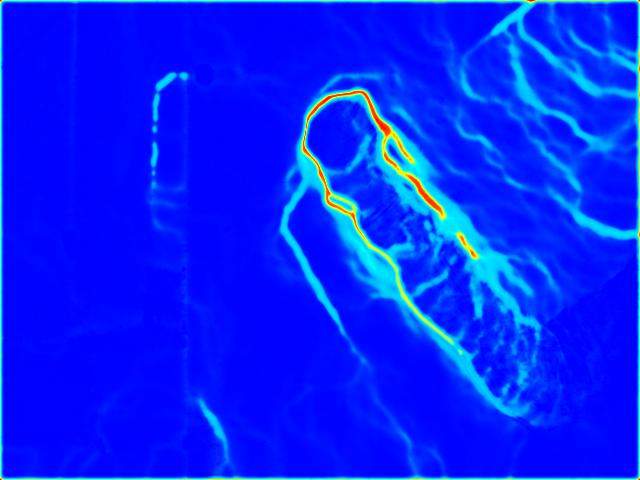}%
\galleryRowReal{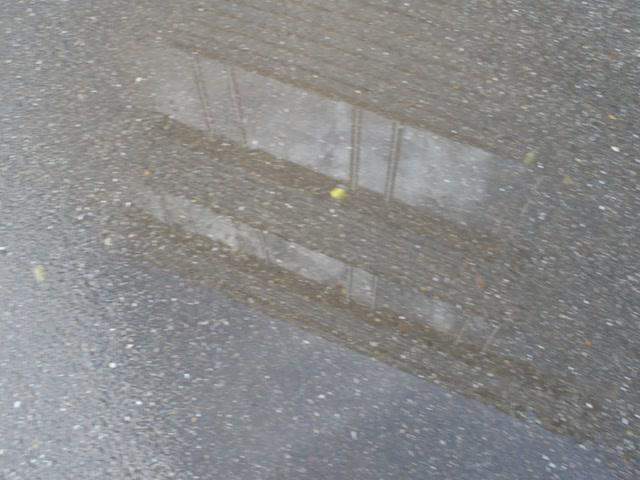}{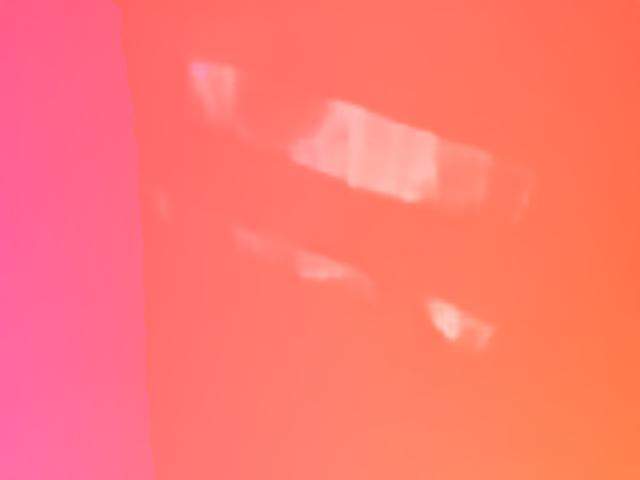}{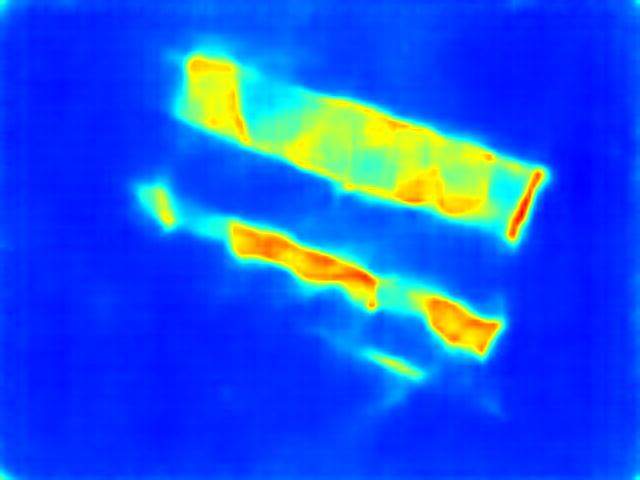}{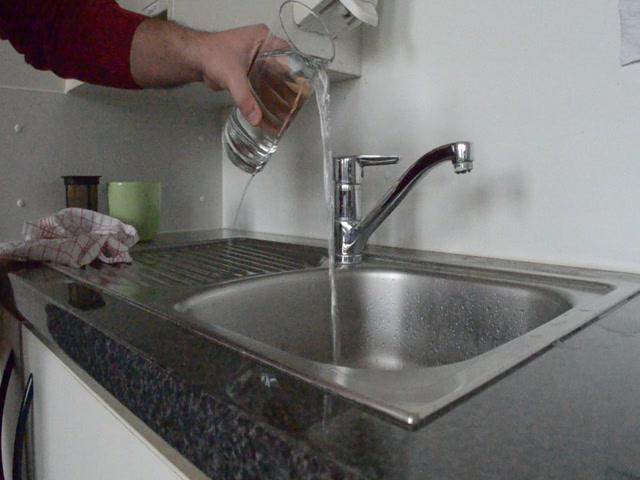}{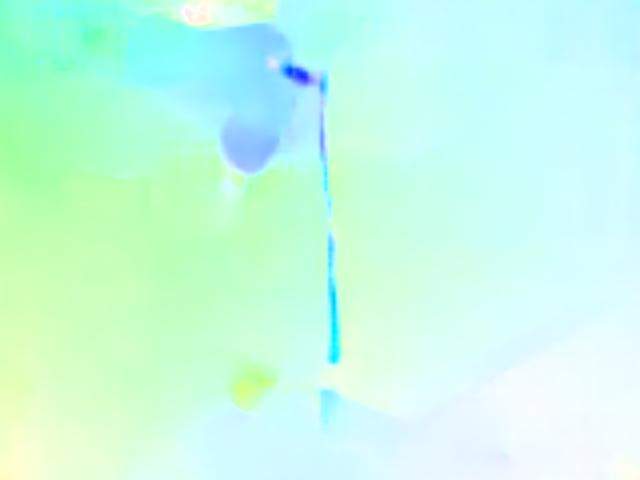}{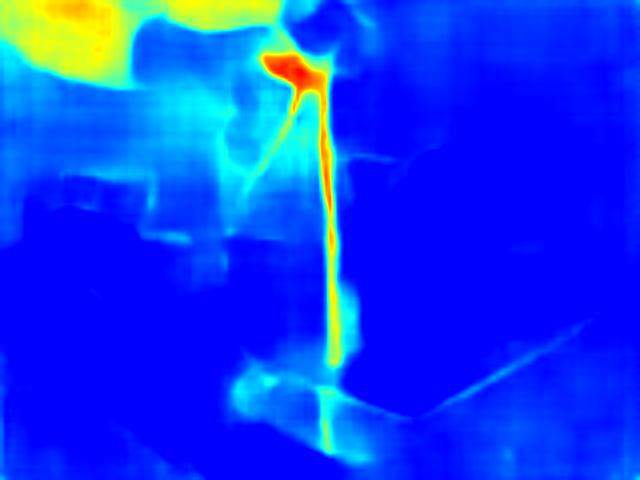}\vspace{3mm}%
\galleryRowReal{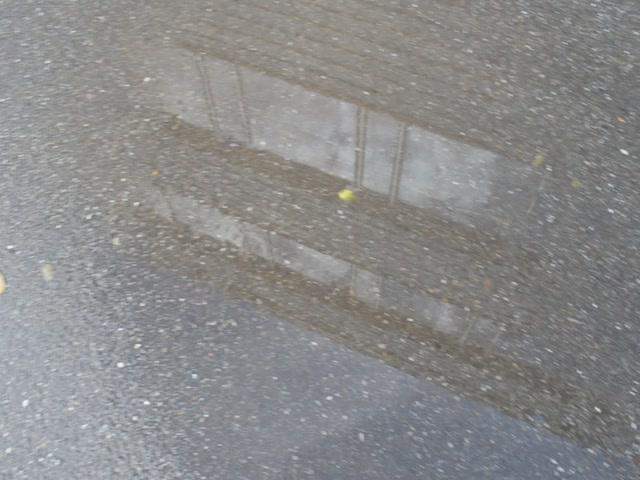}{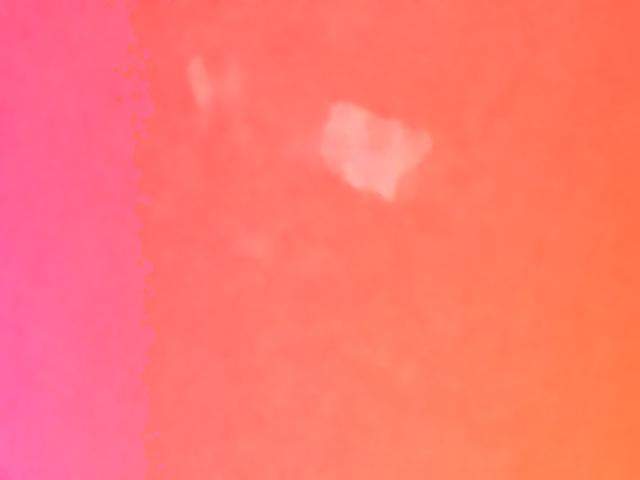}{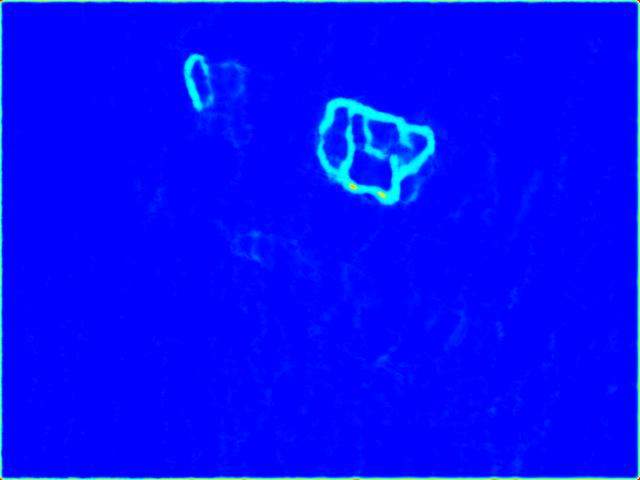}{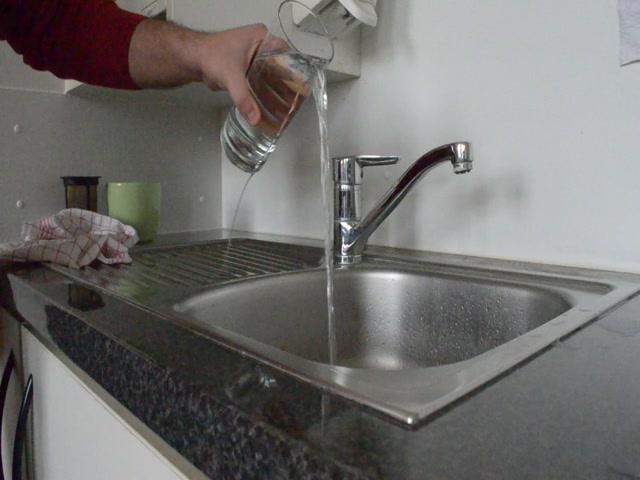}{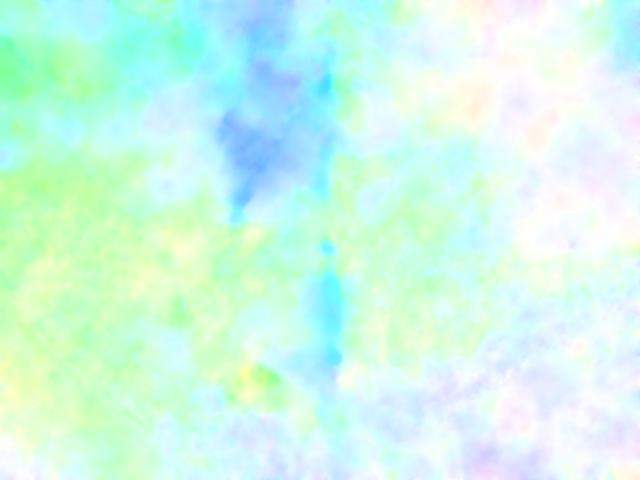}{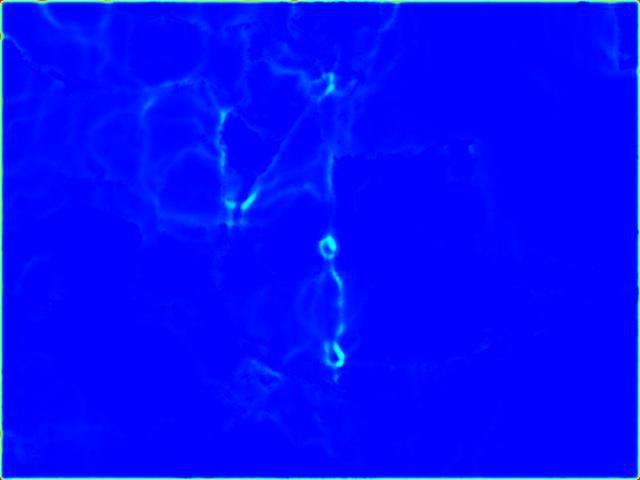}%
\galleryRowReal{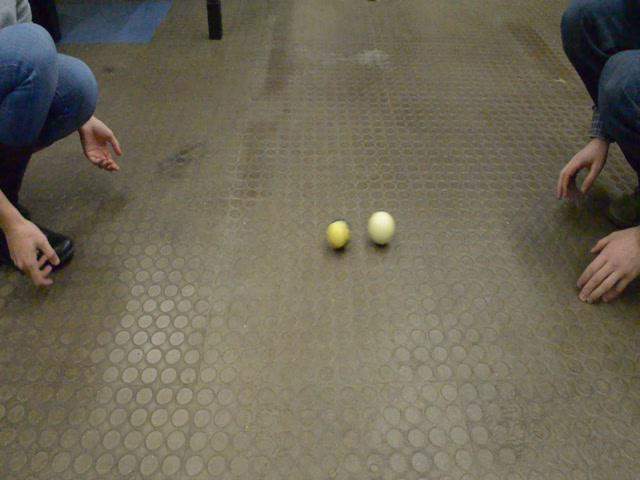}{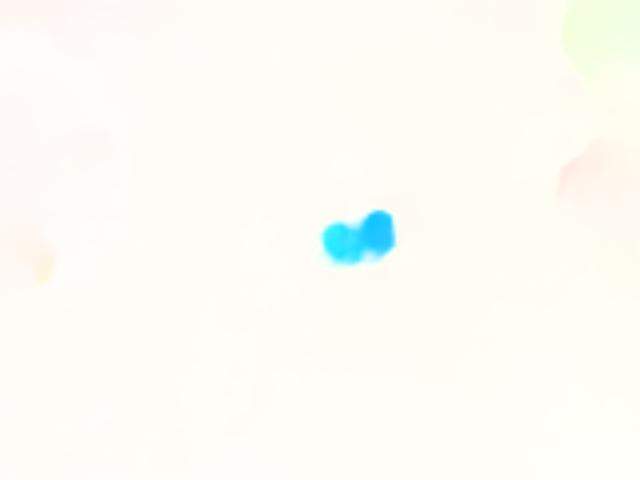}{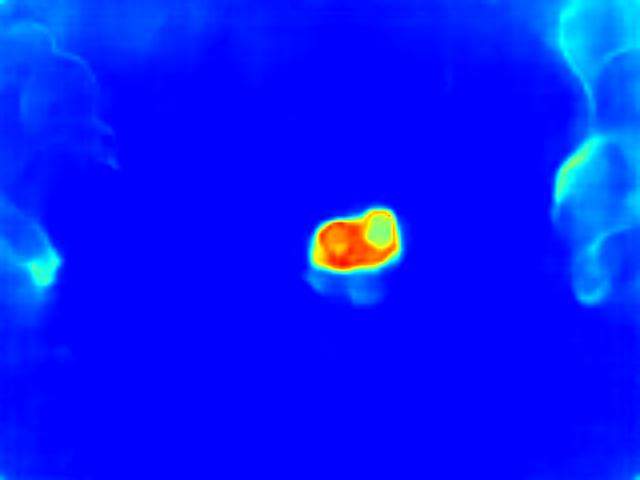}{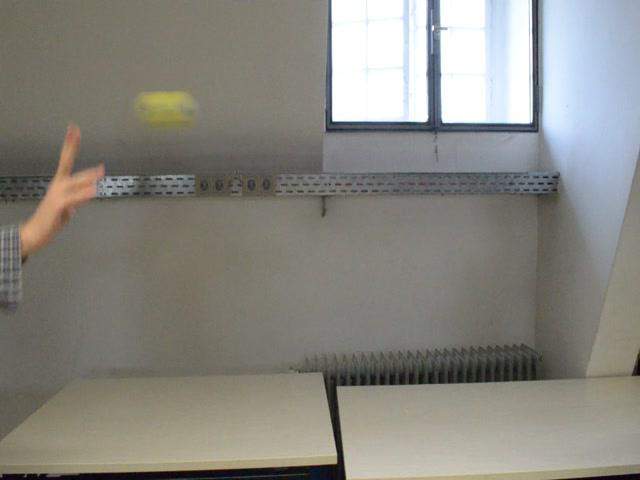}{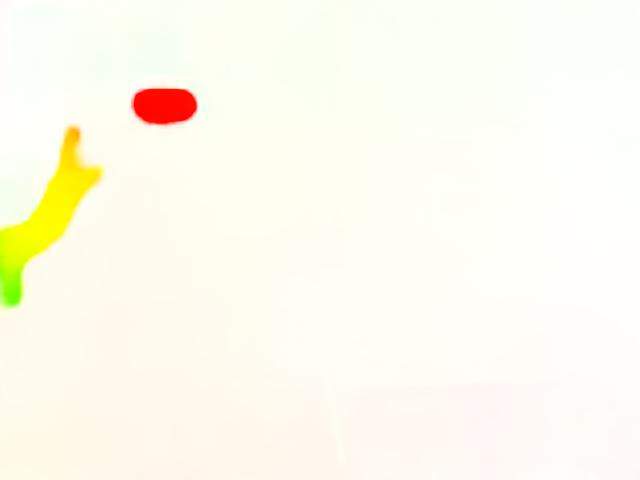}{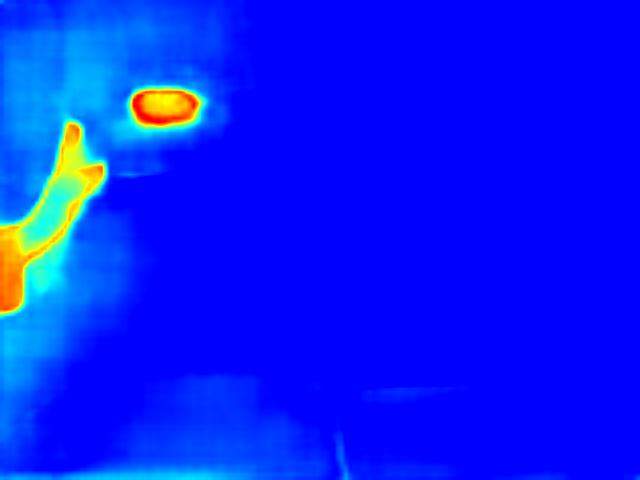}%
\galleryRowReal{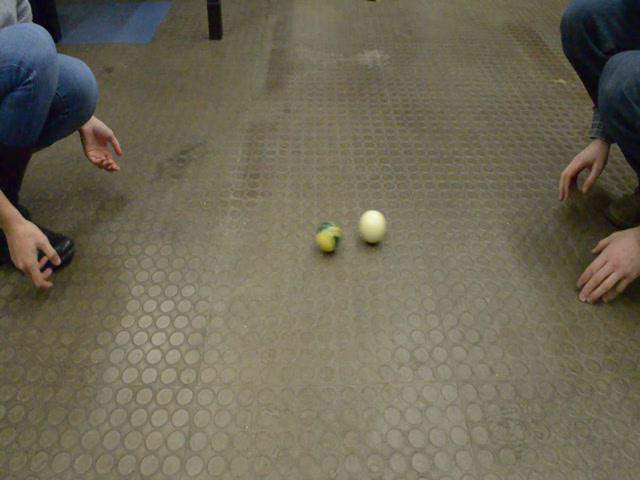}{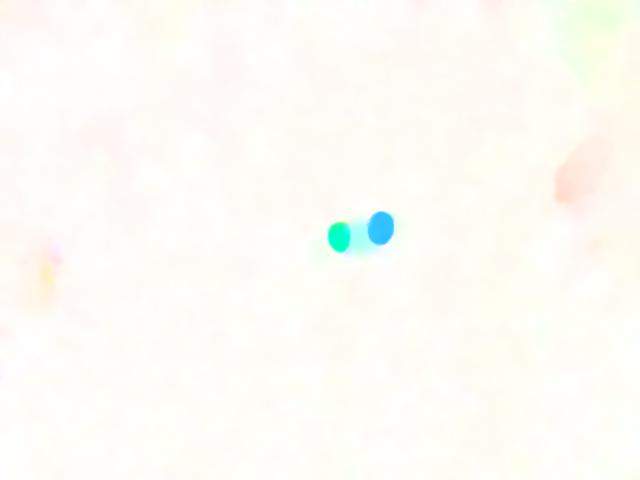}{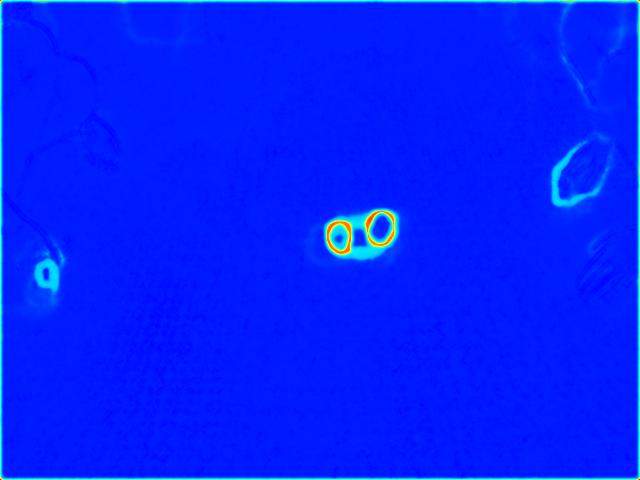}{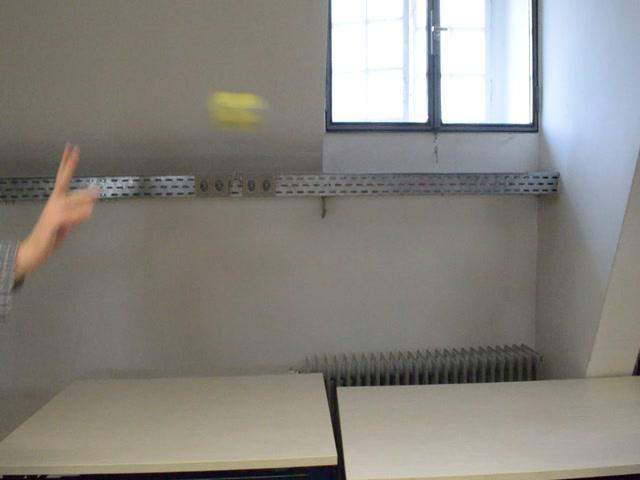}{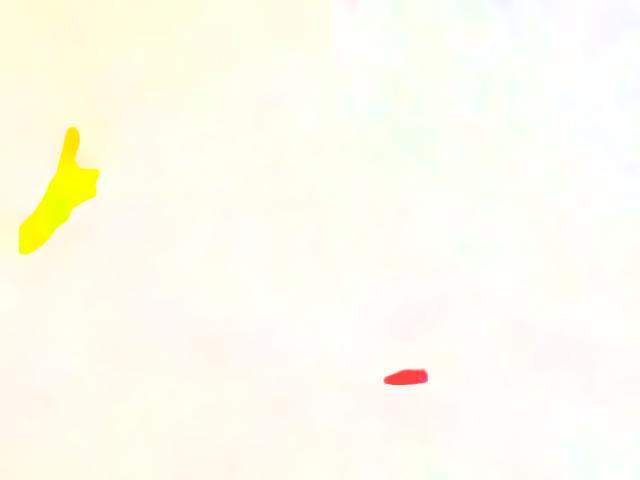}{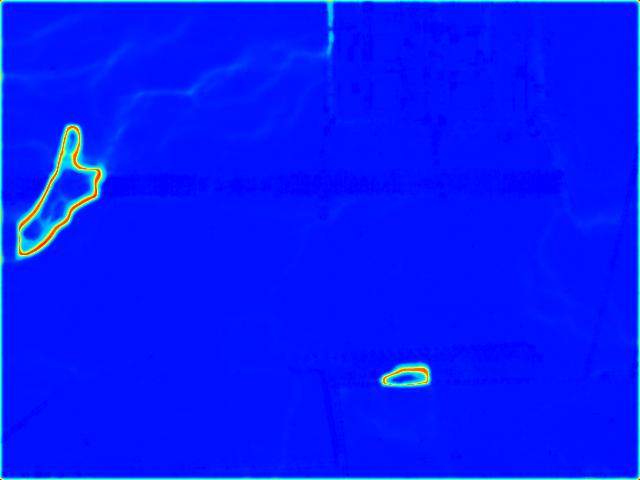}%
    \end{tabular}%
  }%
  \end{center}%
  \caption{Examples from real world data. Examples are arranged in a coarse $4$x$2$ grid, where in each we follow the convention: \textbf{first column:} original image pair, \textbf{second column:} flow predicted by FlowNetH-Pred-Merged and flow predicted by ProbFlow, \textbf{third column:} predicted entropy by FlowNetH-Pred-Merged and predicted entropy by ProbFlow. 
  \textbf{For the full videos of the real world dataset and further comments please see the supplementary video which can also be found on \url{https://youtu.be/HvyovWSo8uE}.}}%
  \label{fig:gallery_real}%
\end{figure*}

\begin{figure*}[t]
  \begin{center}%
    \newcommand{\galleryRowSintel}[6]{ \includegraphics[width=0.2\textwidth]{images/#1} & \includegraphics[width=0.2\textwidth]{images/#2}  & \includegraphics[width=0.2\textwidth]{images/#3} & & \includegraphics[width=0.2\textwidth]{images/#4} & \includegraphics[width=0.2\textwidth]{images/#5}  & \includegraphics[width=0.2\textwidth]{images/#6} \tabularnewline
}
  \resizebox{\linewidth}{!}{%
    \setlength{\tabcolsep}{0.7pt}%
    \begin{tabular}{cccp{2.5mm}ccc}%
\galleryRowSintel{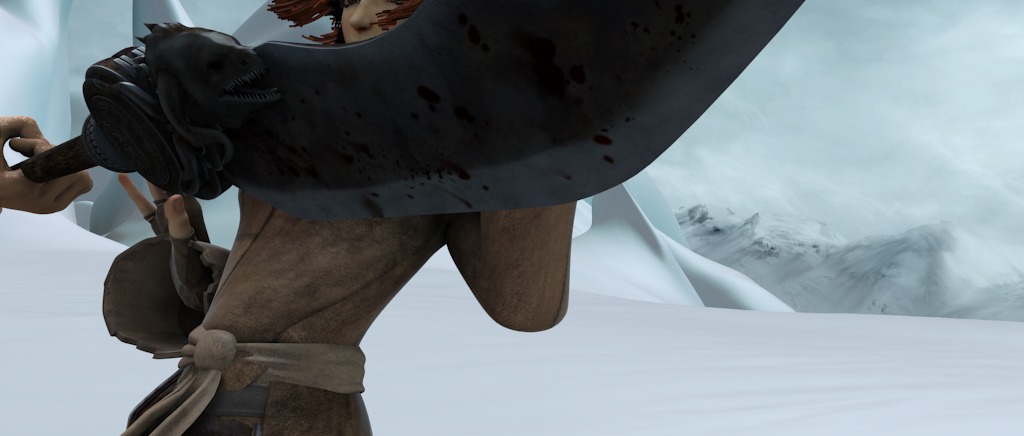}{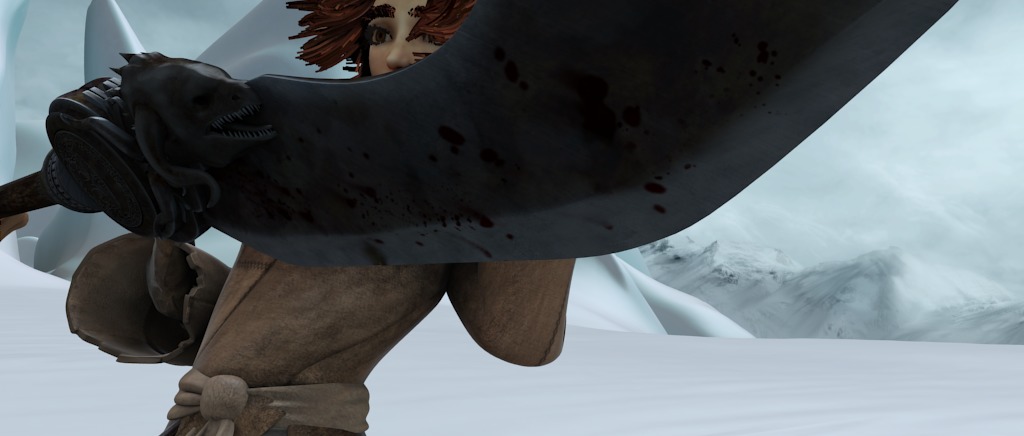}{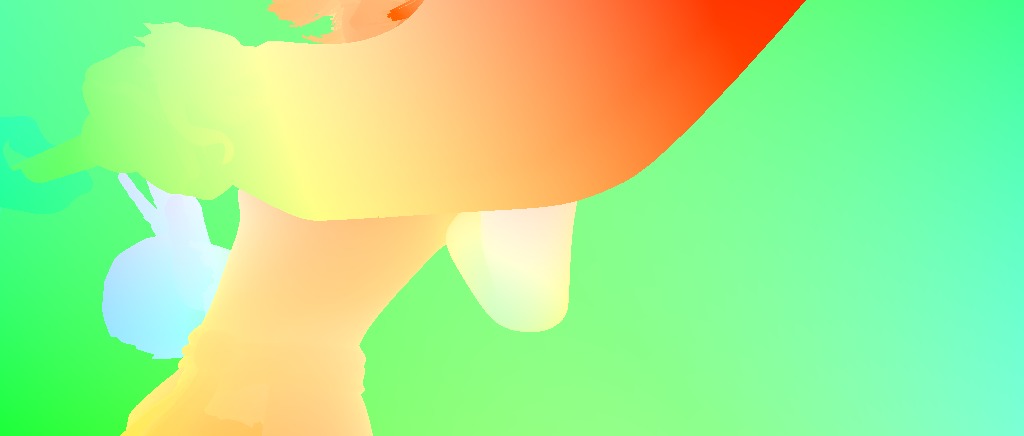}{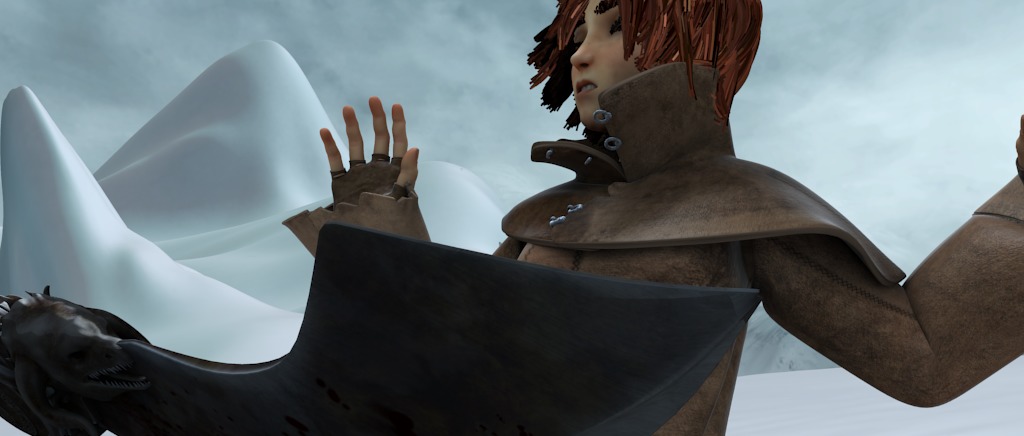}{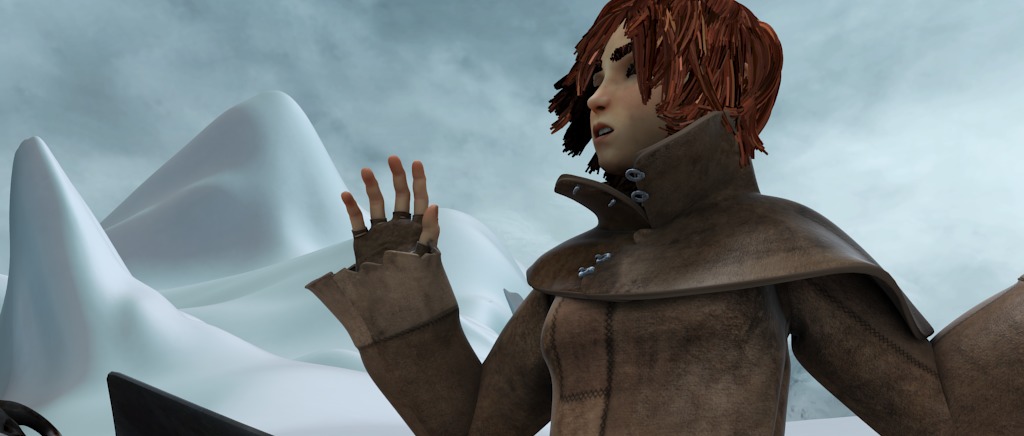}{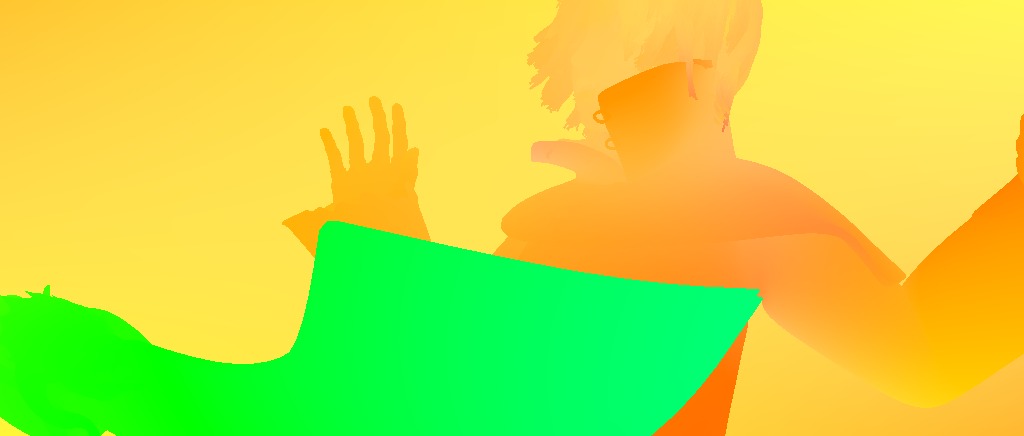}%
\galleryRowSintel{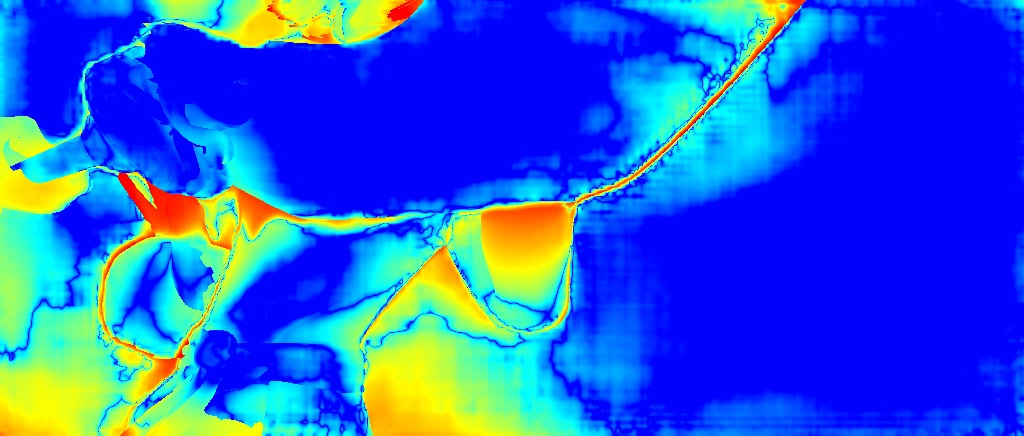}{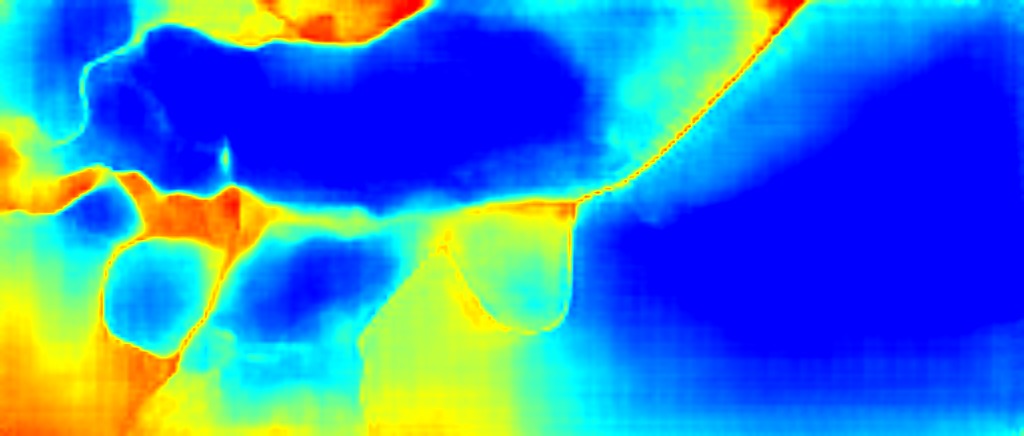}{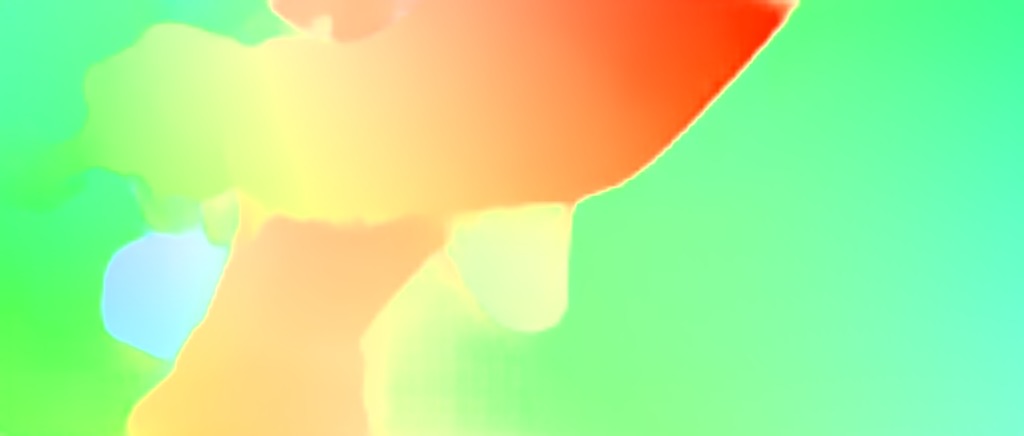}{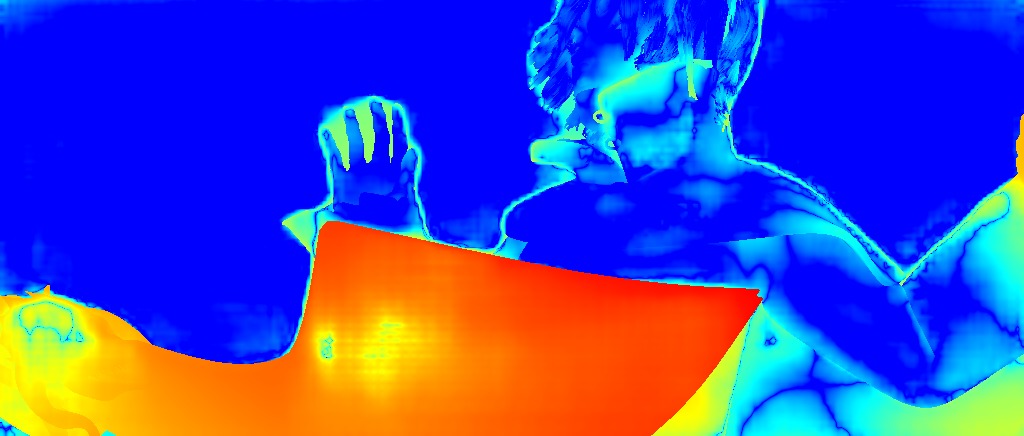}{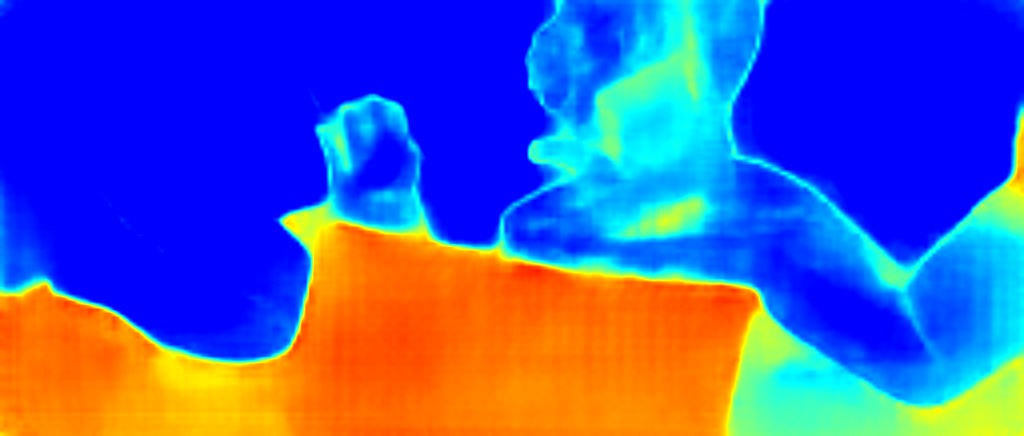}{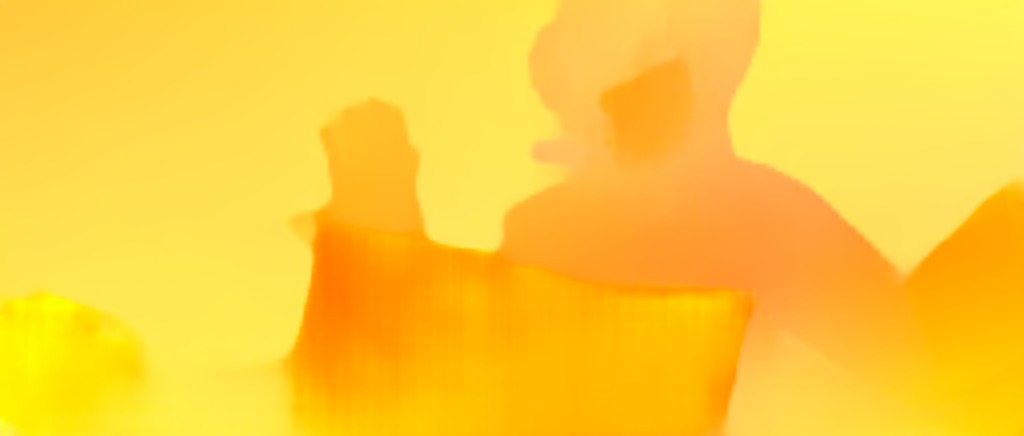}\vspace{3mm}%
\galleryRowSintel{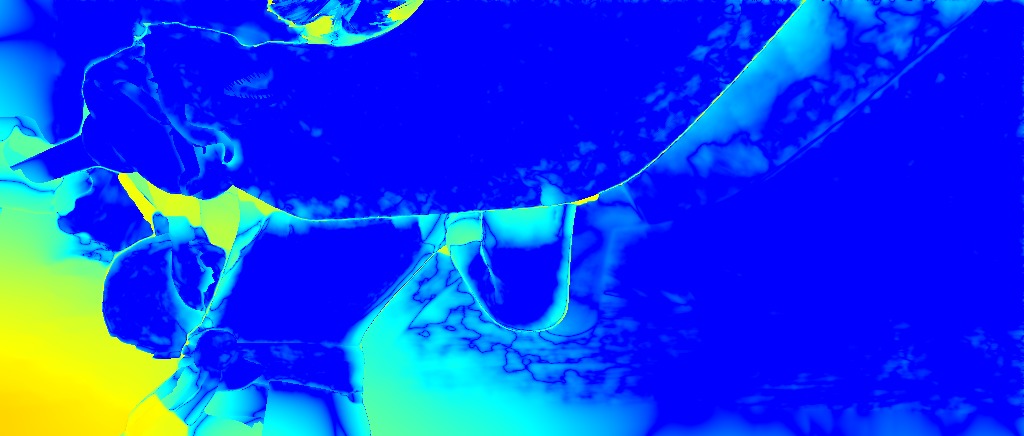}{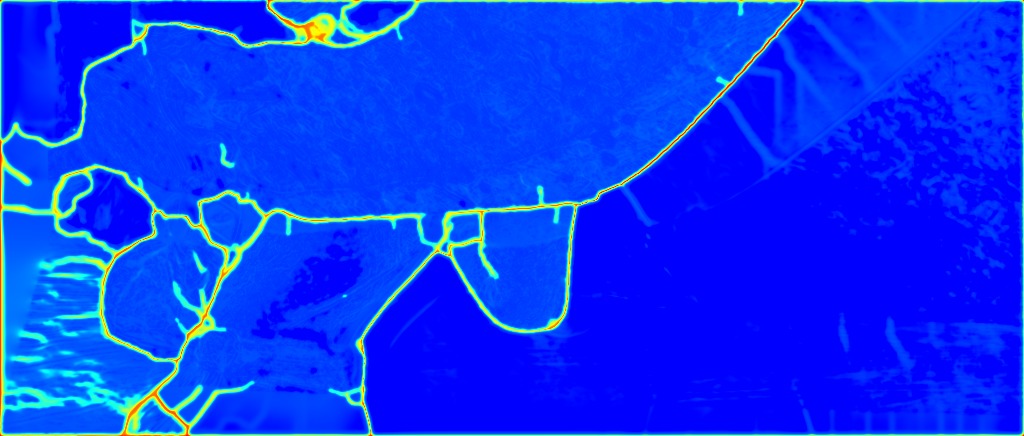}{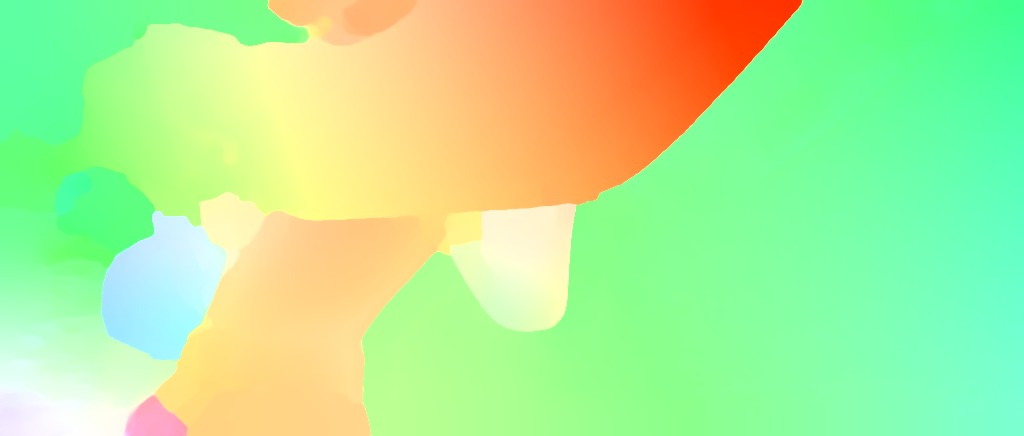}{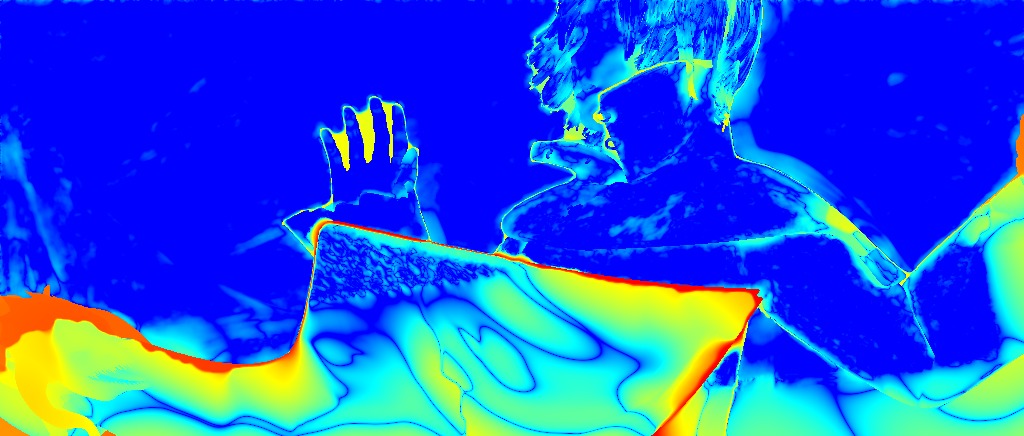}{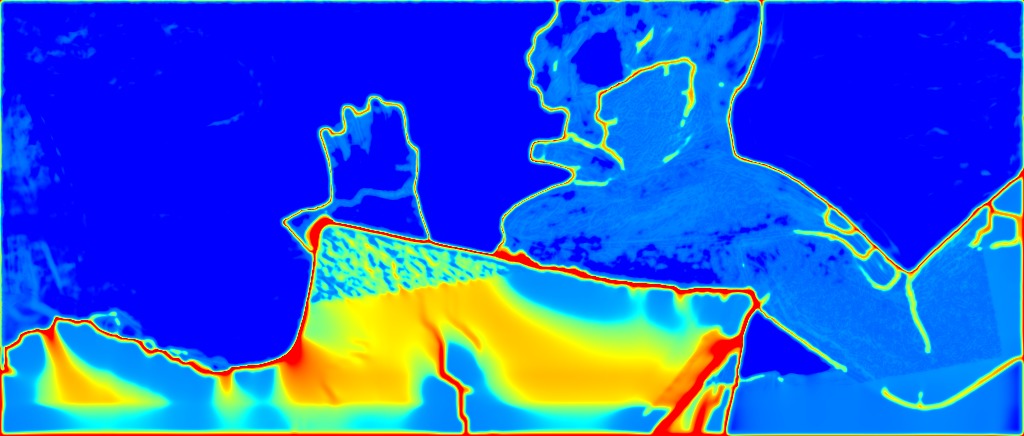}{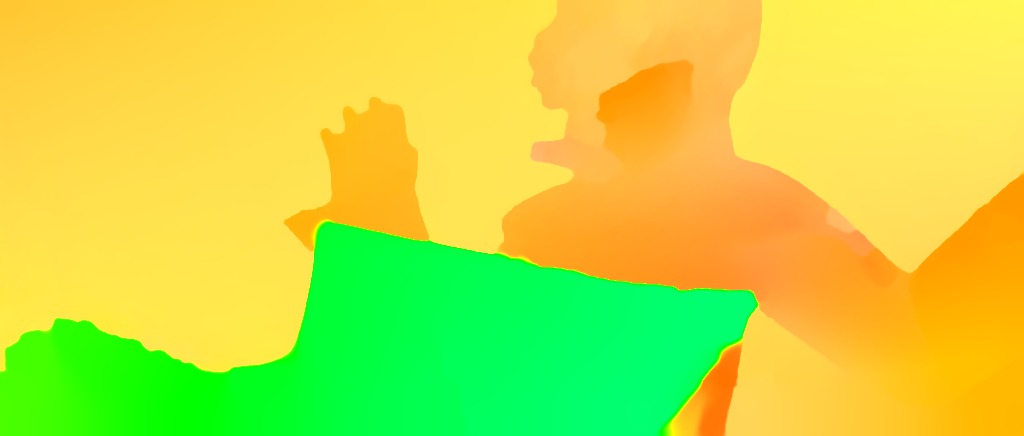}%
\galleryRowSintel{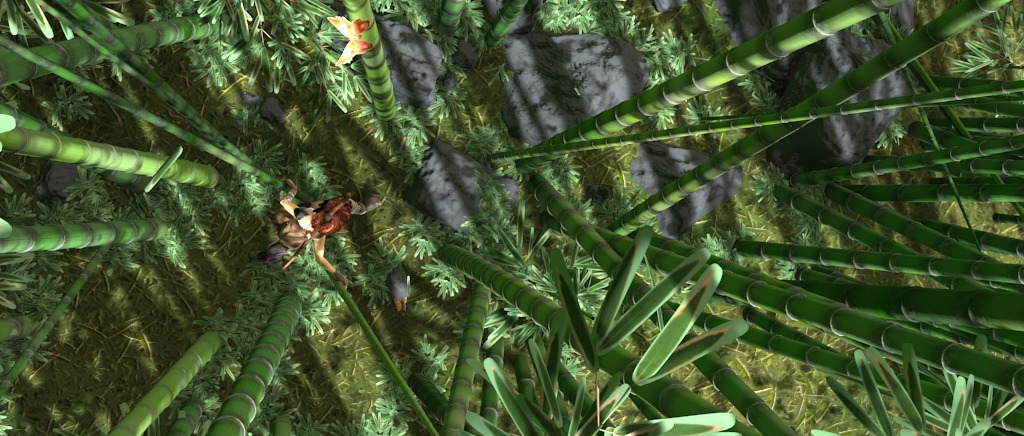}{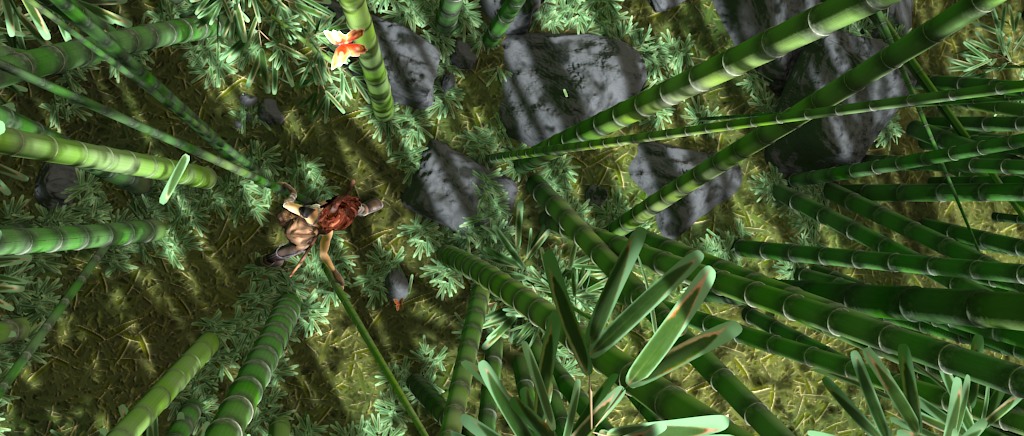}{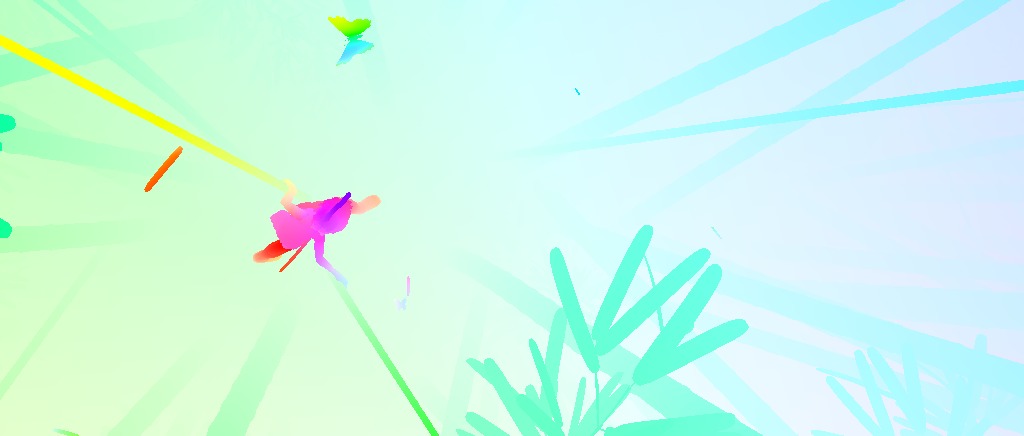}{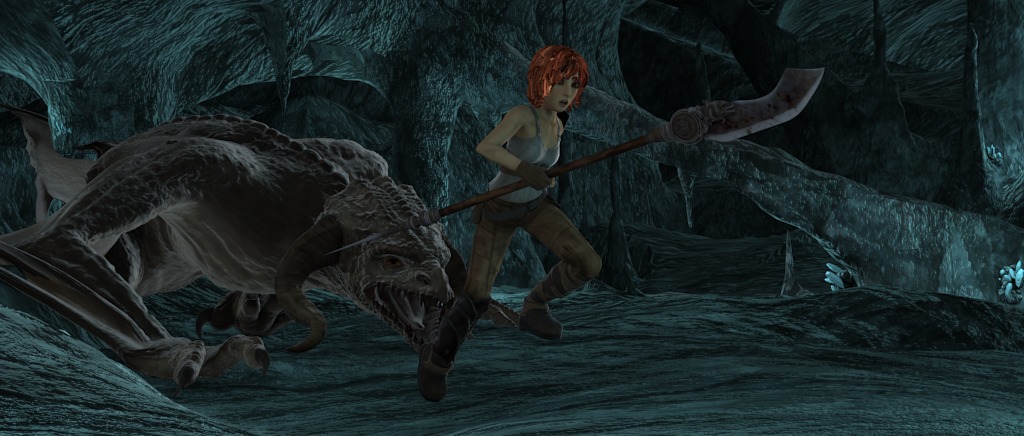}{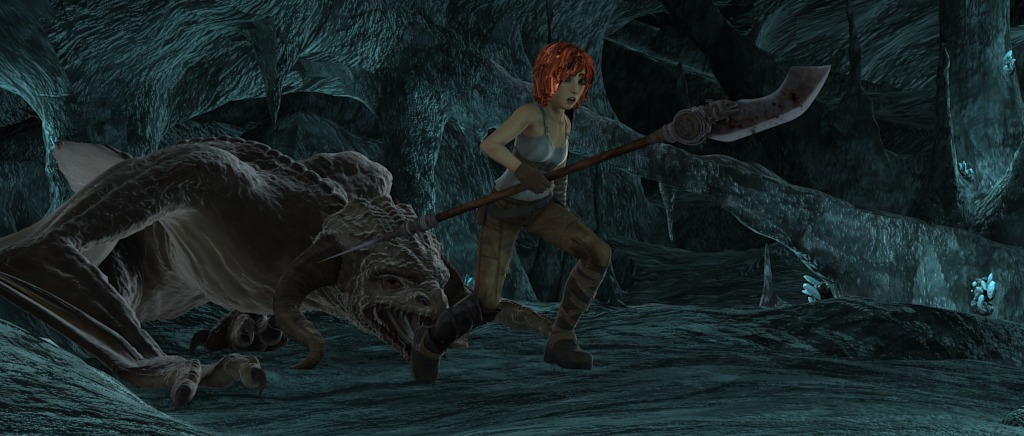}{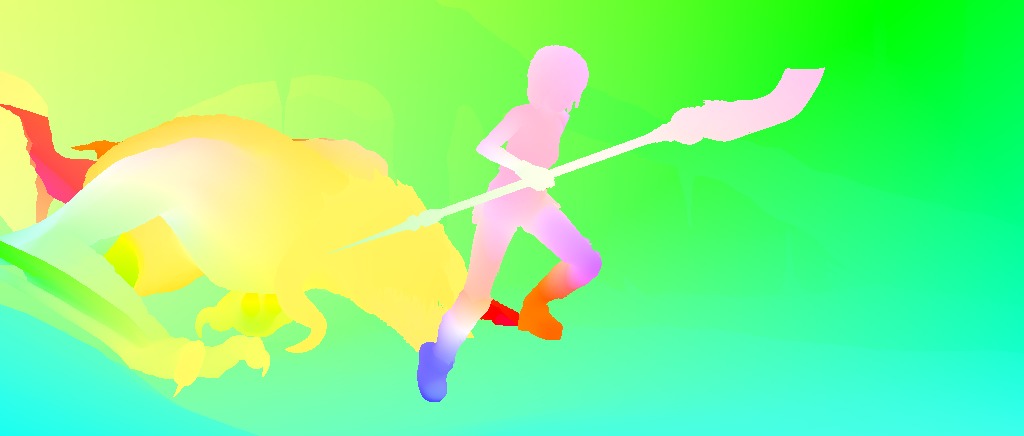}%
\galleryRowSintel{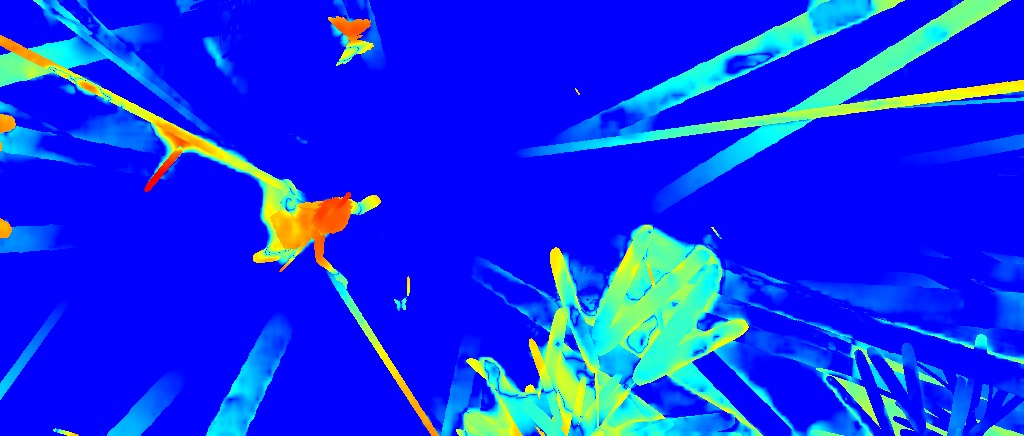}{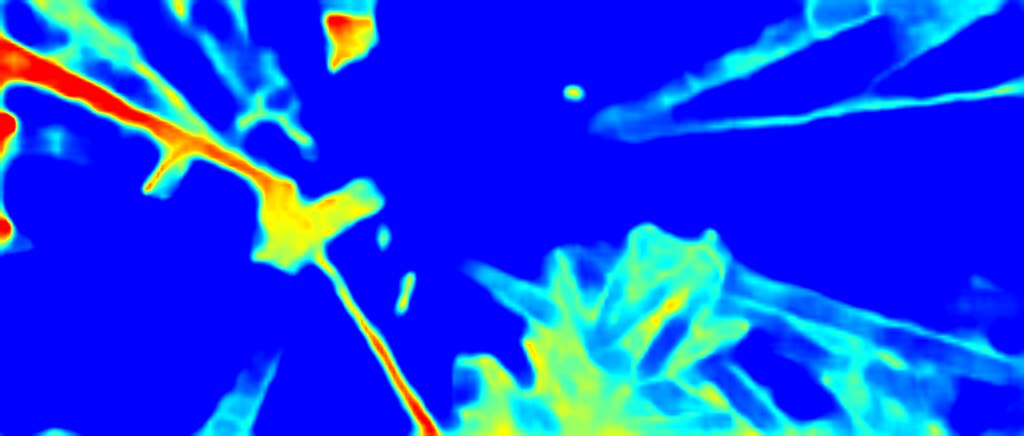}{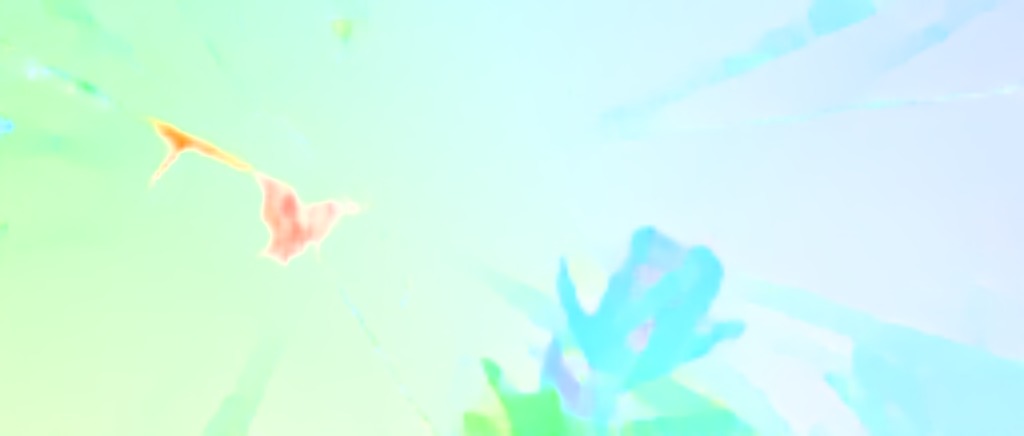}{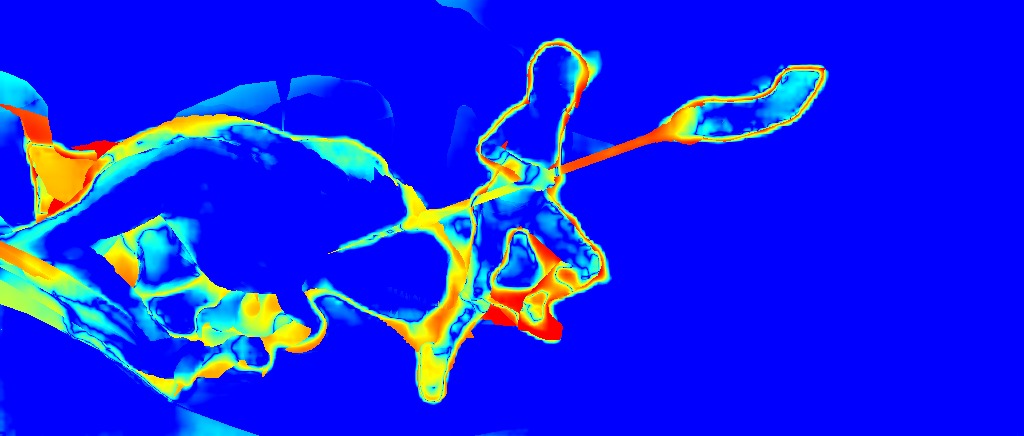}{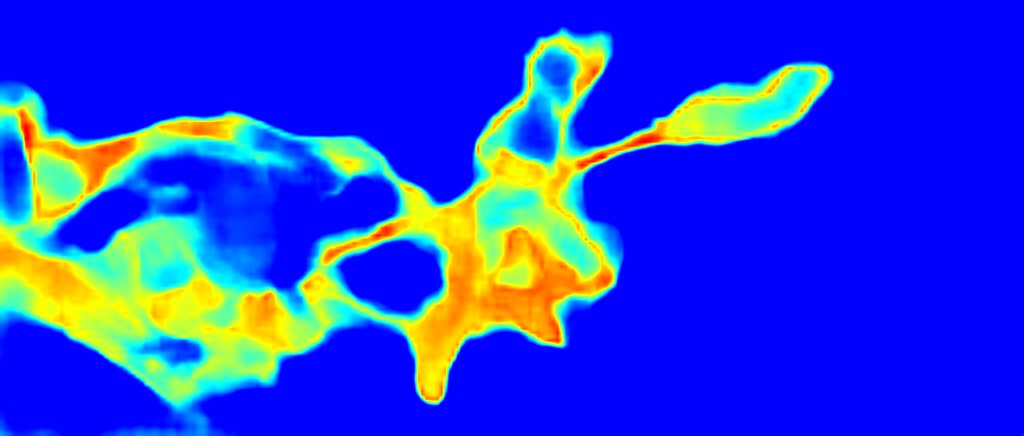}{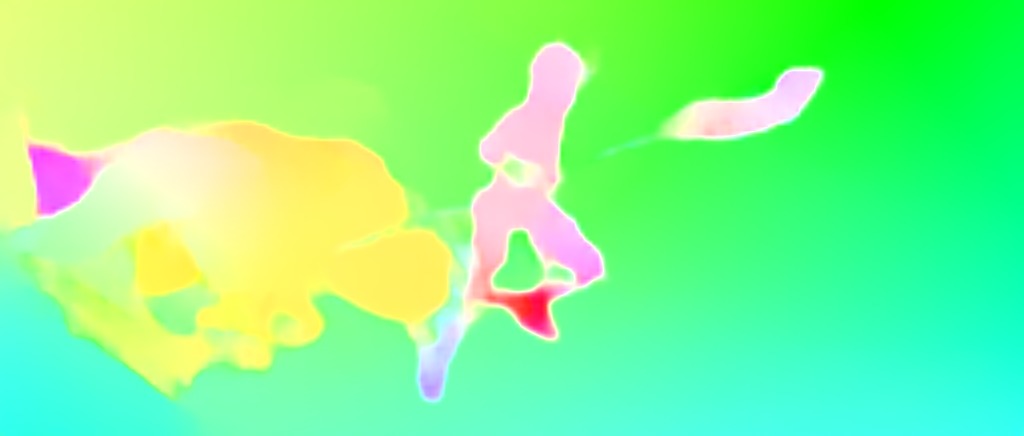}%
\galleryRowSintel{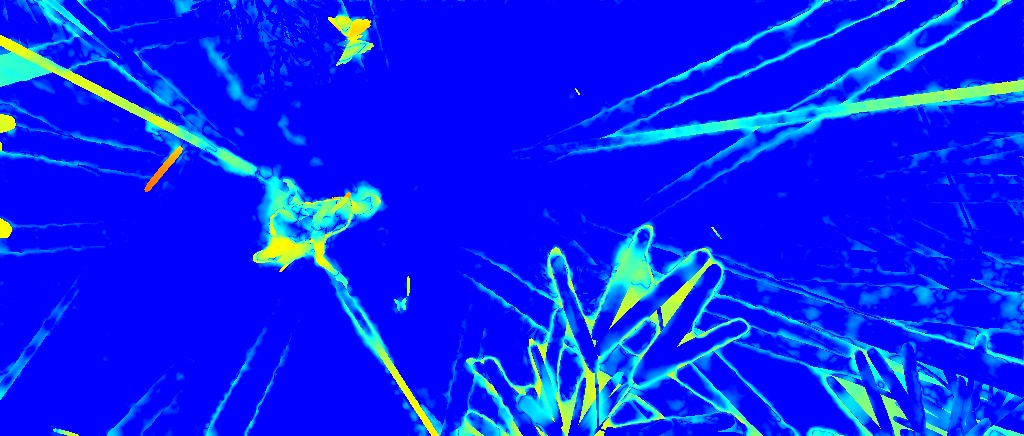}{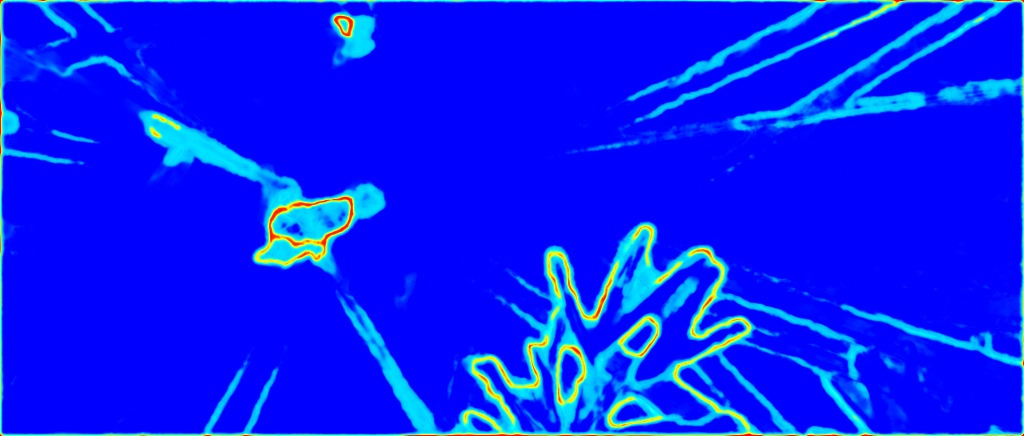}{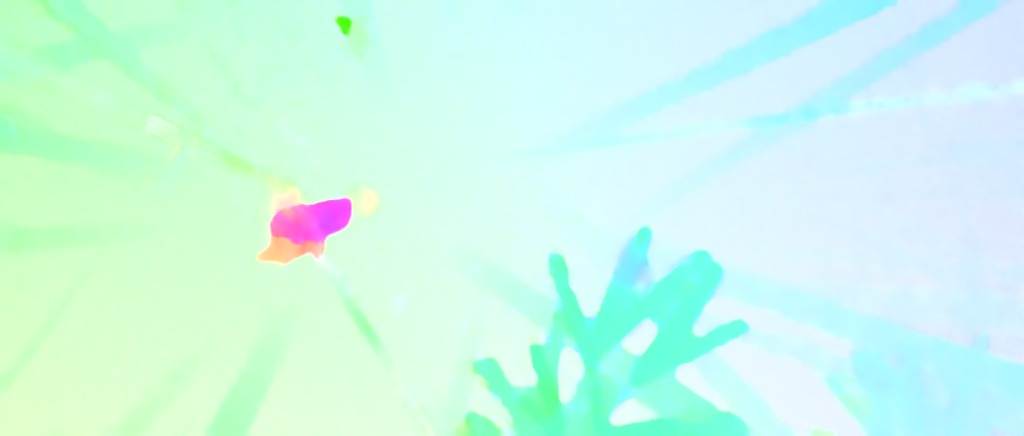}{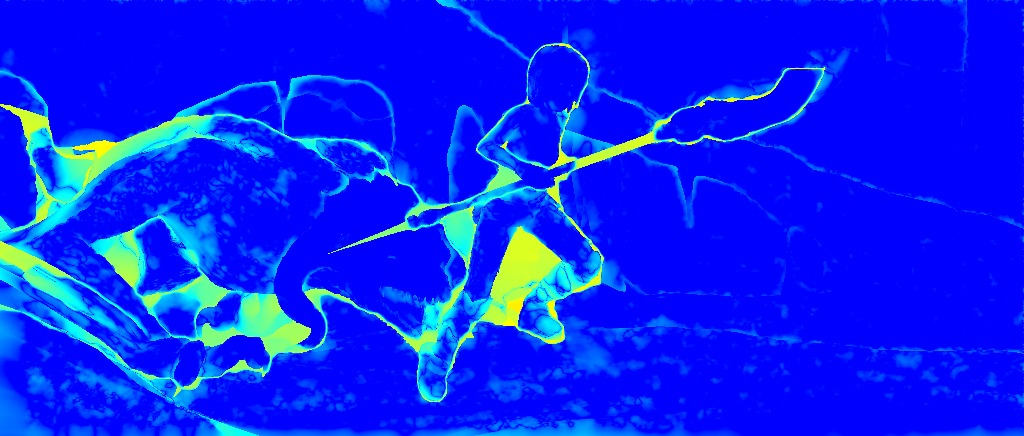}{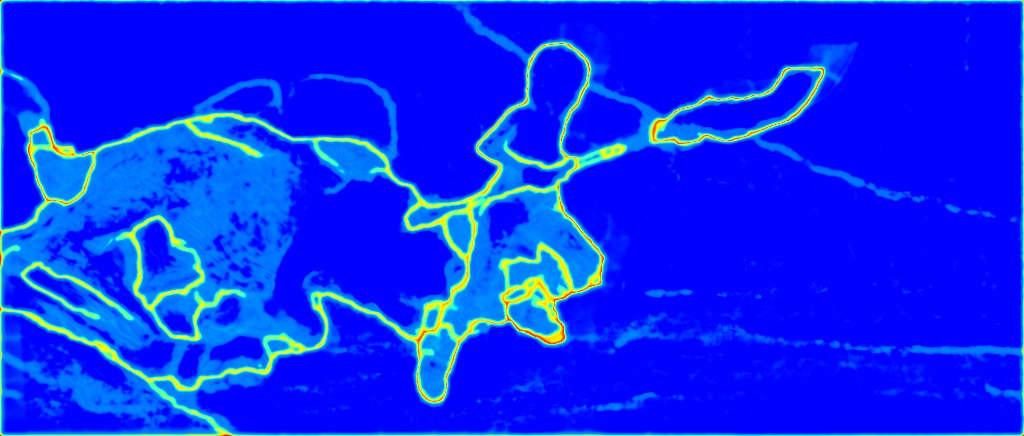}{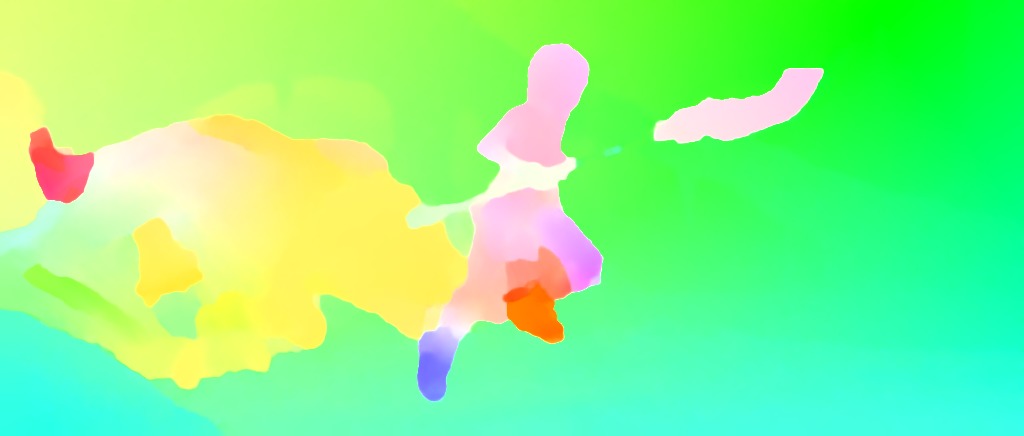}%

    \end{tabular}%
    
  }%
  \end{center}%
  \caption{Four examples from the Sintel train clean dataset for qualitative comparison between FlowNetH-Pred-Merged and ProbFlow. For each example: \textbf{first row} shows the original image sequence followed by its ground truth flow field. \textbf{Second row} shows FlowNetH-Pred-Merged results: oracle entropy (representing the optimal uncertainty), predicted entropy and predicted flow. Similar to the second row, the \textbf{third row} shows the results for ProbFlow. While our method is predicting uncertainties on large areas, ProbFlow shows uncertainties mainly only on the motion or image boundaries and sometimes shows overconfidence in the regions where its prediction is wrong. This is visible e.g. in the upper left example for the lower left corner, where the estimation is wrong, but the uncertainty is low.
  }%
  \label{fig:gallery_sintel2}%
\end{figure*}

\begin{figure*}[t]
  \begin{center}%
    \newcommand{\galleryRowKitti}[6]{ \includegraphics[width=0.2\textwidth]{images/#1} & \includegraphics[width=0.2\textwidth]{images/#2}  & \includegraphics[width=0.2\textwidth]{images/#3} & & \includegraphics[width=0.2\textwidth]{images/#4} & \includegraphics[width=0.2\textwidth]{images/#5}  & \includegraphics[width=0.2\textwidth]{images/#6} \tabularnewline
}
  \resizebox{\linewidth}{!}{%
    \setlength{\tabcolsep}{0.7pt}%
    \begin{tabular}{cccp{2.5mm}ccc}%
\galleryRowKitti{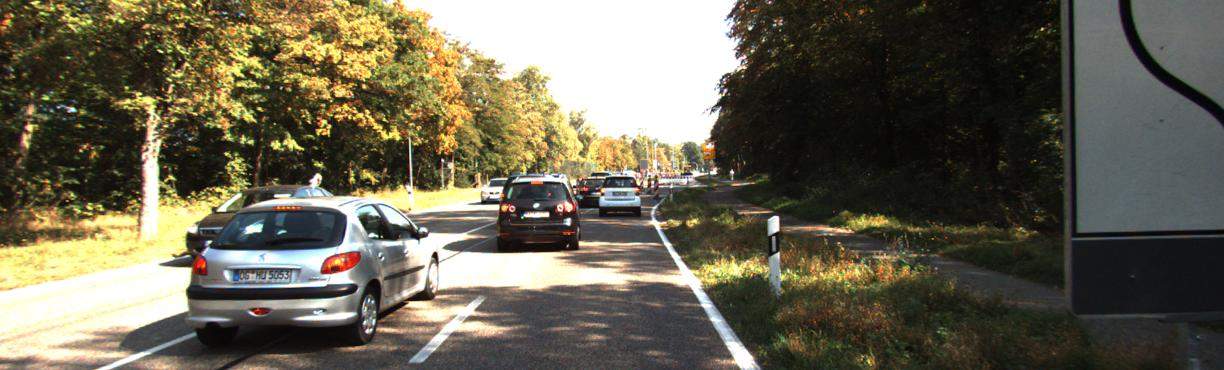}{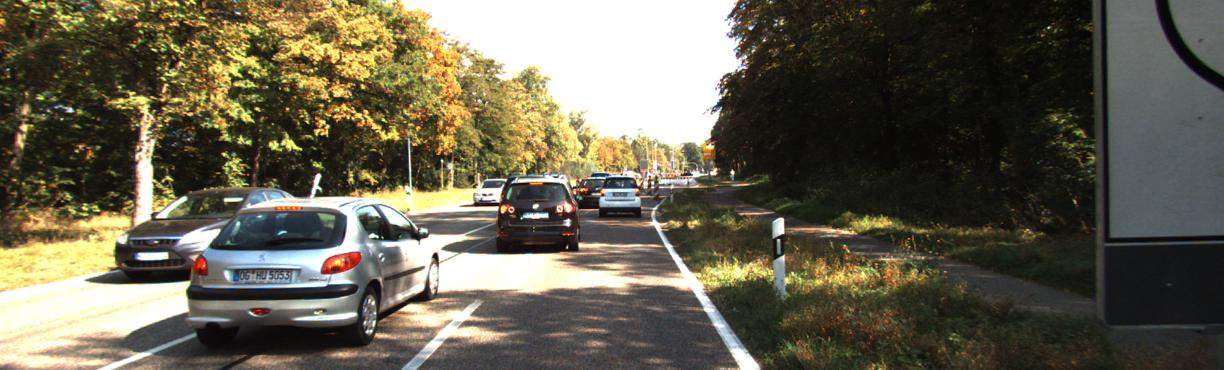}{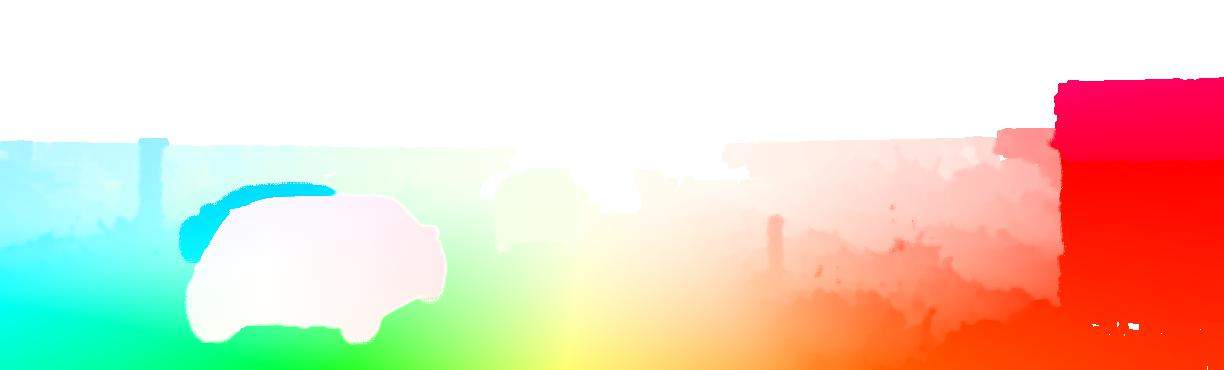}{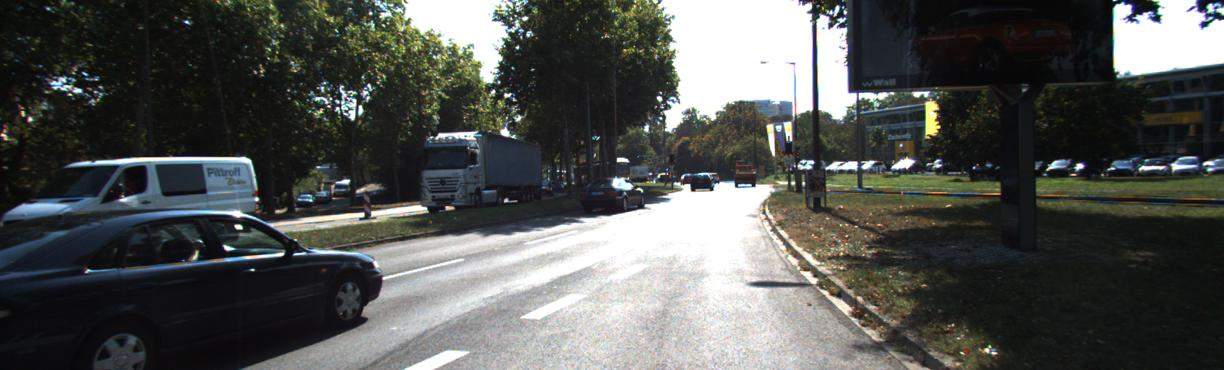}{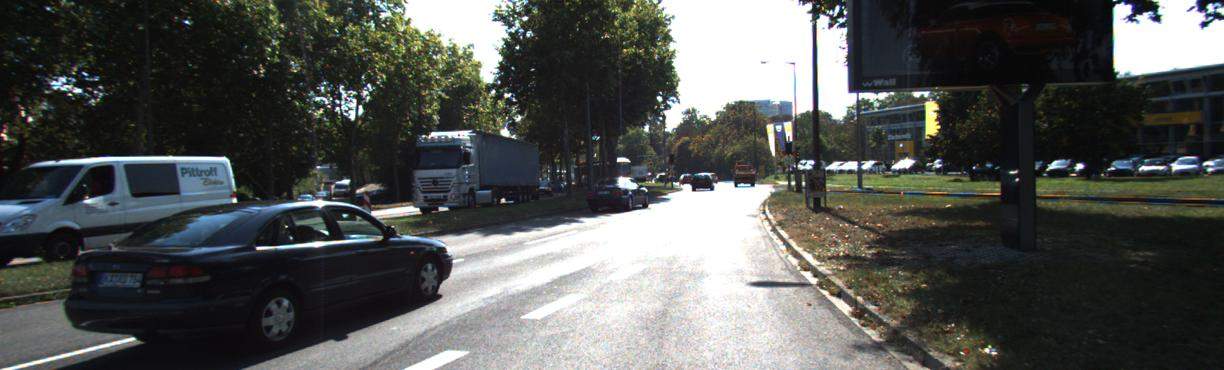}{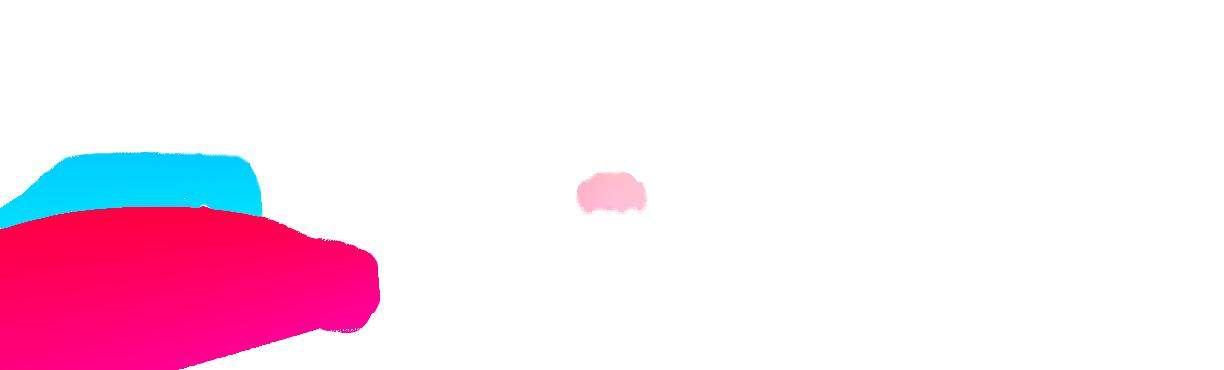}%
\galleryRowKitti{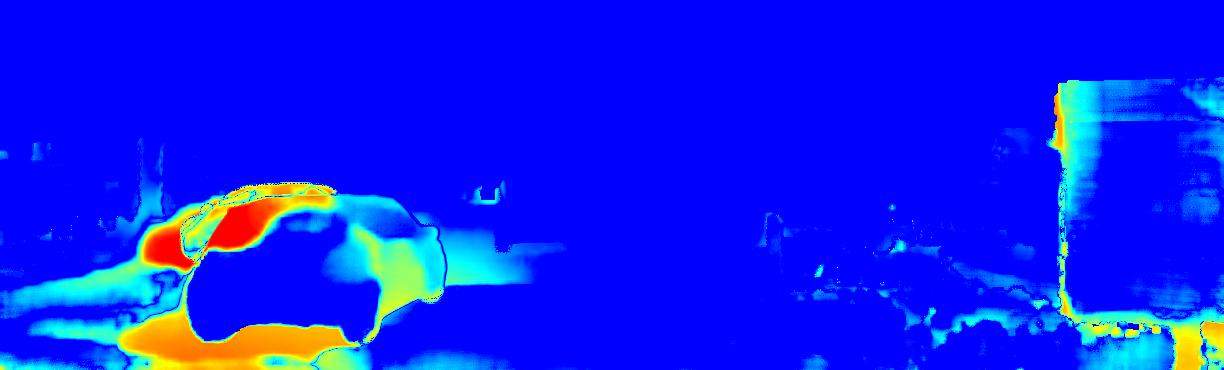}{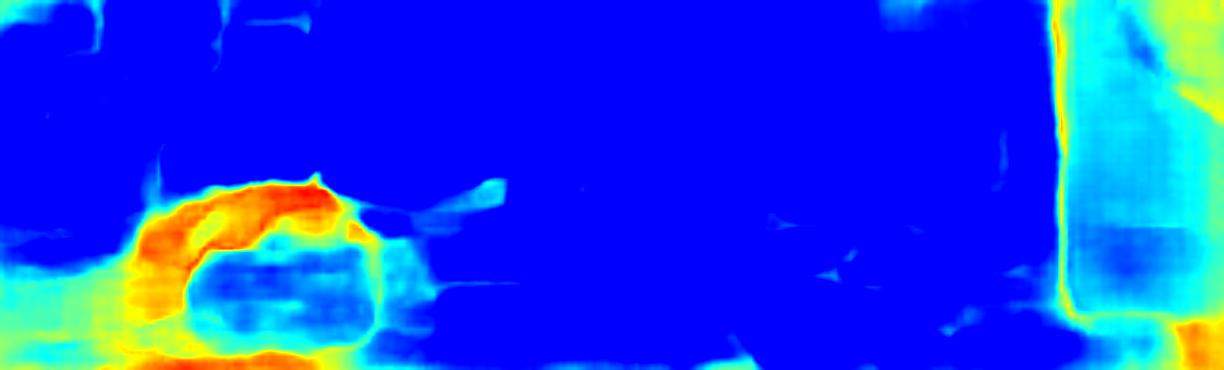}{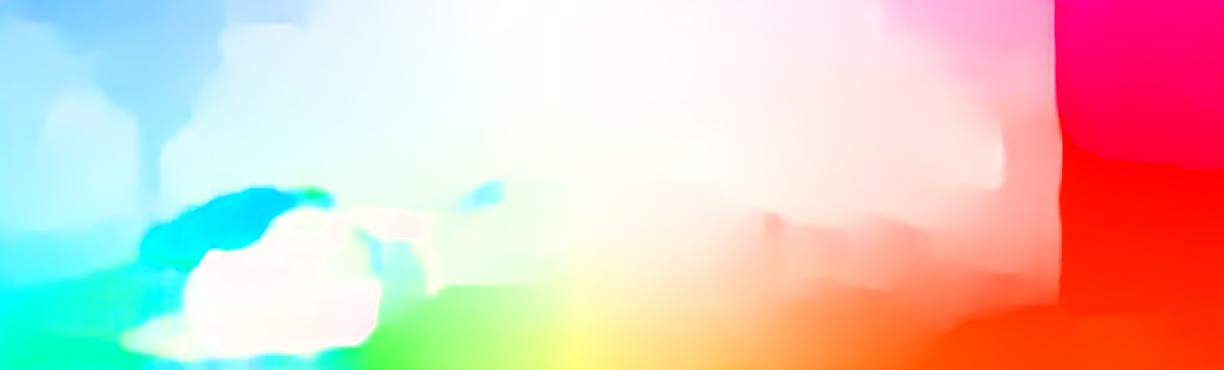}{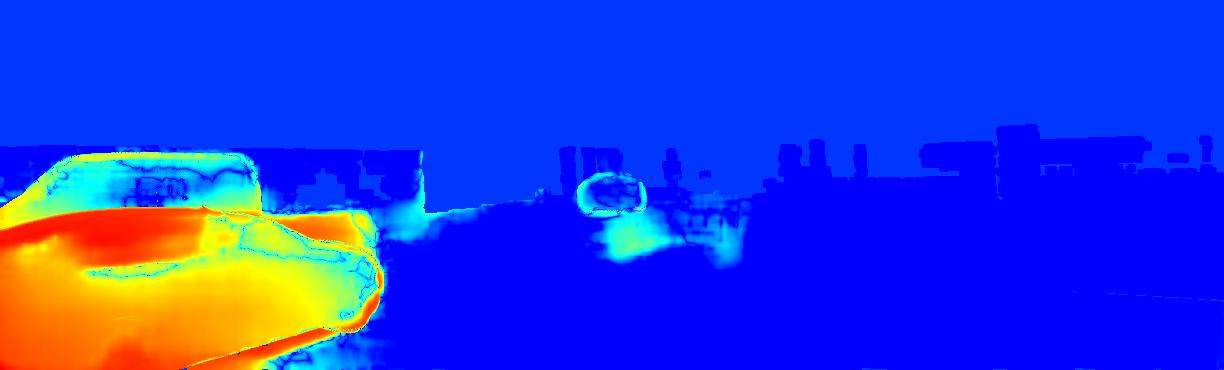}{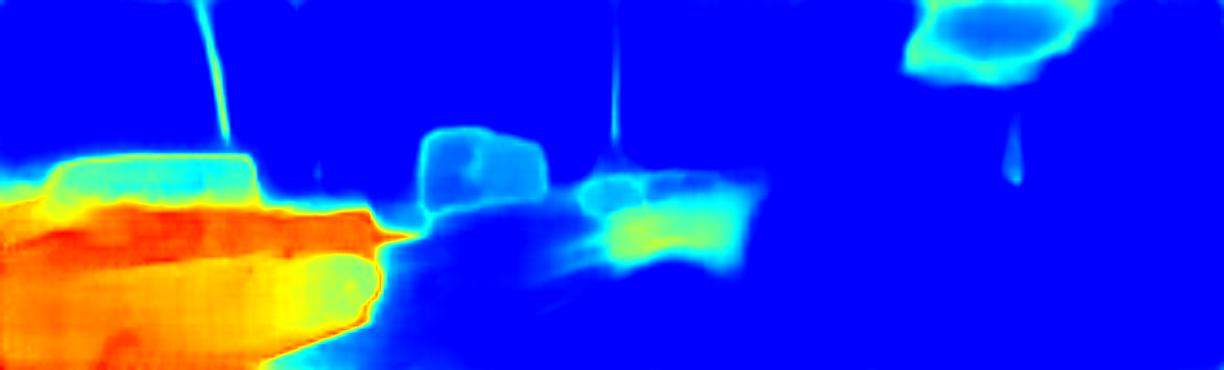}{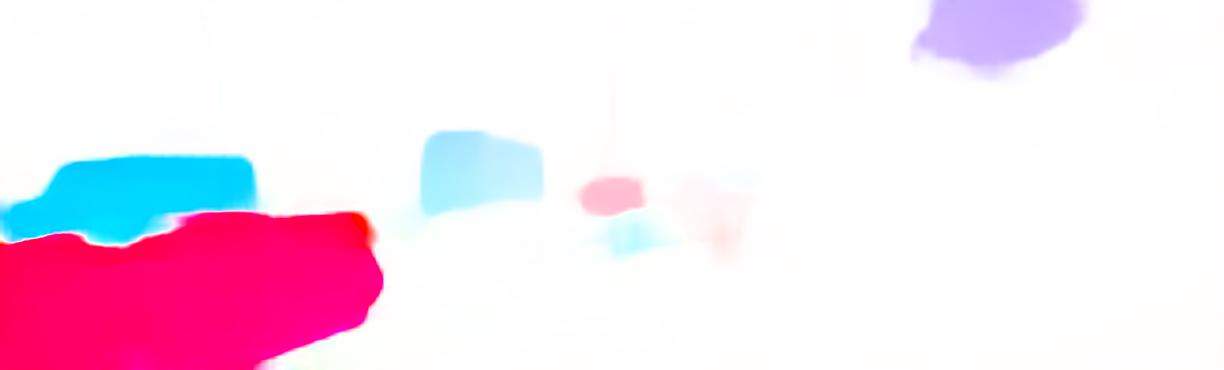}\vspace{3mm}%
\galleryRowKitti{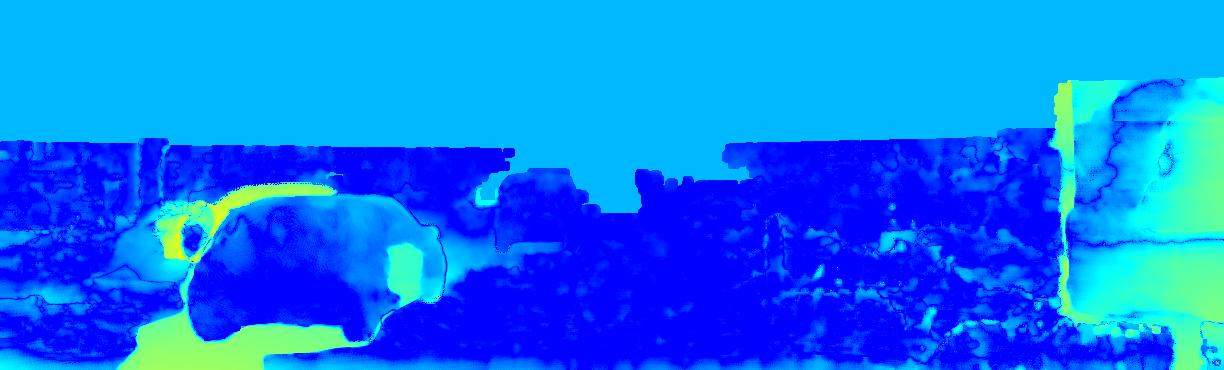}{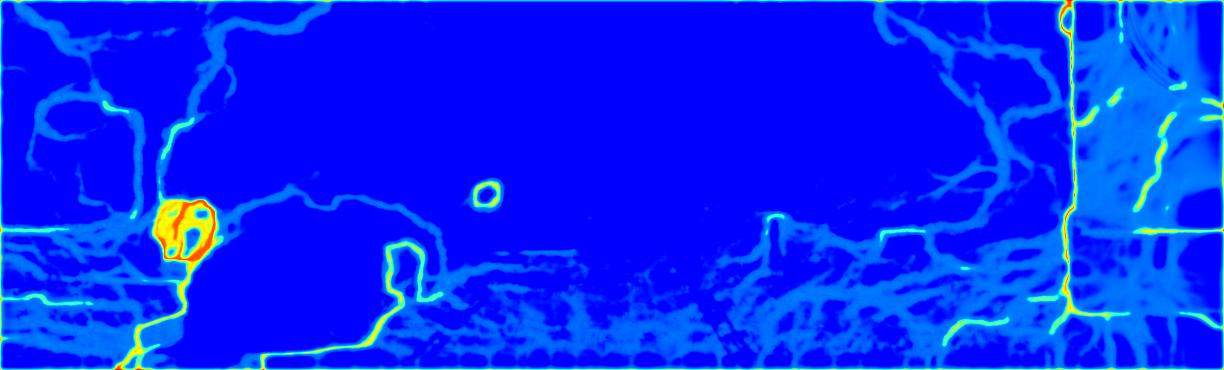}{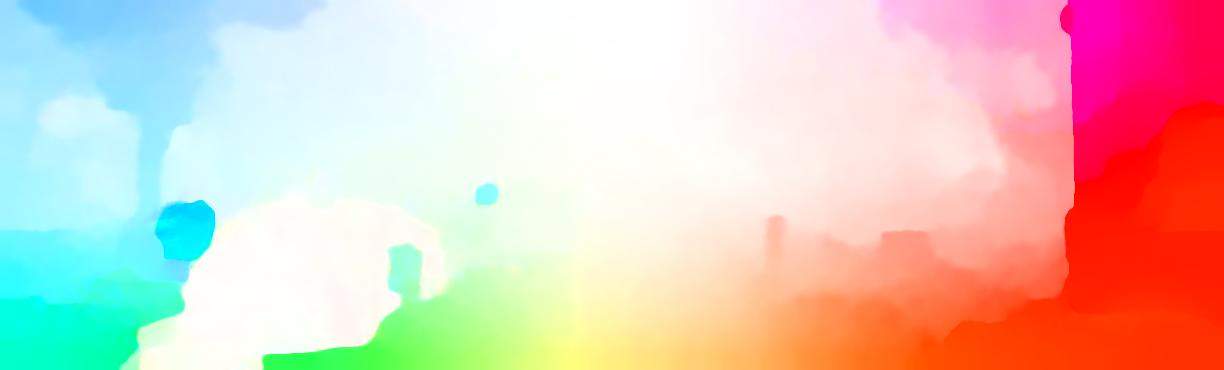}{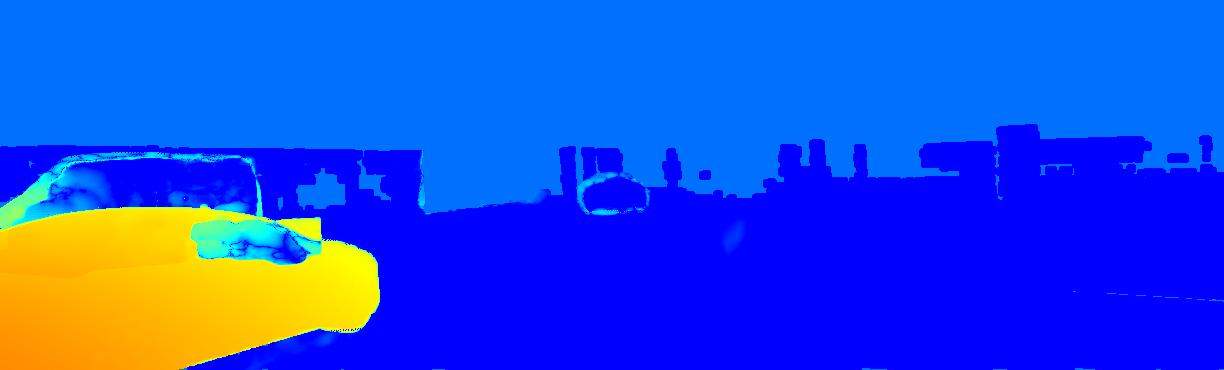}{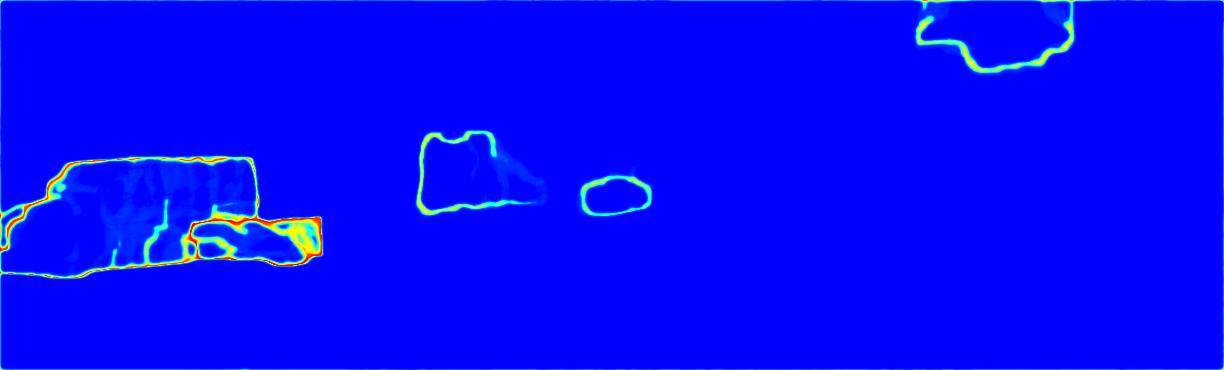}{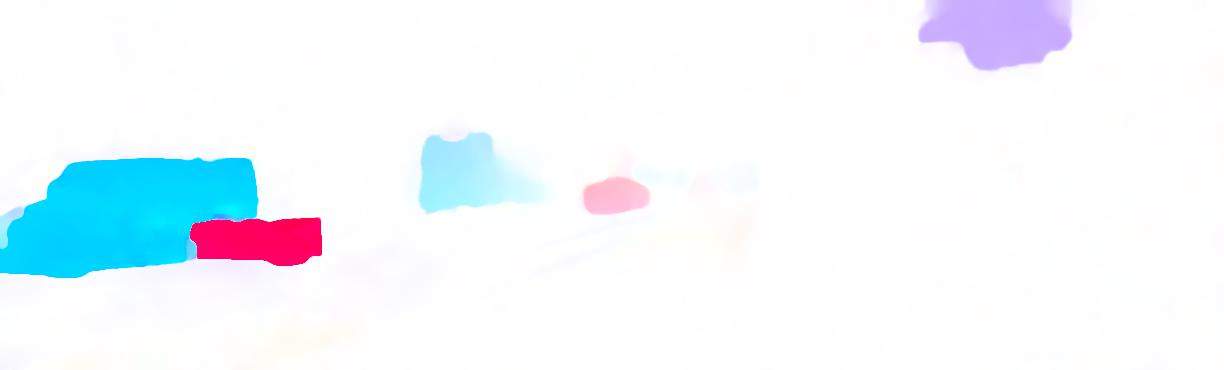}%
\galleryRowKitti{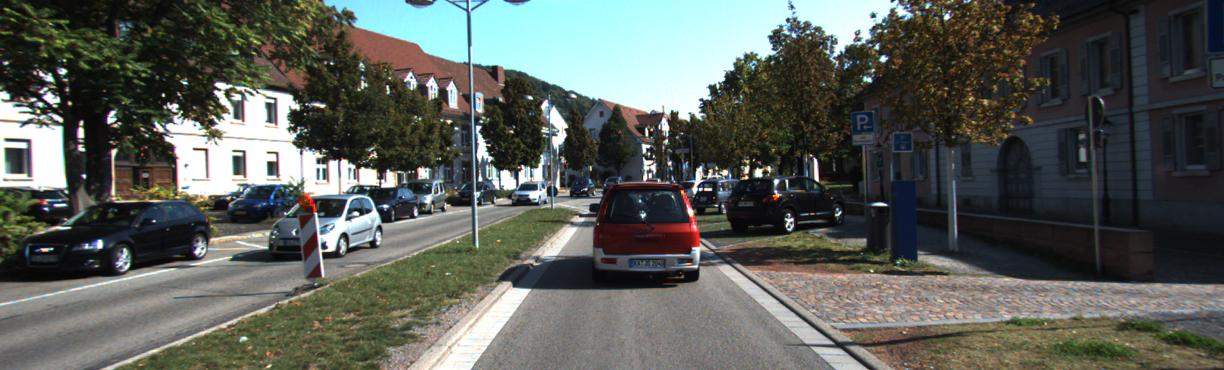}{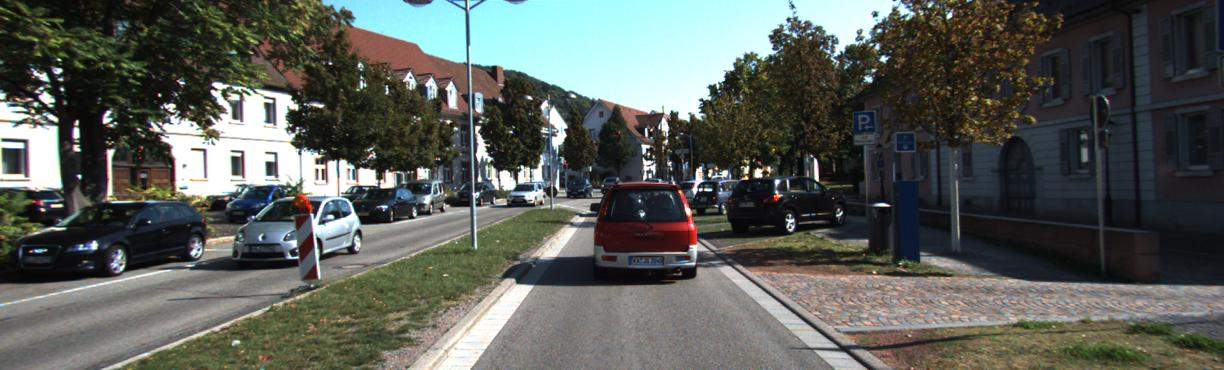}{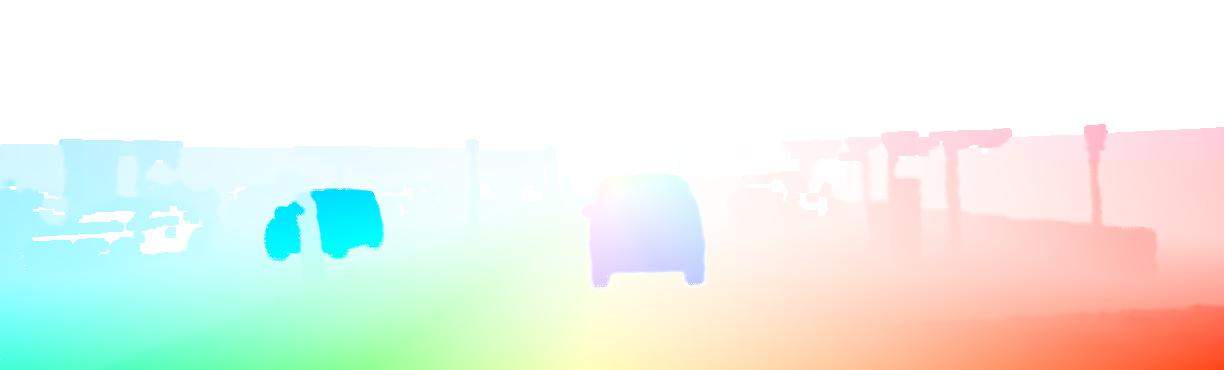}{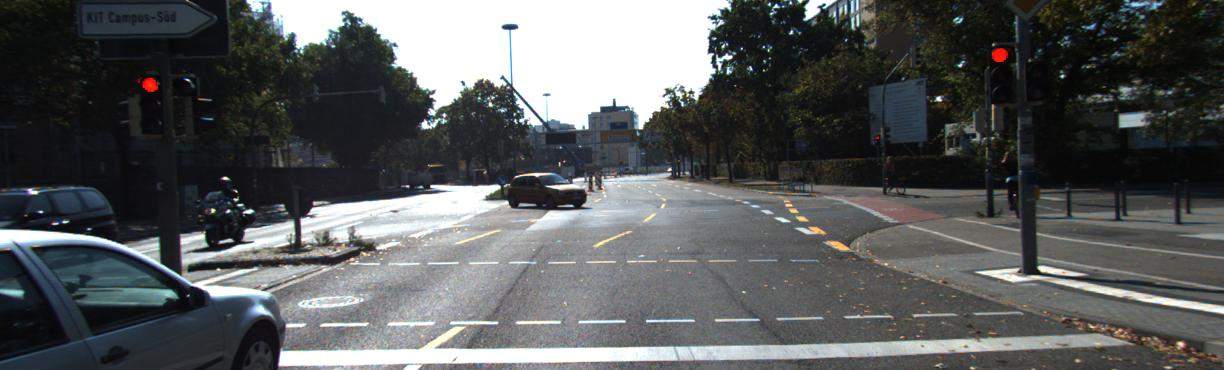}{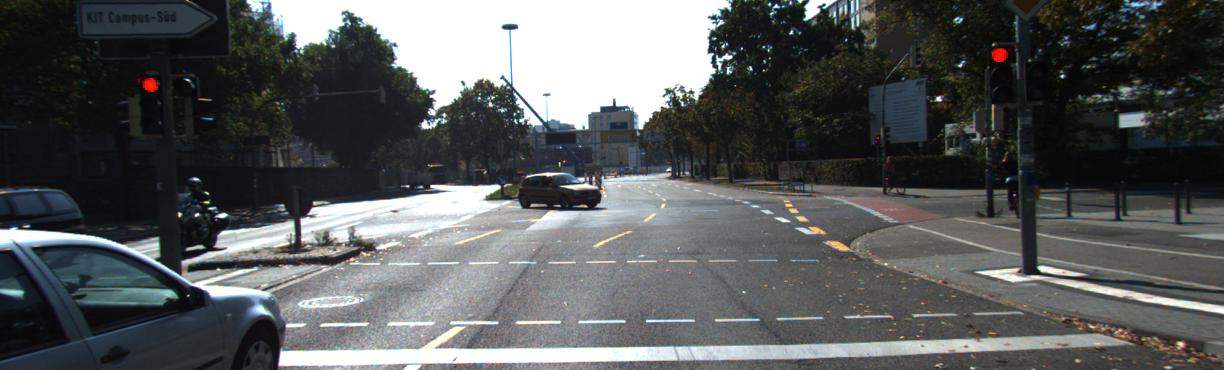}{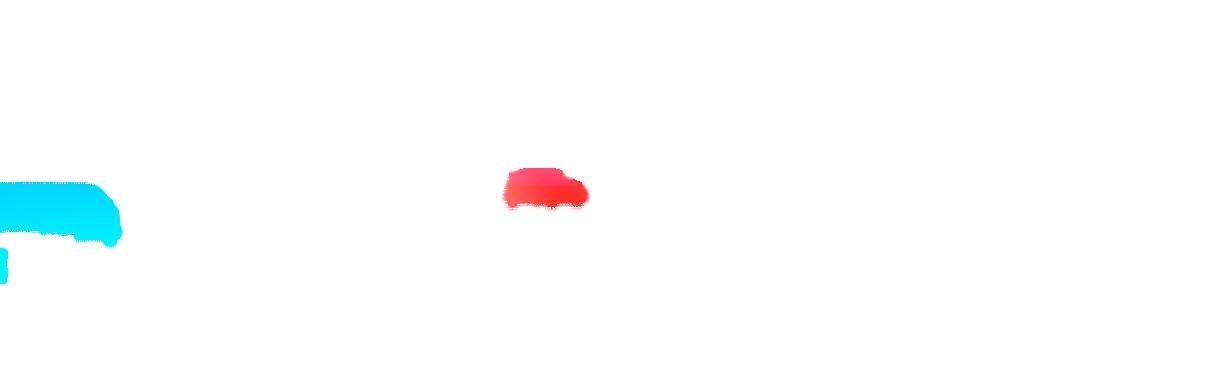}%
\galleryRowKitti{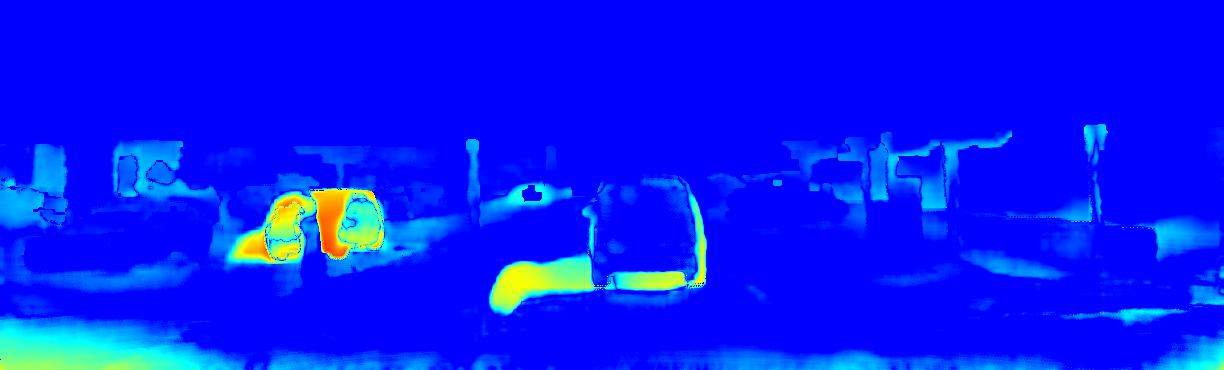}{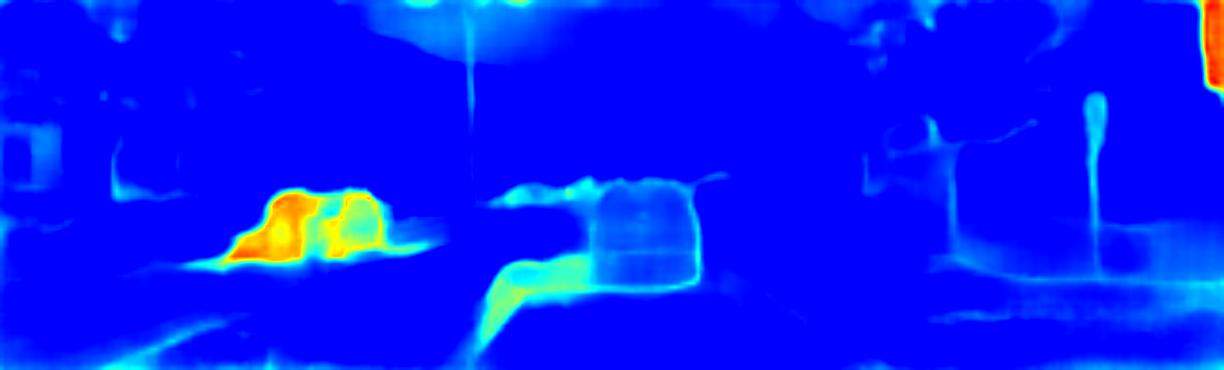}{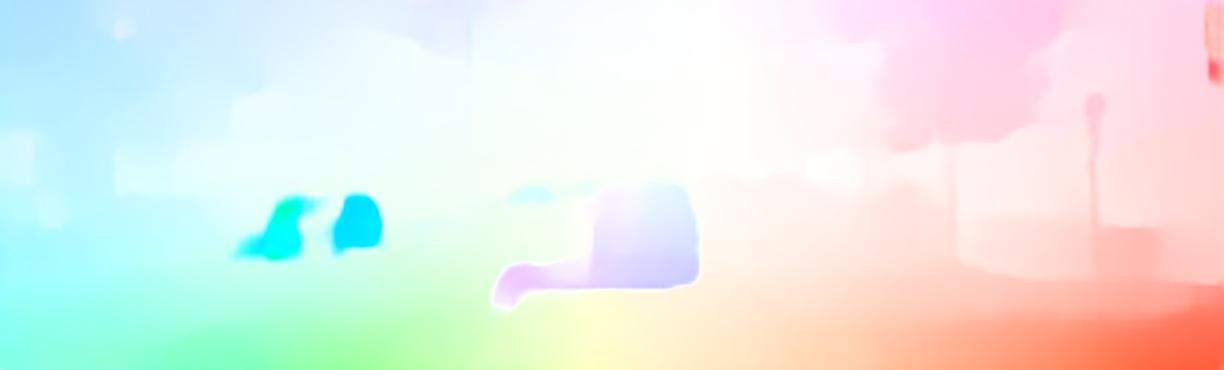}{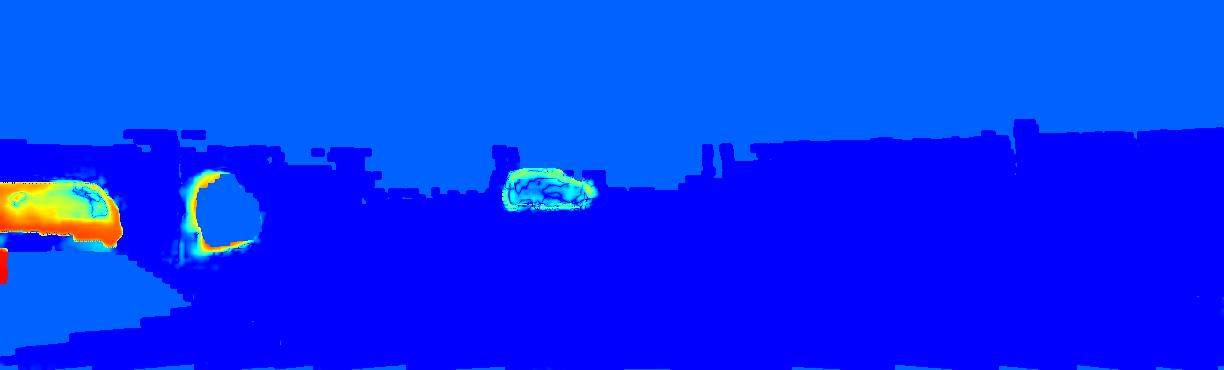}{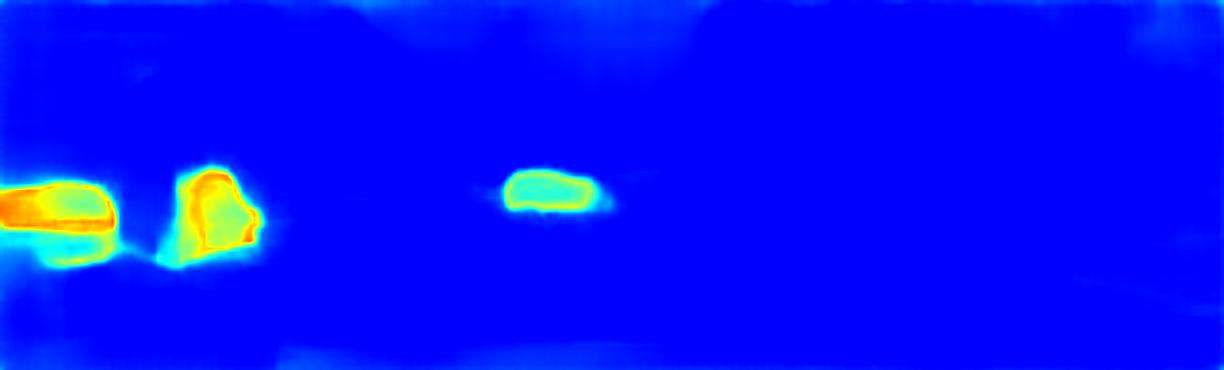}{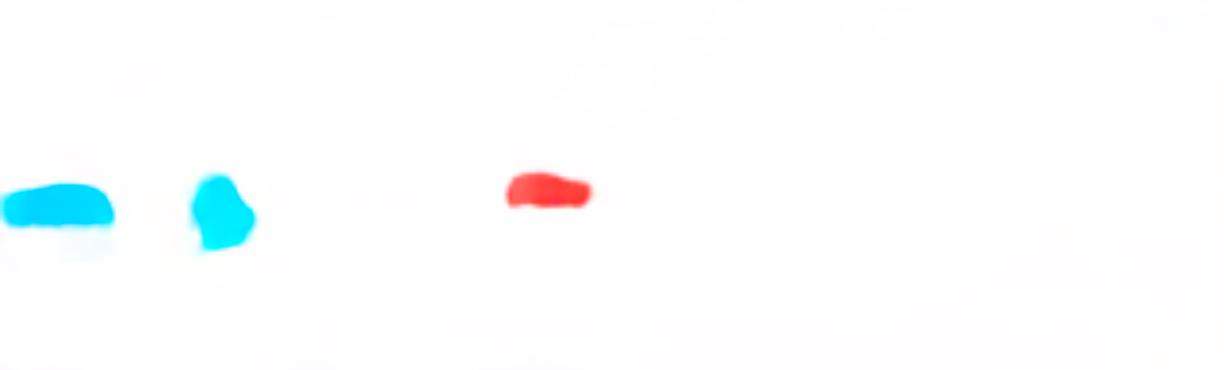}%
\galleryRowKitti{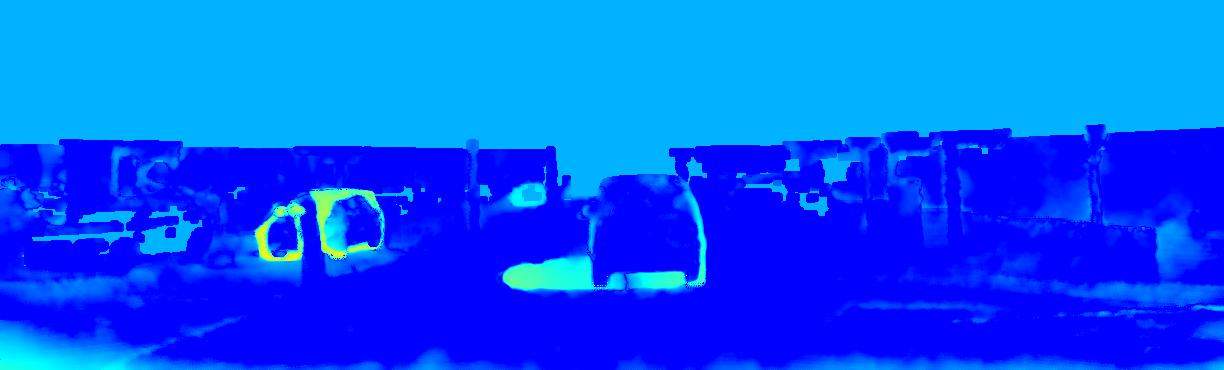}{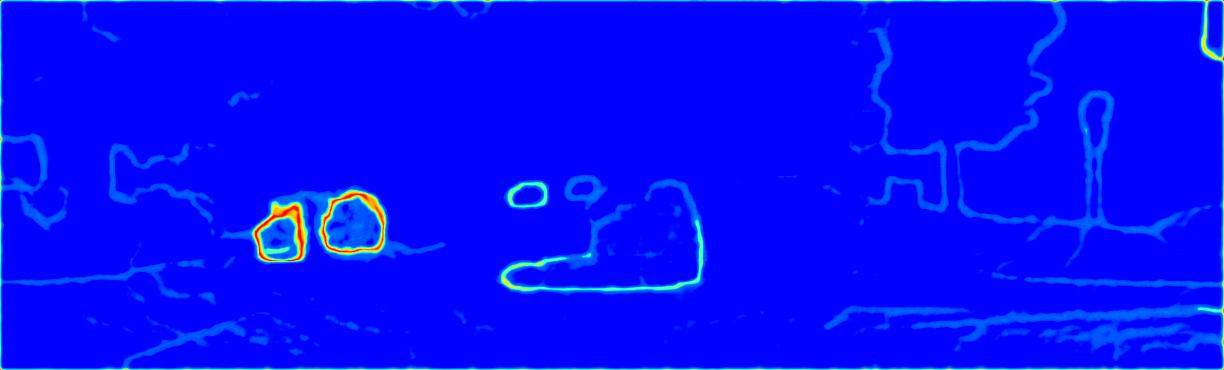}{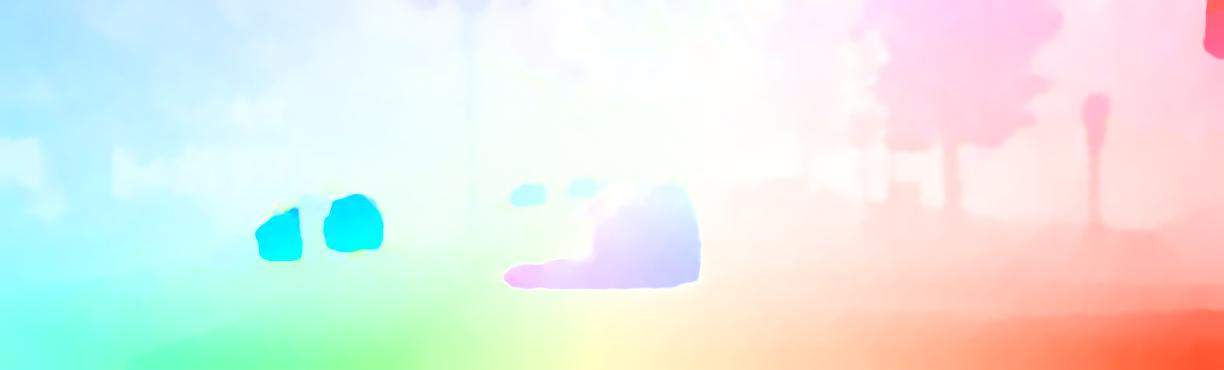}{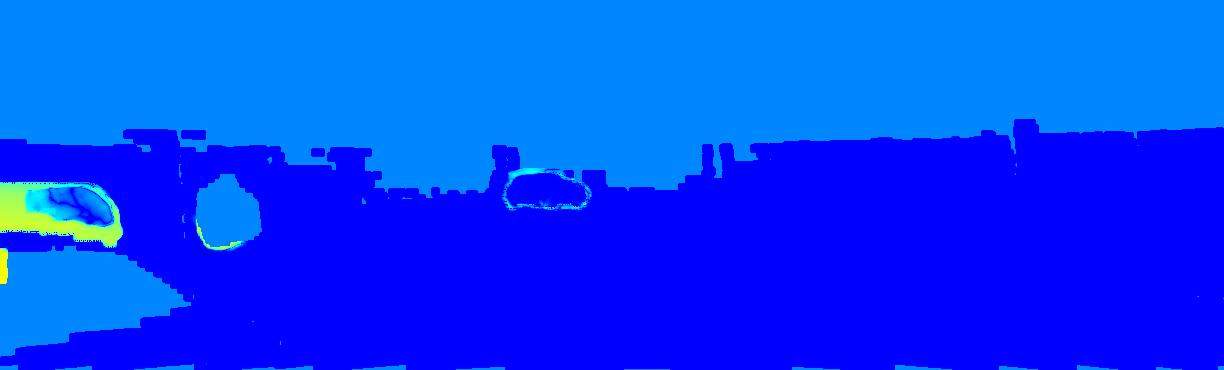}{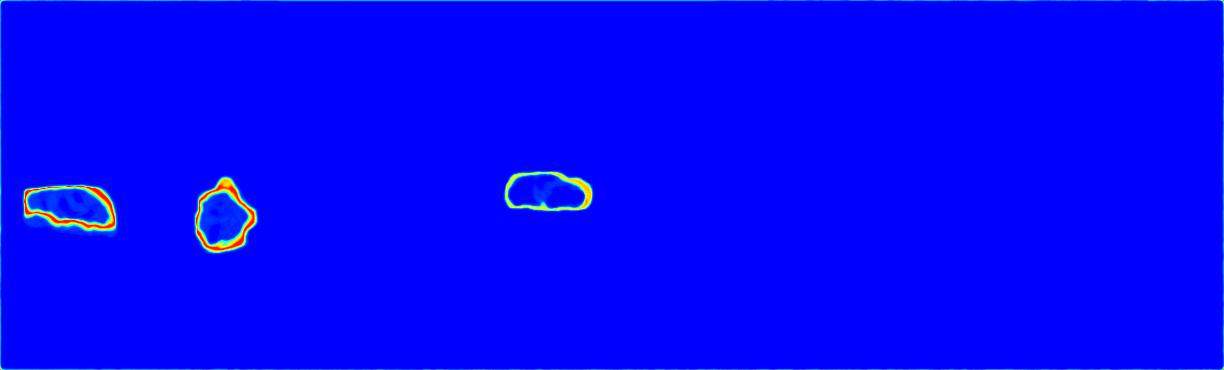}{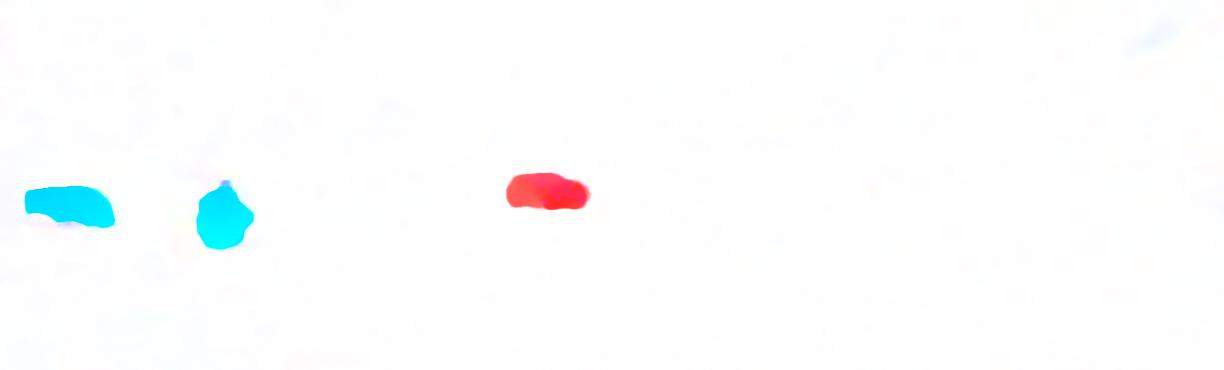}%
    \end{tabular}%
  }%

  \end{center}%
  \caption{Examples from KITTI2012 and KITTI2015 datasets for qualitative comparison between FlowNetH-Pred-Merged and ProbFlow. For each example: \textbf{first row} shows the original image sequence followed by its ground truth flow field (bilinearly interpolated from sparse ground truth). \textbf{Second row} shows FlowNetH-Pred-Merged results: oracle entropy (representing the optimal uncertainty), predicted entropy and predicted flow. Similar to second row, \textbf{third row} shows the results for ProbFlow. Remark: In the sky the groundtruth provided in the datasets are invalid due to data acquisition.}%
  \label{fig:gallery_kitti2}%
\end{figure*}

We provide qualitative results on real world videos, Sintel train clean, KITTI2012 and KITTI2015 datasets for FlowNetH-Pred-Merged and ProbFlow in Figure~\ref{fig:gallery_real}, Figure~\ref{fig:gallery_sintel2} and Figure~\ref{fig:gallery_kitti2}.

At last, we show the outputs for all ensemble members for a simple and a difficult case in Tables~\ref{tab:ex2_1},\ref{tab:ex2_2} and Tables~\ref{tab:ex1_1},\ref{tab:ex1_2}. We note that comparing to the other ensembles, hypothesis from FlowNetH generate the most diverse results.

\clearpage

\vspace*{1cm} 
        \bgroup
        \def\arraystretch{1.0}
        \renewcommand{\tabcolsep}{0.05cm}
        \begin{table*}
        \begin{center}
            \resizebox{\linewidth}{!}{%
                \begin{tabular}{|cccc|cc|}
        
\hline
\multicolumn{4}{|l|}{Data:}& & \\ 
\includegraphics[width=0.1666\linewidth]{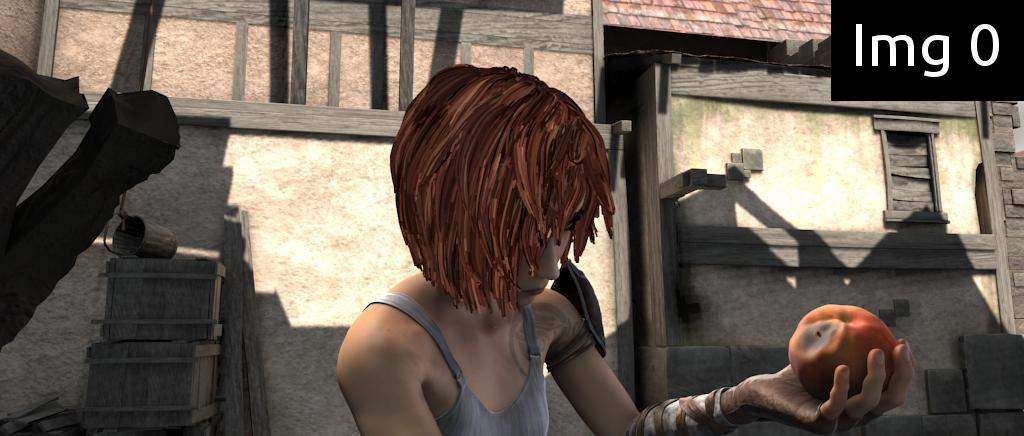}
&
\includegraphics[width=0.1666\linewidth]{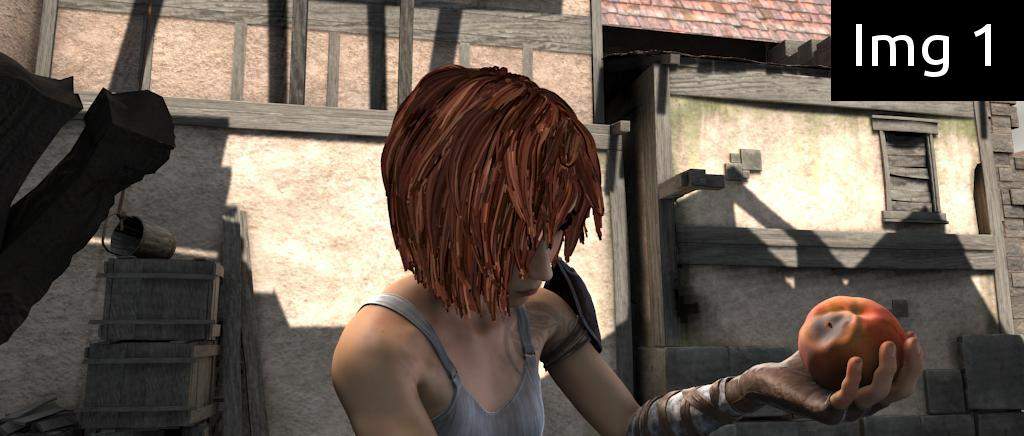}
&
&
&
\includegraphics[width=0.1666\linewidth]{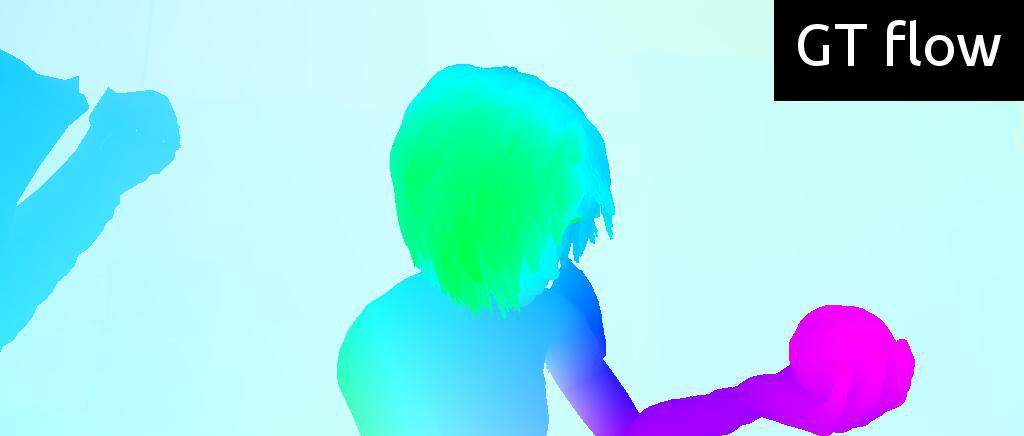}
&
\\

\hline
\hline
\multicolumn{4}{|l|}{FlowNetC Emp:}& & \\ 
&
&
&
&
\includegraphics[width=0.1666\linewidth]{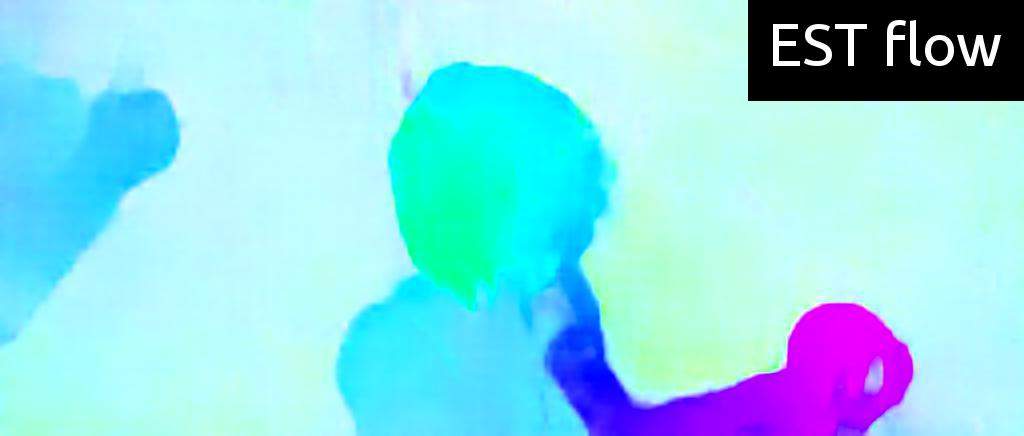}
&
\\

\hline
\hline
\multicolumn{4}{|l|}{FlowNetH Base:}& & \\ 
\includegraphics[width=0.1666\linewidth]{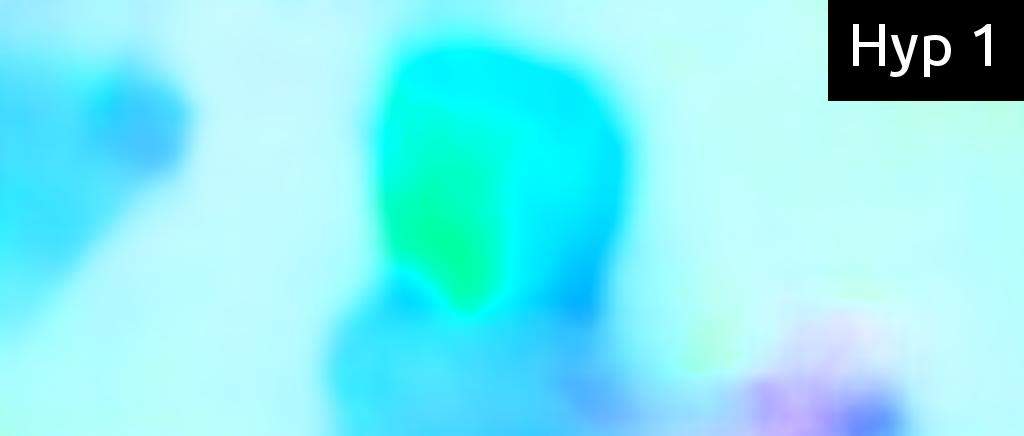}
&
\includegraphics[width=0.1666\linewidth]{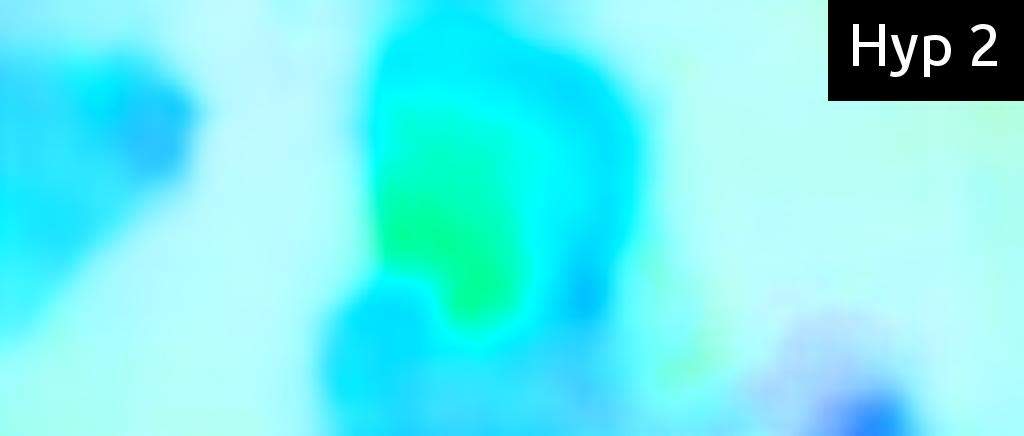}
&
\includegraphics[width=0.1666\linewidth]{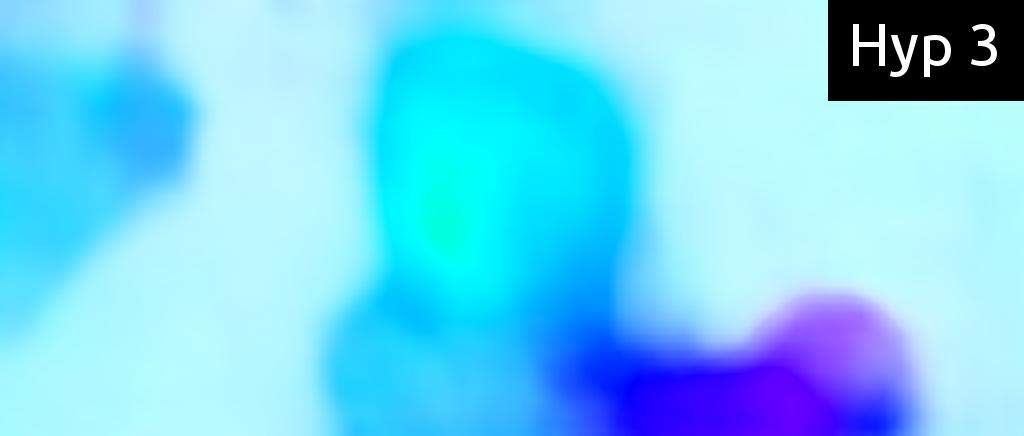}
&
\includegraphics[width=0.1666\linewidth]{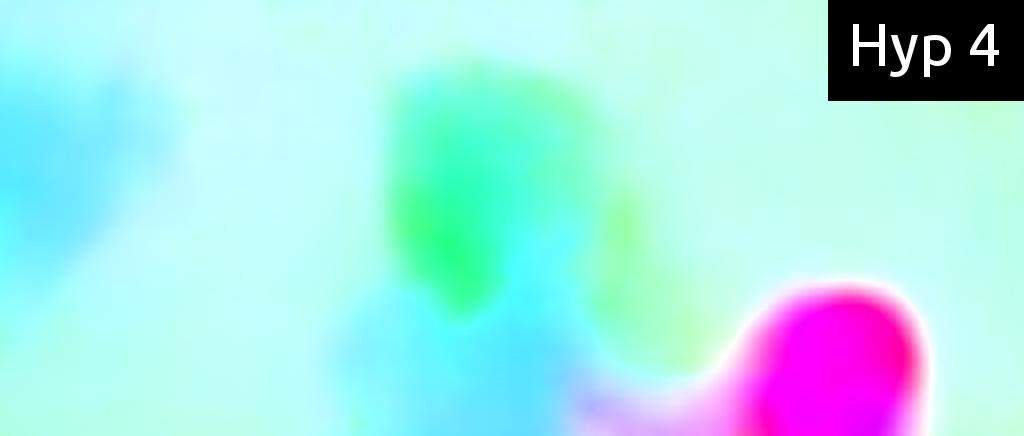}
&
\includegraphics[width=0.1666\linewidth]{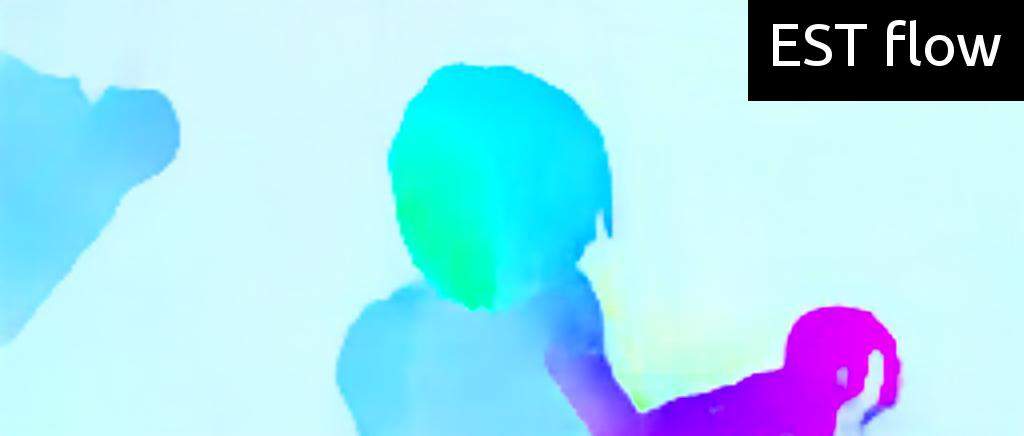}
&
\\

\includegraphics[width=0.1666\linewidth]{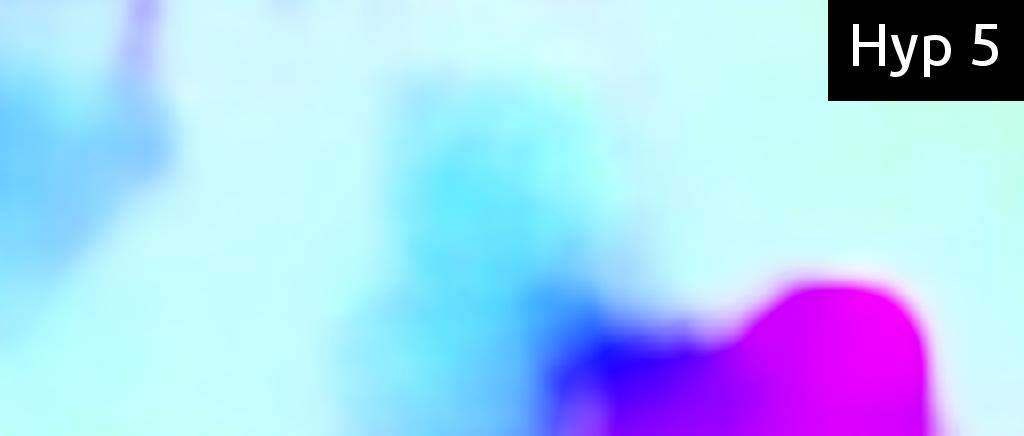}
&
\includegraphics[width=0.1666\linewidth]{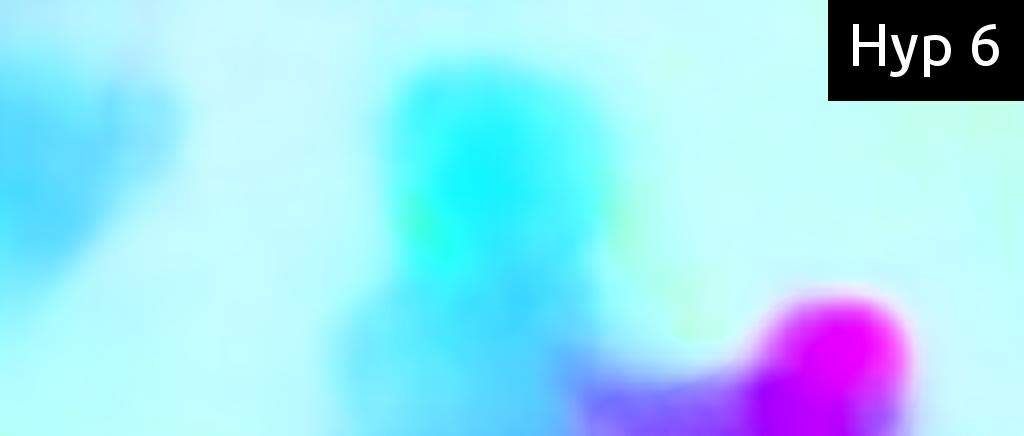}
&
\includegraphics[width=0.1666\linewidth]{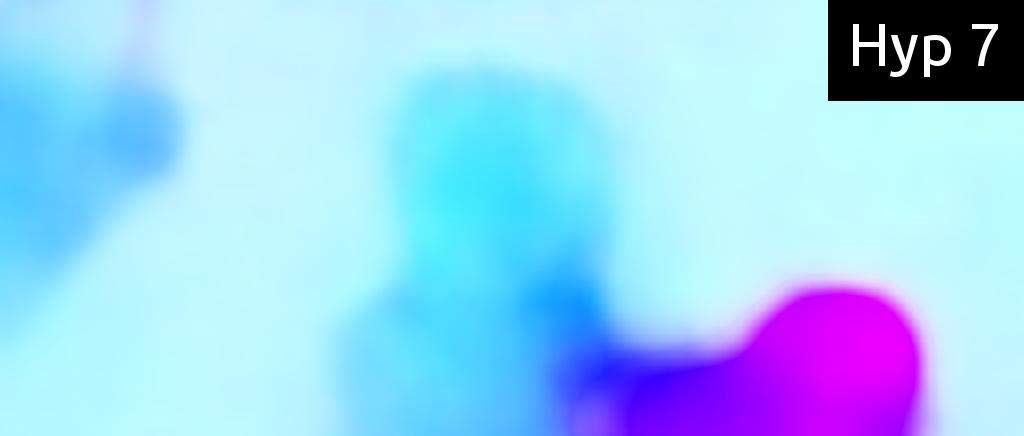}
&
\includegraphics[width=0.1666\linewidth]{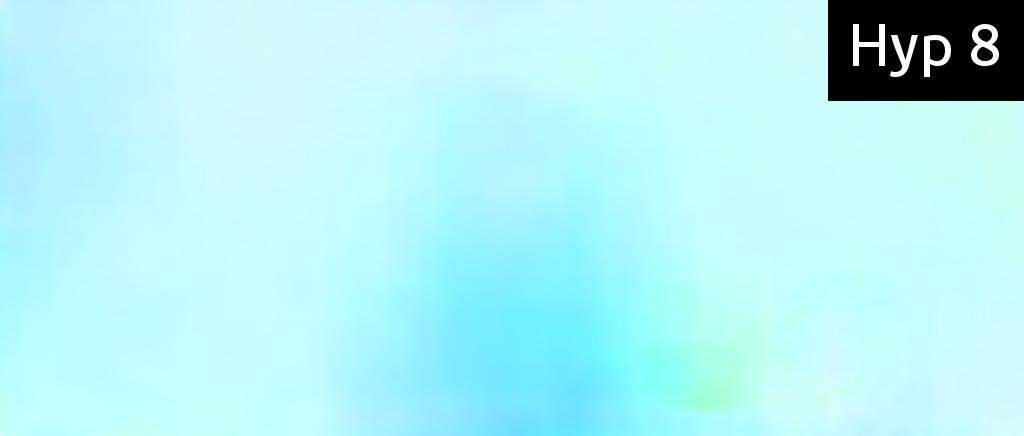}
&
&
\\

\hline
\hline
\multicolumn{4}{|l|}{Dropout Emp:}& & \\ 
\includegraphics[width=0.1666\linewidth]{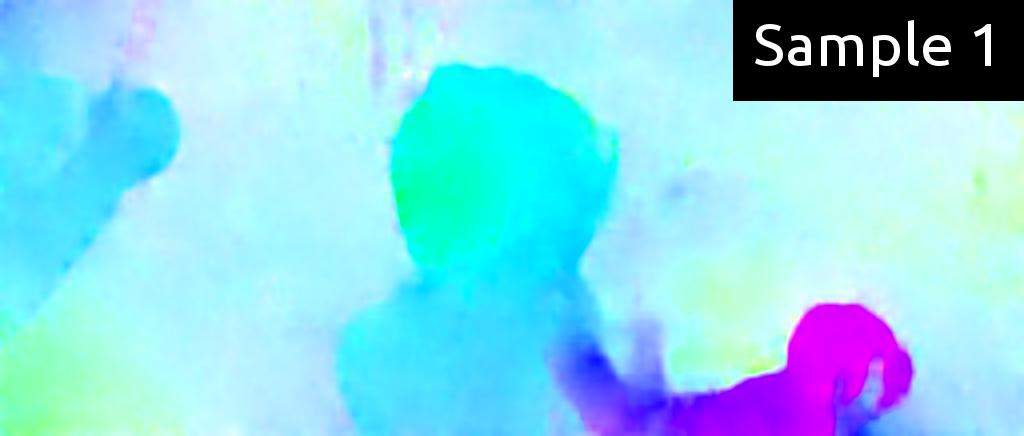}
&
\includegraphics[width=0.1666\linewidth]{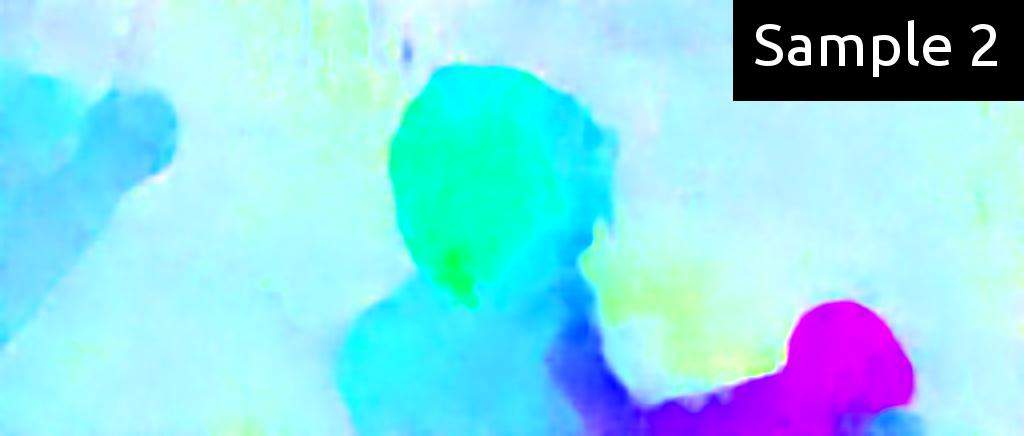}
&
\includegraphics[width=0.1666\linewidth]{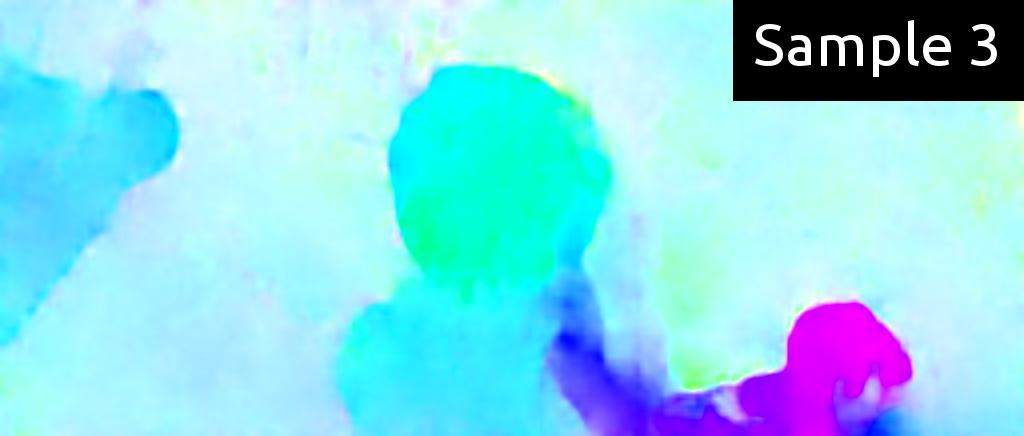}
&
\includegraphics[width=0.1666\linewidth]{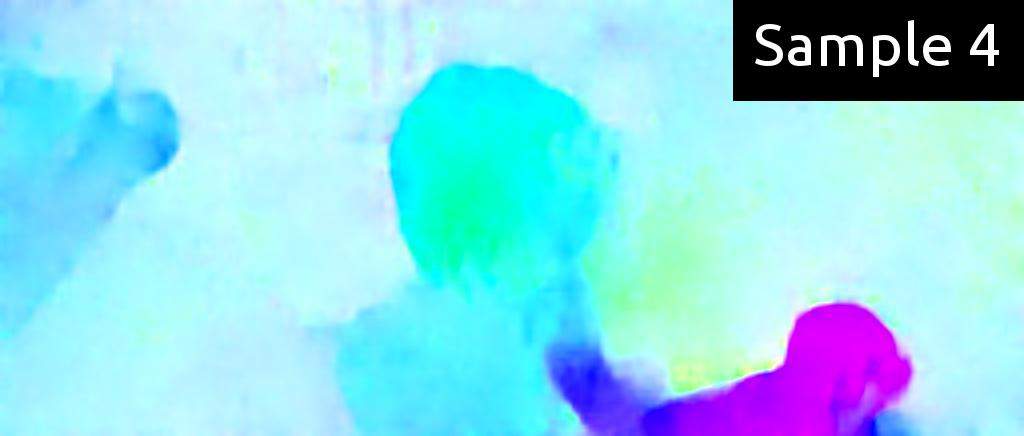}
&
\includegraphics[width=0.1666\linewidth]{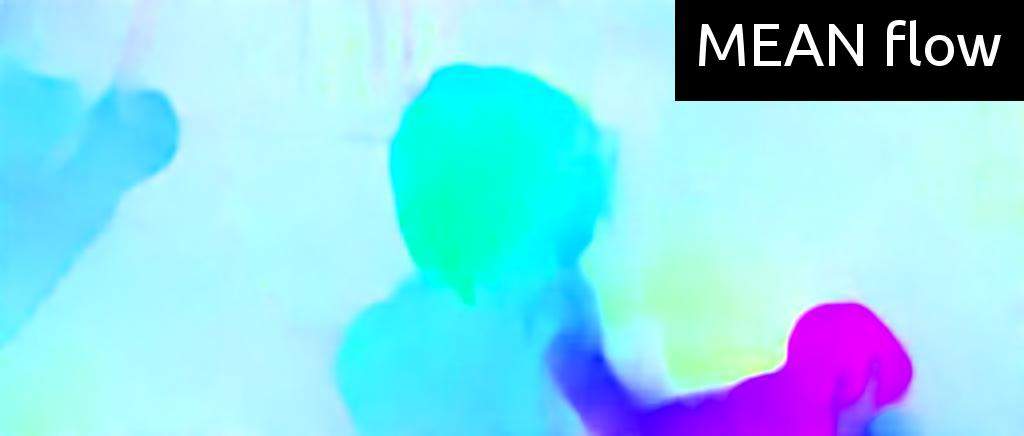}
&
\includegraphics[width=0.1666\linewidth]{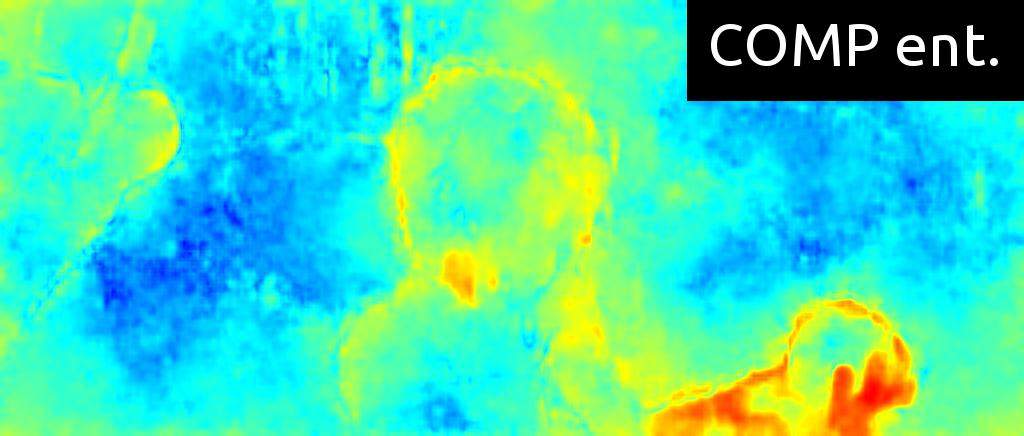}
\\

\includegraphics[width=0.1666\linewidth]{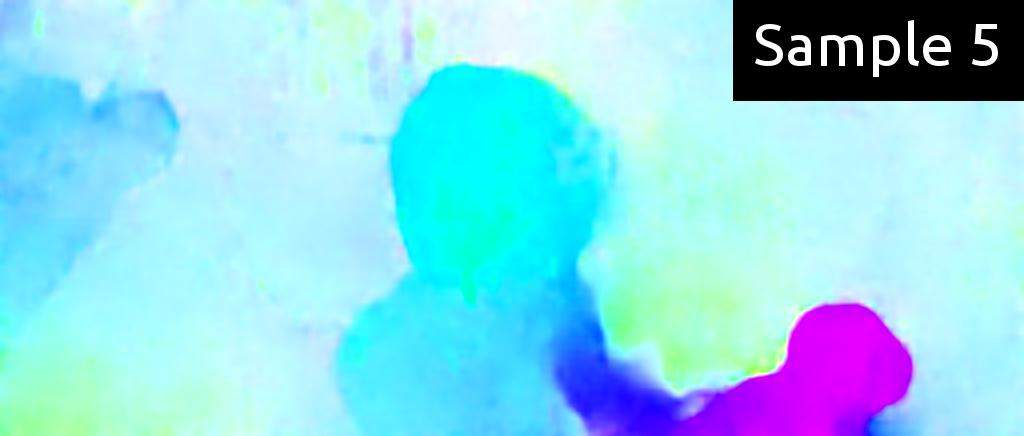}
&
\includegraphics[width=0.1666\linewidth]{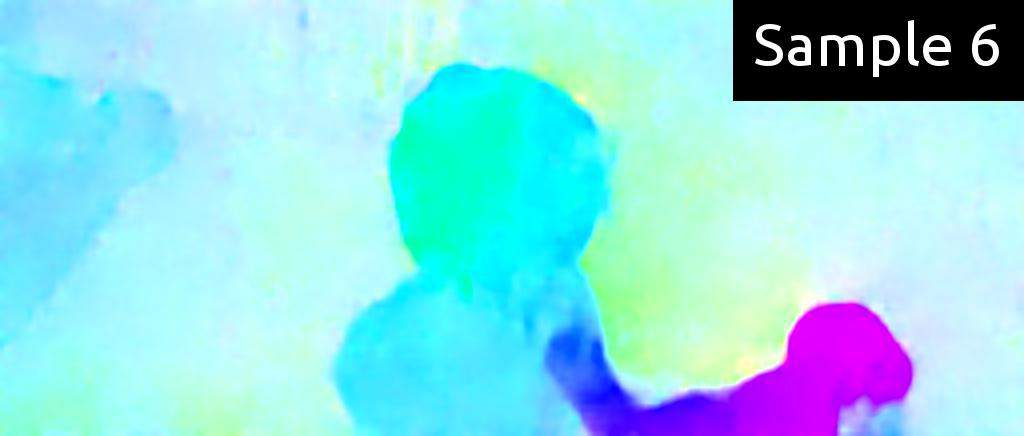}
&
\includegraphics[width=0.1666\linewidth]{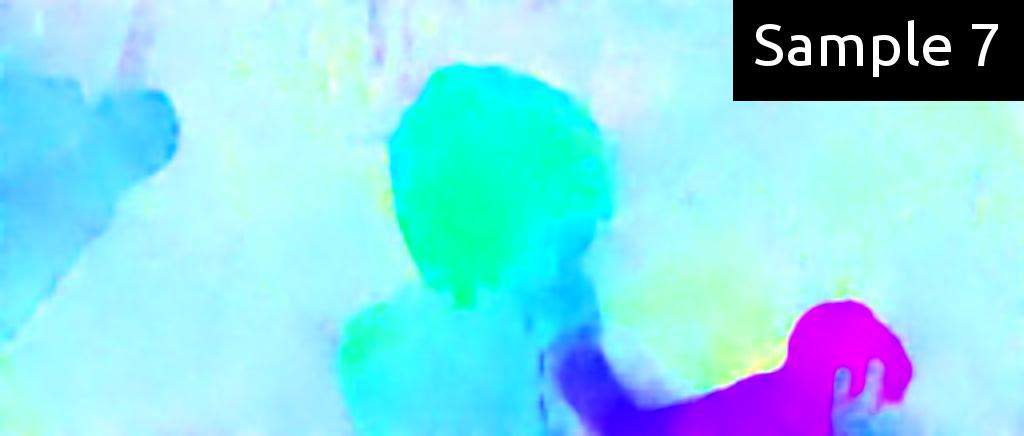}
&
\includegraphics[width=0.1666\linewidth]{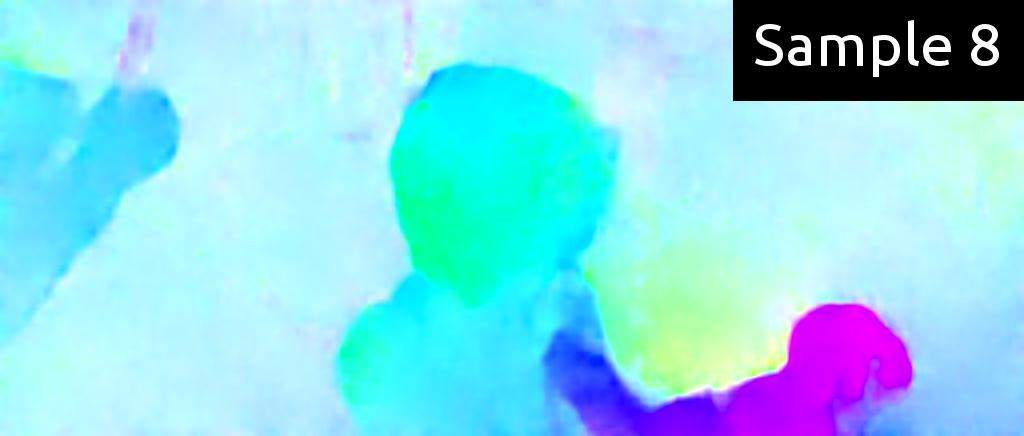}
&
&
\\

\hline
\hline
\multicolumn{4}{|l|}{SGDR Emp:}& & \\ 
\includegraphics[width=0.1666\linewidth]{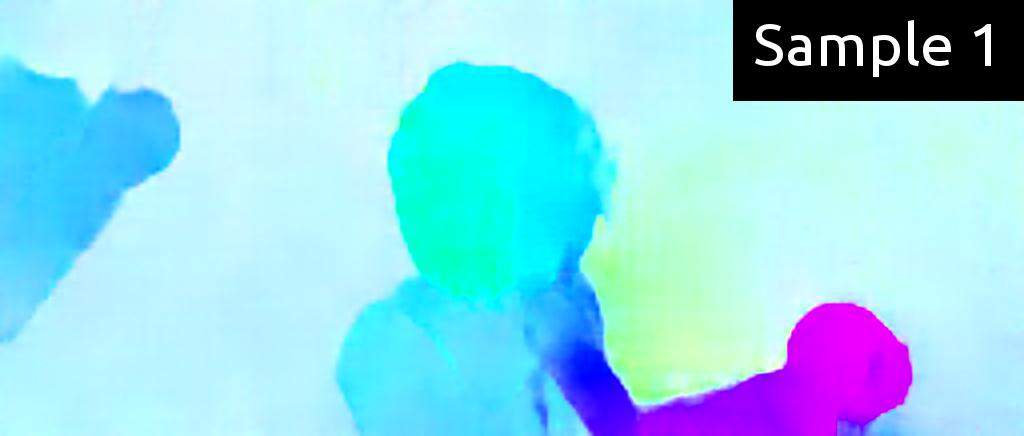}
&
\includegraphics[width=0.1666\linewidth]{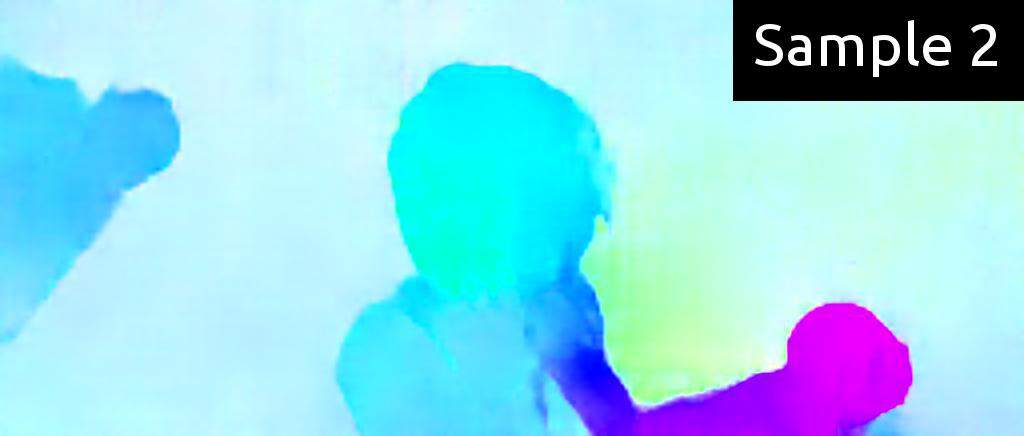}
&
\includegraphics[width=0.1666\linewidth]{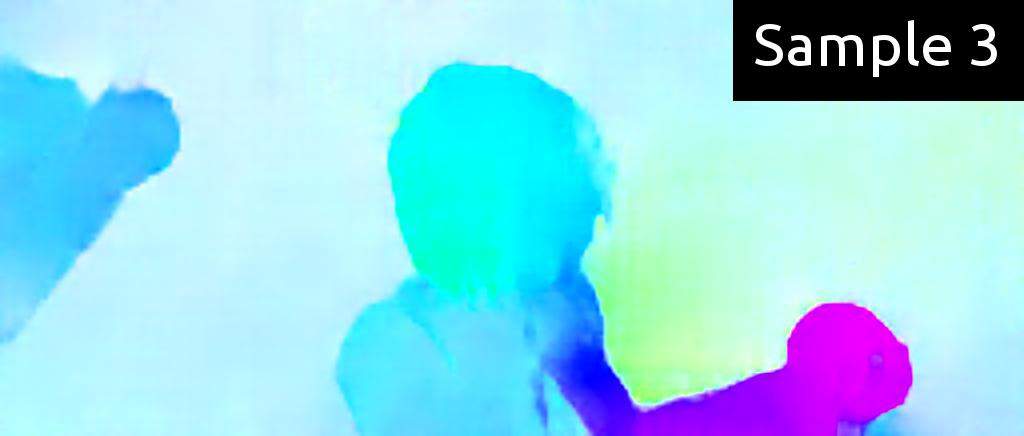}
&
\includegraphics[width=0.1666\linewidth]{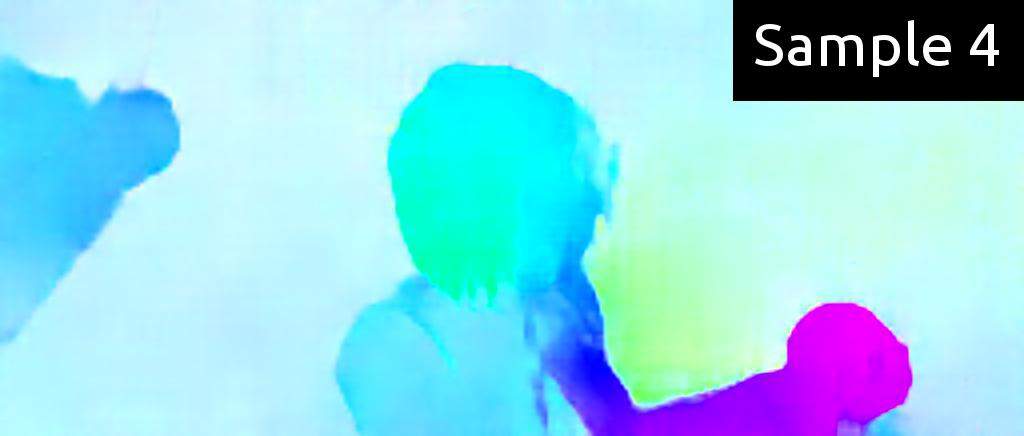}
&
\includegraphics[width=0.1666\linewidth]{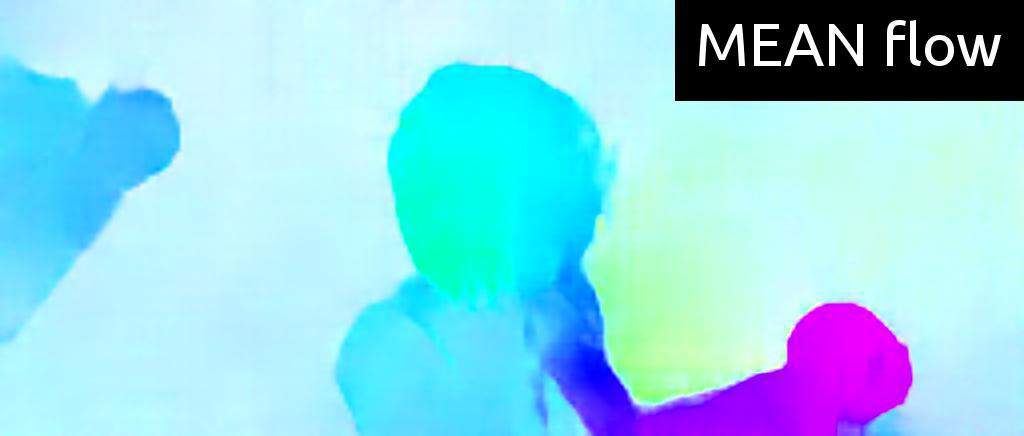}
&
\includegraphics[width=0.1666\linewidth]{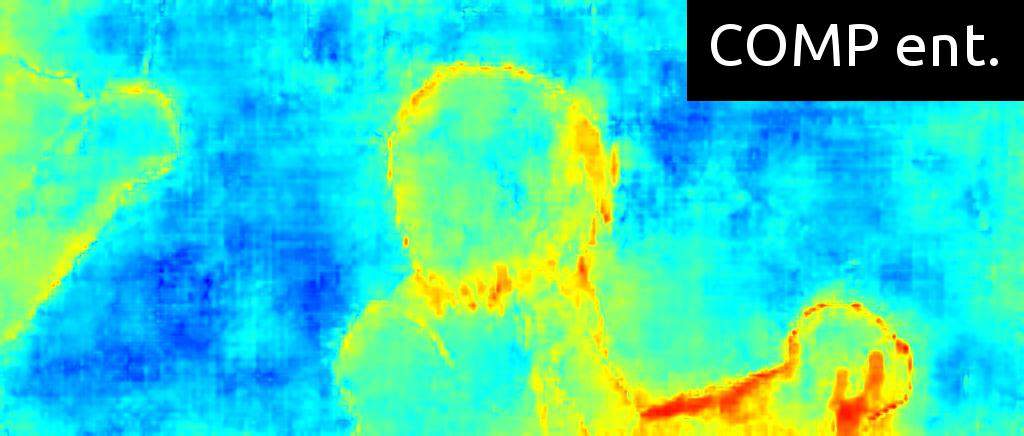}
\\

\includegraphics[width=0.1666\linewidth]{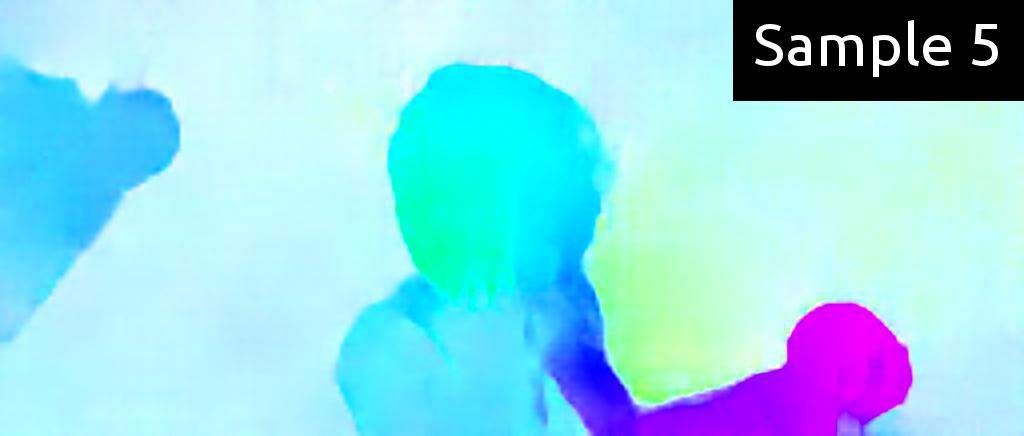}
&
\includegraphics[width=0.1666\linewidth]{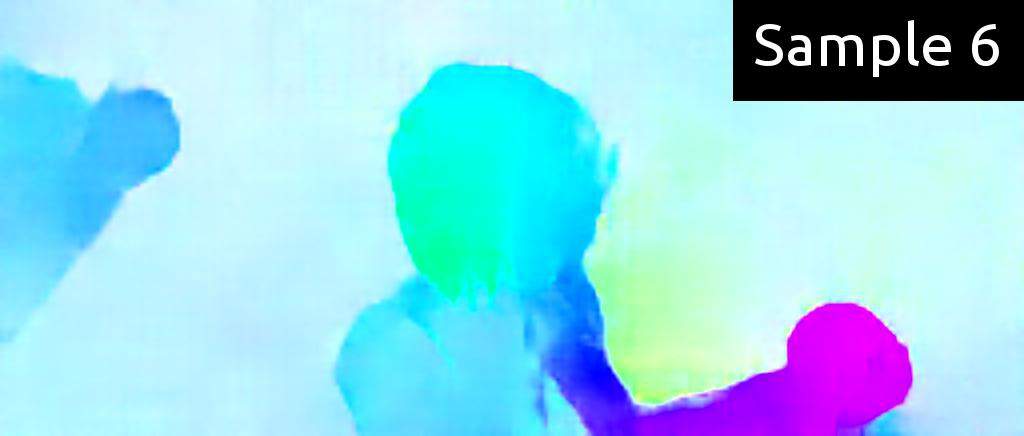}
&
\includegraphics[width=0.1666\linewidth]{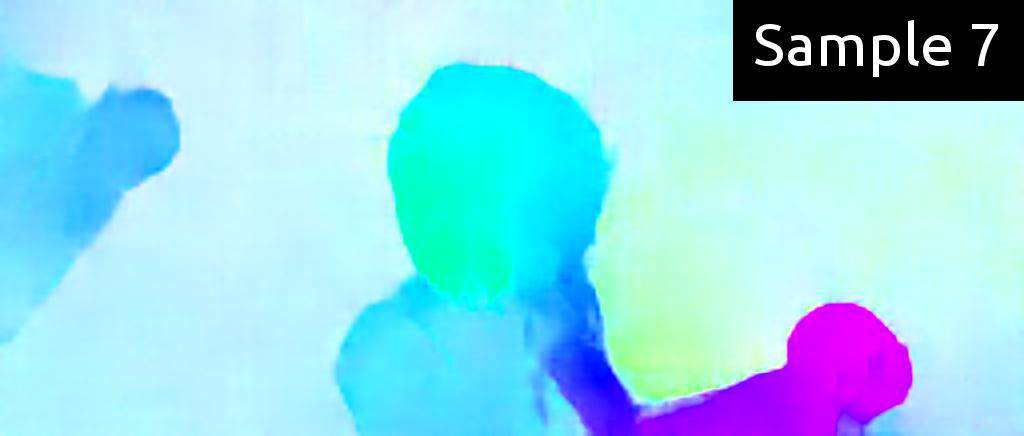}
&
\includegraphics[width=0.1666\linewidth]{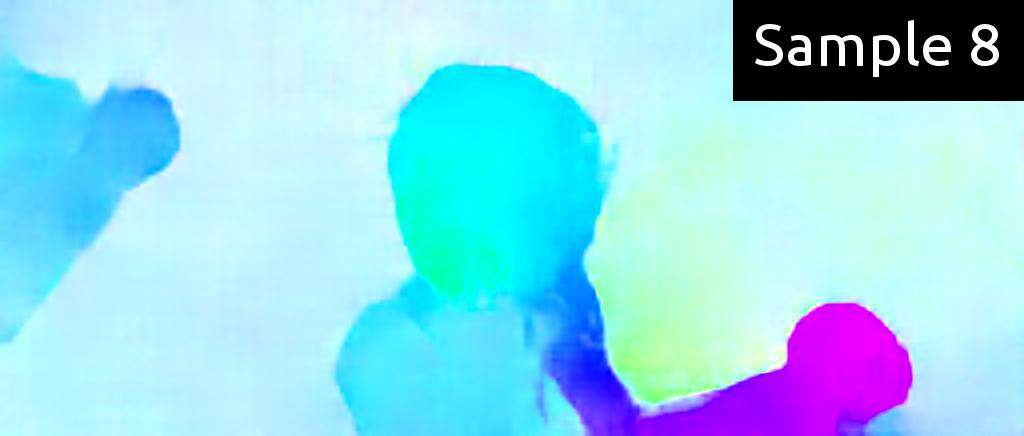}
&
&
\\

\hline
\hline
\multicolumn{4}{|l|}{Bootstrapped Ensemble Emp:}& & \\ 
\includegraphics[width=0.1666\linewidth]{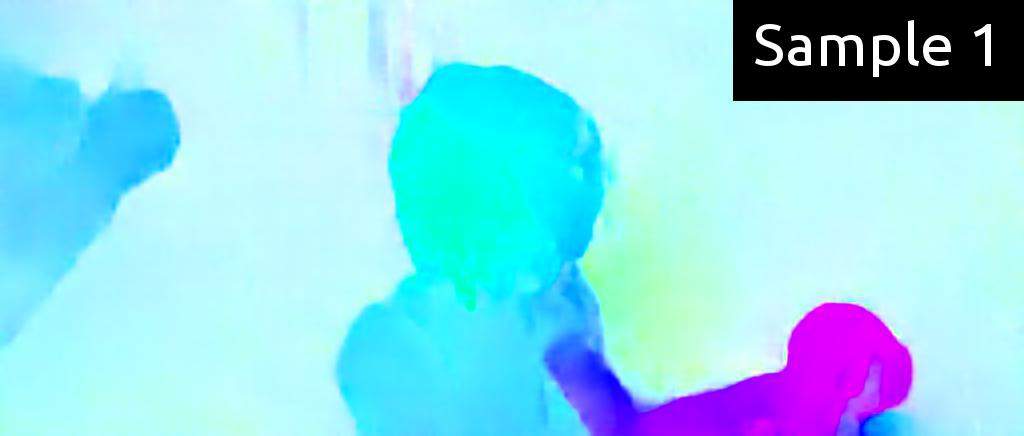}
&
\includegraphics[width=0.1666\linewidth]{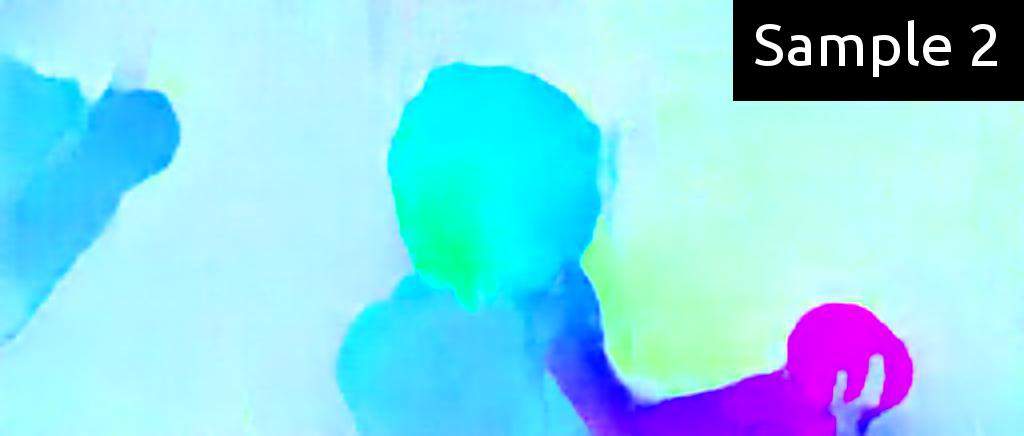}
&
\includegraphics[width=0.1666\linewidth]{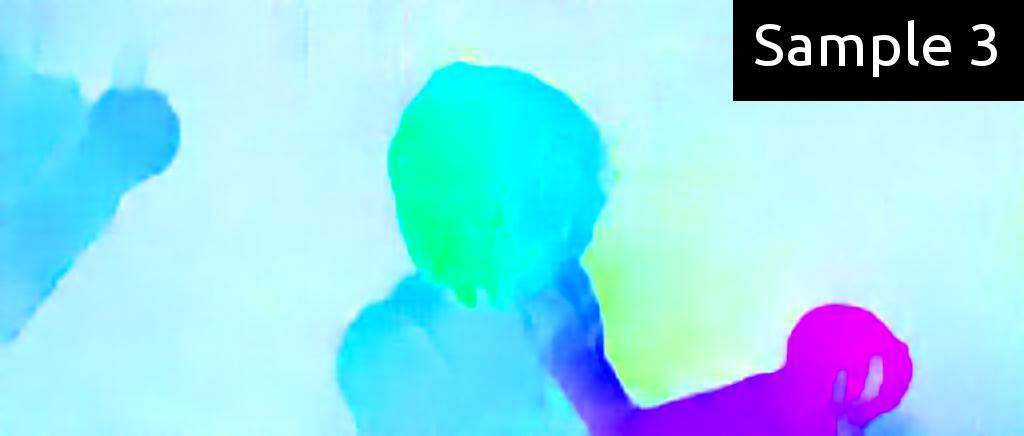}
&
\includegraphics[width=0.1666\linewidth]{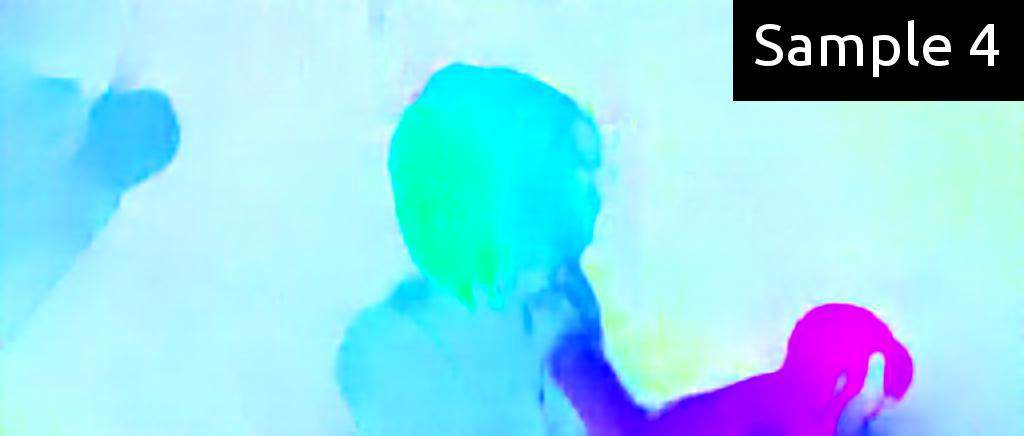}
&
\includegraphics[width=0.1666\linewidth]{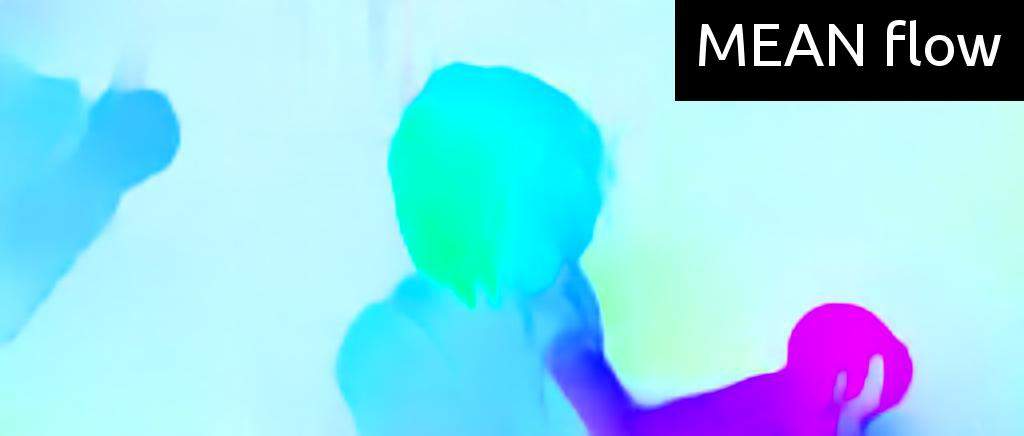}
&
\includegraphics[width=0.1666\linewidth]{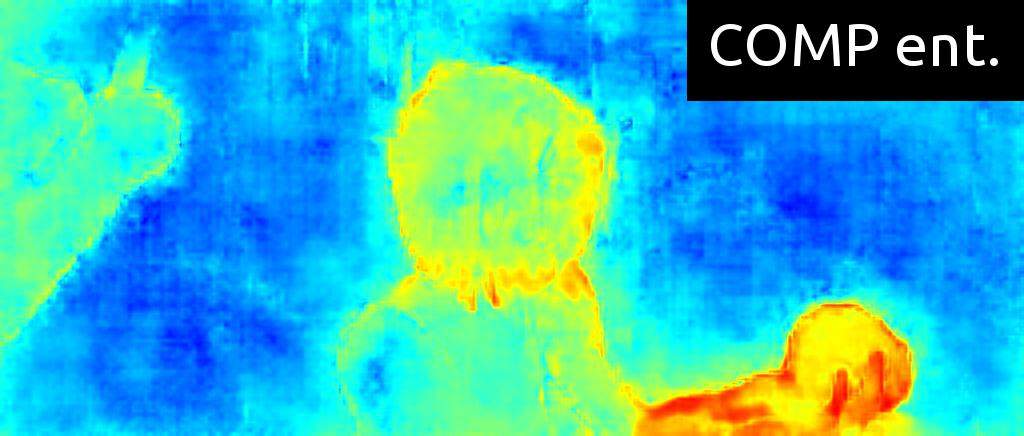}
\\

\includegraphics[width=0.1666\linewidth]{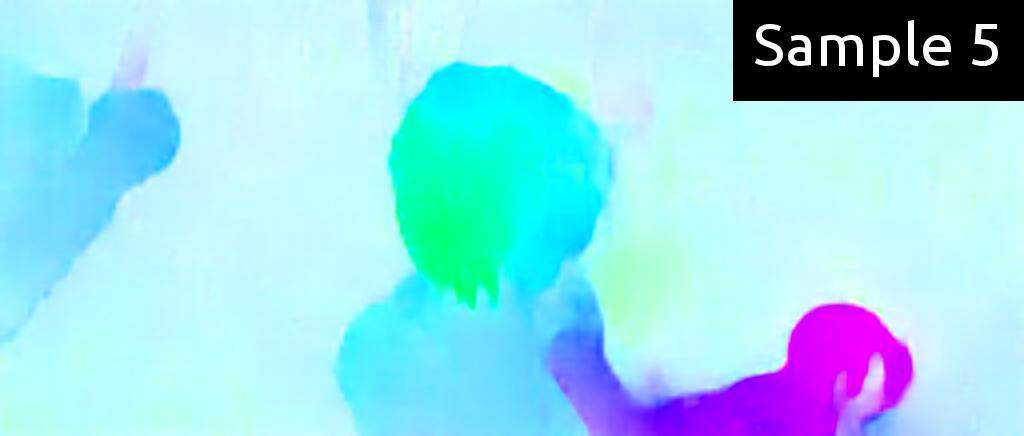}
&
\includegraphics[width=0.1666\linewidth]{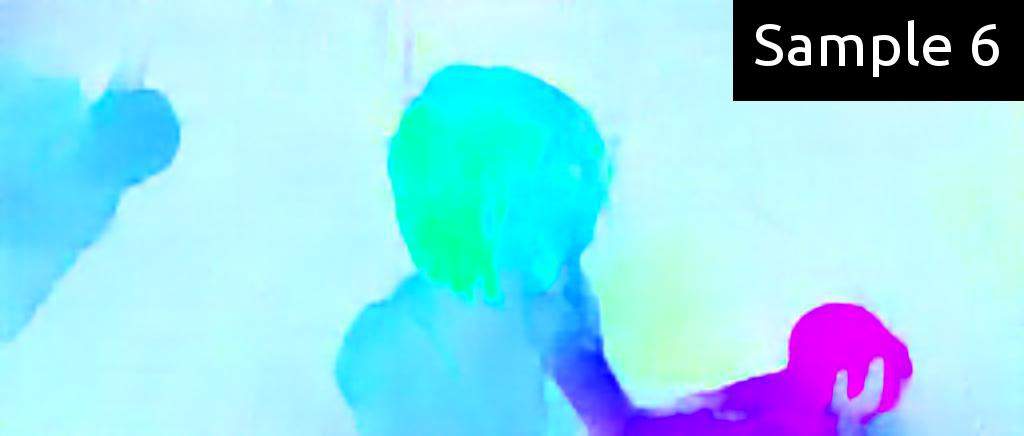}
&
\includegraphics[width=0.1666\linewidth]{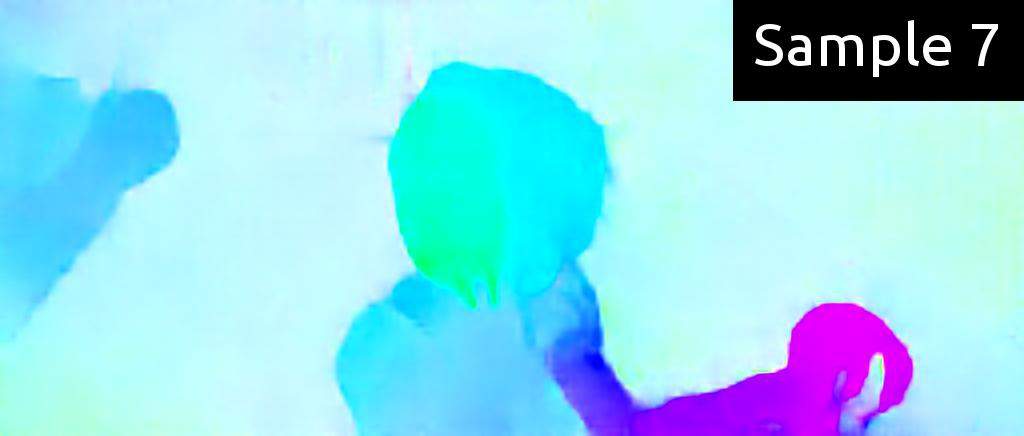}
&
\includegraphics[width=0.1666\linewidth]{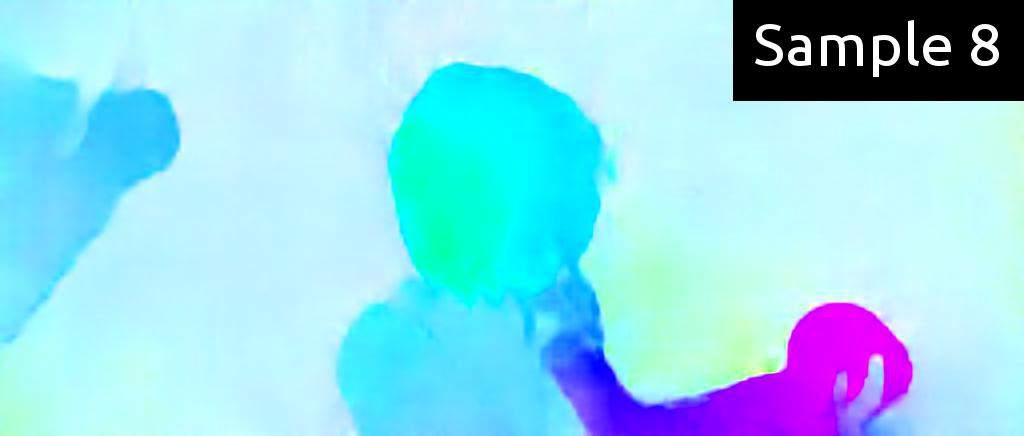}
&
&
\\

\hline

                \end{tabular}
            }
       \end{center}
             \caption{
            In this table we show the outputs of empirical experiments with all presented methods for an easy Sintel example as well as the averaged flows and computed entropies. Because the example is easy, the networks are certain and not much variety is visible in the outputs.
            }
            \label{tab:ex2_1}
        \end{table*}
        \egroup

        \bgroup
        \def\arraystretch{1.0}
        \renewcommand{\tabcolsep}{0.05cm}
        \begin{table*}
        \begin{center}
            \resizebox{\linewidth}{!}{%
                \begin{tabular}{|cccc|cc|}
        
\hline
\multicolumn{4}{|l|}{Data:}& & \\ 
\includegraphics[width=0.1666\linewidth]{images/method-overview/ex2/img0.jpg}
&
\includegraphics[width=0.1666\linewidth]{images/method-overview/ex2/img1.jpg}
&
&
&
\includegraphics[width=0.1666\linewidth]{images/method-overview/ex2/flow_gt.jpg}
&
\\

\hline
\hline
\multicolumn{4}{|l|}{FlowNetC Pred:}& & \\ 
&
&
&
&
\includegraphics[width=0.1666\linewidth]{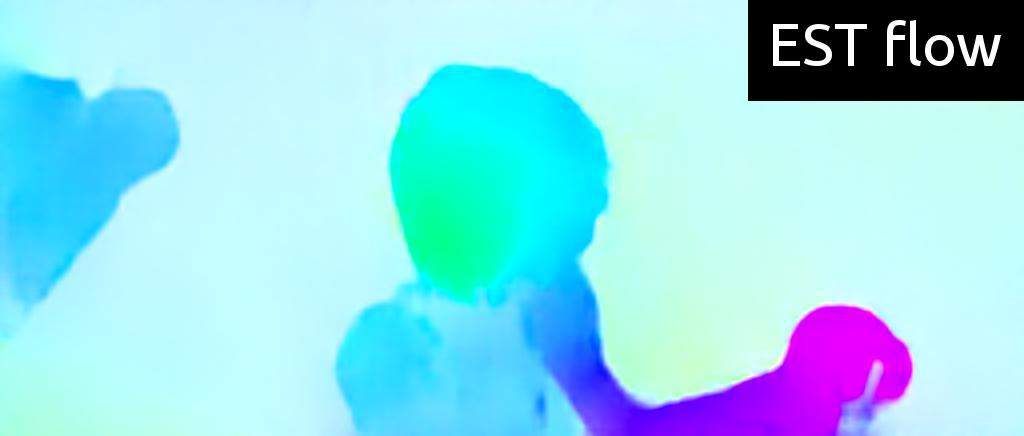}
&
\includegraphics[width=0.1666\linewidth]{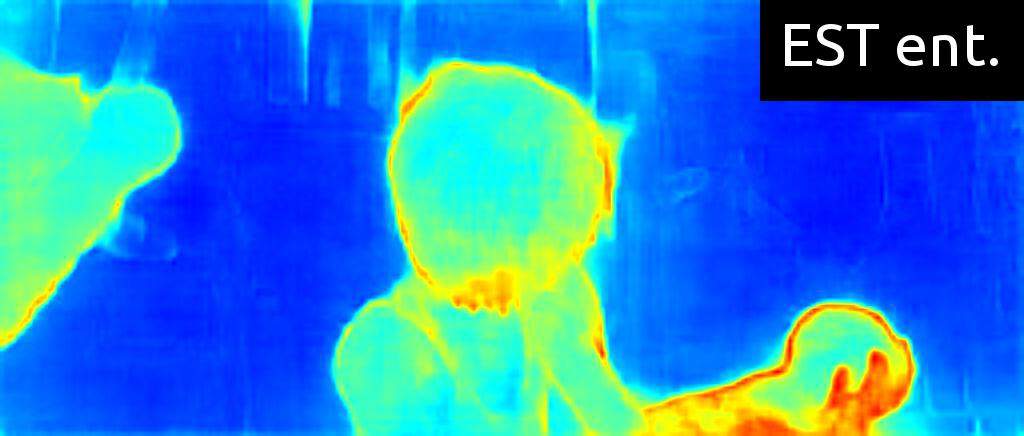}
\\

\hline
\hline
\multicolumn{4}{|l|}{Dropout Pred:}& & \\ 
\includegraphics[width=0.1666\linewidth]{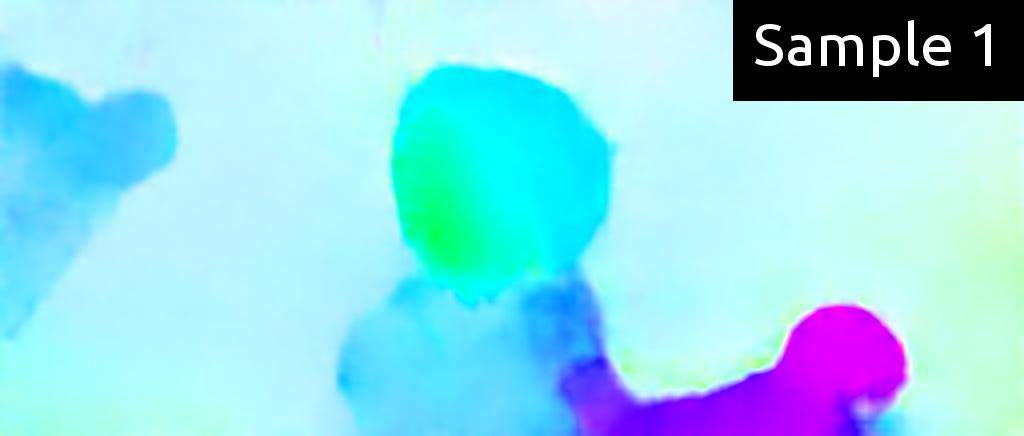}
&
\includegraphics[width=0.1666\linewidth]{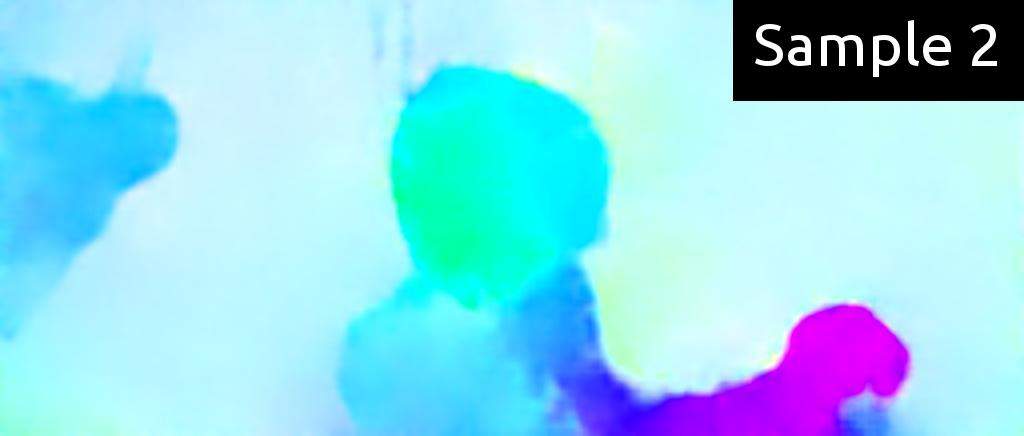}
&
\includegraphics[width=0.1666\linewidth]{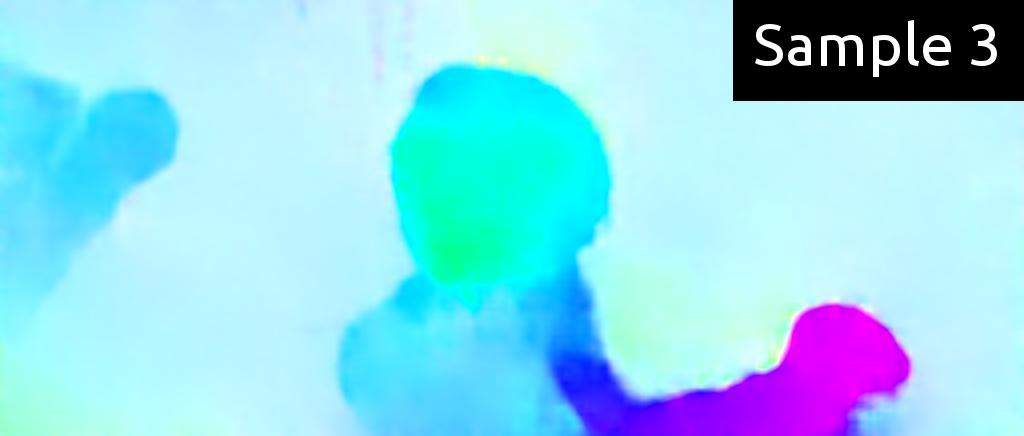}
&
\includegraphics[width=0.1666\linewidth]{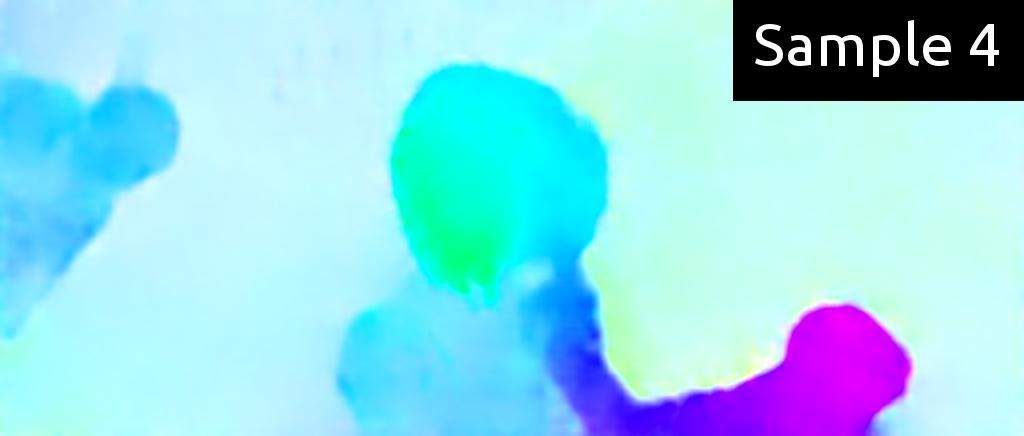}
&
\includegraphics[width=0.1666\linewidth]{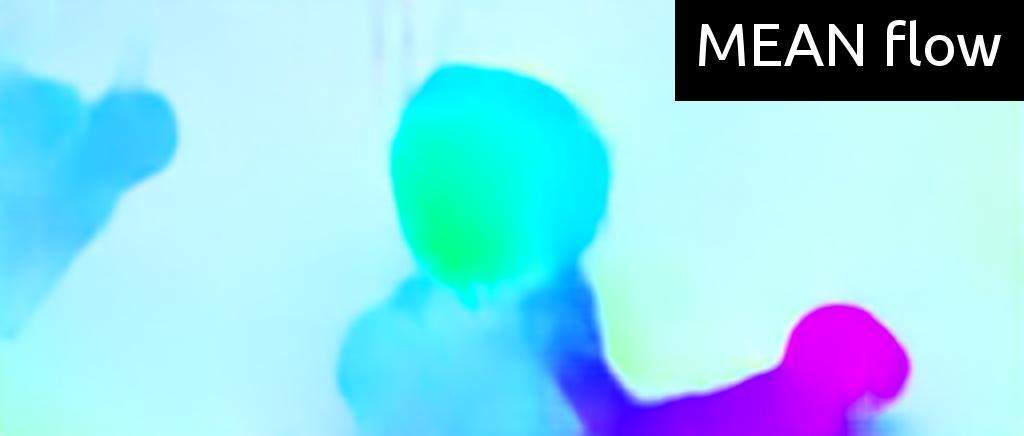}
&
\includegraphics[width=0.1666\linewidth]{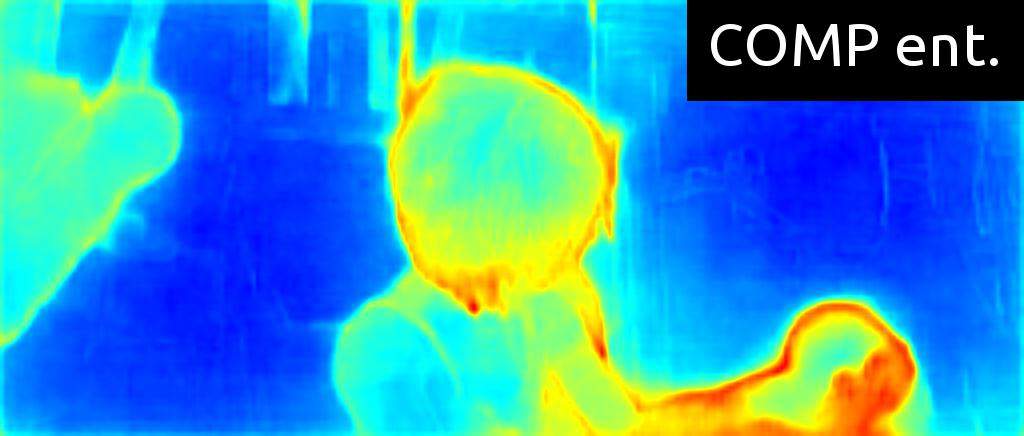}
\\

\includegraphics[width=0.1666\linewidth]{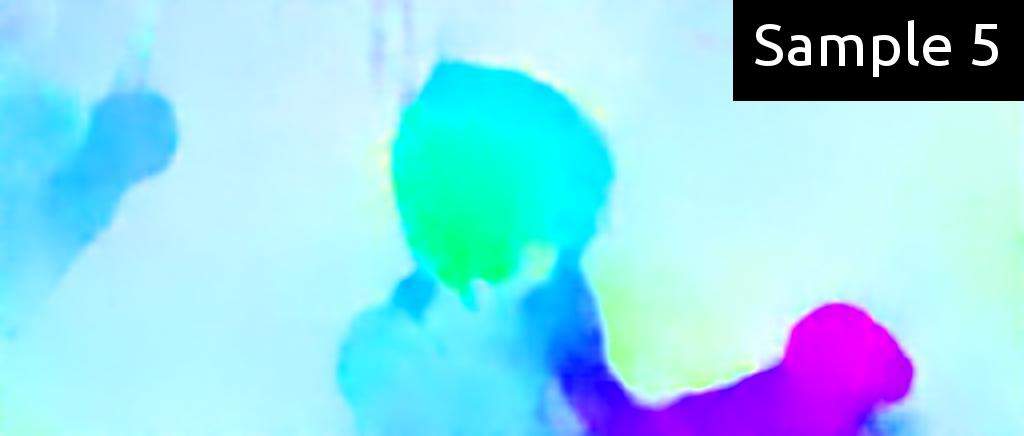}
&
\includegraphics[width=0.1666\linewidth]{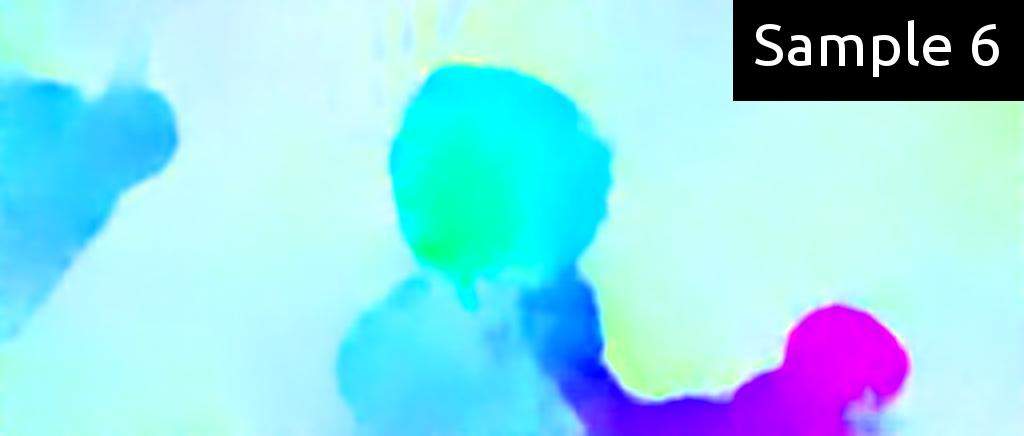}
&
\includegraphics[width=0.1666\linewidth]{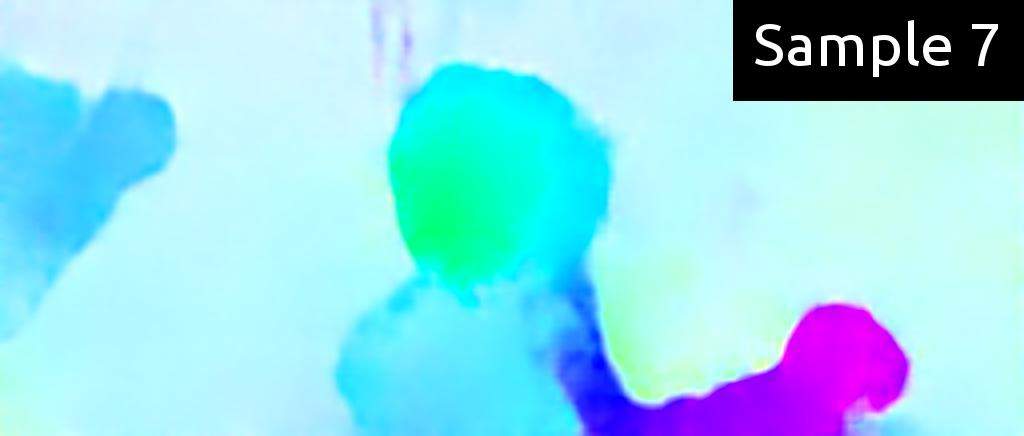}
&
\includegraphics[width=0.1666\linewidth]{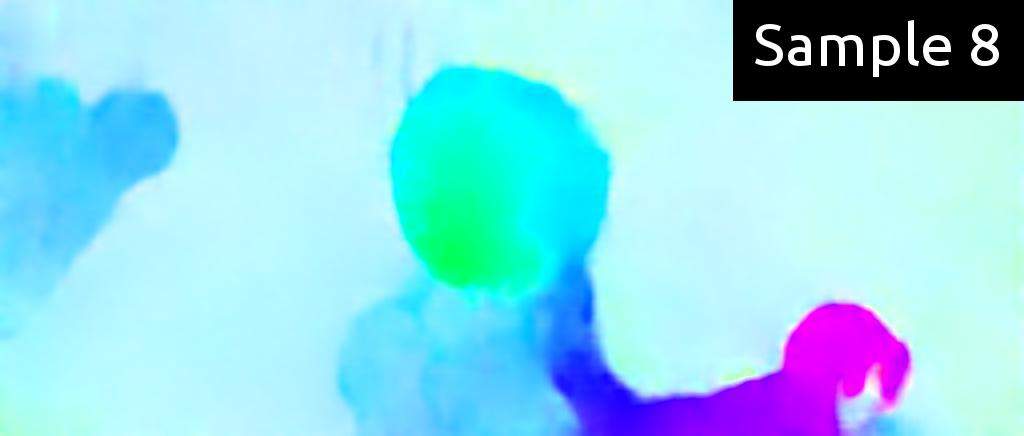}
&
&
\\

\hline
\hline
\multicolumn{4}{|l|}{SGDR Pred:}& & \\ 
\includegraphics[width=0.1666\linewidth]{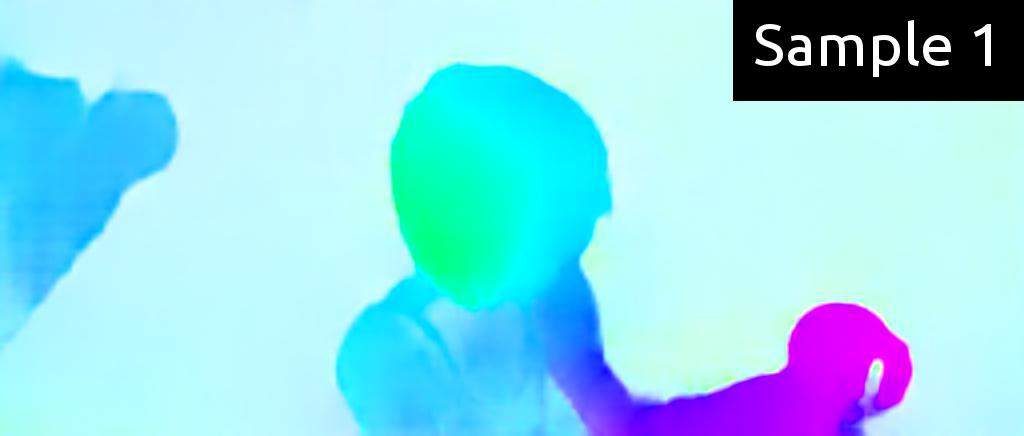}
&
\includegraphics[width=0.1666\linewidth]{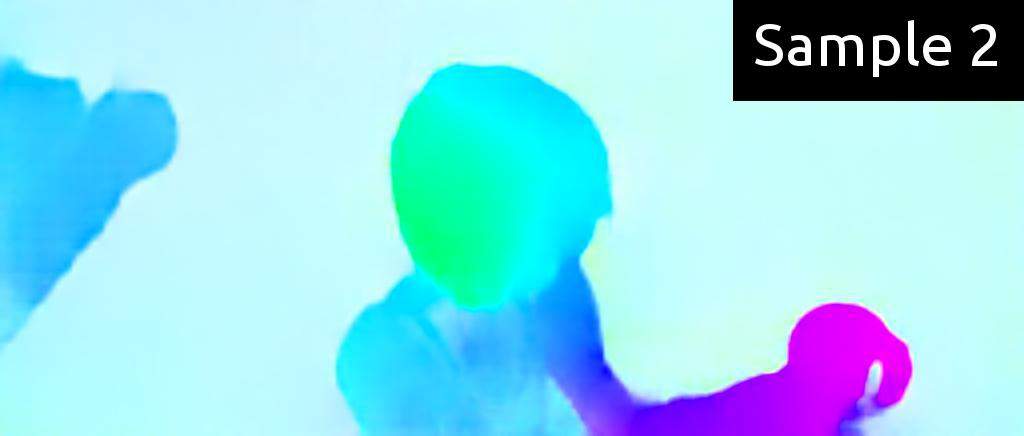}
&
\includegraphics[width=0.1666\linewidth]{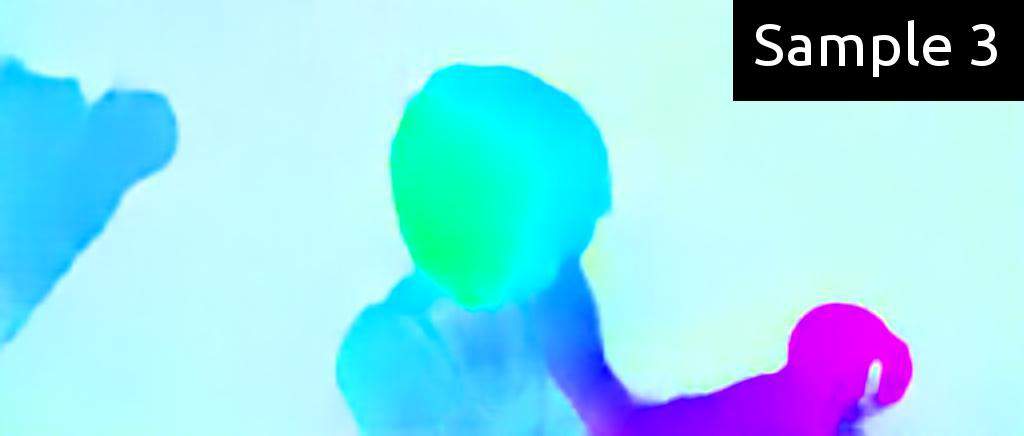}
&
\includegraphics[width=0.1666\linewidth]{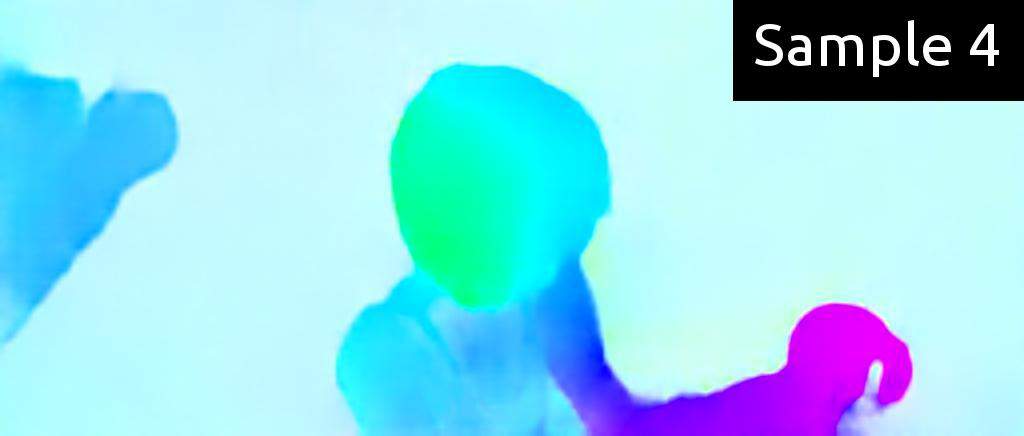}
&
\includegraphics[width=0.1666\linewidth]{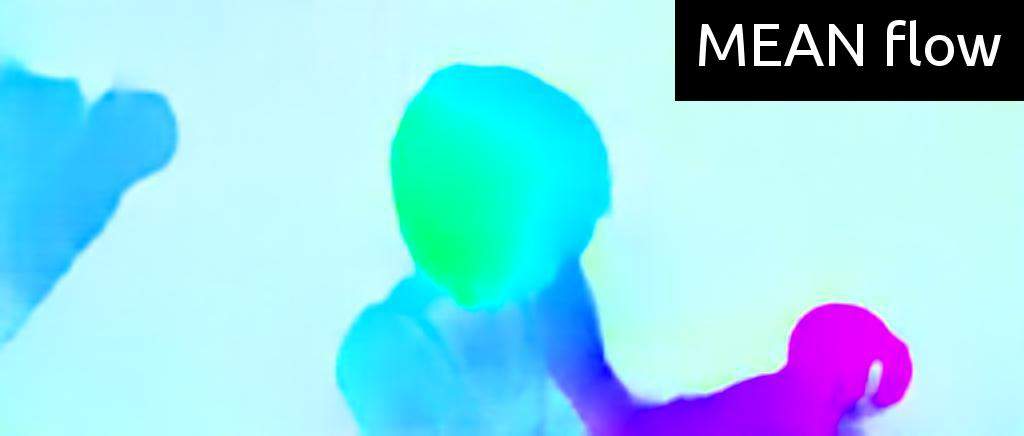}
&
\includegraphics[width=0.1666\linewidth]{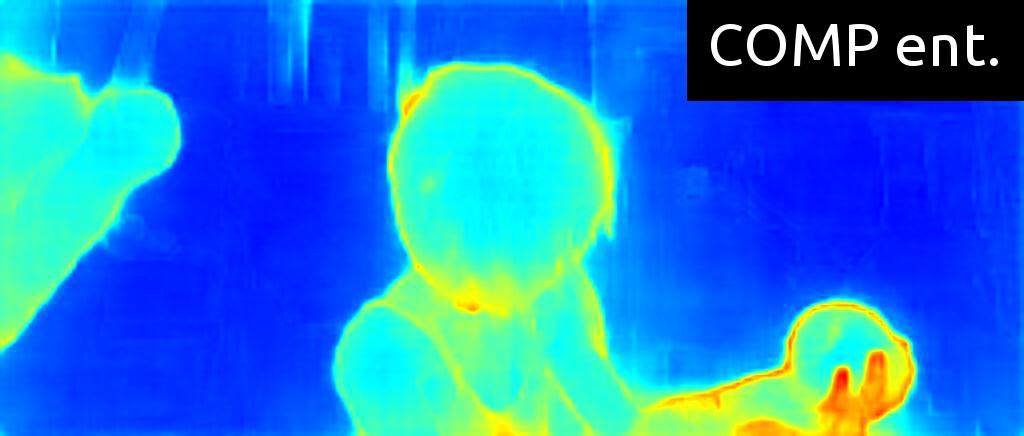}
\\

\includegraphics[width=0.1666\linewidth]{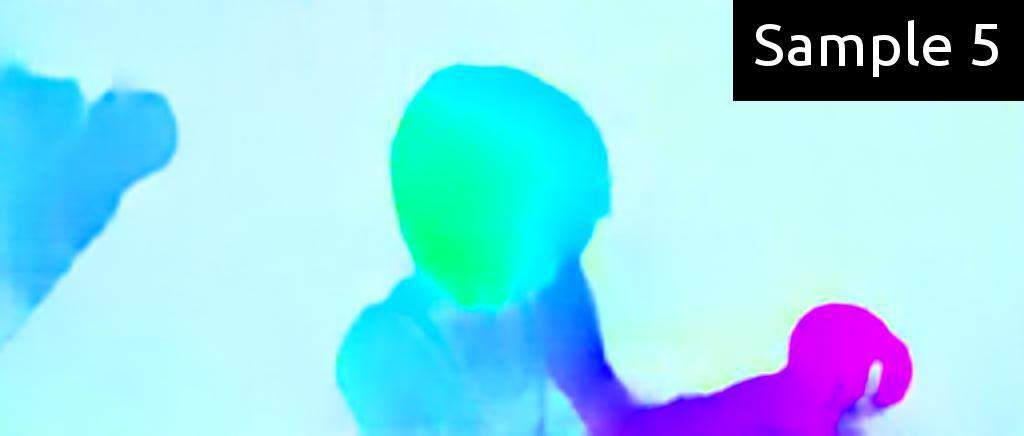}
&
\includegraphics[width=0.1666\linewidth]{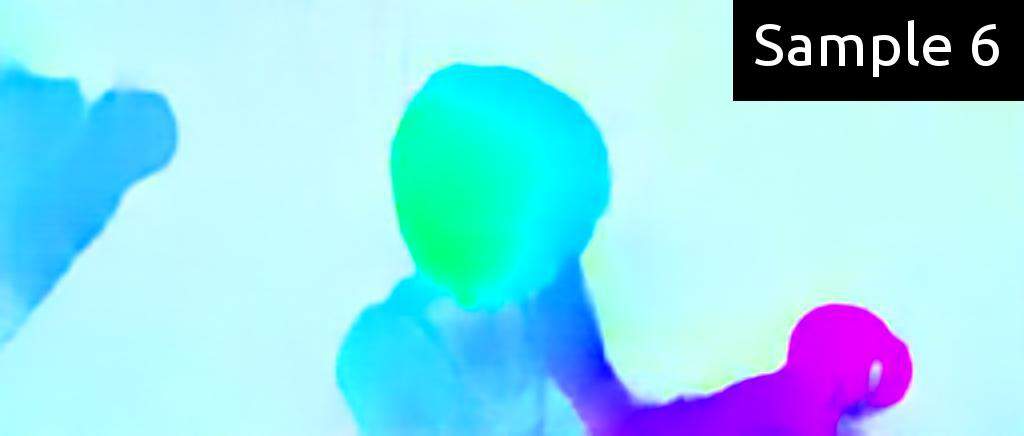}
&
\includegraphics[width=0.1666\linewidth]{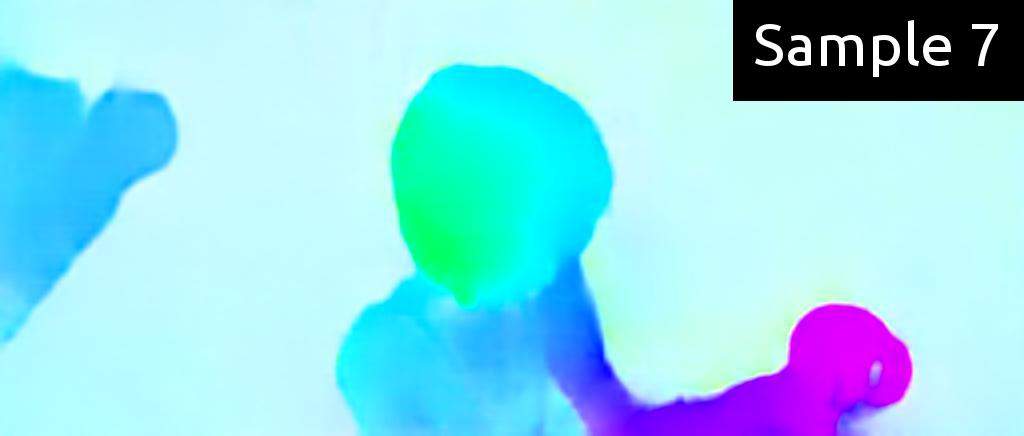}
&
\includegraphics[width=0.1666\linewidth]{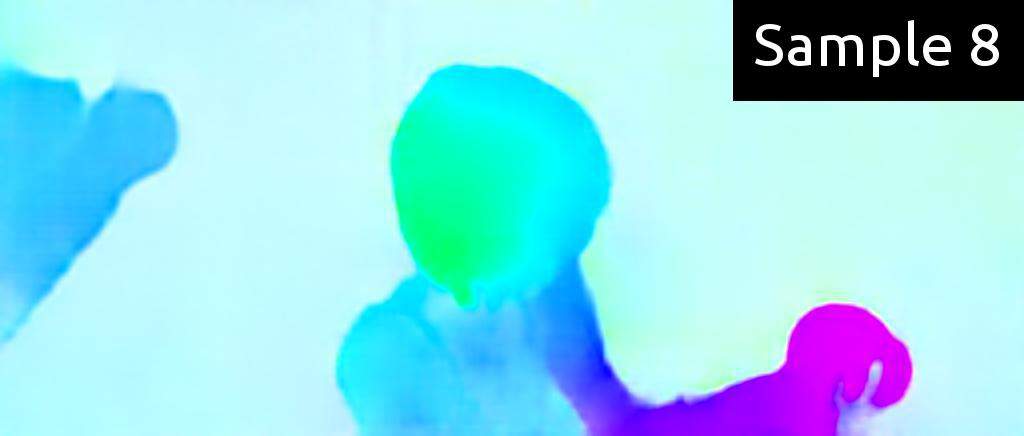}
&
&
\\

\hline
\hline
\multicolumn{4}{|l|}{BootstrappedEnsemble Pred:}& & \\ 
\includegraphics[width=0.1666\linewidth]{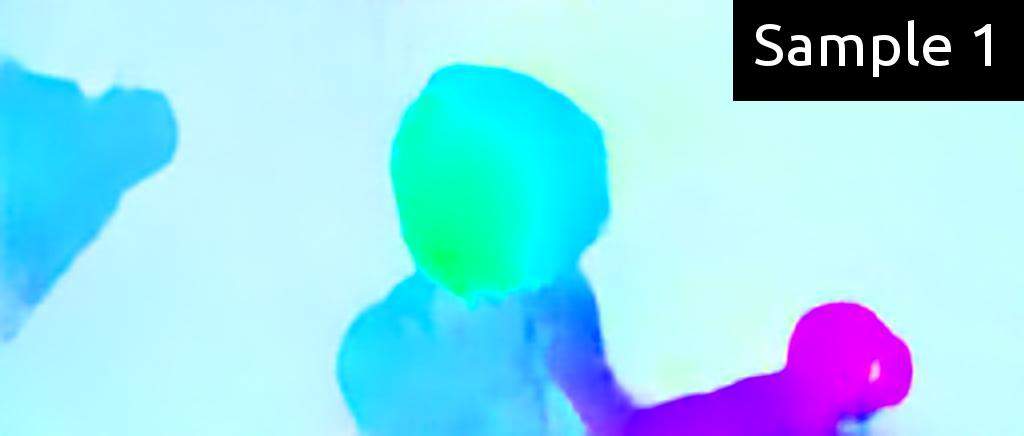}
&
\includegraphics[width=0.1666\linewidth]{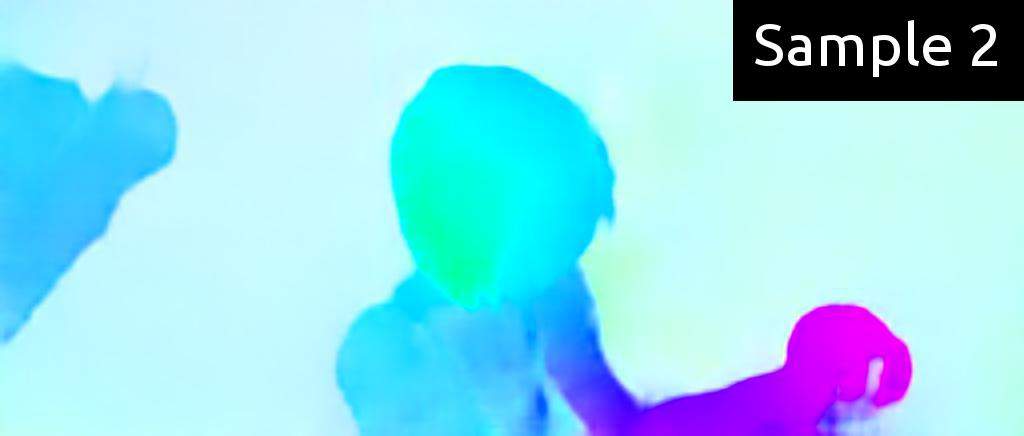}
&
\includegraphics[width=0.1666\linewidth]{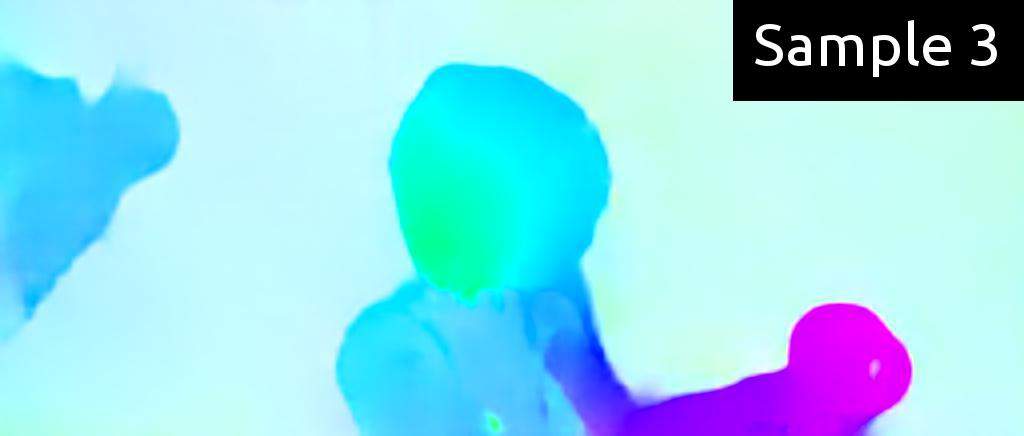}
&
\includegraphics[width=0.1666\linewidth]{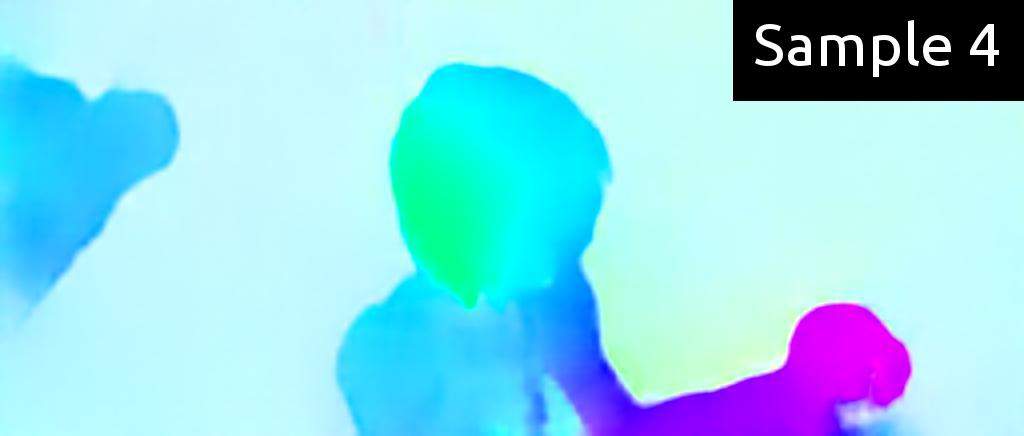}
&
\includegraphics[width=0.1666\linewidth]{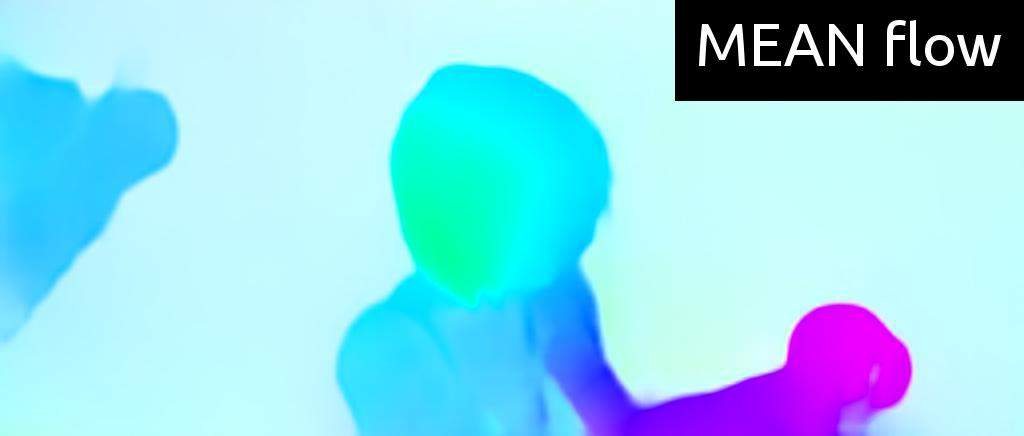}
&
\includegraphics[width=0.1666\linewidth]{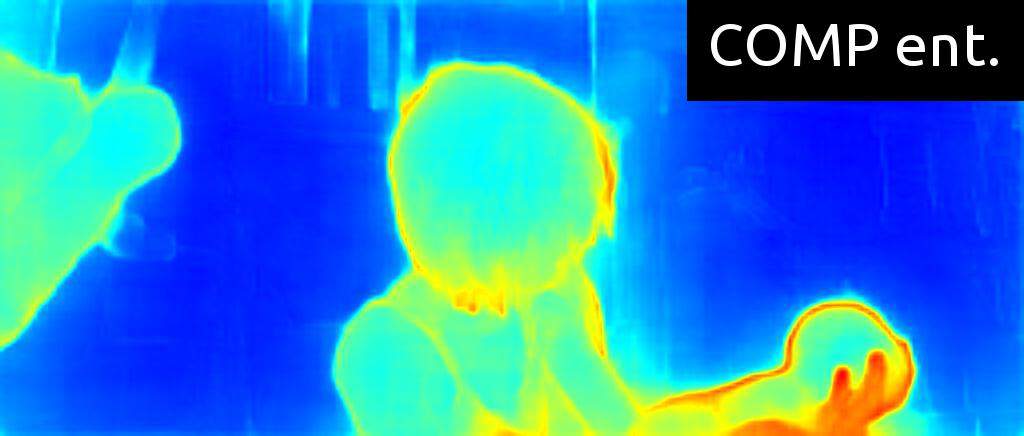}
\\

\includegraphics[width=0.1666\linewidth]{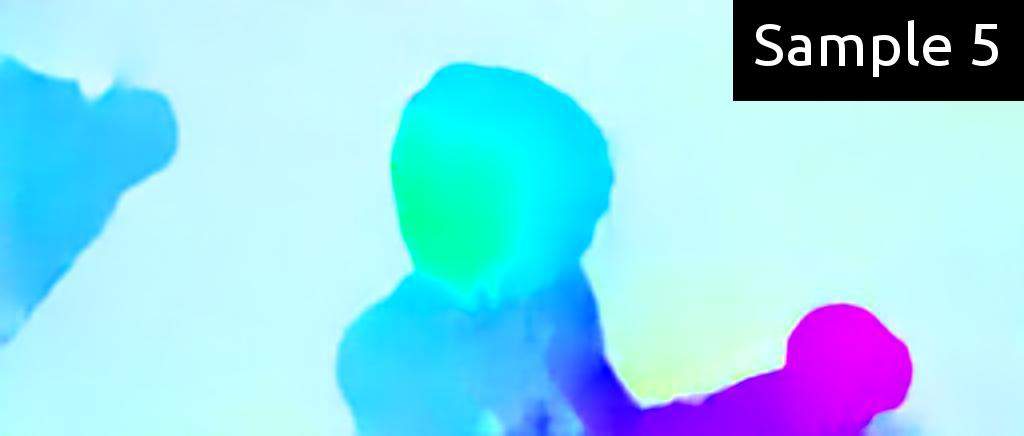}
&
\includegraphics[width=0.1666\linewidth]{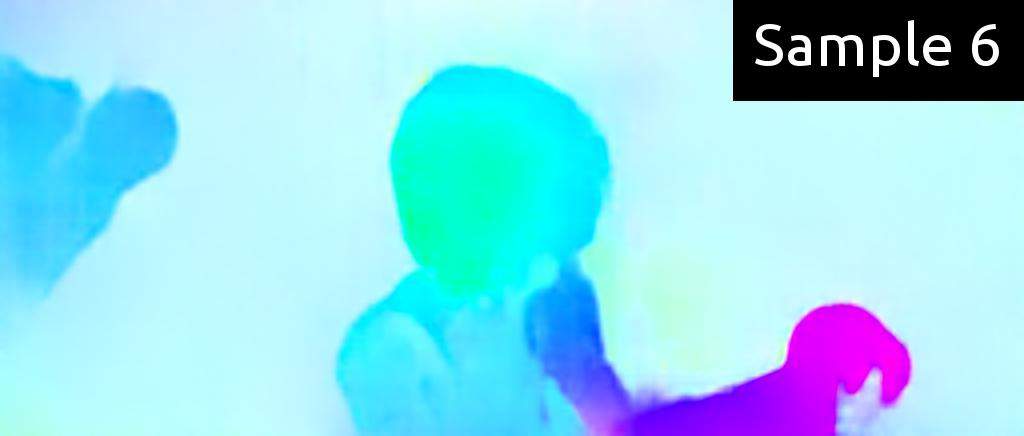}
&
\includegraphics[width=0.1666\linewidth]{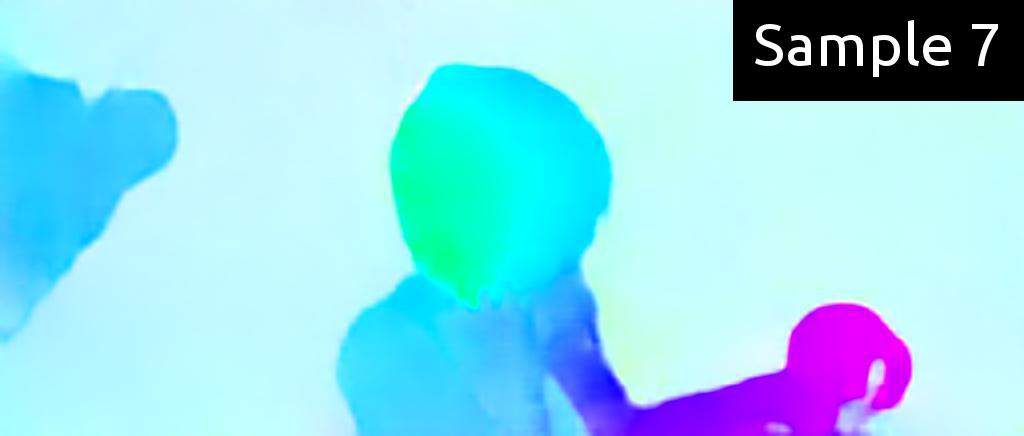}
&
\includegraphics[width=0.1666\linewidth]{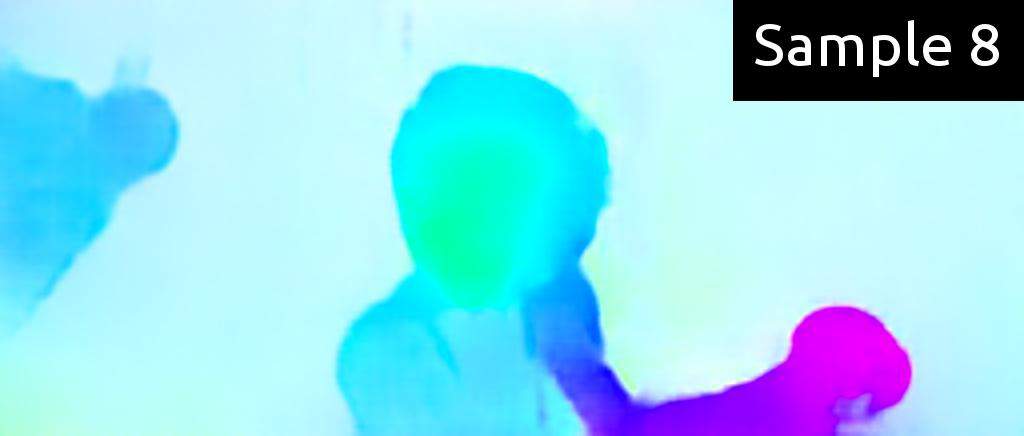}
&
\includegraphics[width=0.1666\linewidth]{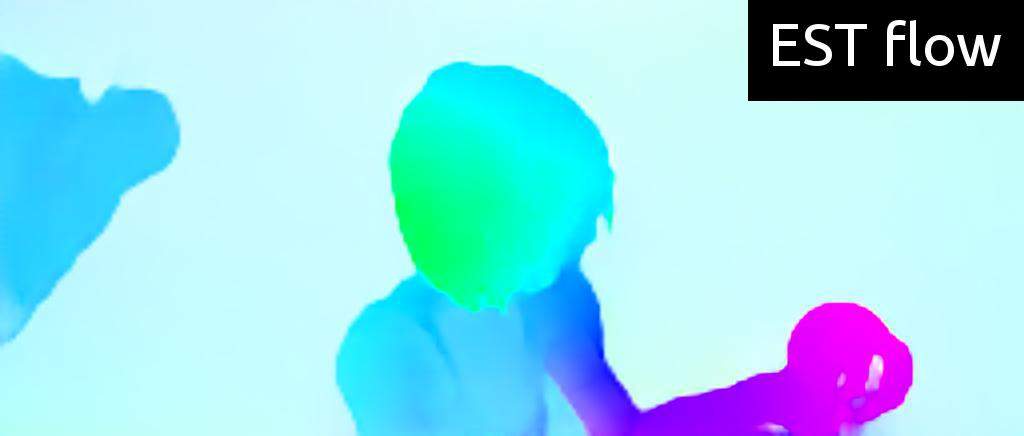}
&
\includegraphics[width=0.1666\linewidth]{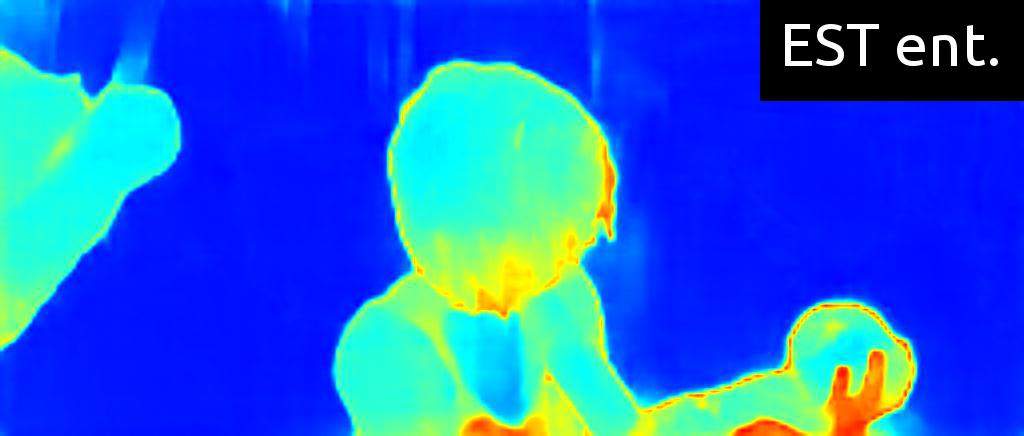}
\\

\hline
\hline
\multicolumn{4}{|l|}{FlowNetH Pred-Merged:}& & \\ 
\includegraphics[width=0.1666\linewidth]{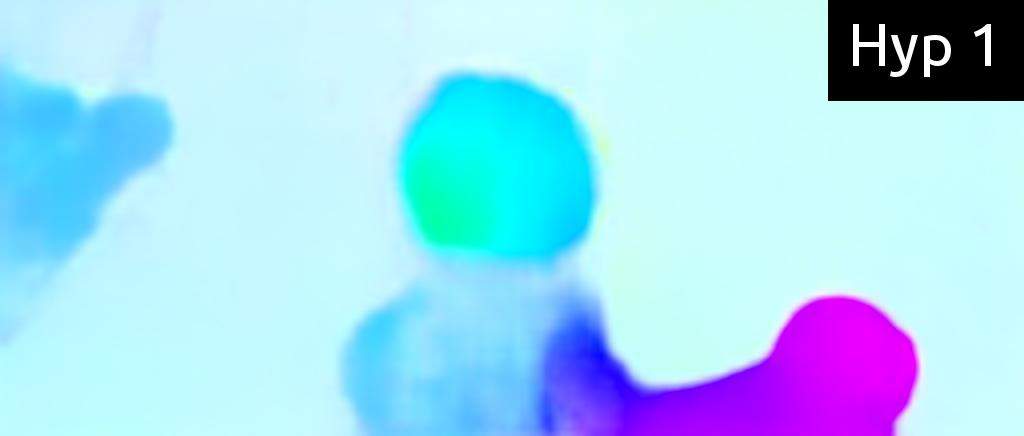}
&
\includegraphics[width=0.1666\linewidth]{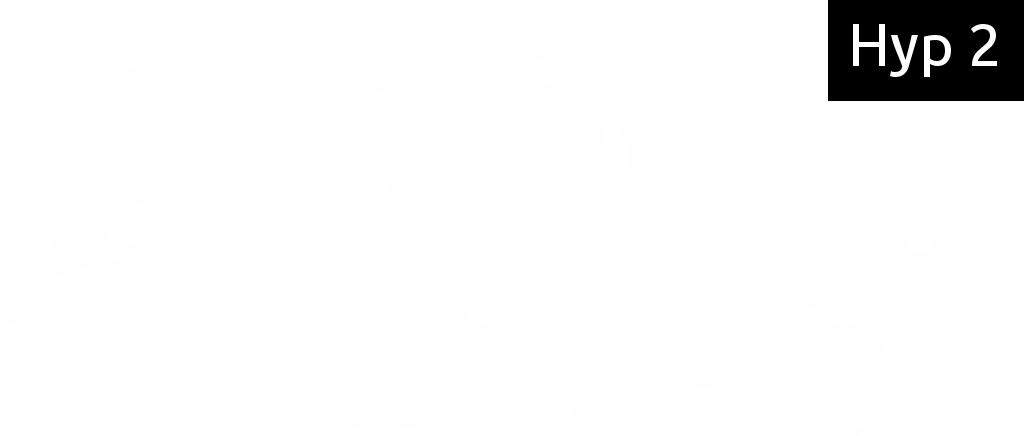}
&
\includegraphics[width=0.1666\linewidth]{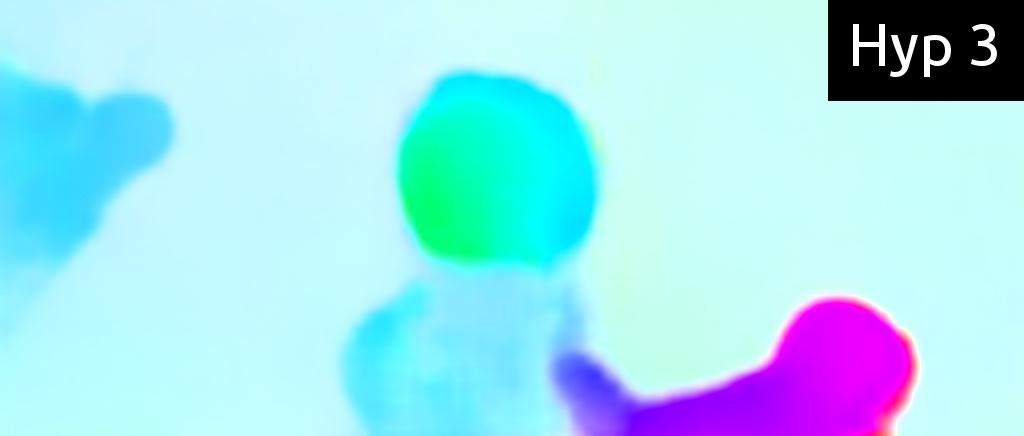}
&
\includegraphics[width=0.1666\linewidth]{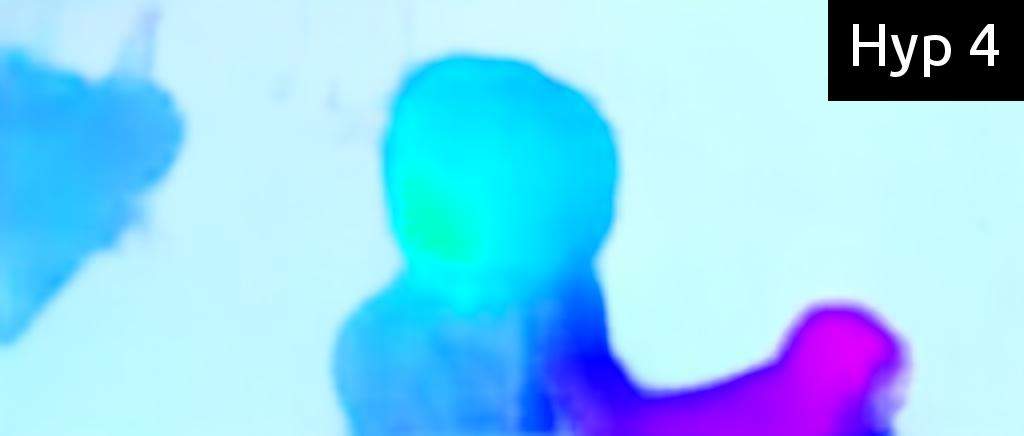}
&
&
\\

\includegraphics[width=0.1666\linewidth]{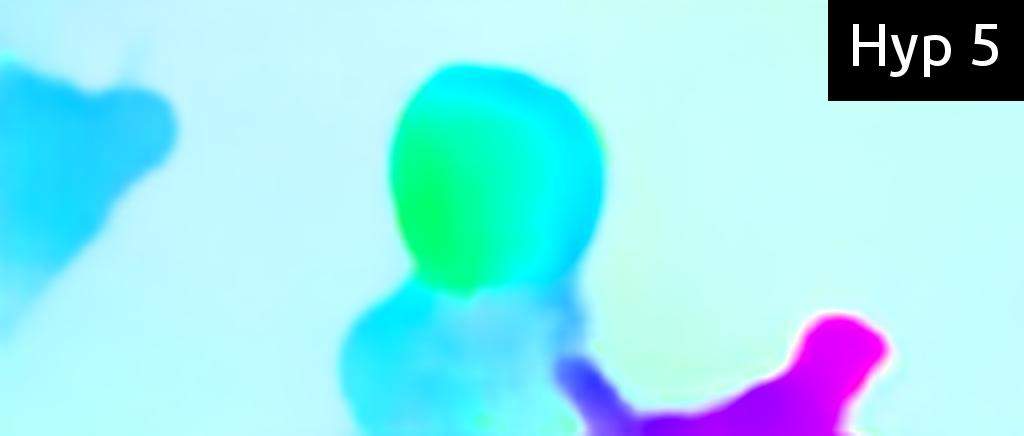}
&
\includegraphics[width=0.1666\linewidth]{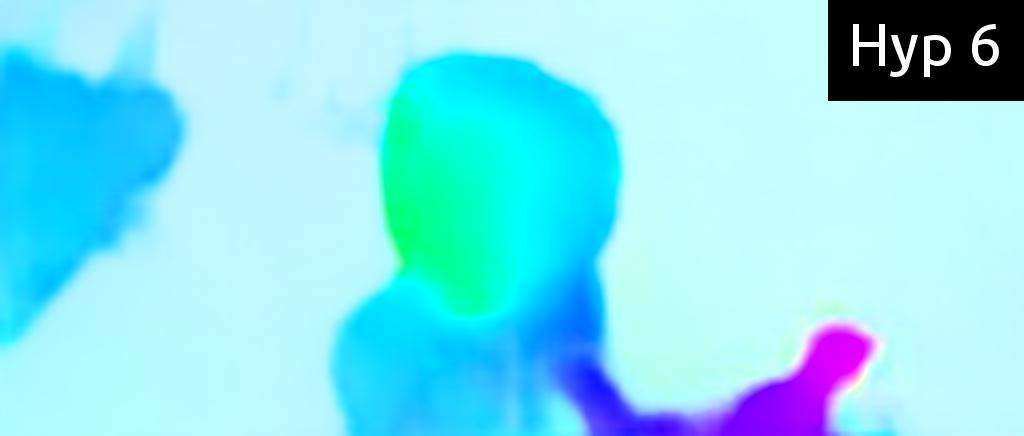}
&
\includegraphics[width=0.1666\linewidth]{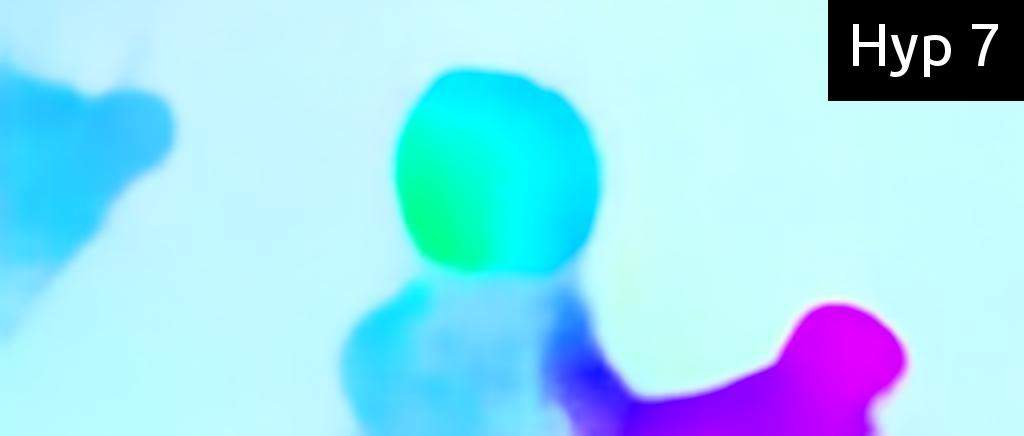}
&
\includegraphics[width=0.1666\linewidth]{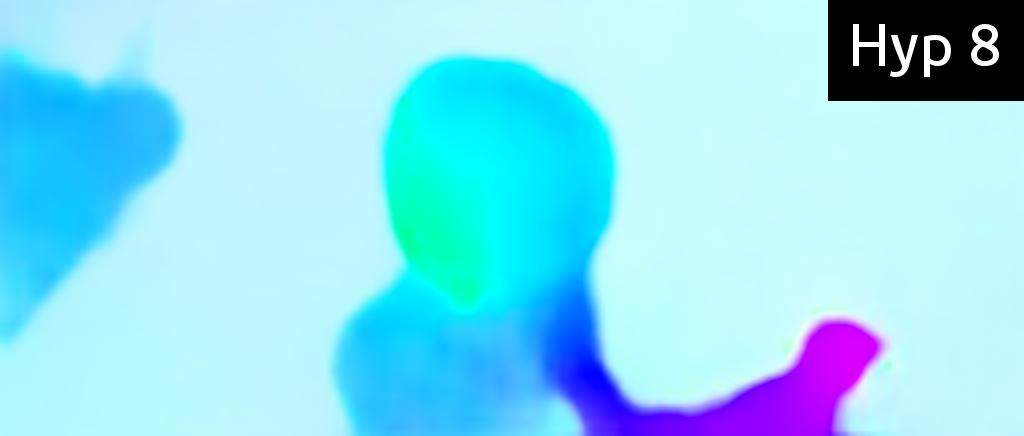}
&
\includegraphics[width=0.1666\linewidth]{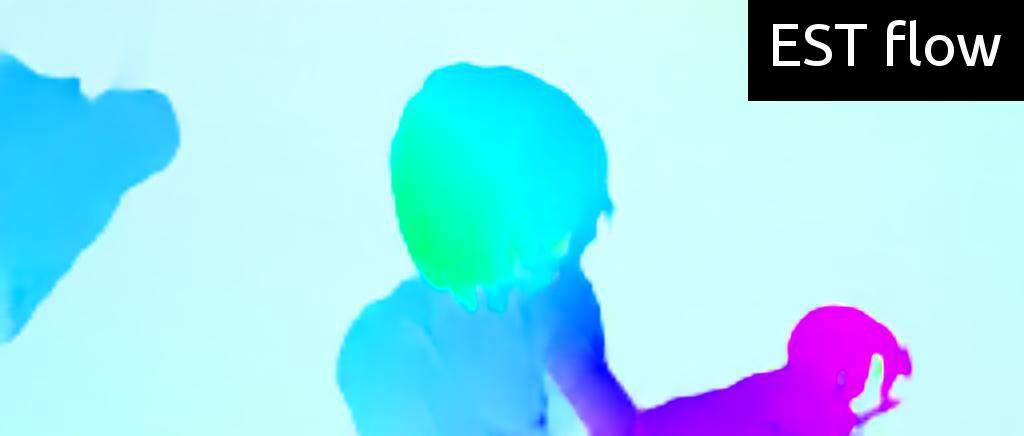}
&
\includegraphics[width=0.1666\linewidth]{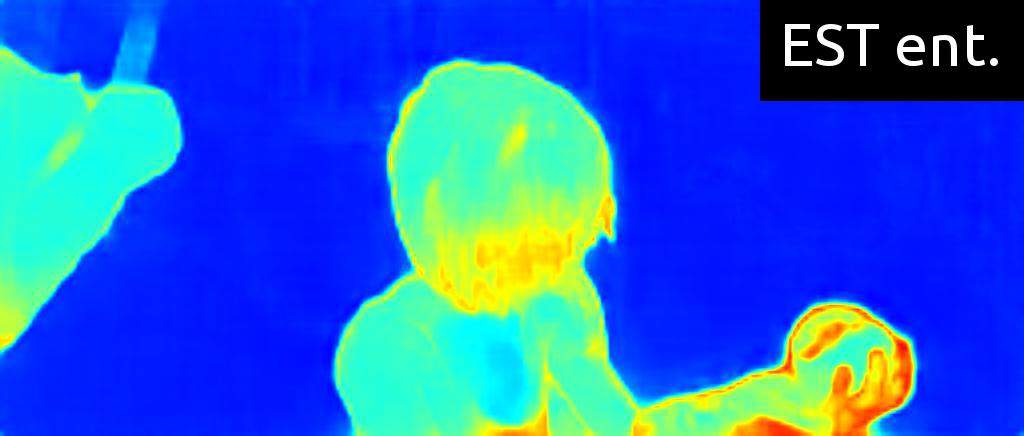}
\\

\hline

                \end{tabular}
            }
       \end{center}
             \caption{
            In this table we show the outputs of predictive experiments with all presented methods for an easy Sintel example as well as the averaged flows and computed entropies. For Bootstrapped-Ensemble-Pred-Merged and FlowNetH-Pred-Merged we show also the estimated flow and estimated entropy as the output of the merging network on top.
            }
            \label{tab:ex2_2}
        \end{table*}
        \egroup

        \bgroup
        \def\arraystretch{1.0}
        \renewcommand{\tabcolsep}{0.05cm}
        \begin{table*}
        \begin{center}
            \resizebox{\linewidth}{!}{%
                \begin{tabular}{|cccc|cc|}
        
\hline
\multicolumn{4}{|l|}{Data:}& & \\ 
\includegraphics[width=0.1666\linewidth]{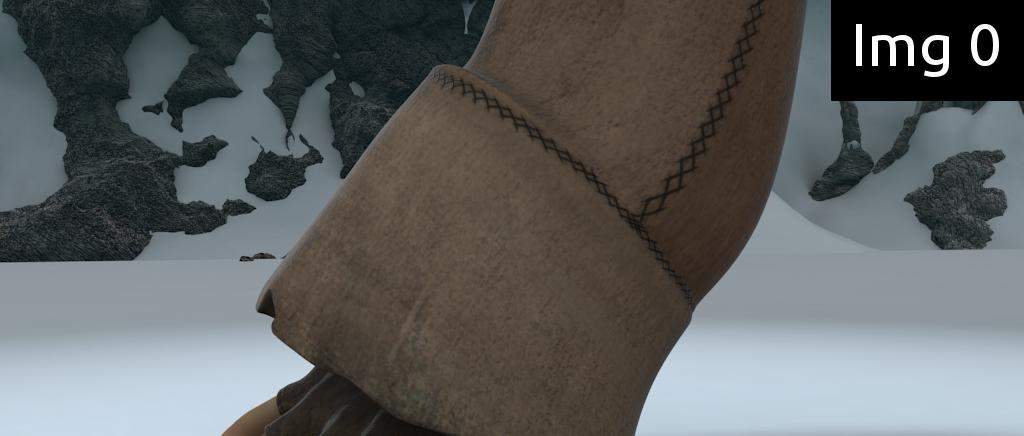}
&
\includegraphics[width=0.1666\linewidth]{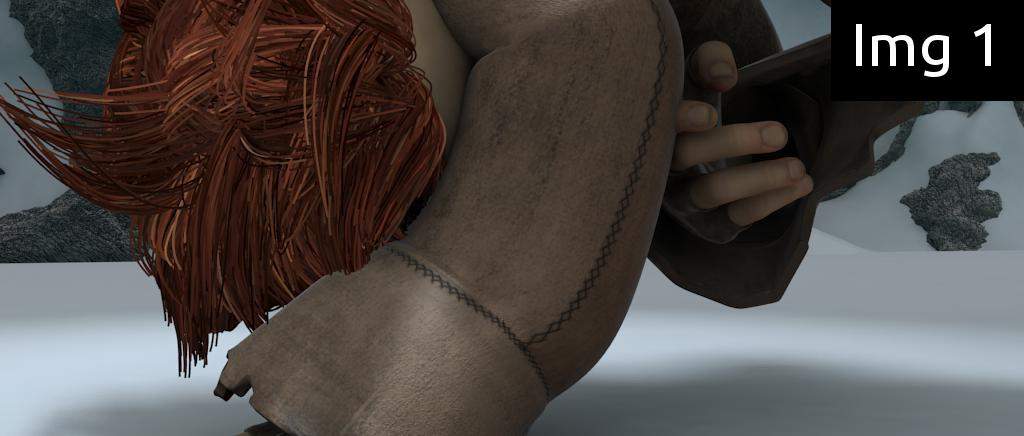}
&
&
&
\includegraphics[width=0.1666\linewidth]{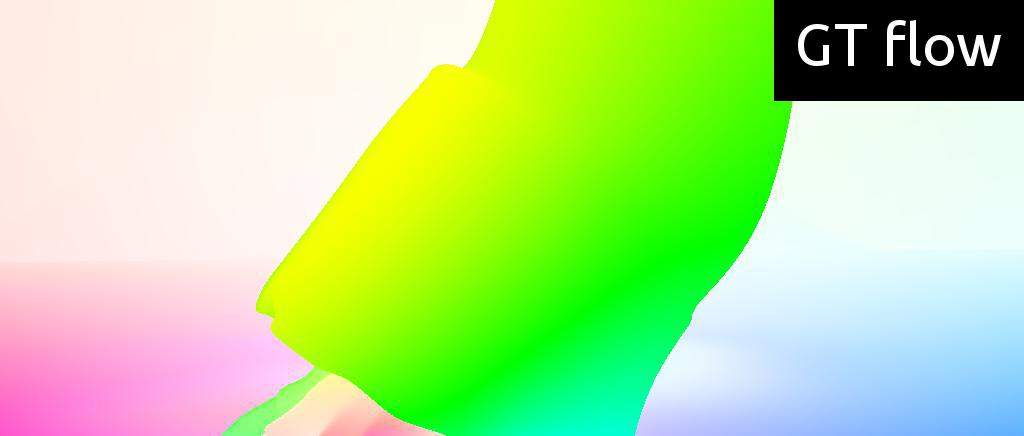}
&
\\

\hline
\hline
\multicolumn{4}{|l|}{FlowNetC Emp:}& & \\ 
&
&
&
&
\includegraphics[width=0.1666\linewidth]{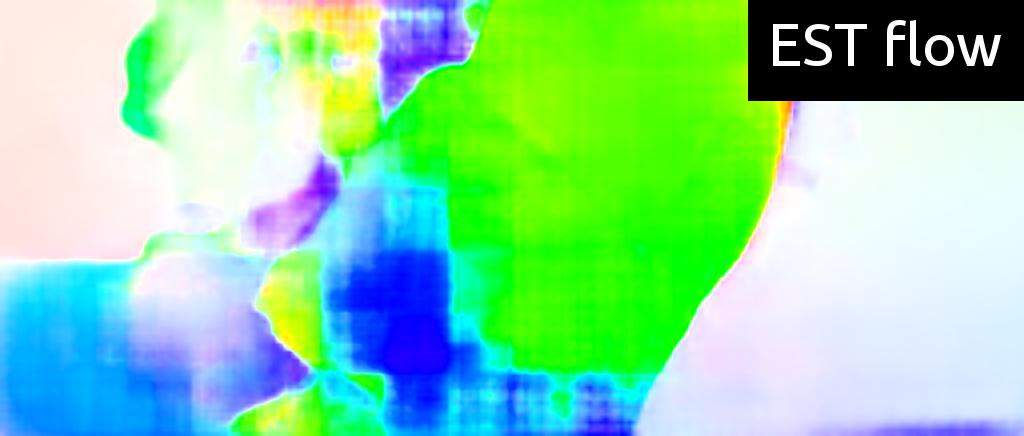}
&
\\

\hline
\hline
\multicolumn{4}{|l|}{FlowNetH Base:}& & \\ 
\includegraphics[width=0.1666\linewidth]{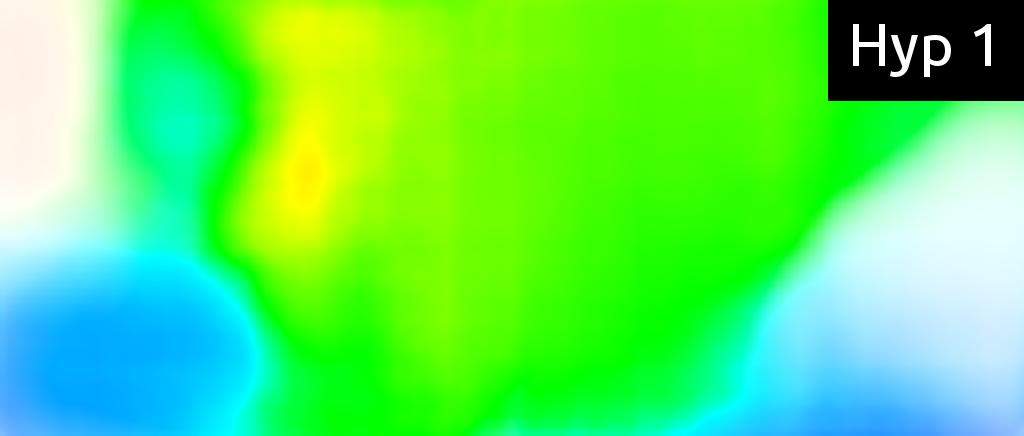}
&
\includegraphics[width=0.1666\linewidth]{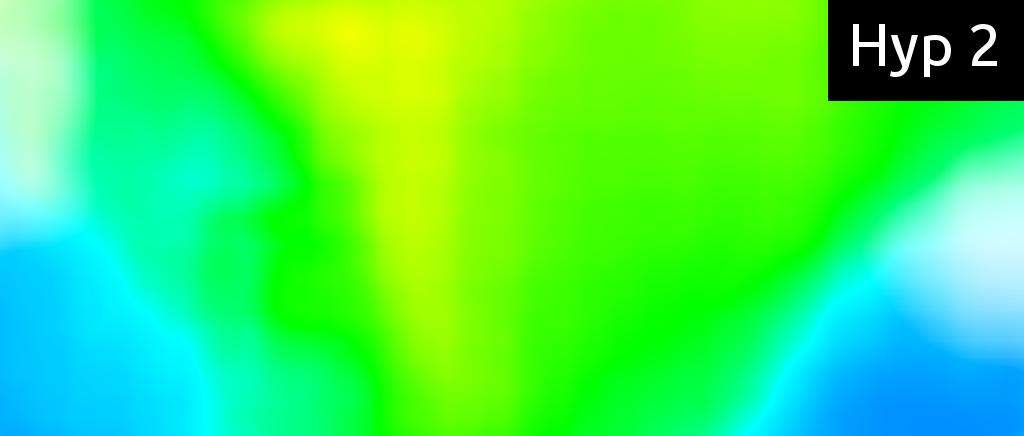}
&
\includegraphics[width=0.1666\linewidth]{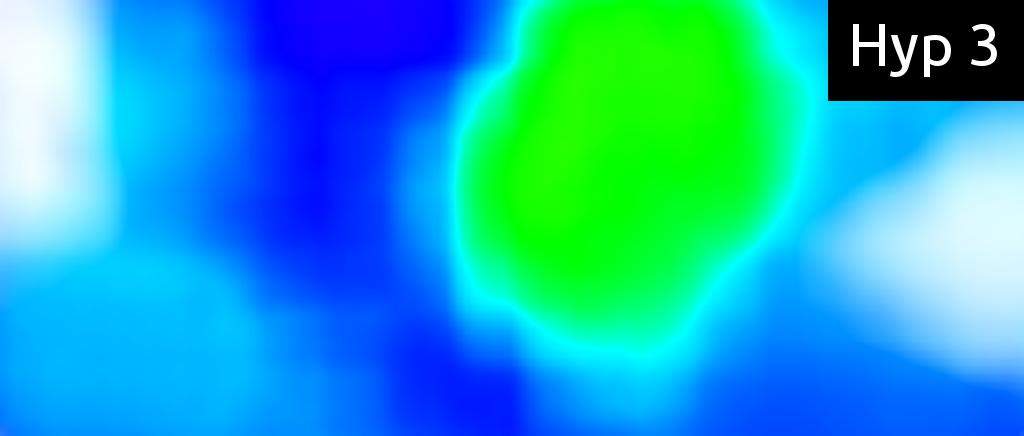}
&
\includegraphics[width=0.1666\linewidth]{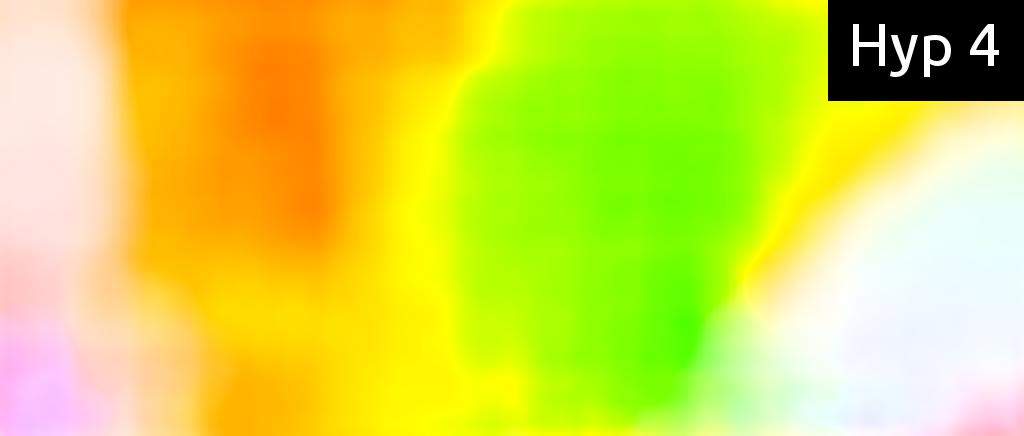}
&
\includegraphics[width=0.1666\linewidth]{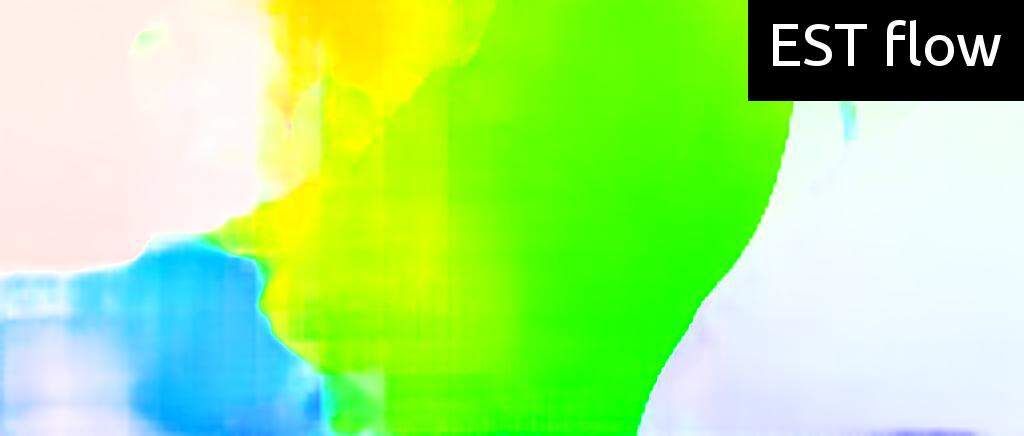}
&
\\

\includegraphics[width=0.1666\linewidth]{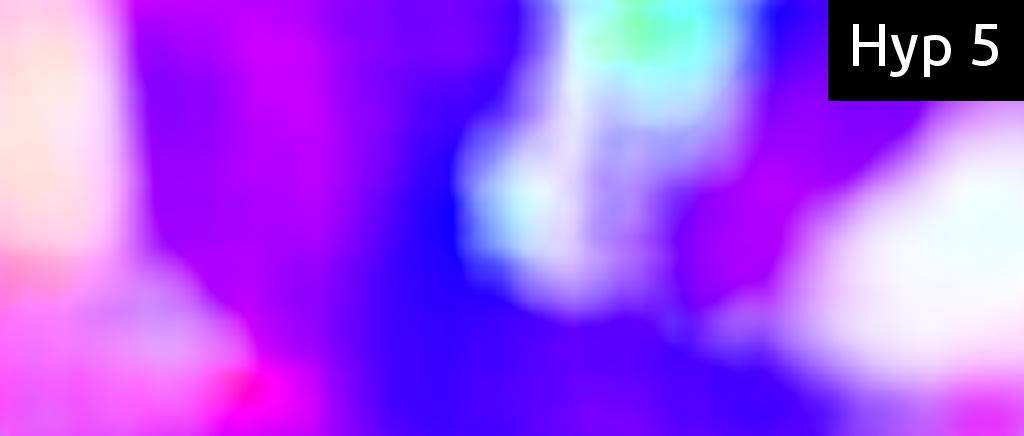}
&
\includegraphics[width=0.1666\linewidth]{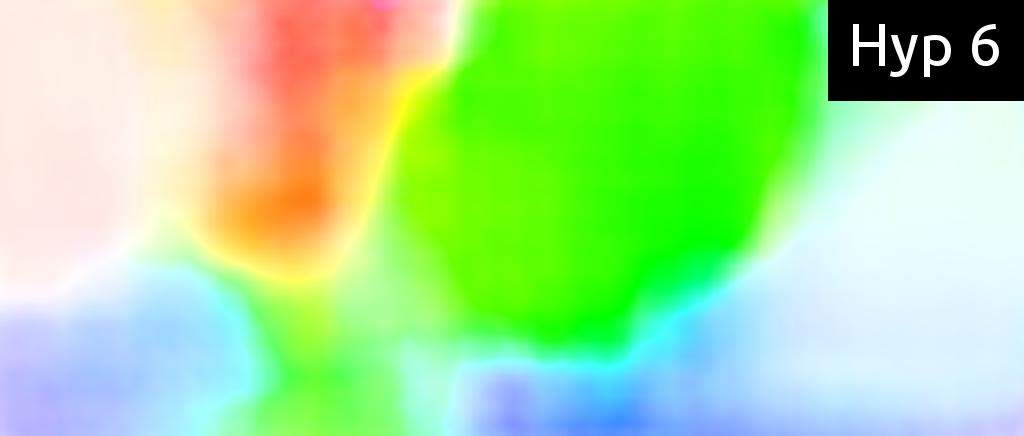}
&
\includegraphics[width=0.1666\linewidth]{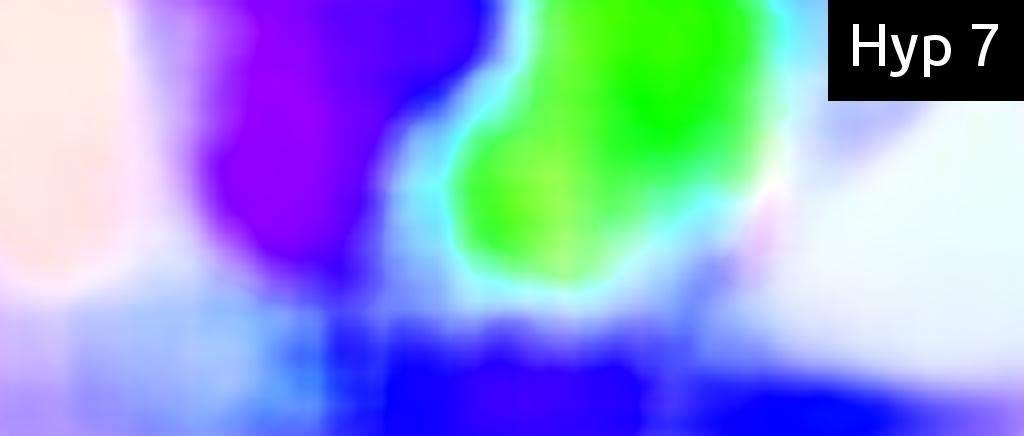}
&
\includegraphics[width=0.1666\linewidth]{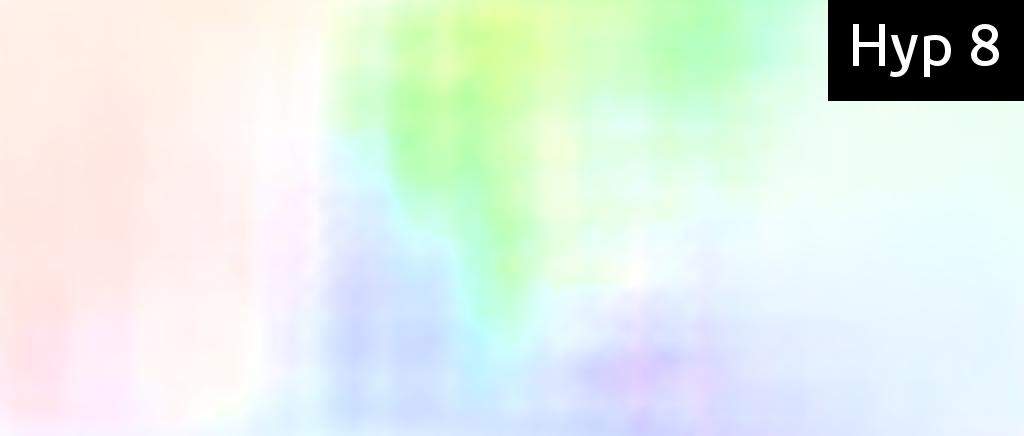}
&
&
\\

\hline
\hline
\multicolumn{4}{|l|}{Dropout Emp:}& & \\ 
\includegraphics[width=0.1666\linewidth]{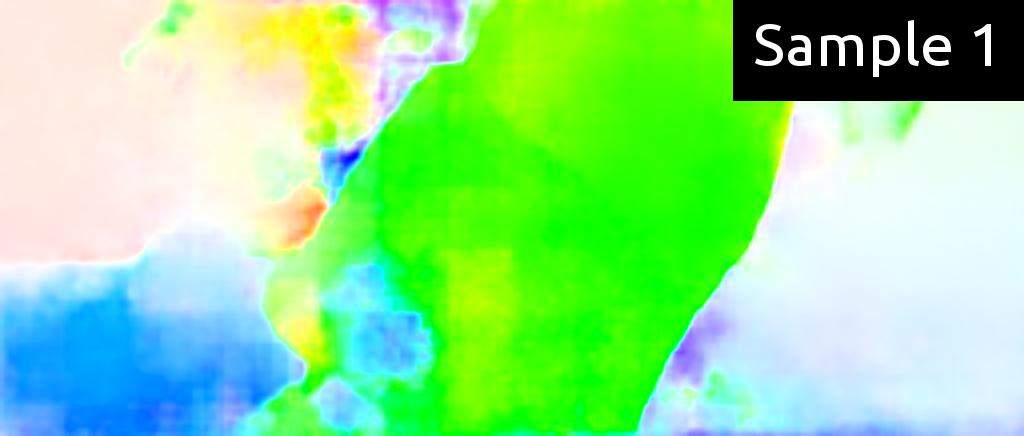}
&
\includegraphics[width=0.1666\linewidth]{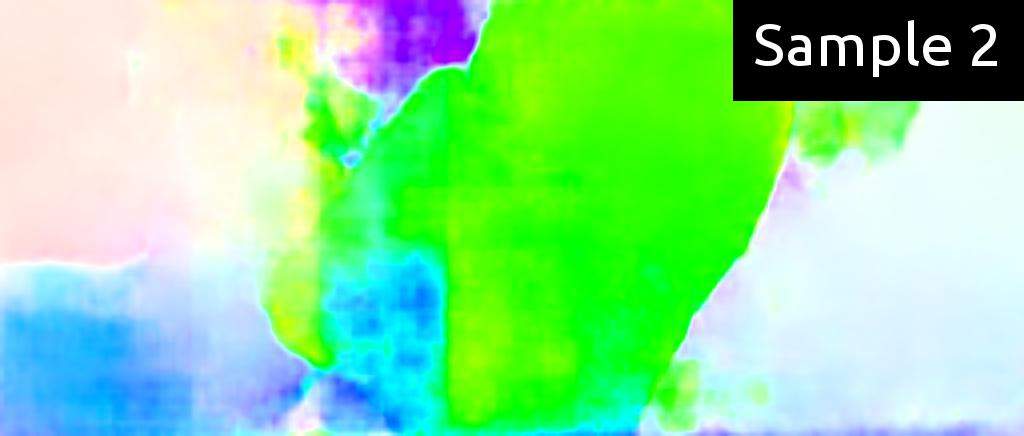}
&
\includegraphics[width=0.1666\linewidth]{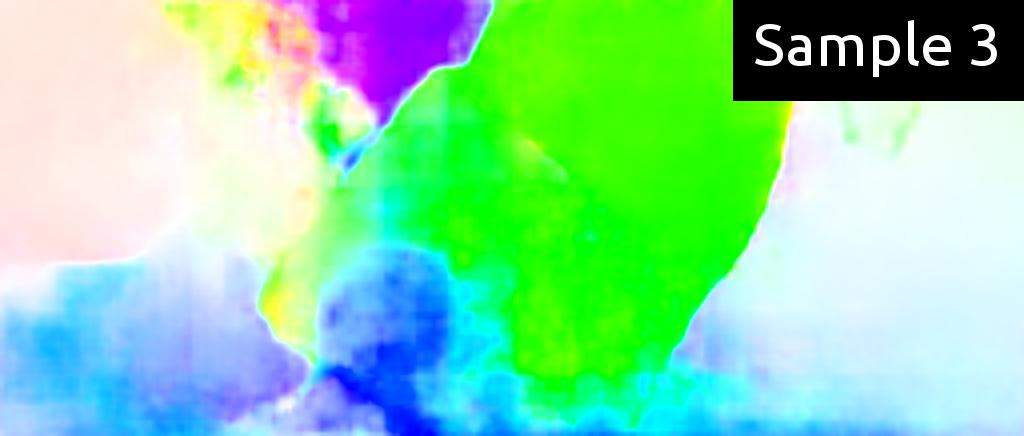}
&
\includegraphics[width=0.1666\linewidth]{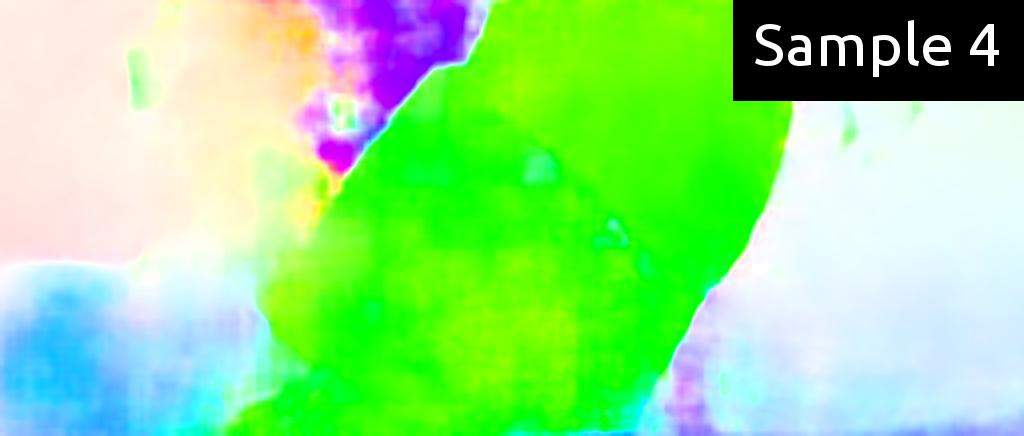}
&
\includegraphics[width=0.1666\linewidth]{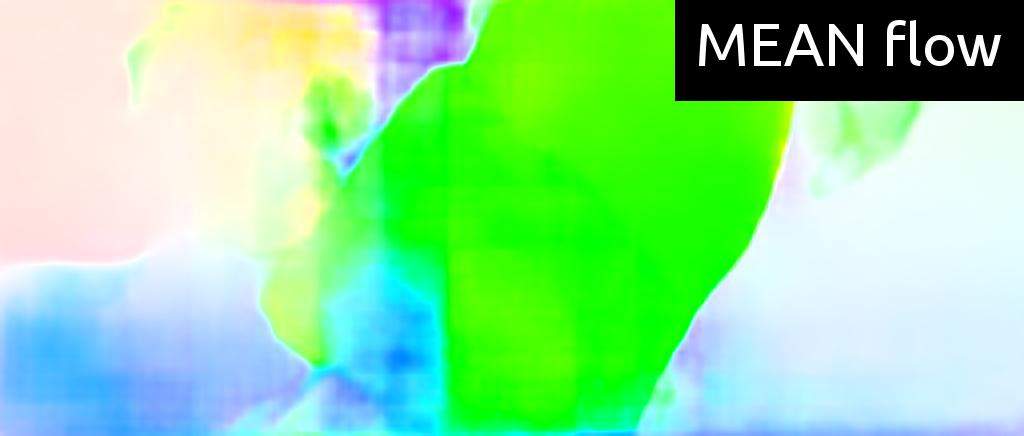}
&
\includegraphics[width=0.1666\linewidth]{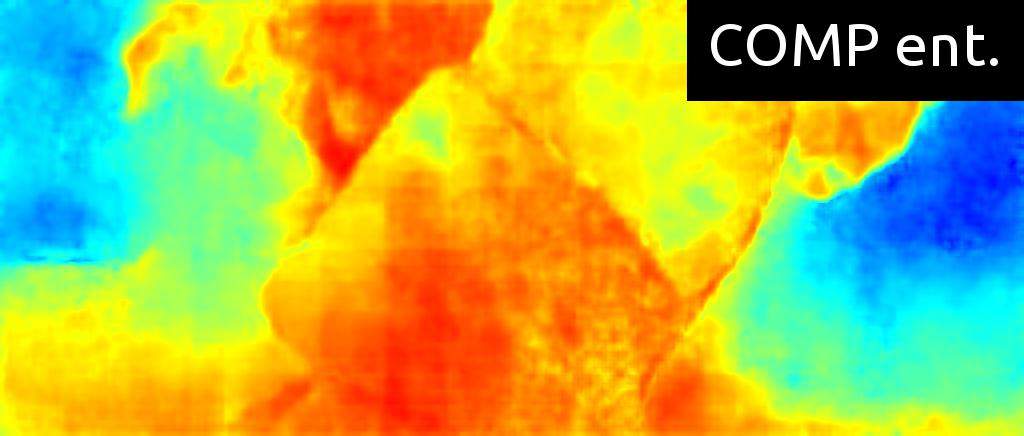}
\\

\includegraphics[width=0.1666\linewidth]{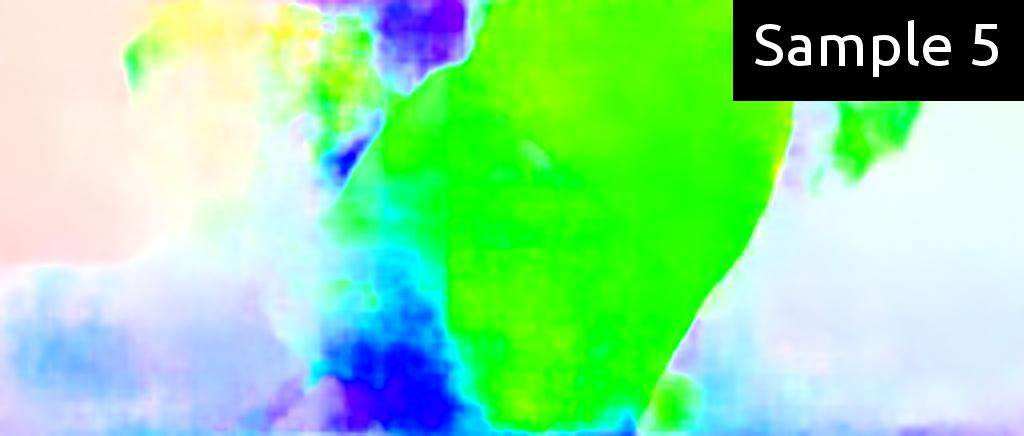}
&
\includegraphics[width=0.1666\linewidth]{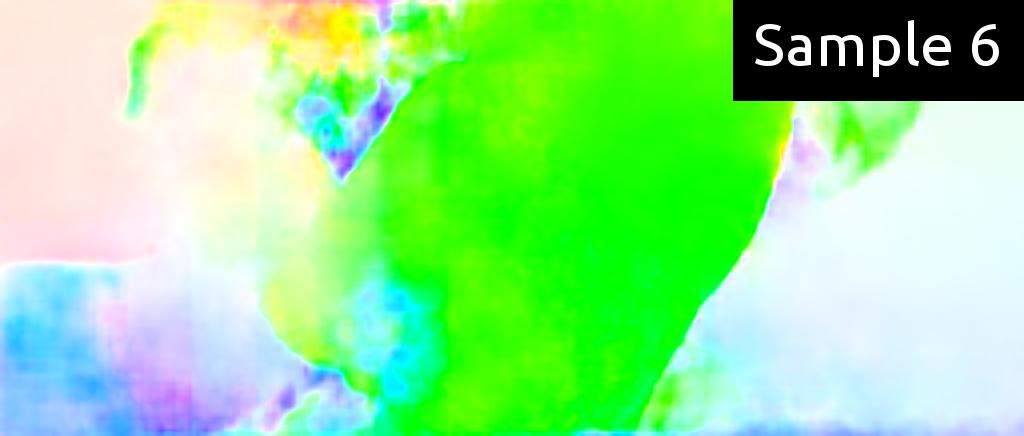}
&
\includegraphics[width=0.1666\linewidth]{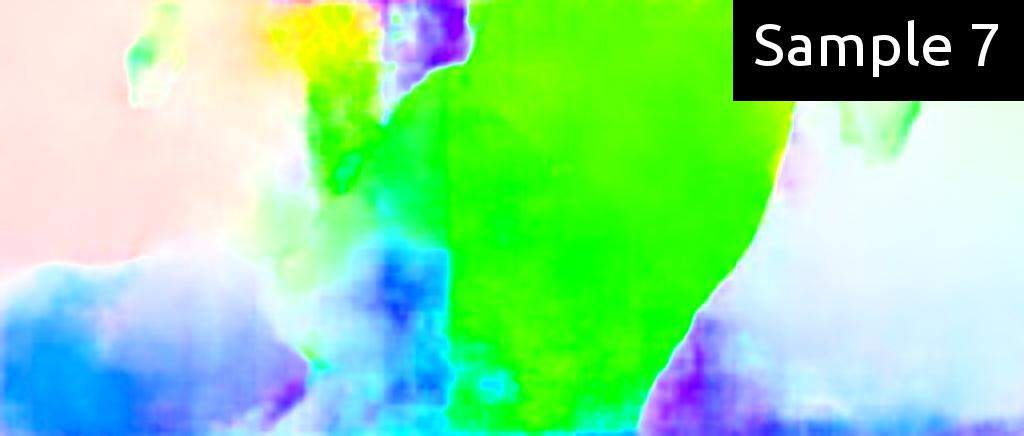}
&
\includegraphics[width=0.1666\linewidth]{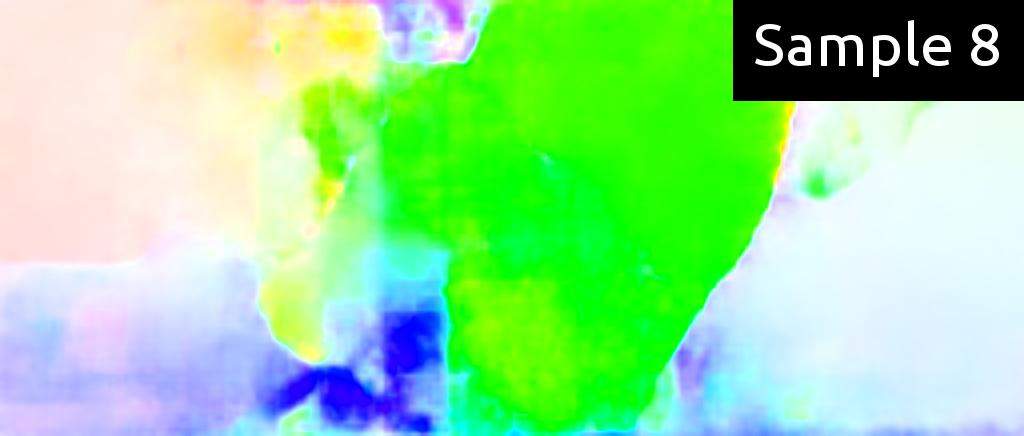}
&
&
\\

\hline
\hline
\multicolumn{4}{|l|}{SGDR Emp:}& & \\ 
\includegraphics[width=0.1666\linewidth]{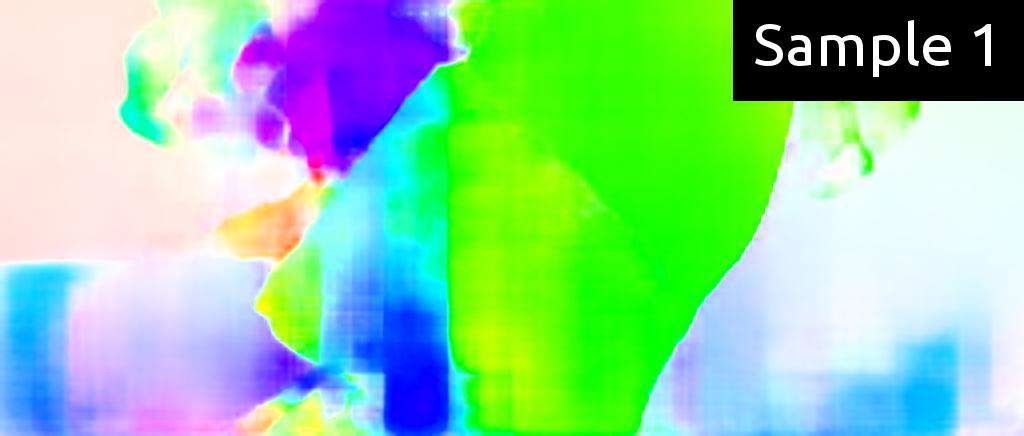}
&
\includegraphics[width=0.1666\linewidth]{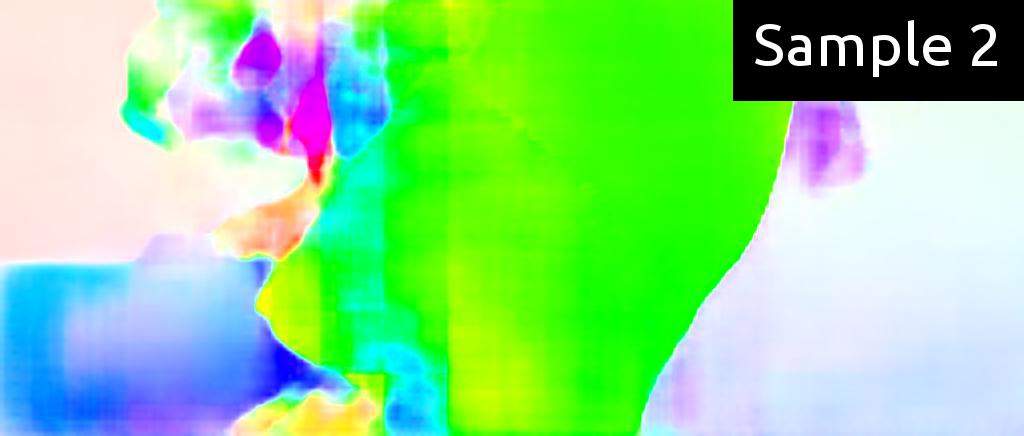}
&
\includegraphics[width=0.1666\linewidth]{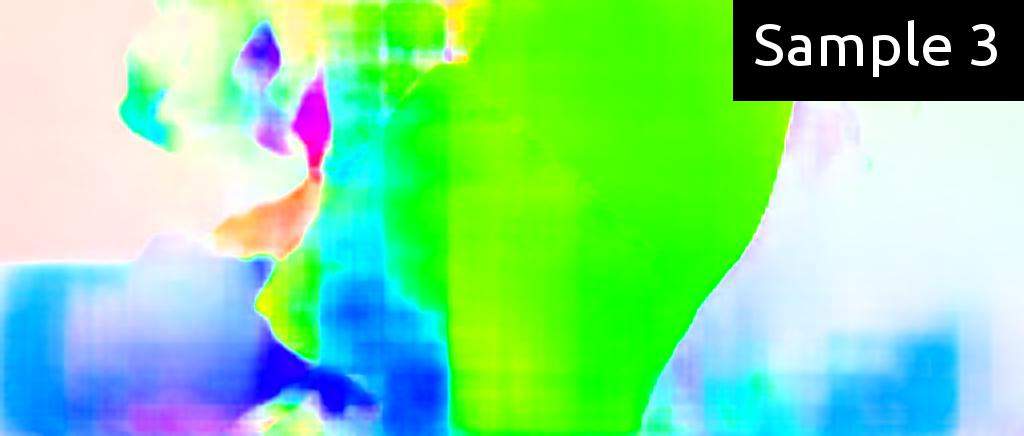}
&
\includegraphics[width=0.1666\linewidth]{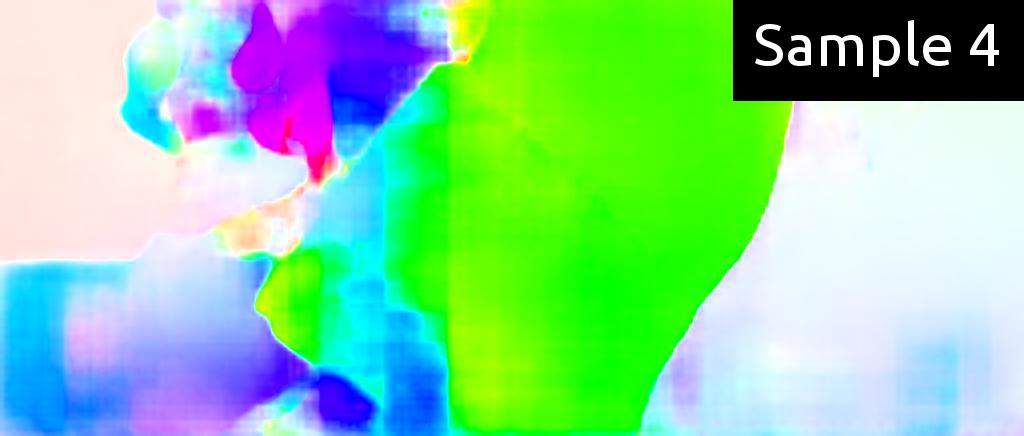}
&
\includegraphics[width=0.1666\linewidth]{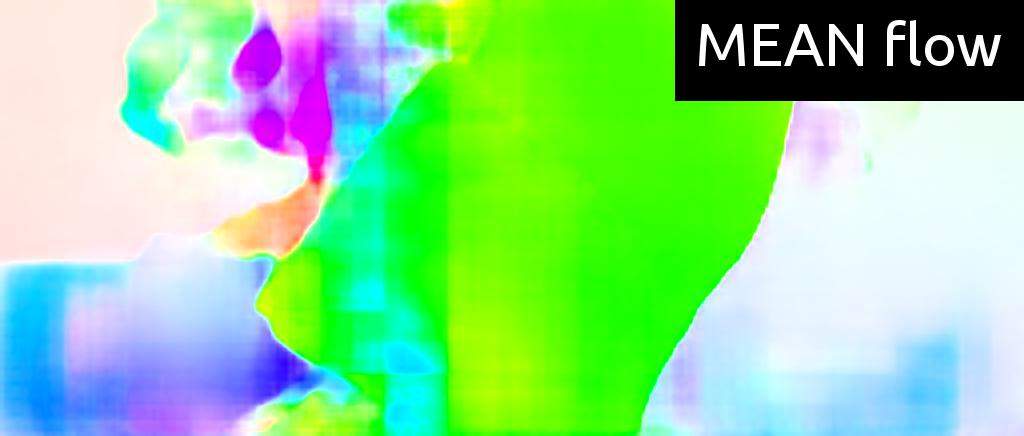}
&
\includegraphics[width=0.1666\linewidth]{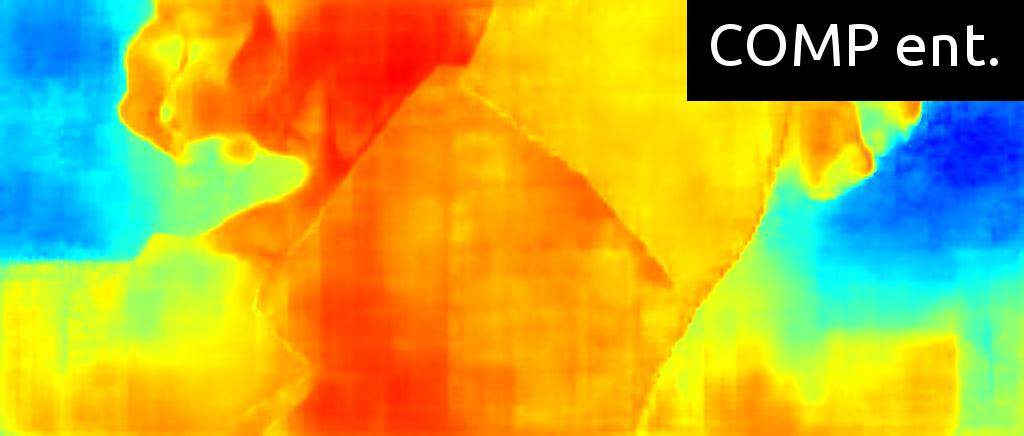}
\\

\includegraphics[width=0.1666\linewidth]{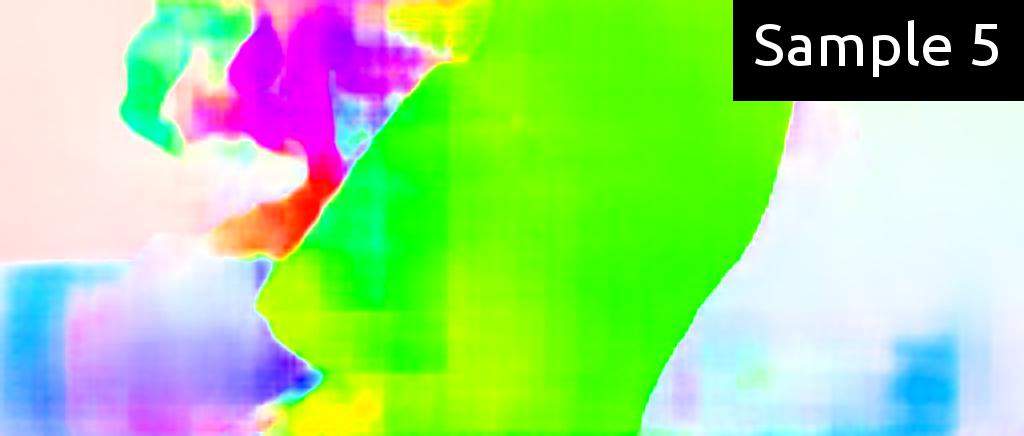}
&
\includegraphics[width=0.1666\linewidth]{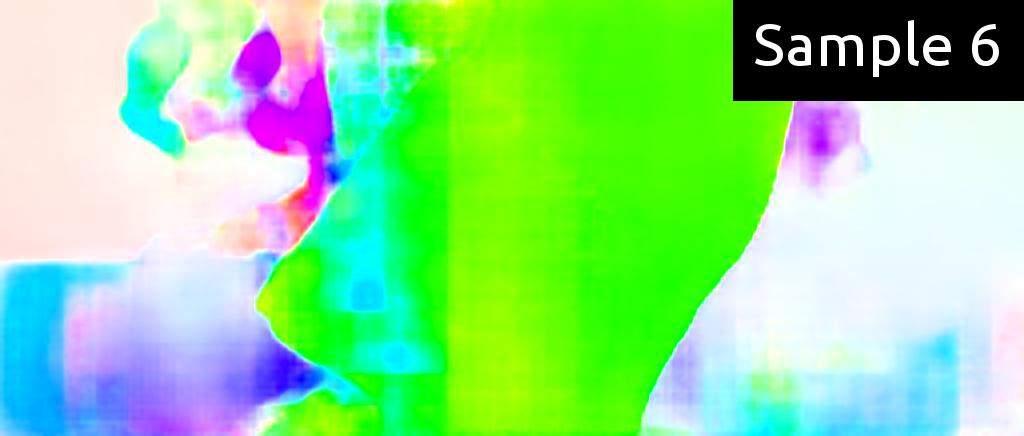}
&
\includegraphics[width=0.1666\linewidth]{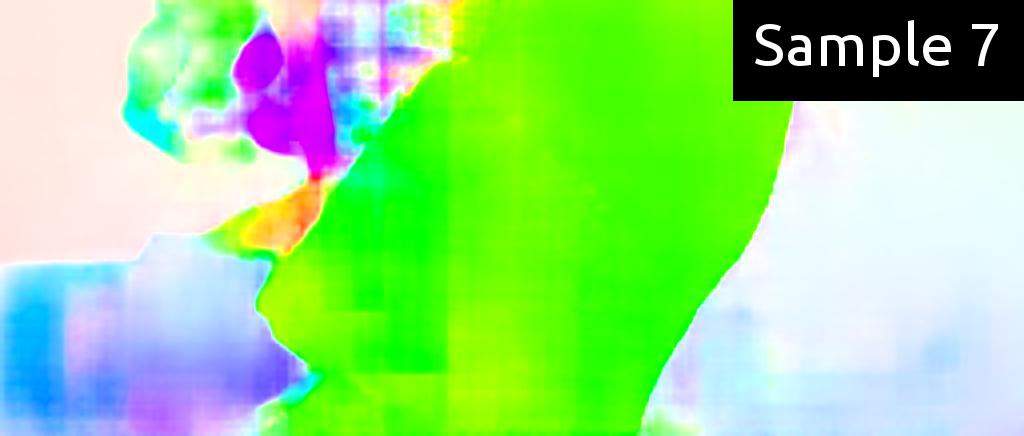}
&
\includegraphics[width=0.1666\linewidth]{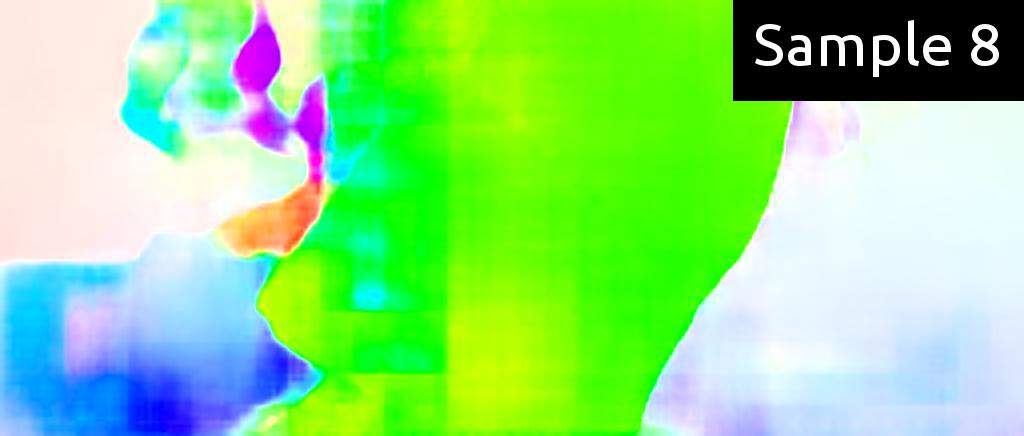}
&
&
\\

\hline
\hline
\multicolumn{4}{|l|}{Bootstrapped Ensemble Emp:}& & \\ 
\includegraphics[width=0.1666\linewidth]{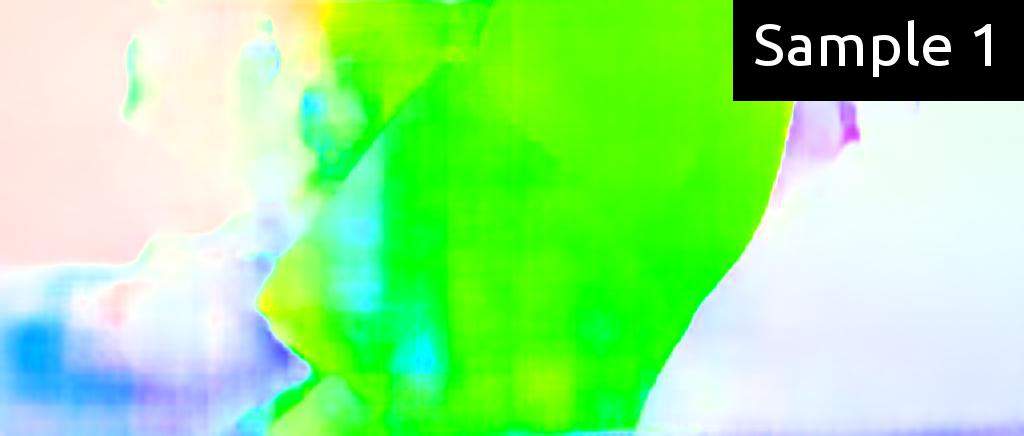}
&
\includegraphics[width=0.1666\linewidth]{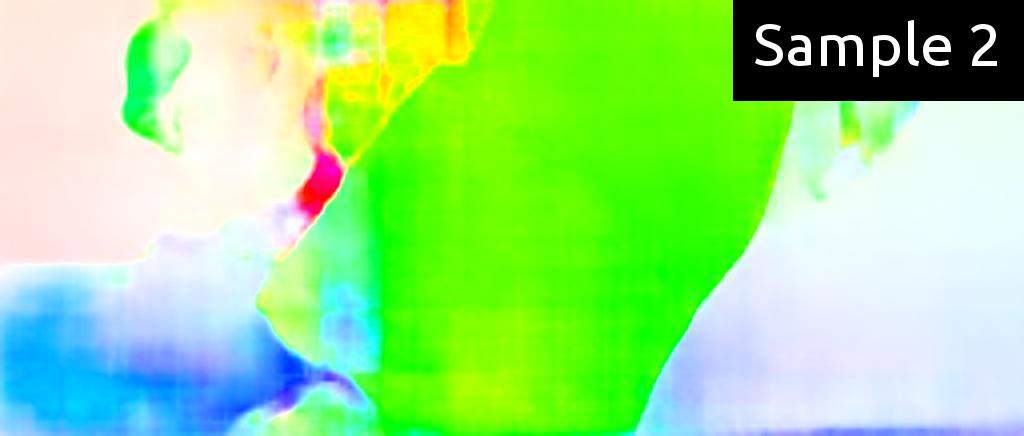}
&
\includegraphics[width=0.1666\linewidth]{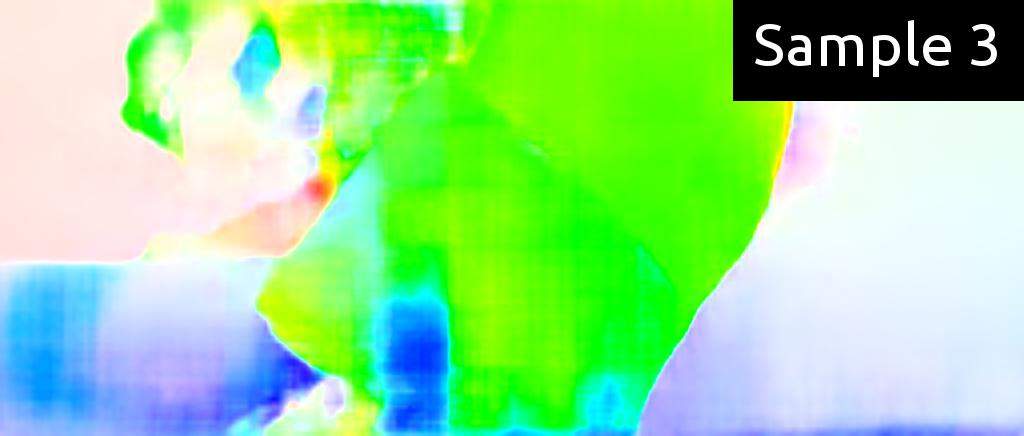}
&
\includegraphics[width=0.1666\linewidth]{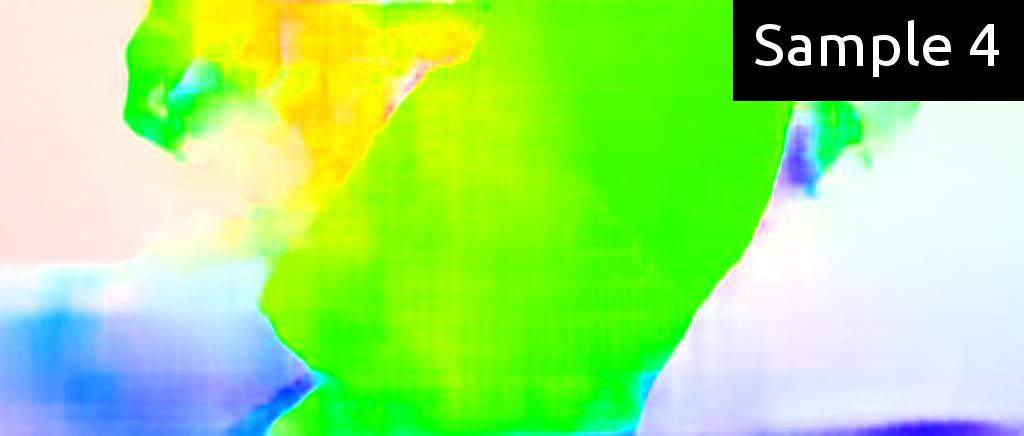}
&
\includegraphics[width=0.1666\linewidth]{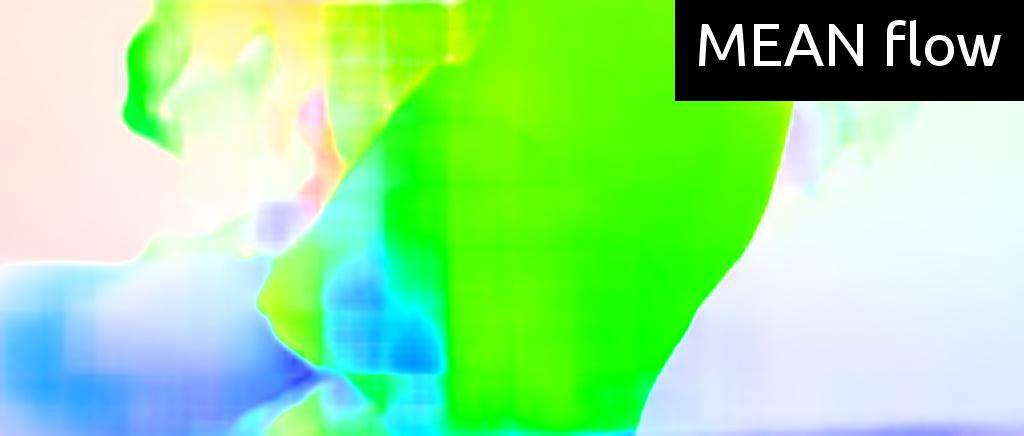}
&
\includegraphics[width=0.1666\linewidth]{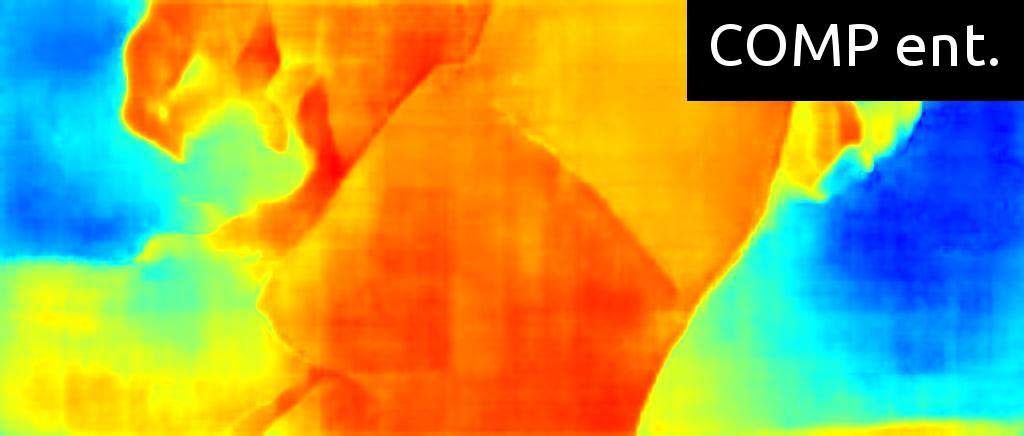}
\\

\includegraphics[width=0.1666\linewidth]{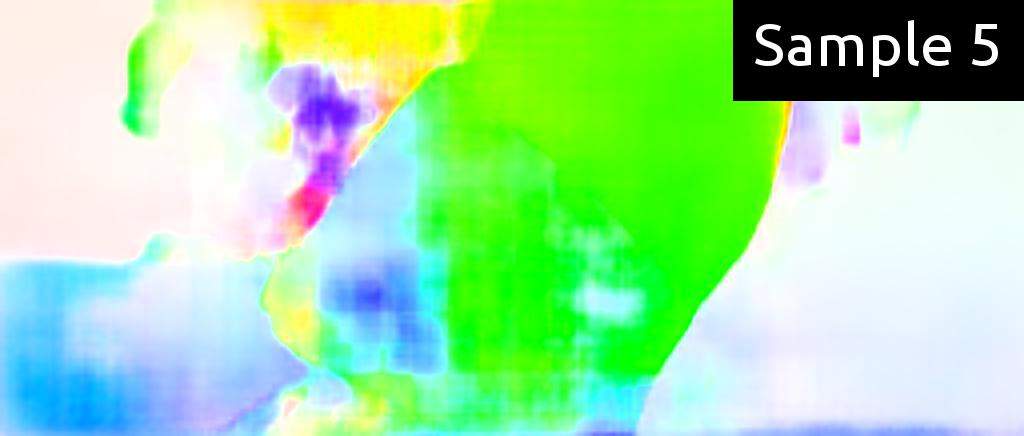}
&
\includegraphics[width=0.1666\linewidth]{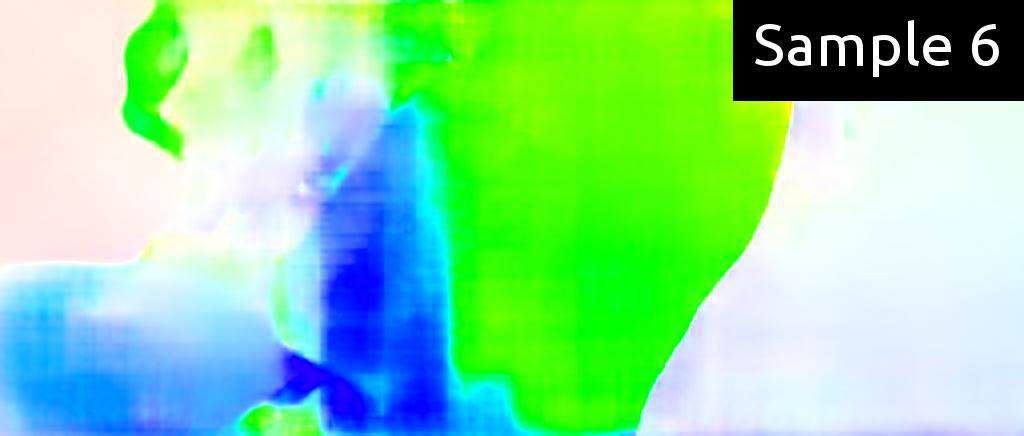}
&
\includegraphics[width=0.1666\linewidth]{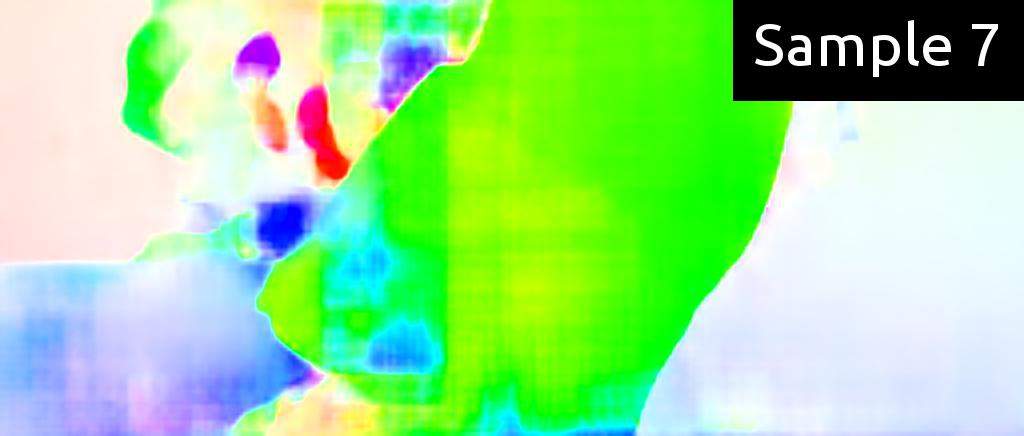}
&
\includegraphics[width=0.1666\linewidth]{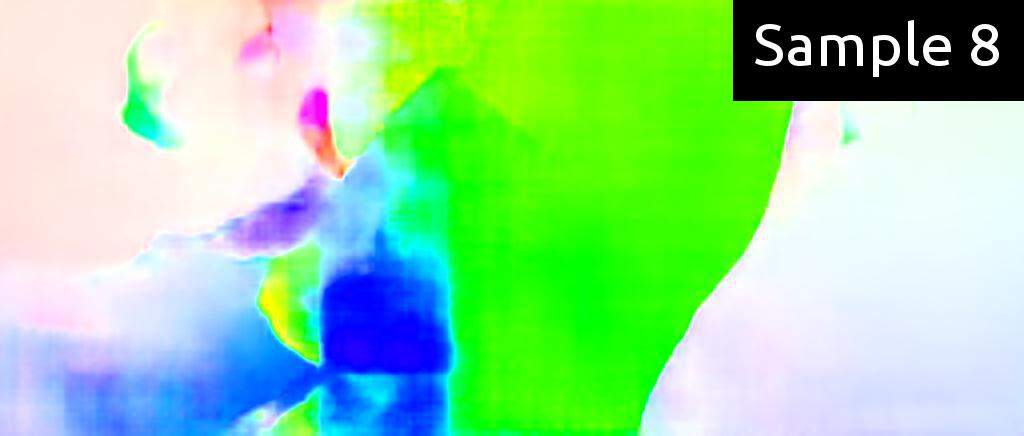}
&
&
\\

\hline

                \end{tabular}
            }
        \end{center}
            \caption{In this table we show the outputs of empirical experiments with all presented methods for a hard Sintel example as well as the averaged flows and computed entropies. Some variety of each method is visible, while FlowNetH provides a different kind of output with much more variety. 
            }
            \label{tab:ex1_1}
        \end{table*}
        \egroup

        \bgroup
        \def\arraystretch{1.0}
        \renewcommand{\tabcolsep}{0.05cm}
        \begin{table*}
        \begin{center}
            \resizebox{\linewidth}{!}{%
                \begin{tabular}{|cccc|cc|}
        
\hline
\multicolumn{4}{|l|}{Data:}& & \\ 
\includegraphics[width=0.1666\linewidth]{images/method-overview/ex1/img0.jpg}
&
\includegraphics[width=0.1666\linewidth]{images/method-overview/ex1/img1.jpg}
&
&
&
\includegraphics[width=0.1666\linewidth]{images/method-overview/ex1/flow_gt.jpg}
&
\\

\hline
\hline
\multicolumn{4}{|l|}{FlowNetC Pred:}& & \\ 
&
&
&
&
\includegraphics[width=0.1666\linewidth]{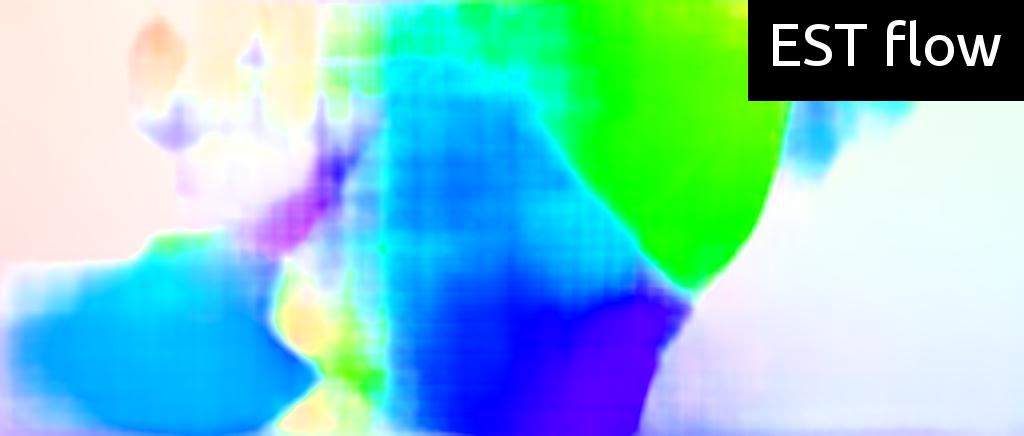}
&
\includegraphics[width=0.1666\linewidth]{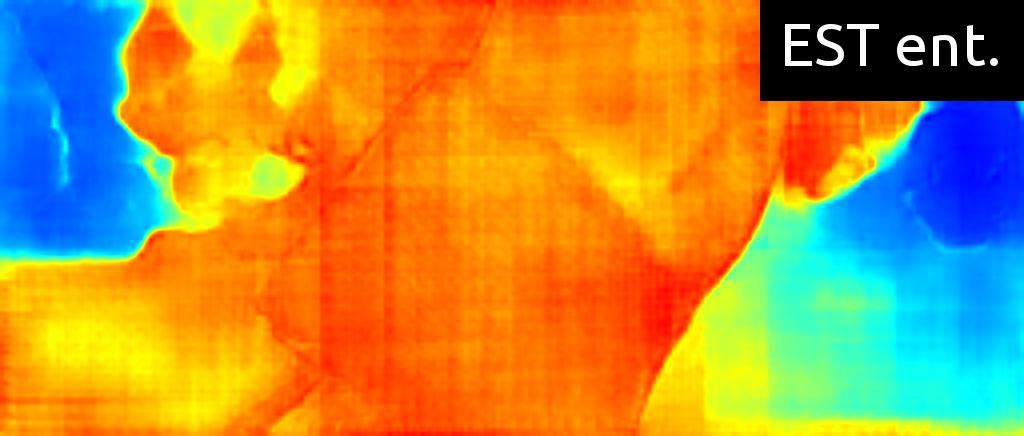}
\\

\hline
\hline
\multicolumn{4}{|l|}{Dropout Pred:}& & \\ 
\includegraphics[width=0.1666\linewidth]{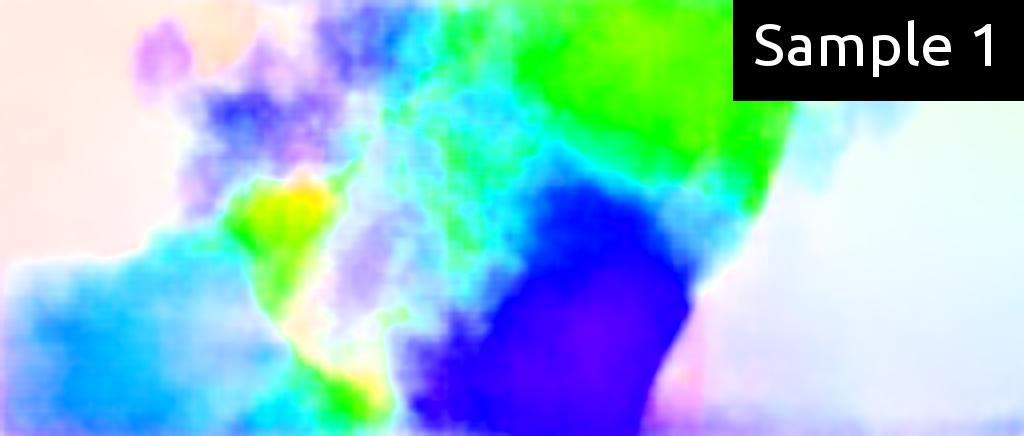}
&
\includegraphics[width=0.1666\linewidth]{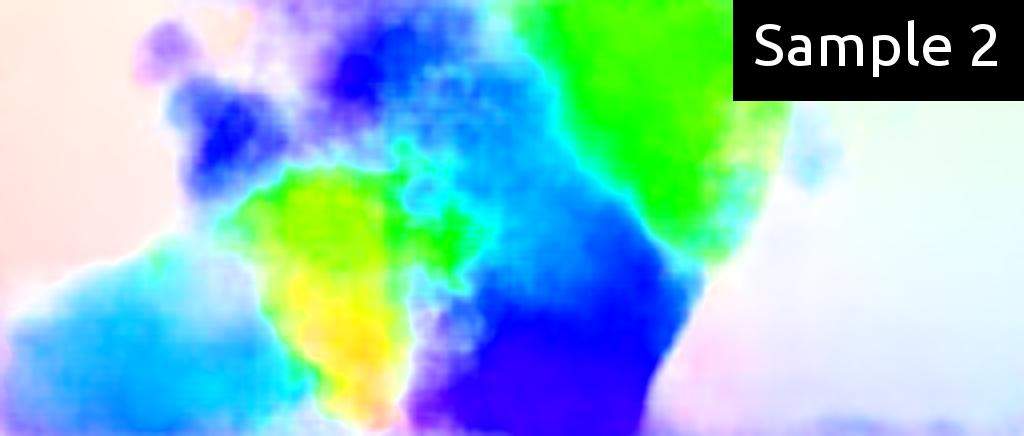}
&
\includegraphics[width=0.1666\linewidth]{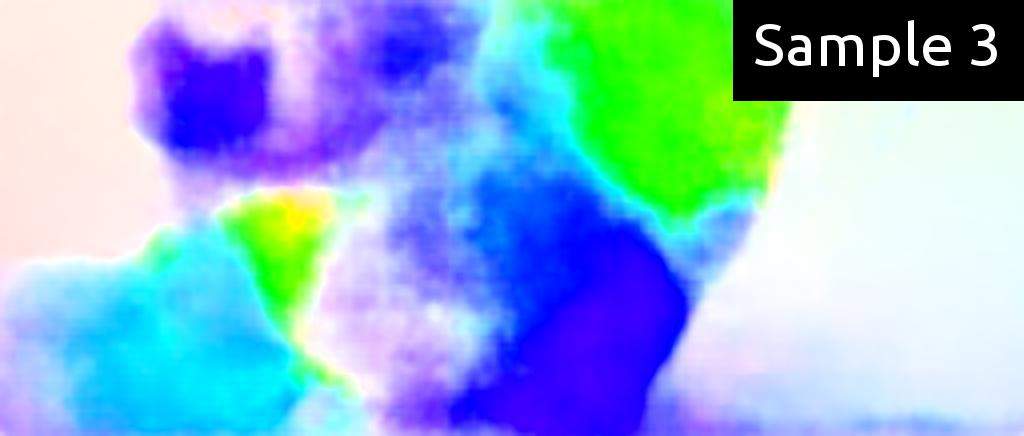}
&
\includegraphics[width=0.1666\linewidth]{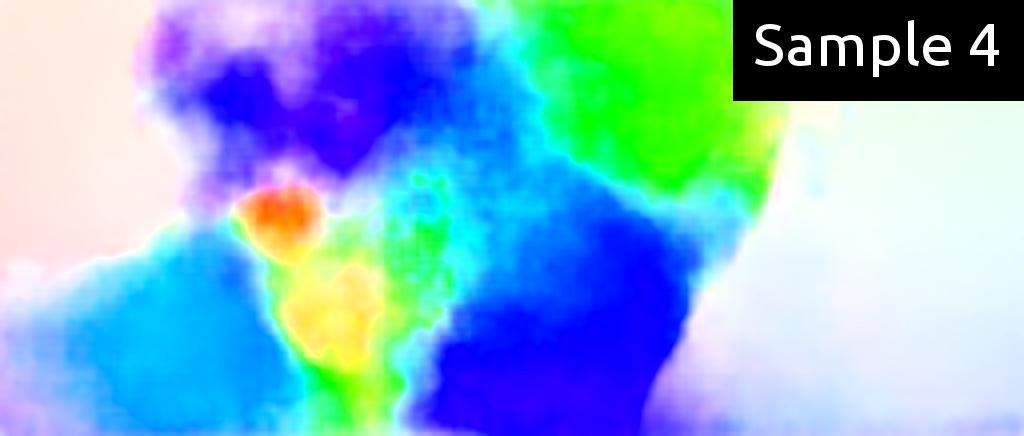}
&
\includegraphics[width=0.1666\linewidth]{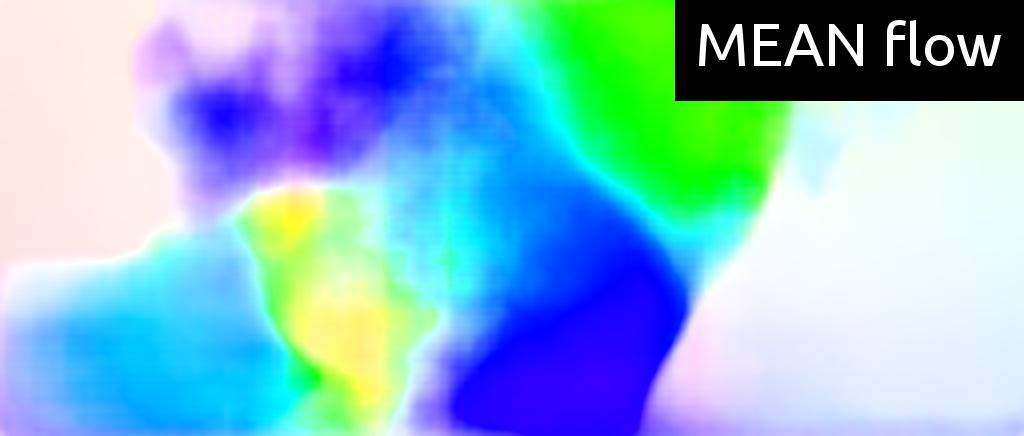}
&
\includegraphics[width=0.1666\linewidth]{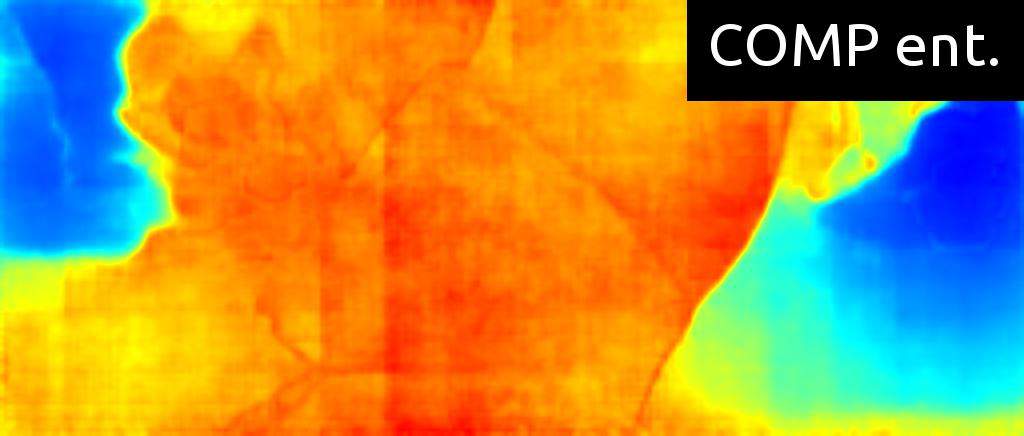}
\\

\includegraphics[width=0.1666\linewidth]{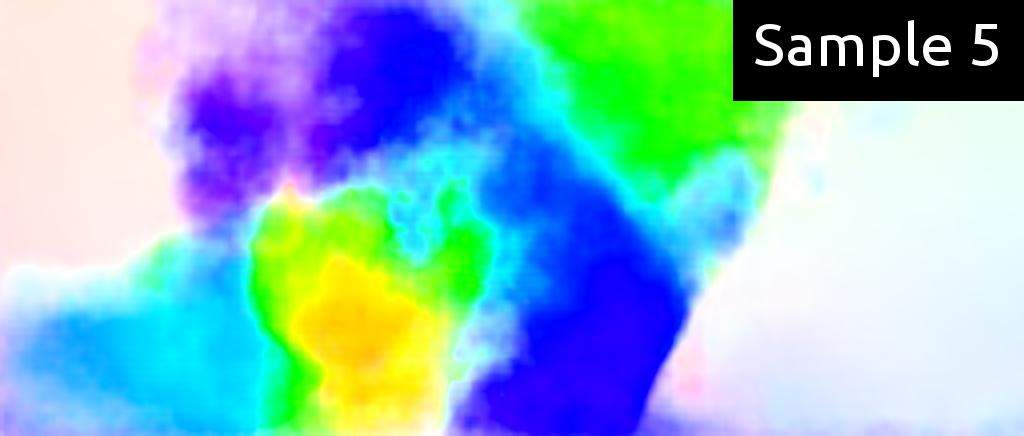}
&
\includegraphics[width=0.1666\linewidth]{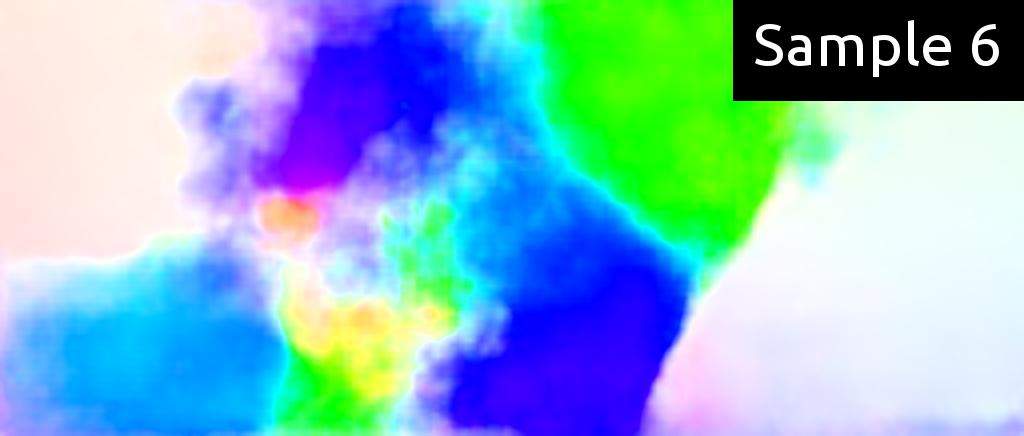}
&
\includegraphics[width=0.1666\linewidth]{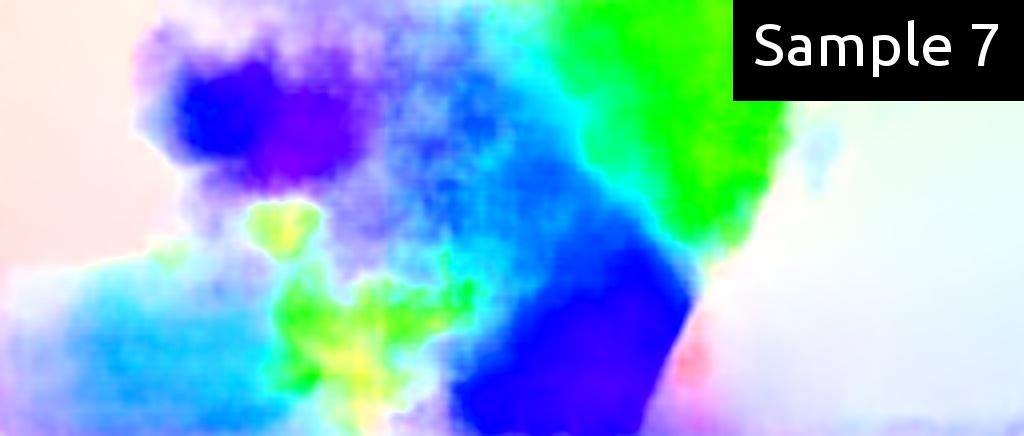}
&
\includegraphics[width=0.1666\linewidth]{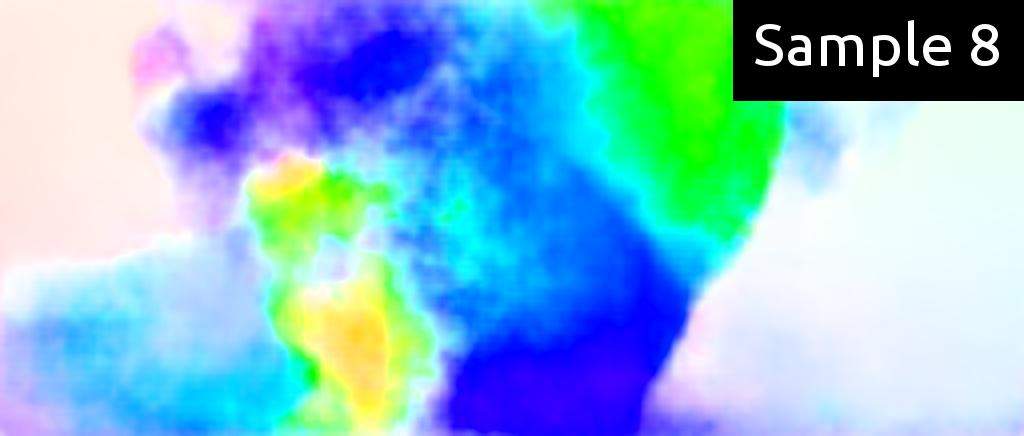}
&
&
\\

\hline
\hline
\multicolumn{4}{|l|}{SGDR Pred:}& & \\ 
\includegraphics[width=0.1666\linewidth]{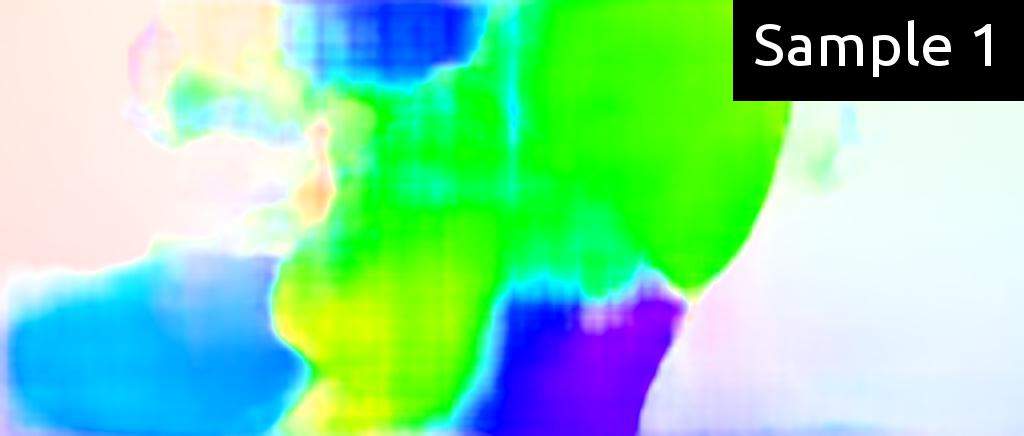}
&
\includegraphics[width=0.1666\linewidth]{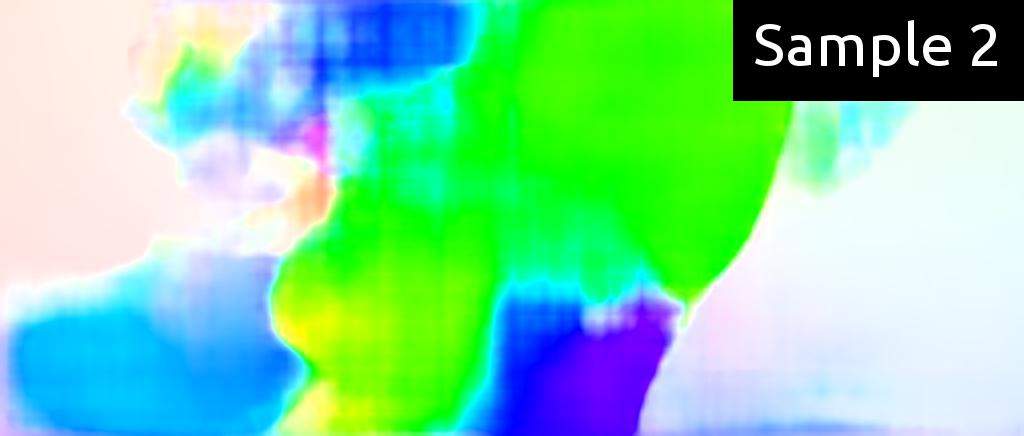}
&
\includegraphics[width=0.1666\linewidth]{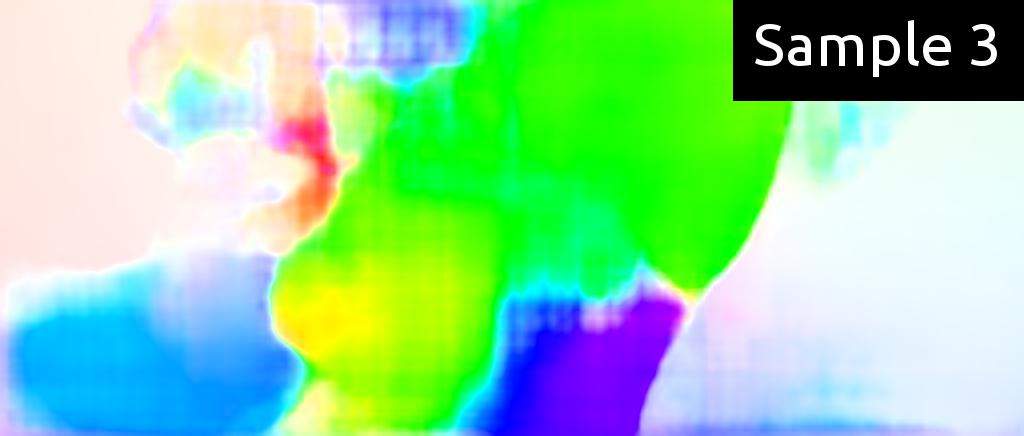}
&
\includegraphics[width=0.1666\linewidth]{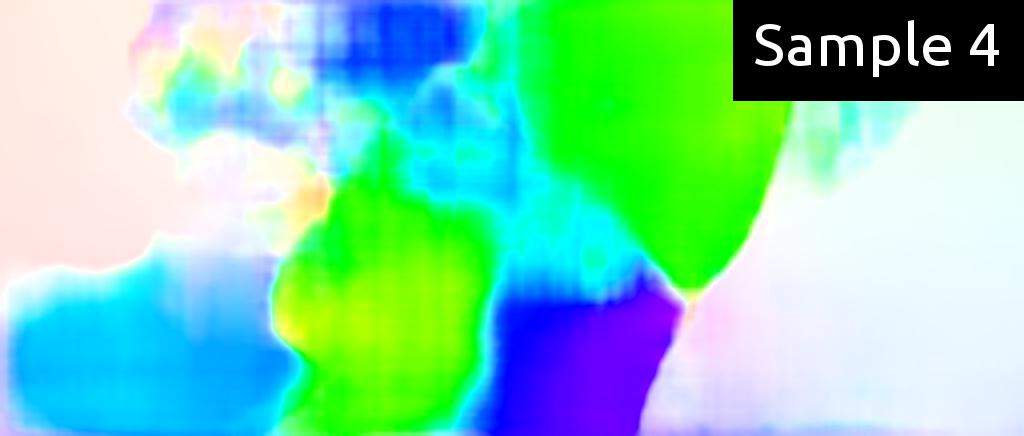}
&
\includegraphics[width=0.1666\linewidth]{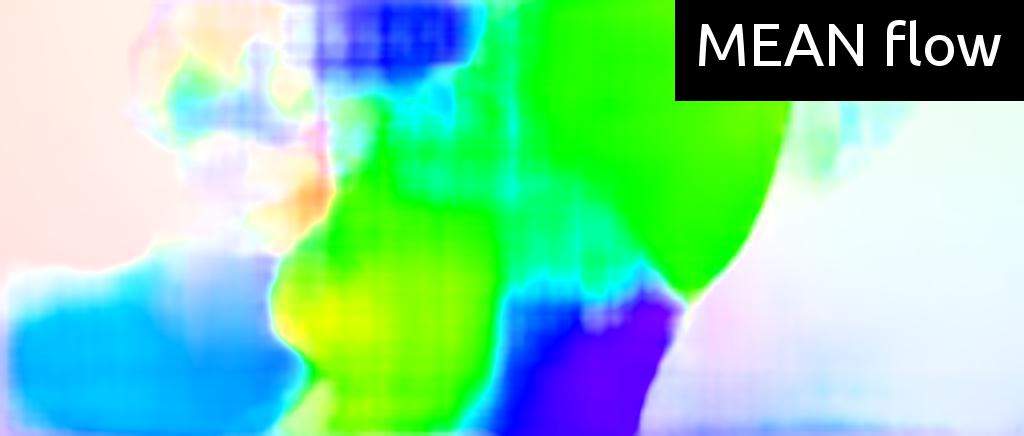}
&
\includegraphics[width=0.1666\linewidth]{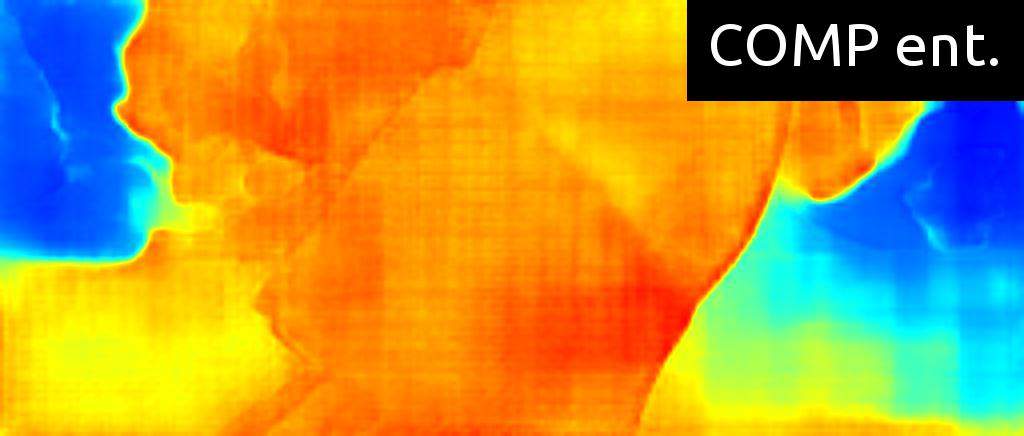}
\\

\includegraphics[width=0.1666\linewidth]{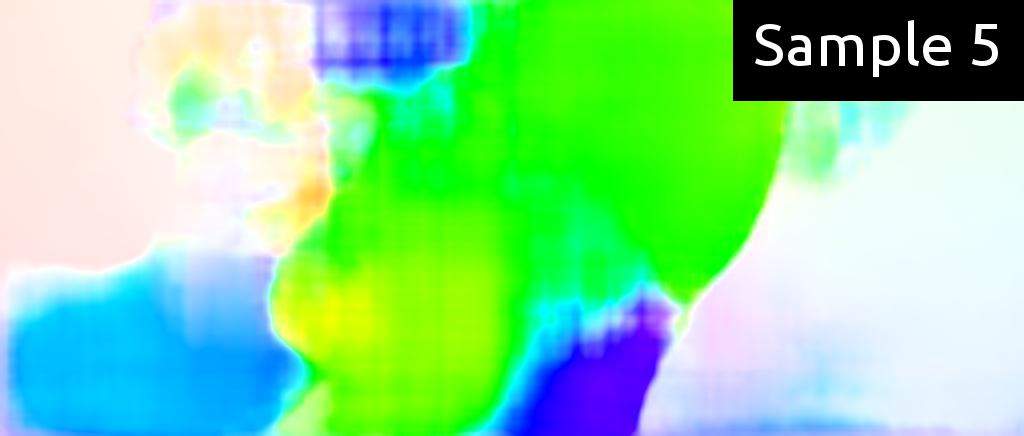}
&
\includegraphics[width=0.1666\linewidth]{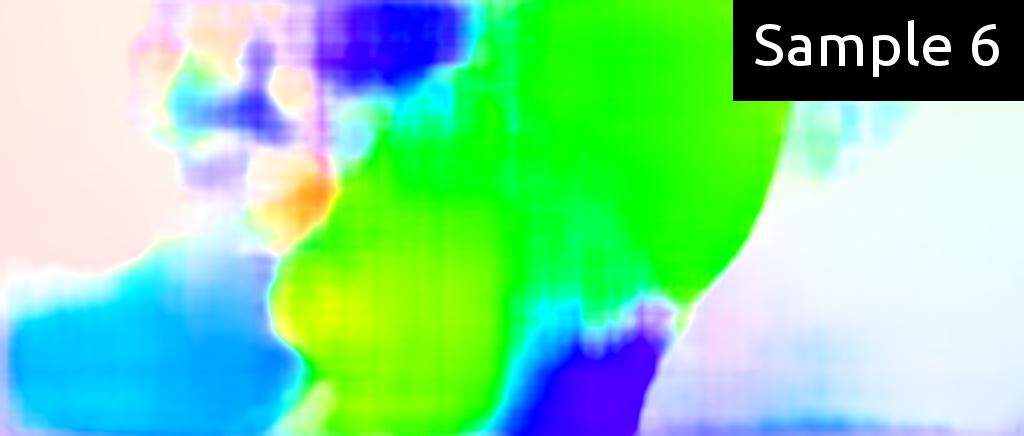}
&
\includegraphics[width=0.1666\linewidth]{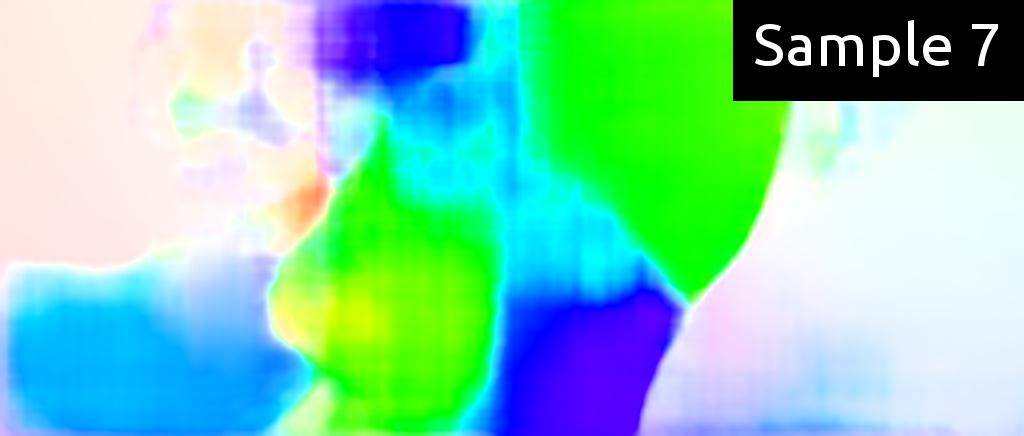}
&
\includegraphics[width=0.1666\linewidth]{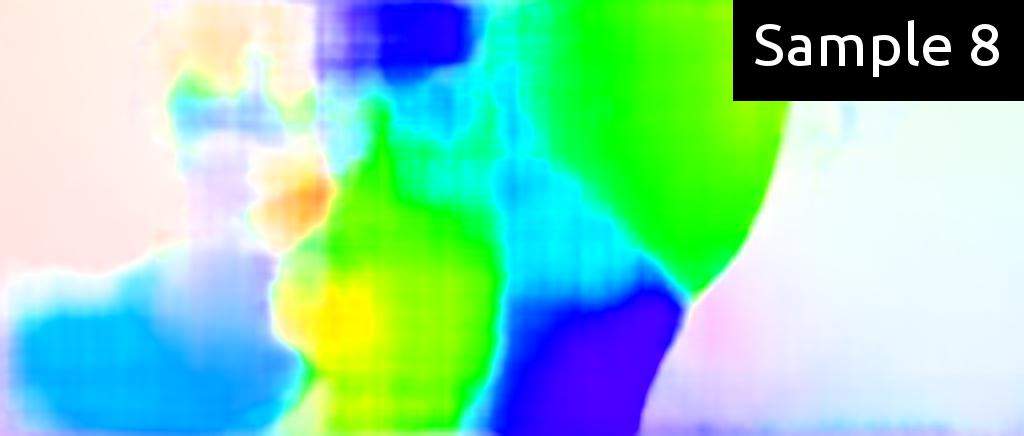}
&
&
\\

\hline
\hline
\multicolumn{4}{|l|}{BootstrappedEnsemble Pred:}& & \\ 
\includegraphics[width=0.1666\linewidth]{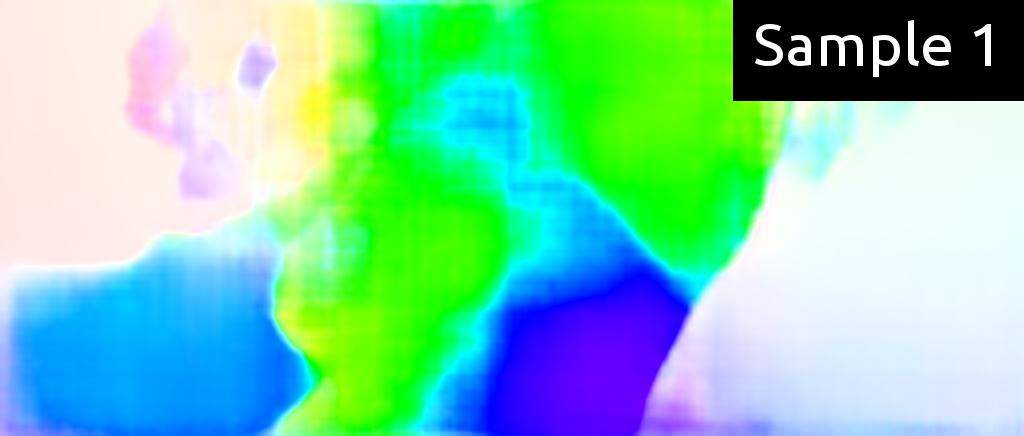}
&
\includegraphics[width=0.1666\linewidth]{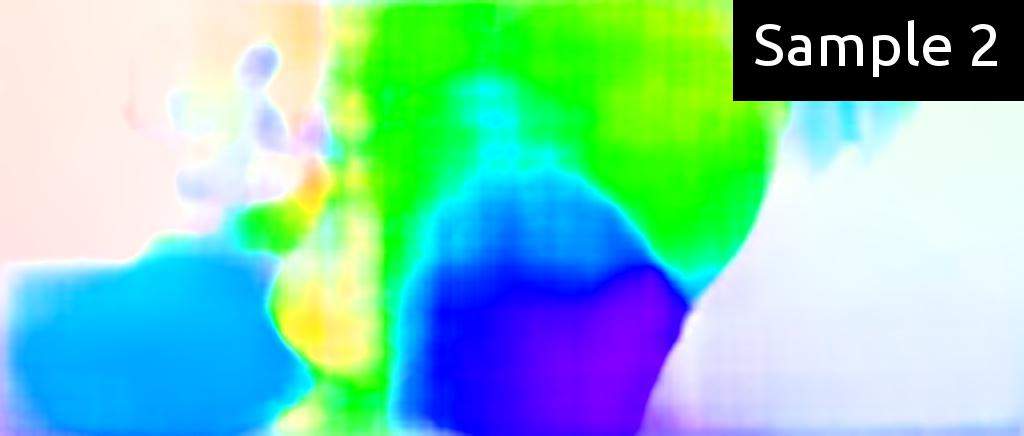}
&
\includegraphics[width=0.1666\linewidth]{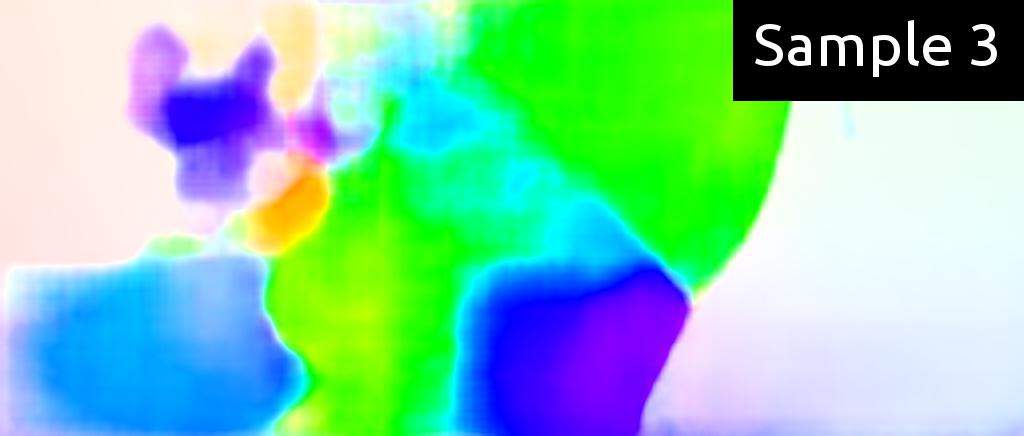}
&
\includegraphics[width=0.1666\linewidth]{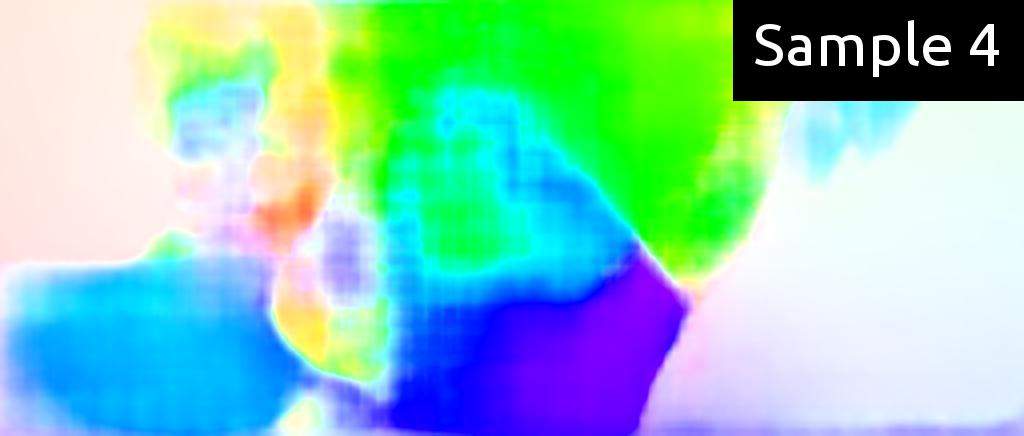}
&
\includegraphics[width=0.1666\linewidth]{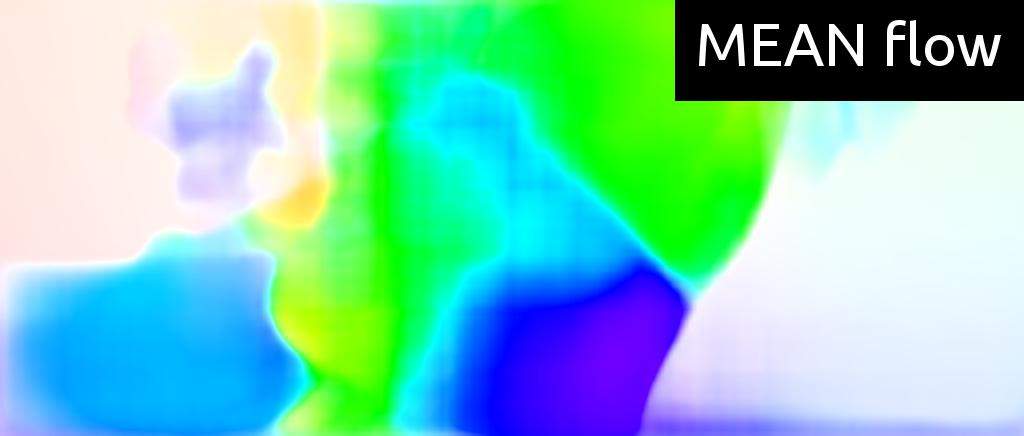}
&
\includegraphics[width=0.1666\linewidth]{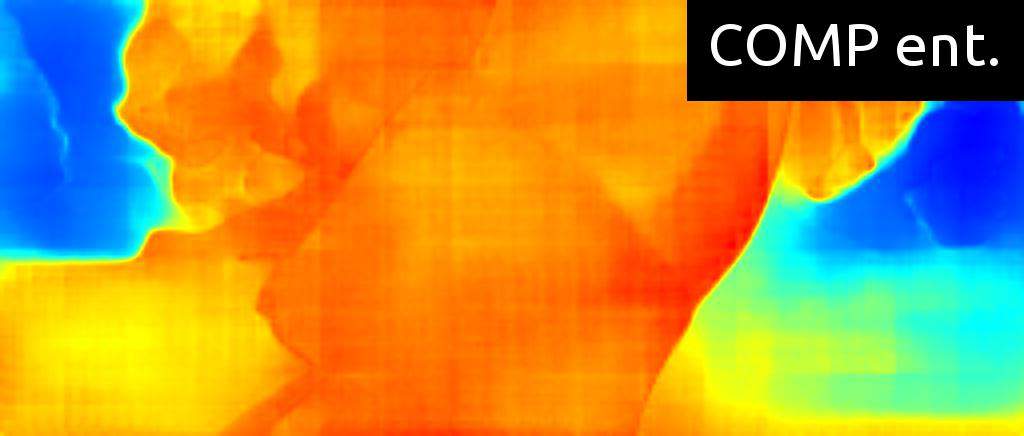}
\\

\includegraphics[width=0.1666\linewidth]{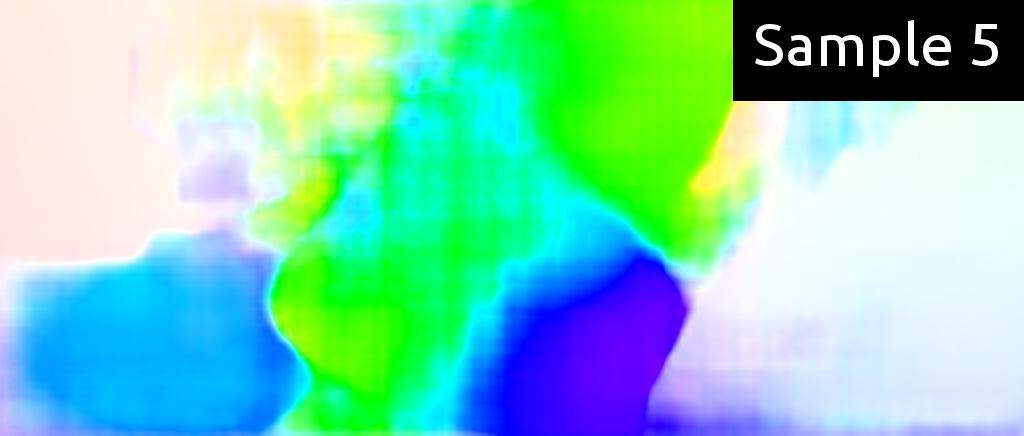}
&
\includegraphics[width=0.1666\linewidth]{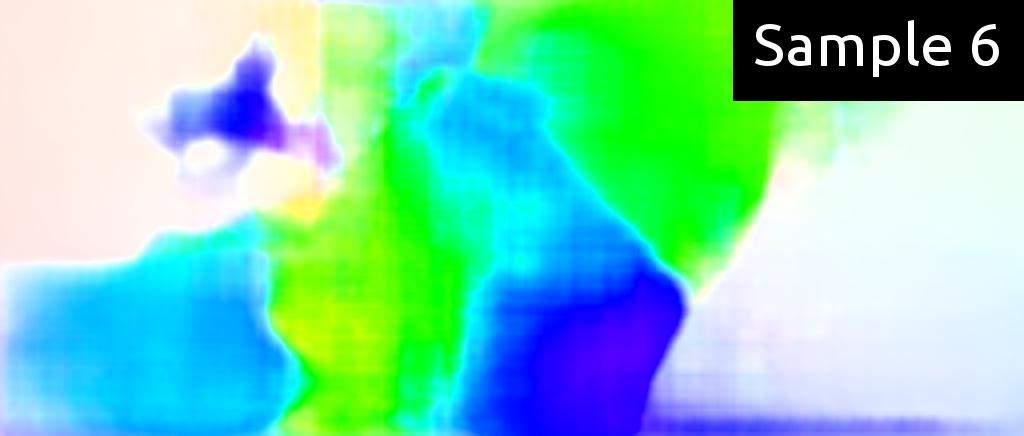}
&
\includegraphics[width=0.1666\linewidth]{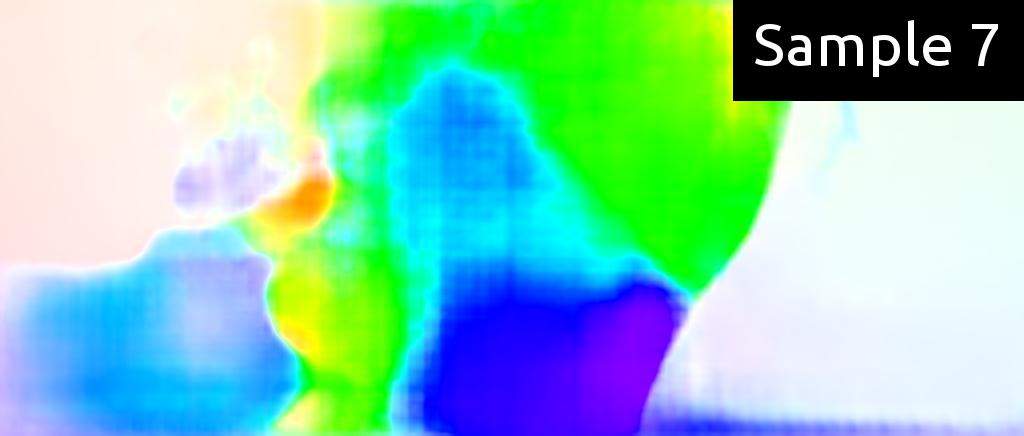}
&
\includegraphics[width=0.1666\linewidth]{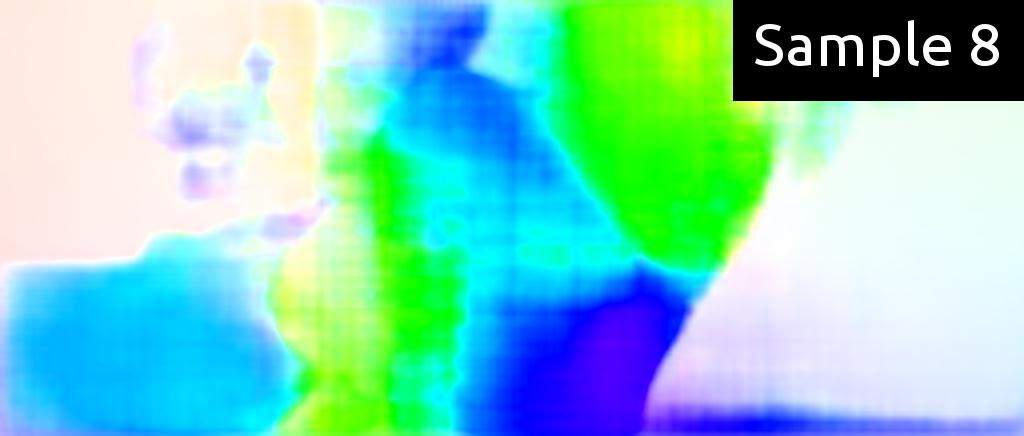}
&
\includegraphics[width=0.1666\linewidth]{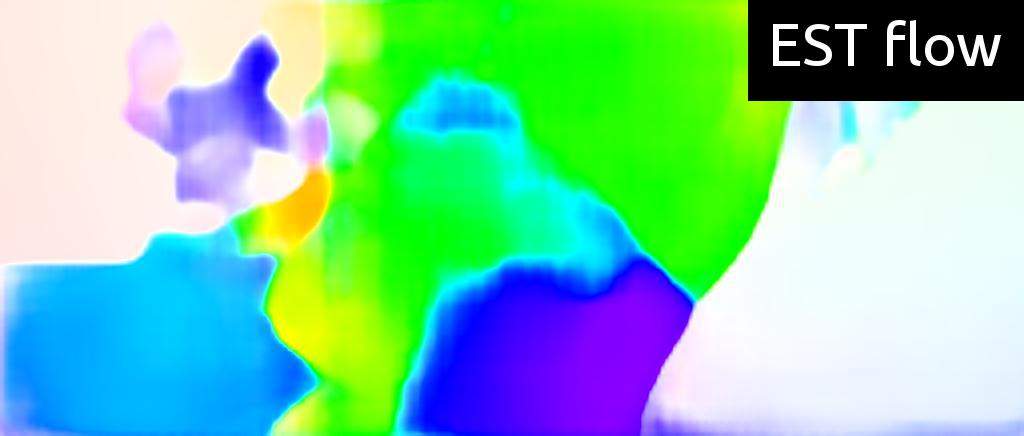}
&
\includegraphics[width=0.1666\linewidth]{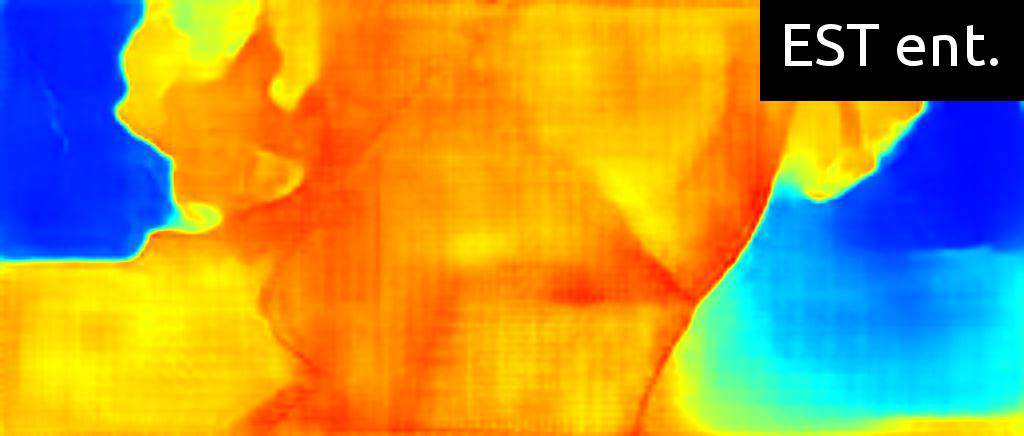}
\\

\hline
\hline
\multicolumn{4}{|l|}{FlowNetH Pred-Merged:}& & \\ 
\includegraphics[width=0.1666\linewidth]{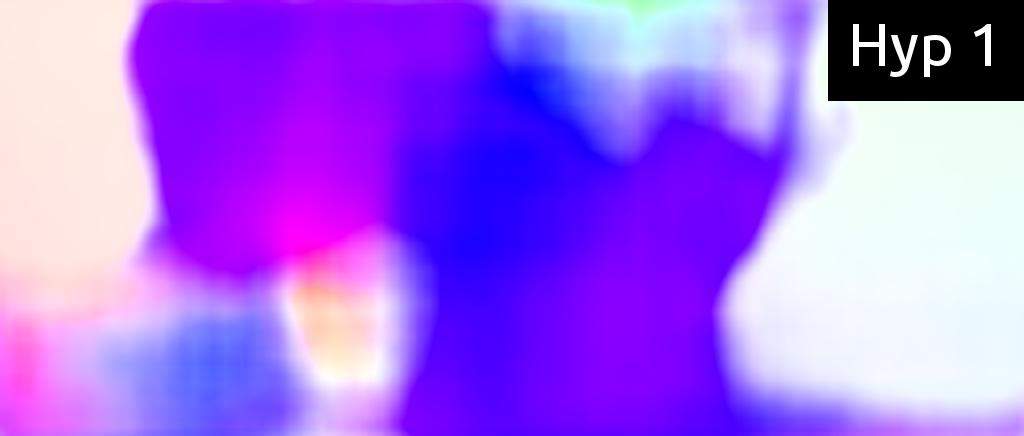}
&
\includegraphics[width=0.1666\linewidth]{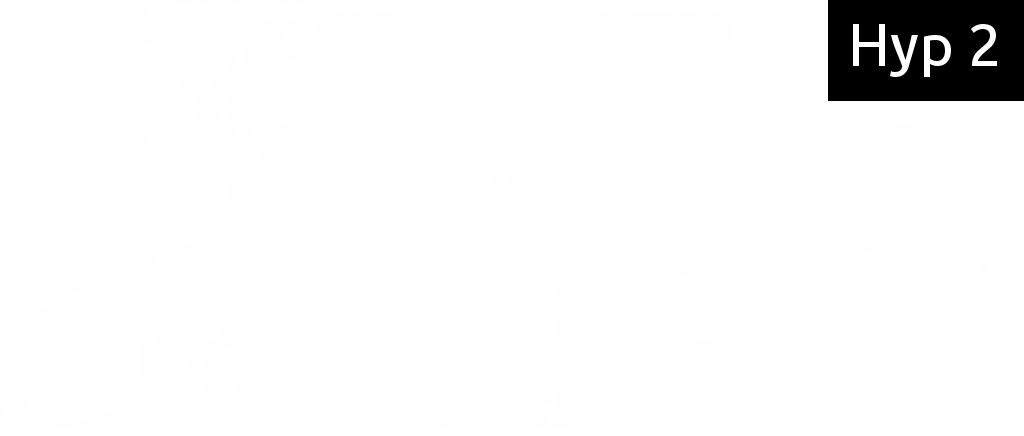}
&
\includegraphics[width=0.1666\linewidth]{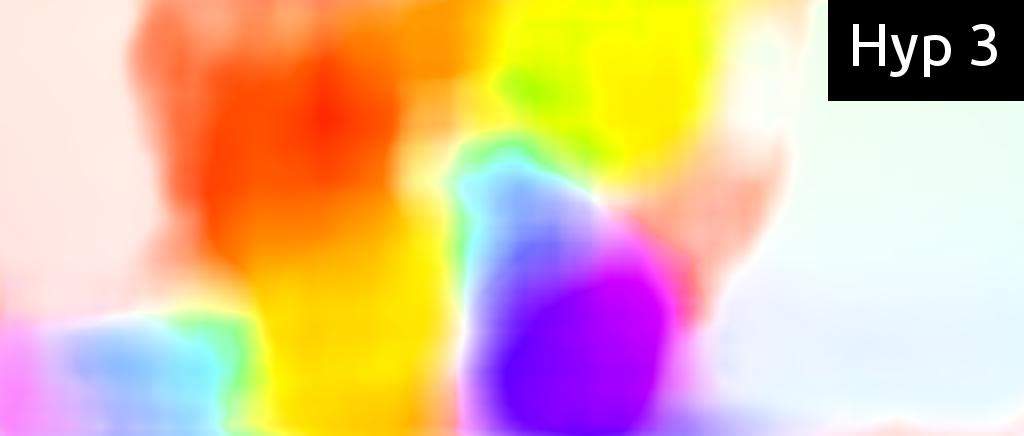}
&
\includegraphics[width=0.1666\linewidth]{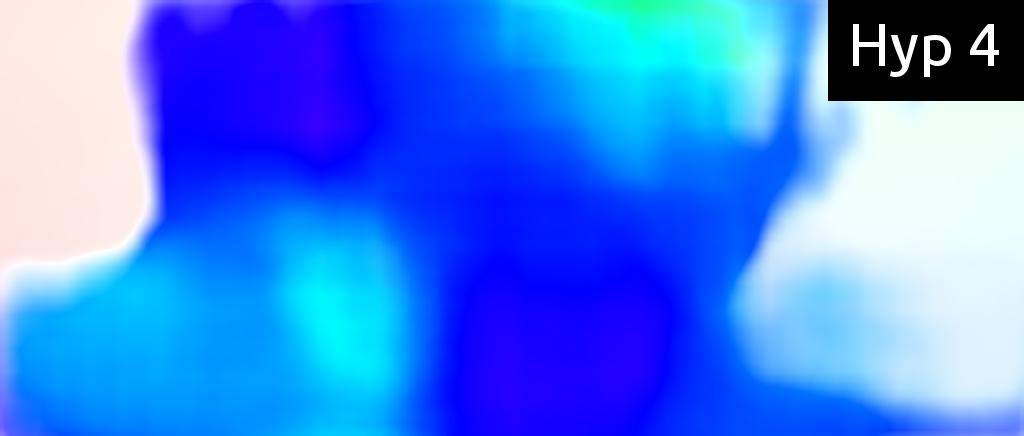}
&
&
\\

\includegraphics[width=0.1666\linewidth]{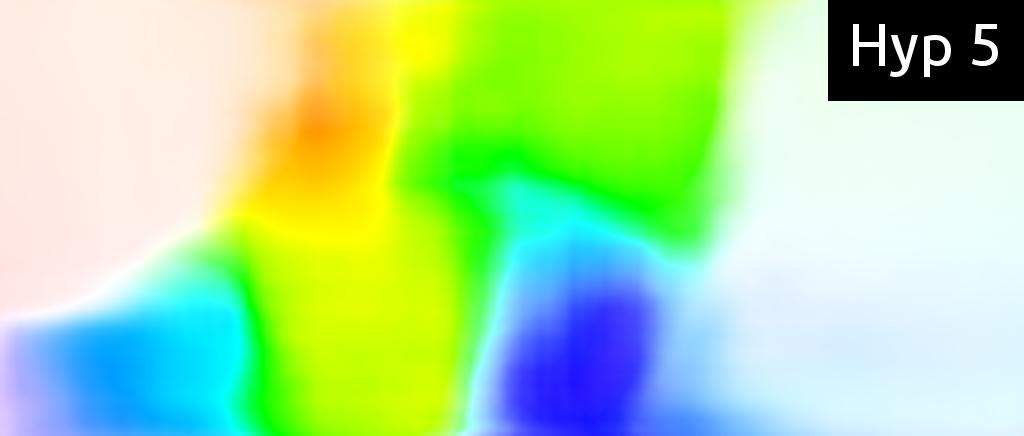}
&
\includegraphics[width=0.1666\linewidth]{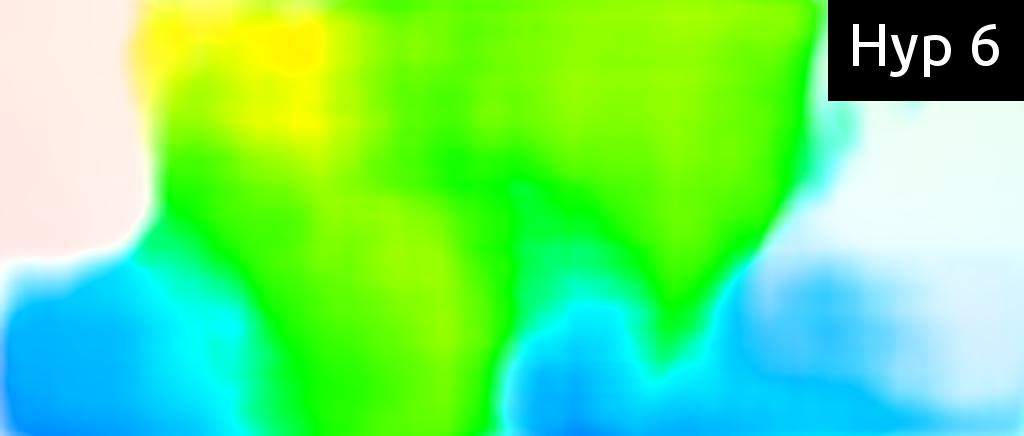}
&
\includegraphics[width=0.1666\linewidth]{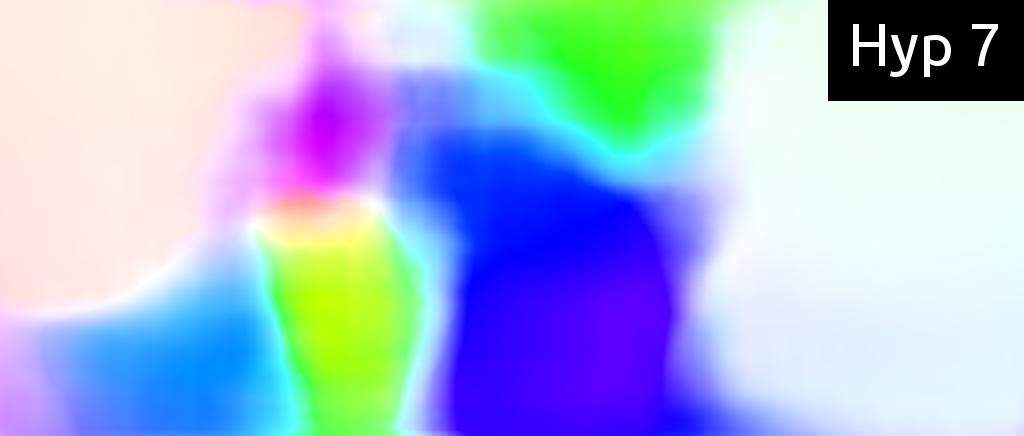}
&
\includegraphics[width=0.1666\linewidth]{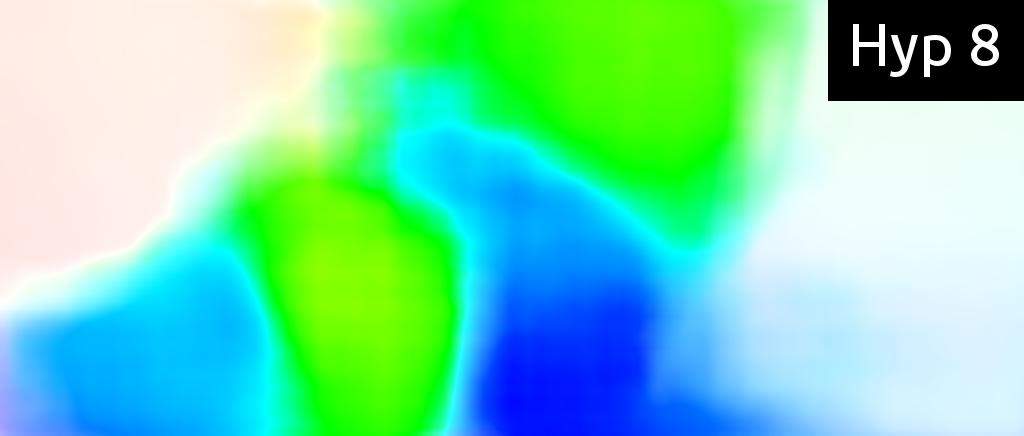}
&
\includegraphics[width=0.1666\linewidth]{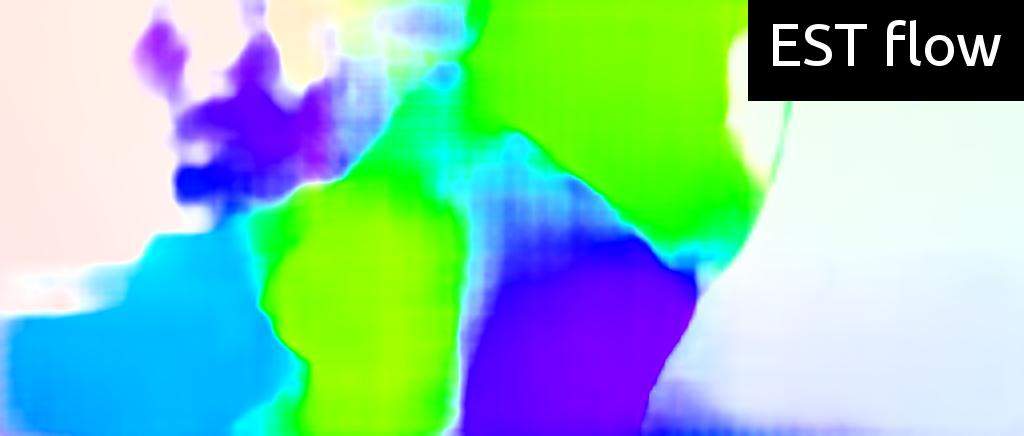}
&
\includegraphics[width=0.1666\linewidth]{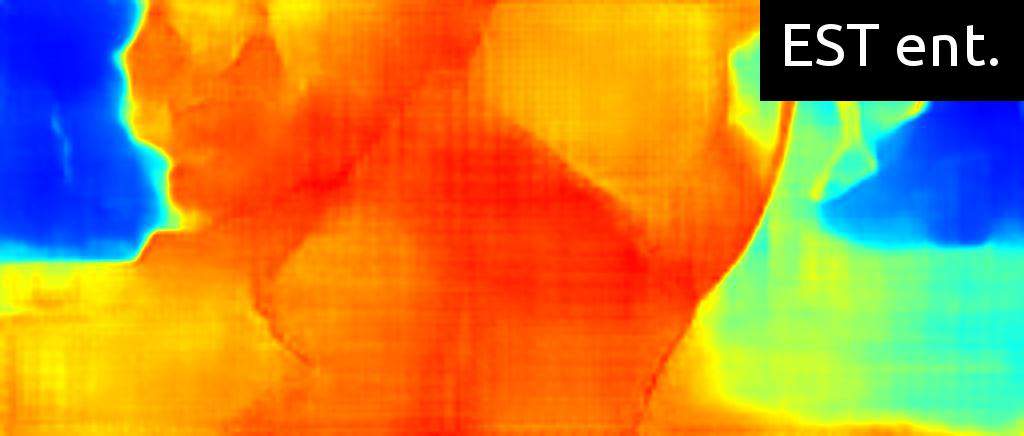}
\\

\hline

                \end{tabular}
            }
       \end{center}
             \caption{
            In this table we show the outputs of predictive experiments with all presented methods for a hard Sintel example, as well as the averaged flows and computed entropies. For BootstrappedEnsemble-Pred-Merged and FlowNetH-Pred-Merged we show also the estimated flow and estimated entropy as the output of the merging network on top. The hypothesis estimated by FlowNetH-Pred-Merged is the most diverse one. In the second hypothesis, the motion predicted is very small and could be corresponding to the background.
            \label{tab:ex1_2}
            }
        \end{table*}
        \egroup

\end{document}